%% file: main.tex
\documentclass[11pt]{article}

\newif\ifarrreview
\ifdefined\ARRreview \arrreviewtrue \fi
\ifarrreview
  \usepackage[review]{acl}
\else
  \usepackage[preprint]{acl}
\fi

\usepackage{times}
\usepackage{latexsym}
\usepackage[T1]{fontenc}
\usepackage[utf8]{inputenc}
\usepackage{microtype}
\usepackage{inconsolata}
\usepackage{graphicx}
\usepackage{booktabs}
\usepackage{multirow}
\usepackage{makecell}
\usepackage{array}
\usepackage{amsmath}
\usepackage{amssymb}
\usepackage{xcolor}
\usepackage{enumitem}

\newcommand{\crit}[1]{\texttt{#1}}

\newcolumntype{L}[1]{>{\raggedright\arraybackslash}p{#1}}


\makeatletter
\newcommand{\holdtopfloats}{\global\@topnum\z@}
\makeatother

\makeatletter
\AtBeginDocument{%
  \let\jev@aclmaketitle\maketitle
  \def\maketitle{{%
    \def\@xfootnotenext[##1]{%
      \begingroup
        \csname c@\@mpfn\endcsname ##1\relax
        \unrestored@protected@xdef\@thefnmark{\thempfn}%
      \endgroup
      \@footnotetext}%
    \jev@aclmaketitle}}%
}
\makeatother

\newcommand{\dg}{\rlap{\textsuperscript{\dag}}}
\newcommand{\ddg}{\rlap{\textsuperscript{\ddag}}}
\newcommand{\scm}{\rlap{\textsuperscript{\S}}}

\title{Decision Models vs.\ LLMs as Rubric Judges:\\Cheaper, Faster, and Wrong in the Same Places}

\author{Delip Rao\thanks{Corresponding author.} \\
  University of Pennsylvania \\
  \texttt{delip@upenn.edu} \\\And
  Chris Callison-Burch \\
  University of Pennsylvania \\
  \texttt{ccb@upenn.edu} \\}

\begin{document}
\maketitle

\begin{abstract}
LLM judges now score outputs against rubrics, in benchmarks and as training rewards. Decision models, classifier-like alternatives, return probabilities over permitted answers with a confidence score but no free text. Two such models are Jev and an open reproduction, Perplexity's Decider. We compare each with three flash-tier LLM judges on nine panels from seven benchmarks with human judgments, giving every judge identical criterion texts. The LLM judges run in two setups, reading a whole rubric at once, as decision models do, or one criterion at a time. Our findings suggest both decision models can often stand in for them. The LLM judges cost 16 to 357 times as much and take 5 to 350 times as long, yet in both setups each decision model trails them significantly in at most 6 of 27 comparisons. The two kinds of judge err alike, agreeing more with one another than with the human-annotated judgment labels on ordinal criteria. More than 85\% of flash-tier LLM verdicts on either decision model's most confident errors repeat them, where independent errors would give about half. Each decision model's confidence ranks its errors on most ordinal panels, so each should make a natural first stage of a cascade deferring uncertain verdicts to an LLM judge. Yet with flash-tier fallbacks, such cascades gained a mere 2.7 points at best over the strongest single judge, even with oracle thresholds, which suggests they mainly save cost. A cascade needs a fallback that is right where its first stage is wrong, but the LLM judges are mostly wrong in the same places as Jev, and the Decider is already the strongest single judge on most panels. Even each vendor's higher-tier model repeats those errors, and as a fallback it lifted no cascade more than 1.7 points above both tiers' strongest single judge. Holding across setups, effort levels, and tiers, these findings suggest designing future decision models afresh to get right what LLM judges get wrong, and vice versa.
\end{abstract}

\holdtopfloats
\begin{figure}[t]
\centering
\includegraphics[width=\columnwidth]{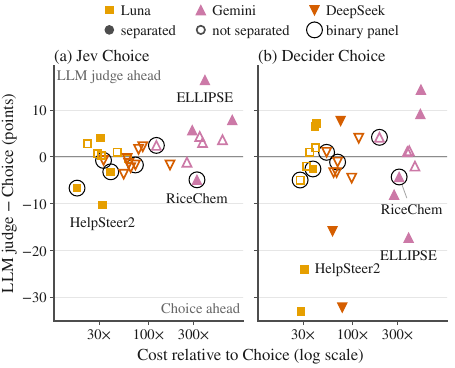}
\caption{LLM-based rubric judges against the Choice framing of (a) Jev and (b) Perplexity's Decider (Section~\ref{sec:jev}; \texttt{pplx-decider-v1.1-27b}). In most comparisons with Jev, and in about half with the Decider, the LLM judges are not separated (Section~\ref{sec:metrics}) from the decision model in accuracy, though they cost far more. Similar results are observed for Jev with holistic rubric judges (Figure~\ref{fig:whole-rubric}).}
\label{fig:parity}
\end{figure}

\section{Introduction}
\label{sec:intro}

LLM judges now return verdicts on the individual criteria of rubrics, for benchmarks~\citep{arora2025healthbench} and as training rewards~\citep{gunjal2026rubrics}, and are validated against human labels~\citep{zheng2023judging,bavaresco2025llms}. We call the judged submission a \emph{unit} and a (unit, criterion) combination a \emph{pair}. A per-criterion grader sends each pair as its own request~\citep{rao2026autorubric}, which multiplies the requests by the number of criteria but raises cost far less (Appendix~\ref{app:wholerubric}).

A \emph{typed classifier} answers a question about a structured input with a probability distribution over a fixed set of permitted answers and writes no free text. Jev, from TypeSafe, is one; its creators call it a ``System One'' model\footnote{The name comes from the System~1/System~2 distinction in \citet{kahneman2011thinking} between fast, automatic judgment and slow, deliberate reasoning. Vendor documentation: \url{https://docs.typesafe.ai}, accessed 23 September 2026.} and a ``decision model'', the term we use for the class. Perplexity's Decider (\texttt{pplx-decider-v1.1-27b}), which a community index lists among open reproductions of Jev, is a second decision model. Jev, the original, is our reference, and we do not compare the two with each other. A decision model judges every criterion of a unit in one request, with a probability or confidence for each answer, so it could replace an LLM judge or pick the pairs that need one. It returns only probabilities, whereas LLM judges reason before they answer. That difference in design suggests the two kinds of judge would err in different places, and a cascade can beat the LLM judge it defers to only if they do.

We set both decision models against three \emph{flash-tier} LLM judges, from tiers their vendors call fast and lower-priced (Appendix~\ref{app:protocol}): GPT-5.6 Luna, Gemini 3.8 Flash, and DeepSeek V4.1 Flash (Luna, Gemini, and DeepSeek). Earlier work also chose smaller judges, partly to cut cost~\citep{verga2024replacing,chlapanis2025greekbarbench,xie2026small}. The judges all receive the same 5,003 pairs from nine \emph{panels}, sampled from seven public benchmarks with human judgments (Table~\ref{tab:panels}). Two are \emph{binary panels}, with only binary criteria, and seven are \emph{ordinal panels}, which rate most criteria on ordered levels. Every judge reads the same criterion texts, fixed in advance with no examples, rewording, or tuning. We run the LLM judges one criterion at a time, on whole rubrics as the decision models read them, at high reasoning effort, and in each vendor's higher tier. The answers to our three questions, given below for each decision model, hold in all four conditions, with each condition's own counts (Table~\ref{tab:conditions}):
\begin{enumerate}[label=\textbf{RQ\arabic*.},leftmargin=*,nosep]
\item Can a typed classifier stand in for LLM judges, and at what cost and speed?
\item How different are the classifier's verdicts from the LLM judges'?
\item How well does a confidence-based cascade from the classifier to an LLM judge work?
\end{enumerate}

\textbf{RQ1: Often, and far more cheaply.} Against per-criterion LLM judges, Jev differs significantly in only 8 of 27 paired accuracy comparisons, leading mostly on binary panels and trailing only on ordinal ones, and reaches parity (within 5 points) in 8 more (Figure~\ref{fig:parity}; Section~\ref{sec:parity}). The Decider differs significantly in 13 (one of them also at parity), leading in 8, mostly on ordinal panels, and trailing in 5, so it is not significantly worse in 22. Across both judging modes, the LLM judges' costs were 16 to 325 times Jev's and their wall times 28 to 350 times. Against the Decider, their costs were 18 to 357 times its own and their wall times 5.4 to 67.5 times.

\textbf{RQ2: Less than their designs suggest.} With either decision model among them, the judges agree with one another more than with the labels on each ordinal panel. On six of the seven, Jev and every LLM judge place units lower on average than the raters did, a pattern unstated rating conventions could produce (Section~\ref{sec:shared}). The Decider is the exception. In its Choice framing it places units lower on only four panels, and in its Score framing it is about as accurate on ELLIPSE as the raters are against one another. The judges make \emph{correlated errors}, the ``same places'' of our title. When one judge is wrong, another gives the same wrong verdict more often than independent errors would. Of the LLM verdicts on Jev's most confident errors, 96.0\% repeat its answer, where independence predicts 50.3\%; on the Decider's, 86.5\% do, against 47.9\%.

\textbf{RQ3: Well for cost, not for accuracy.} Both decision models' confidence ranks their errors on six ordinal panels, as a first stage requires, and Jev's probabilities are calibrated on the binary panels. But a cascade gains over its fallback only on kept pairs that its first stage alone gets right. With Jev first, correlated errors leave few. A Decider-first cascade meets a different limit, since on six of nine panels the best single judge it must beat is already one of the Decider's own framings. Cascades cut cost, yet even with oracle thresholds, Jev-first cascades beat the best single judge by at most 2.7 points with flash-tier fallbacks, and Decider-first ones by at most 1.7. With higher-tier fallbacks and oracle thresholds, the largest gain over either tier's best single judge is 1.7 points for Jev-first cascades and 1.4 for Decider-first ones (Section~\ref{sec:routing}).

\section{Related Work}
\label{sec:related}

LLM judges show position, verbosity, and self-enhancement biases, yet GPT-4 agrees with human preferences about as often as humans do~\citep{zheng2023judging}, and LLMs given the evaluators' instructions rate consistently with experts~\citep{chiang2023can}; on JUDGE-BENCH, which includes both USR sets, reliability varies with property and annotator expertise~\citep{bavaresco2025llms}. Uncertain labels can make a machine look human-level~\citep{elangovan2025beyond}. High agreement can hide score differences~\citep{thakur2025judging}, and LLM essay graders score below humans~\citep{kundu2024are} and penalize minor errors that humans discount~\citep{mathew2026llms}, so we report the offset.

Prometheus and Prometheus~2 train open evaluators on user-supplied rubrics~\citep{kim2024prometheus,kim2024prometheus2}, and CheckEval's binary checklists raise agreement~\citep{lee2025checkeval}. HealthBench refined its criteria for its grader~\citep{arora2025healthbench}, and LLM-Rubric maps answer distributions to each human's labels~\citep{hashemi2024llmrubric}, a layer that would absorb an offset; we fit no such layer.

Across more than 350 models, larger and more accurate models make highly correlated errors, even across providers~\citep{kim2025correlated}. Mistakes grow more alike with capability~\citep{goel2025great}, and juries of smaller judges from different vendors can reduce bias~\citep{verga2024replacing}. Closed-set probabilities carry information about correctness~\citep{guo2017calibration,kadavath2022language} and serve to abstain~\citep{geifman2017selective} or to escalate to costlier models~\citep{chen2024frugalgpt}; Trust or Escalate sets its threshold on a calibration set to guarantee human agreement~\citep{jung2025trust}.

\section{Judges, Panels, and Protocol}
\label{sec:setup}

All judges run in the open-source AutoRubric framework~\citep{rao2026autorubric}, the \emph{harness}. An ordinal criterion's options are ordered levels, and its pairs are \emph{ordinal pairs}.

\subsection{The decision models}
\label{sec:jev}

A Jev request carries one question per criterion and a record (the vendor's ``state'') holding the unit's input and submission, as an LLM judge receives them. Jev never sees the harness's system prompt and sees options in scale order. It costs \$0.042 per million input tokens; output is free.

The Decider, whose version 1.1 was released on 5 October 2026, is an open-weight fine-tune of Qwen3.8-27B (its training data are described under Limitations) and takes requests of the same shape through Perplexity's Decisions API at \$0.02 per million input tokens, output free.\footnote{\raggedright\url{https://docs.perplexity.ai/docs/decisions/quickstart}, accessed 6 October 2026.} It answered every request on 7 October 2026, with 8 requests in flight under the vendor's limit of 10 requests per second. Minutes later the requests for the 106 repeat units of Section~\ref{sec:llm} were sent again, to give the Decider its own run-to-run baseline.

Both decision models offer three \emph{primitives}, each with a \emph{decoding rule}: Noul returns the probability of yes, thresholded at 0.5; Choice a distribution over option labels, read by argmax; and Score the expected level over ordered level descriptions, rounded to the closest level, ties to even (Appendix~\ref{app:choicescore}). A \emph{framing} is a primitive with our question wording (Table~\ref{tab:judge-inputs}).

On the binary panels \emph{Jev Choice} asks Choice over MET, UNMET, and CANNOT\_ASSESS, and \emph{Jev Noul} asks Noul, both through the \emph{wrapper}, a fixed instruction with the harness's MET and UNMET definitions. On the ordinal panels Jev Choice gets the criterion sentence as instruction and the LLM judges' option texts, the \emph{NA option} (``Cannot assess / not applicable'') included. \emph{Jev Score} reads the same texts as level descriptions without that option and therefore cannot abstain. On the 364 \emph{embedded binary pairs} of the FED and USR panels, both use Jev Noul. Jev Choice and the LLM judges are the \emph{matched judges}, and with Jev Score the \emph{five judges}. Jev's \emph{confidence}, used only to rank pairs, is the confidence that Choice and Score return, or $2\,|P(\text{yes}) - 0.5|$ for Noul (Appendix~\ref{app:confidence}). \emph{Decider Choice}, \emph{Decider Score}, and \emph{Decider Noul} are these framings sent unchanged to the Decider, whose confidence comes from the same answer fields as Jev's. The Decider gets its own copy of the main analyses, with its own matched judges, five judges, and confidence. Unless a sentence names the Decider, ``matched judges'', ``five judges'', and ``confidence'' refer to Jev's copy, as do ``every judge'' and ``no judge'' in statements of results.

Our expectations were formed on Jev (Appendix~\ref{app:binaryobs}). Of the decision models now served through public APIs, such as Fastino's GLiDE and Cloudflare's Clef, we chose the Decider because it heads the community Decision Index\footnote{\raggedright\url{https://huggingface.co/spaces/multimodalart/jev-decision-index}, edition 0.3, data of 6 October 2026, accessed the same day.} (first of 112 entries at 62.75, with Jev, \texttt{jev-1.13.0}, shown tied for second at 60.11) and discloses its weights, backbone, and training data. That index scores models on classification benchmarks, which need not predict how well a model judges rubrics. We report each decision model only against the LLM judges (Appendix~\ref{app:decider}).

\subsection{The LLM judges}
\label{sec:llm}

Each LLM judge runs through the harness (DeepSeek via OpenRouter; Appendix~\ref{app:protocol}) with its default system prompts and an undated model identifier, at temperature 0 except for Luna's enforced 1.0 (Table~\ref{tab:judge-inputs}), and returns a verdict and an explanation, not a probability. A logged seed shuffles each pair's options against position bias~\citep{wang2024large}. The \emph{main runs} (19 September 2026) set no reasoning effort, so vendor defaults applied: medium for Luna and Gemini, high on DeepSeek's own API, and undocumented on OpenRouter (Appendix~\ref{app:effort}).

There are two \emph{judging modes}. With \emph{per-criterion judgments}, the harness default and the mode of our main runs, the judge takes one criterion of a unit at a time; with \emph{whole-rubric (holistic) judgments} it reads the unit's whole rubric at once and still returns one verdict per criterion. Both practices exist~\citep{liu2023geval,lin2023llmeval}. Whole-rubric judgments make the LLM judges comparable with the decision models; per-criterion judgments stay primary, following work that recommends judging criteria separately to avoid the halo effects documented in human rating~\citep{thorndike1920constant,cooper1981ubiquitous} and related effects in LLM judges~\citep{stureborg2024large,hu2024llm,feuer2025judgment}.

The whole-rubric runs (27 September, eight days later) kept the option orders and sent 8 requests at a time, like the decision models; \emph{repeat runs} resent the whole-rubric requests for a seeded 10\% of units (106) as a run-to-run baseline (Appendix~\ref{app:wholerubric}). The same day, \emph{effort runs} repeated the per-criterion judgments at high reasoning effort on all nine panels, and DeepSeek's also at medium. An effort calibration on the repeat units supports reading Luna's and Gemini's main runs as medium; with DeepSeek's medium run they form the \emph{medium condition}, and the high runs form the \emph{high condition} (Appendix~\ref{app:effort}). On 2 October the \emph{higher-tier runs} repeated the per-criterion judgments with each flash-tier judge's higher-priced sibling, GPT-5.6 Terra (Terra), Gemini 3.1 Pro, or DeepSeek V4 Pro, keeping option orders, temperatures, and requests in flight and leaving reasoning at vendor defaults. These \emph{higher-tier judges} make up the \emph{higher-tier condition} (Appendix~\ref{app:highertier}).

\subsection{Panels and labels}
\label{sec:panels}

The binary panels hold RiceChem's chemistry answers~\citep{sonkar2024automated} and HealthBench's chatbot replies~\citep{arora2025healthbench}. The ordinal panels hold ELLIPSE's learner essays~\citep{crossley2023english}, FED's dialogue turns (FED-Turn) and whole dialogues (FED-Dialogue)~\citep{mehri2020unsupervised}, HelpSteer2's assistant replies~\citep{wang2024helpsteer2}, LFQA's answers to ELI5 questions~\citep{kamoda2025quantifying}, and USR's dialogue replies~\citep{mehri2020usr} on Topical-Chat~\citep{gopalakrishnan2019topicalchat} (USR-TC) and PersonaChat~\citep{zhang2018personalizing} (USR-PC). Each ordinal panel is a 10\% or 20\% sample of units, stratified where possible and drawn with a fixed seed before any judging (Table~\ref{tab:panel-construction}). The binary panels keep their benchmarks' checklists, with question-specific presence checks on RiceChem and consensus behavior checks on HealthBench. On the ordinal panels the judges read our \emph{rubric versions}, \ifarrreview in the supplementary material\else released\fi{} (Appendix~\ref{app:release}); the raters had each benchmark's own instructions. No judge sees the labels, the benchmarks' overall scores, or the strata.

\label{sec:labels}An ordinal pair's \emph{human label} is its raters' mean level rounded to the nearest level, halves up (46.8\% of ELLIPSE labels are half-way), and a binary pair's the rater majority, or RiceChem's teaching-assistant label. \emph{Unanimous pairs}, whose raters all agree, and \emph{split pairs} are reported apart, as rater disagreement can be signal~\citep{plank2022problem}. The \emph{rater reference} scores each rater against the rounded mean of the others on the six ordinal panels with several raters; its target averages one rating fewer than the label, so it does not bound a judge. Two \emph{no-read baselines} ignore the unit: HelpSteer2's constant predictor (60.0\% exact accuracy) and LFQA's provenance baseline, which sees only whether ChatGPT or a Reddit user answered (68.1\%).

\subsection{Metrics and statistics}
\label{sec:metrics}

\newcommand{\mk}[1]{\rlap{\textsuperscript{#1}}}
\begin{table*}[t]
\centering
\footnotesize
\renewcommand{\arraystretch}{0.92}
\setlength{\tabcolsep}{3.8pt}
\begin{tabular}{@{}l cc @{\hspace{10pt}} ccccccc@{}}
\toprule
 & \multicolumn{2}{c}{Binary panels} & \multicolumn{7}{c}{Ordinal panels} \\
\cmidrule(lr){2-3}\cmidrule(l){4-10}
 & RiceChem & HealthBench & ELLIPSE & \makecell{FED-\\Turn} & \makecell{FED-\\Dialogue} & HelpSteer2 & LFQA & \makecell{USR-\\TC} & \makecell{USR-\\PC} \\
\midrule
\multicolumn{10}{@{}l}{\textit{(a) Accuracy (\%): verdict accuracy (binary) or exact accuracy (ordinal)}} \\
Jev Choice & \textbf{81.0} & 77.1 & 13.4 & 56.2 & 48.6 & 48.6 & 68.7 & 62.0 & 56.4 \\
Decider Choice & \underline{80.3} & 75.4 & 47.1 & 50.1 & 41.1 & \underline{63.1} & 69.5 & 60.6 & 57.8 \\
Luna & 77.8\mk{\dag*} & 70.4\dg & 14.1\mk{**} & 57.7\mk{**} & 46.7\mk{*} & 39.2\mk{\ddag**} & \textbf{71.6} & 62.5 & \underline{\textbf{60.6}}\dg \\
Gemini & 76.1\mk{\ddag**} & \underline{\textbf{79.6}} & \textbf{29.8}\mk{\ddag**} & \underline{60.6}\mk{**} & \underline{\textbf{56.9}}\mk{\ddag**} & \textbf{54.6}\mk{\dag**} & 67.4 & 65.6 & \underline{\textbf{60.6}} \\
DeepSeek & 79.2 & 76.4 & 15.0\mk{**} & 59.5\mk{**} & 46.9 & 46.8\mk{**} & 65.0 & 58.8 & 57.8 \\
Jev Score & -- & -- & 15.3\scm & \textbf{66.7}\scm & 51.6\scm & 44.4 & 69.7 & \textbf{66.2}\scm & 55.0 \\
Decider Score & -- & -- & \underline{51.0}\scm & 48.0 & 37.8 & 55.3\scm & \underline{71.7} & \underline{69.4}\scm & 48.9\scm \\
\addlinespace[2pt]
Rater reference & -- & -- & 51.4 & 63.8 & 64.9 & -- & 55.4 & 61.7 & 72.6 \\
No-read baseline & -- & -- & -- & -- & -- & 60.0 & 68.1 & -- & -- \\
\midrule
\multicolumn{10}{@{}l}{\textit{(b) Agreement on ordinal pairs: mean QWK}} \\
Judge--judge (Jev set) & -- & -- & 0.38 & 0.57 & 0.61 & 0.66 & 0.70 & 0.65 & 0.61 \\
Judge--label (Jev set) & -- & -- & 0.18 & 0.39 & 0.35 & 0.31 & 0.46 & 0.53 & 0.50 \\
Judge--judge (Decider set) & -- & -- & 0.34 & 0.55 & 0.55 & 0.55 & 0.58 & 0.64 & 0.63 \\
Judge--label (Decider set) & -- & -- & 0.25 & 0.38 & 0.32 & 0.35 & 0.45 & 0.52 & 0.51 \\
Rater reference & -- & -- & 0.50 & 0.36 & 0.29 & -- & 0.39 & 0.46 & 0.52 \\
\bottomrule
\end{tabular}
\caption{Main results (per-criterion LLM judges). Judge--judge and judge--label: mean pairwise QWK among a decision model's four matched judges (its set) and their mean QWK with the labels, on the ordinal pairs all four scored. Bold: best of Jev's five judges in its column; underlined: best of the Decider's five. \dag\ marks separation from Jev Choice and * from Decider Choice (embedded binary pairs included), and \ddag\ and ** add correction for that decision model's 27 comparisons; \S: a Score framing separated from its model's Choice. Each judge's $\kappa$ and Jev Noul: Table~\ref{tab:binary}; each judge's mean QWK: Table~\ref{tab:secondary}; offsets: Table~\ref{tab:offsets}; costs: Table~\ref{tab:cost}; more on the Decider: Appendix~\ref{app:decider}.}
\label{tab:main}
\end{table*}

\emph{Exact accuracy} is the share of ordinal pairs predicted at the label's level, and \emph{mean QWK} averages per-criterion quadratic-weighted kappas~\citep{cohen1968weighted}. A judge's \emph{offset} is its mean predicted minus labeled level, and its \emph{gap} its exact accuracy minus the rater reference's. Binary criteria get \emph{verdict accuracy}, Cohen's $\kappa$~\citep{cohen1960coefficient}, and the \emph{MET-rate offset} in points. As chance-level exact accuracy varies with the number of levels, panels are compared by mean QWK, offset, or gap.

Paired comparisons use the pairs both judges scored, resampling units 10,000 times for percentile intervals and Holm-corrected bootstrap $p$-values (Appendix~\ref{app:tests}). Two judges are \emph{separated}, or differ significantly, when the 95\% interval of their accuracy difference excludes zero, at \emph{parity} when the 90\% interval lies within 5 points of zero (a post-hoc margin; Table~\ref{tab:clustered} gives the smallest each comparison supports), and otherwise \emph{inconclusive}. In each condition, a decision model's 27 comparisons with the LLM judges form their own Holm family.

Abstentions and \emph{judge-call failures}, pairs left without a usable verdict, leave the denominators, so pair sets differ between tables (Appendix~\ref{app:tests}). Cost is list price, the Decider's as documented the day before its run, and DeepSeek's its OpenRouter bill (Appendix~\ref{app:cost}). Level shifts, cascades, juries, and other counterfactuals are \emph{post-hoc analyses} of the recorded verdicts, with no judge rerun.

\section{Decision Models Can Often Replace an LLM Judge}
\label{sec:parity}

\subsection{Accuracy in paired comparisons}
\label{sec:accuracy}

Of Jev Choice's 27 comparisons with the LLM judges, 8 are separated, 8 at parity, and 11 inconclusive (Figure~\ref{fig:parity}a; Table~\ref{tab:conditions}). Three of its four leads fall on binary panels and all four deficits on ordinal ones. Four separations remain significant after correction, and resampling the prompts, questions, or contexts that units share leaves eight (Appendix~\ref{app:tests}).

On no ordinal panel is Jev Choice the most accurate matched judge; Gemini is, or ties, everywhere but LFQA (Table~\ref{tab:main}). Jev Score is separated ahead of Jev Choice on four ordinal panels, never behind, and is the most accurate of Jev's five judges on FED-Turn and USR-TC (Appendix~\ref{app:choicescore}). HelpSteer2's constant predictor beats each of Jev's five judges, and on LFQA no judge, the Decider's framings included, differs significantly from the provenance baseline in exact accuracy (Appendix~\ref{app:conventions}). Decider Choice leads the LLM judges in 8 comparisons, mostly on ordinal panels, and trails them in 5 (Figure~\ref{fig:parity}b; Table~\ref{tab:main}; Appendix~\ref{app:decider}).

\subsection{Decision models are cheaper and faster than every LLM judge}
\label{sec:price}

Over the nine panels Jev Choice cost \$0.063 (Table~\ref{tab:cost}), Luna 29 times as much, Gemini 325 times, and DeepSeek 66 times; Jev Choice took 29 seconds, Luna 859, Gemini 950, and DeepSeek 6,298 (30 to 220 times as long). Each Jev framing sent 8 requests at a time and each LLM judge 16 to 32 on the ordinal panels, so these ratios understate Jev's advantage at equal concurrency. Table~\ref{tab:conditions} gives the multiples against both decision models in all four conditions, including the whole-rubric runs, which matched Jev's concurrency. Perplexity's rate limit, not the 8 requests in flight, sets most of Decider Choice's wall time (Appendix~\ref{app:decider}).

\subsection{Judging mode, reasoning effort, and model tier}
\label{sec:whole-rubric}
\label{sec:effort}
\label{sec:highertier}

\begin{table}[!t]
\centering
\scriptsize
\renewcommand{\arraystretch}{0.88}
\setlength{\tabcolsep}{2pt}
\begin{tabular}{@{}l r r r r@{}}
\toprule
 & \makecell[r]{Per-\\criterion} & \makecell[r]{Whole-\\rubric} & \makecell[r]{High\\effort} & \makecell[r]{Higher\\tier} \\
\midrule
\multicolumn{5}{@{}l}{\emph{RQ1}} \\
Cost ($\times$ Choice) & 29/325/66 & 16/182/54 & 36/675/93 & 233/984/67 \\
\quad Decider & 32/357/73 & 18/200/59 & 40/742/102 & 255/1080/74 \\
Wall time ($\times$ Choice) & 30/33/220 & 28/45/350 & 43/117/-- & 35/162/-- \\
\quad Decider & 6/6/42 & 5/9/67 & 8/23/-- & 7/31/-- \\
Sep./parity/inconcl. & 8/8/11 & 7/9/11 & 5/11/11 & 6/9/12 \\
\quad Decider & 13/5/9 & 14/5/8 & 10/5/12 & 13/4/10 \\
Leads (binary)/deficits & 4 (3)/4 & 4 (3)/3 & 3 (2)/2 & 2 (2)/4 \\
\quad Decider & 8 (2)/5 & 8 (2)/6 & 7 (1)/3 & 7 (1)/6 \\
\multicolumn{5}{@{}l}{\emph{RQ2}} \\
Judge--judge $>$ label & 7 & 7 & 7 & 7 \\
\quad Decider & 7 & 7 & 7 & 7 \\
Offsets all negative & 6 & 6 & 6 & 6 \\
\quad Decider & 4 & 4 & 4 & 4 \\
Confident repeats & 96.0 (50.3) & 96.8 (50.7) & 97.2 (51.2) & 94.8 (52.0) \\
\quad Decider & 86.5 (47.9) & 88.1 (48.9) & 86.9 (47.3) & 79.8 (47.8) \\
\multicolumn{5}{@{}l}{\emph{RQ3}} \\
Gain: cross-fitted/oracle & 1.5/2.0 & 2.5/2.6 & 2.4/2.7 & 3.7/3.7 \\
\quad Decider & 1.2/1.7 & 0.8/1.7 & 0.8/1.1 & 1.4/1.8 \\
Either tier & 1.7/2.0 & -- & 1.5/2.0 & 1.3/1.7 \\
\quad Decider & 1.3/1.7 & -- & 0.8/1.1 & 0.4/1.4 \\
Cost under half & 7 & 4 & 6 & 7 \\
\quad Decider & 6 & 4 & 6 & 6 \\
\bottomrule
\end{tabular}
\caption{Answers by LLM-judge condition, the decision models' verdicts fixed. Each row gives Jev's copy and the indented row below it the Decider's. Triples: the OpenAI/Google/DeepSeek judges, multiples of the decision model's Choice cost (the DeepSeek judges' as billed) and wall time (concurrency: Section~\ref{sec:price}); DeepSeek's high and DeepSeek V4 Pro's wall times are unmeasured. A comparison both separated and at parity counts once, as separated; one of the Decider's per-criterion and one of its higher-tier separations are such. Leads on binary panels in parentheses. RQ2 (Section~\ref{sec:shared}): ordinal panels with judge--judge above judge--label QWK and with every matched offset negative; LLM verdicts (\%) repeating the decision model's confident errors (baseline). RQ3 (Section~\ref{sec:routing}): largest gain over the condition's best single judge or over that of either tier (picked from the first stage's framings and the LLM judges of both tiers; not computed for whole-rubric runs), in points; panels where a cross-fitted cascade matching the best LLM judge costs under half of it.}
\label{tab:conditions}
\end{table}

Neither judging mode, reasoning effort, nor model tier changes the answers to RQ1 to RQ3. Only the numbers behind them differ between the columns of Table~\ref{tab:conditions}. The runs of each condition, its changes in the LLM judges' accuracy, and Jev's copy of its comparisons are in Appendices~\ref{app:wholerubric} (whole-rubric judgments), \ref{app:effort} (high effort), and~\ref{app:highertier} (higher tier), with the Decider's copy in Appendix~\ref{app:decider}.

\section{Decision Models and LLM Judges Err Alike}
\label{sec:shared}

\subsection{A shared shortfall on ordinal criteria}
\label{sec:agreement}

On all seven ordinal panels, in either decision model's copy, agreement among the matched judges exceeds their agreement with the labels, and even the least-agreeing pair of judges beats the best judge--label agreement on five panels in Jev's copy (Tables~\ref{tab:main}b and~\ref{tab:wr-downstream}b). Label noise or a common offset could produce this pattern. Every judge of either copy falls short of the rater reference on ELLIPSE, FED-Dialogue, and USR-PC but exceeds it on LFQA, and in both copies the matched judges' mean QWK with the labels trails the reference's only on ELLIPSE and USR-PC.

In Jev's copy every matched judge's offset is negative on the six ordinal panels other than USR-TC, as are 31 of the five judges' 35 offsets (Tables~\ref{tab:secondary}h and~\ref{tab:offsets}). ELLIPSE's are largest. Its labels put 36.3\% of pairs at levels 4--5, and none of Jev's five judges more than 4.2\% (Figure~\ref{fig:location}a). Decider Choice places units lower than the raters on FED-Turn, FED-Dialogue, and USR-PC and higher on HelpSteer2, LFQA, and USR-TC. On ELLIPSE it puts 10.7\% of pairs at levels 4--5, and the sign of its offset there depends on the rounding rule of Section~\ref{sec:labels}. The offset is $-0.46$ levels against our half-up labels, $-0.23$ against the unrounded rater mean, and $+0.004$ against labels rounded down.

\label{sec:location}Criteria with larger negative offsets also fall further below the rater reference, a link that disappears once each criterion's predictions get a leave-one-out shift (Appendix~\ref{app:offset}).

\label{sec:contrasts}\label{sec:conventions}\label{sec:alternatives}Rating conventions that guided the raters but are absent from our criterion texts may underlie the shortfall; we inferred candidate conventions from the judges' errors, so this account is exploratory (Appendix~\ref{app:conventions}). The account fits Jev and the three LLM judges but not the Decider, whose Score framing reaches 51.0\% exact accuracy on ELLIPSE against the rater reference's 51.4\%.

\subsection{The LLM judges repeat the decision models' confident errors}
\label{sec:shared-errors}

We count an LLM verdict that repeats a decision model's wrong answer as a \emph{repeated error}, and the share of repeated errors on its wrong pairs measures correlated errors. Since judges sharing an offset would often agree anyway, the \emph{independence baseline} is each LLM judge's rate of giving that wrong answer on other pairs with the same criterion and label. Over all of Jev Choice's errors the share beats the baseline on every ordinal panel, by 18.8 to 34.2 points on six and 8.6 on ELLIPSE, where the judges' negative offsets explain most repeated errors (Table~\ref{tab:confidence}c). The \emph{confident-error sample} holds a Choice framing's 12 most confident errors per ordinal panel. For Jev, 242 of the 252 LLM verdicts on these 84 pairs (96.0\%) repeat its answer, against 50.3\% expected, and about as many in the other conditions (Table~\ref{tab:conditions}). HealthBench shows the same (Appendix~\ref{app:shared}). Rater disagreement may explain repeats on split pairs, which hold 62 of Jev's 72 confident errors on multi-rater panels, but not on unanimous pairs, such as a USR-TC pair whose context holds no fact (Appendix~\ref{app:conventions}).

The LLM judges repeat the Decider's answer in 218 of their 252 verdicts on its confident-error sample (86.5\%), against 47.9\% expected, far above baseline on six panels (Appendix~\ref{app:decider}). On ELLIPSE the share stays at or below baseline in every condition. In the main runs the LLM judges are wrong on all 36 verdicts there, missing the same pairs in the same direction, but 20 of the 36 lie further below the label than the Decider's answer. On HealthBench 44.4\% of the LLM verdicts on Decider Noul's confident errors repeat its answer, fewer than half but more than the 24.8\% expected.

\section{Correlated Errors Undo the Cascade Advantage}
\label{sec:routing}

A \emph{cascade} lets a cheap judge answer every pair and sends the \emph{deferred pairs}, whose confidence falls below a threshold, to a \emph{fallback judge}; the cheap judge's verdicts stand on the \emph{kept pairs}. An \emph{oracle threshold} is chosen on the pairs it is evaluated on, so its gains are optimistic; a \emph{cross-fitted threshold} is chosen on a random half of the units and evaluated on the other. A cheap judge whose confidence tracks its errors is the natural first stage, but the cascade rises above its fallback only through kept pairs that the cheap judge gets right and the fallback misses, and for Jev these are few, at most 14 per ordinal panel in that panel's best cascade (Section~\ref{sec:shared-errors}; Table~\ref{tab:cascade-split}).

\subsection{The decision models' confidence ranks their errors}
\label{sec:confidence}

Jev Choice's errors gather among its less confident verdicts on every ordinal panel but ELLIPSE (Figure~\ref{fig:confidence}a); on those six, the area under the receiver operating characteristic curve (AUROC) of its confidence for correctness is 0.57 to 0.70. On ELLIPSE it is 0.49, its error rate flat across bands (Appendix~\ref{app:confidence}). On the binary panels, the only ones where we measure calibration, Jev's P(MET) separates MET from UNMET at an AUROC of 0.80 to 0.86 with an expected calibration error (ECE) of 0.056 to 0.096 (Table~\ref{tab:binary}). Yet repeated errors, and their excess over the baseline, peak on every ordinal panel in Jev Choice's top confidence band (80.5\% to 92.8\% against 33.6\% to 63.7\%; Figure~\ref{fig:confidence}b); these are the verdicts a cascade keeps. For Decider Choice the AUROC is 0.60 to 0.78 on the same six ordinal panels and 0.52 on ELLIPSE, where its error rate is flat too. The Decider's P(MET) has an AUROC of 0.78 to 0.89 on the binary panels, and Decider Noul is calibrated on RiceChem (ECE 0.084) but overconfident at both ends on HealthBench (ECE 0.148).

\subsection{Cascades lower cost but add little accuracy}
\label{sec:cascades}

\begin{table}[t]
\centering
\scriptsize
\renewcommand{\arraystretch}{0.88}
\setlength{\tabcolsep}{2pt}
\begin{tabular}{@{}l rrrr @{\hspace{7pt}} rrrr@{}}
\toprule
 & \multicolumn{4}{c}{Jev first} & \multicolumn{4}{c}{Decider first} \\
\cmidrule(lr){2-5}\cmidrule(l){6-9}
Panel & Best & Oracle & \makecell[r]{Cross-\\fitted} & \makecell[r]{Cost\\(\%)} & Best & Oracle & \makecell[r]{Cross-\\fitted} & \makecell[r]{Cost\\(\%)} \\
\midrule
RiceChem & 81.0 & $-0.4$ & $-1.0$ & 6.6 & 80.3 & $+0.7$ & $+0.4$ & 5.3 \\
HealthBench & 79.6 & $+2.0$ & $+1.5$ & 15.6 & 79.6 & $0.0$ & $+0.3$ & 97.3 \\
\midrule
ELLIPSE & 29.8 & $0.0$ & $-0.1$ & 99.8 & 51.0 & $-3.7$ & $-3.8$ & 0.3 \\
FED-Turn & 70.7 & $-5.2$ & $-7.3$ & 45.5 & 65.2 & $+0.2$ & $-0.3$ & 66.3 \\
FED-Dialogue & 59.7 & $0.0$ & $-0.9$ & 77.9 & 59.9 & $0.0$ & $-2.1$ & 81.8 \\
HelpSteer2 & 54.3 & $+0.6$ & $-0.9$ & 35.6 & 63.1 & $+0.9$ & $-0.3$ & 0.4 \\
LFQA & 71.5 & $+0.6$ & $+0.6$ & 39.0 & 72.0 & $+1.7$ & $+1.2$ & 17.9 \\
USR-TC & 70.6 & $+0.3$ & $-0.7$ & 47.5 & 75.0 & $-3.9$ & $-6.1$ & 28.1 \\
USR-PC & 74.6 & $+1.3$ & $+0.8$ & 31.3 & 74.3 & $+1.0$ & $+0.7$ & 27.4 \\
\bottomrule
\end{tabular}
\caption{Post-hoc cascades with each decision model first. Best: accuracy (\%) of the best single judge. For Jev-first cascades it is Jev Choice on RiceChem, Jev Score on FED-Turn, Luna on LFQA, and Gemini elsewhere; for Decider-first ones it is Decider Choice on RiceChem and HelpSteer2, Decider bare Noul (Noul without the wrapper, Table~\ref{tab:judge-inputs}; tied with Gemini) on HealthBench, Decider Score on ELLIPSE, LFQA, and USR-TC, and Gemini elsewhere. Oracle, cross-fitted: cascade minus best (points). Cost: share of the best LLM judge's cost, the decision model's run included, for the cross-fitted cascade that matches that judge, DeepSeek on RiceChem, Luna on LFQA, and Gemini elsewhere. Halvings and held-out accuracy: Table~\ref{tab:cascade-halvings} and Appendix~\ref{app:decider}; pair set: Appendix~\ref{app:tests}; more: Appendix~\ref{app:cascades}.}
\label{tab:cascade}
\end{table}

We replay cascades with each decision model first, its Choice framing on the ordinal panels and its Noul framing on the binary ones (Appendix~\ref{app:cascades}), against the \emph{best single judge}, chosen from the LLM judges and that decision model's framings. Cross-fitted thresholds use 50 halvings, each half choosing its own fallback and best single judge.

With cross-fitted thresholds no Jev-first cascade's average gain over the best single judge exceeds 1.5 points (HealthBench; Table~\ref{tab:cascade}); these cascades trail it on six of nine panels, by 0.1 to 7.3 points. Oracle thresholds lift the largest gain to 2.0 points (HealthBench), but neither it nor the next, 1.3 (USR-PC), is significant in a kept-pair sign test (Table~\ref{tab:cascade-split}).

Decider-first gains stay small for a different reason, since one of the Decider's framings is itself the best single judge on six of nine panels (Appendix~\ref{app:cascades}). On the two ordinal panels where Decider Choice is more accurate than all three LLM judges, its cascades lead their fallbacks by 17.4 points on ELLIPSE and 8.9 on HelpSteer2, through kept pairs that it gets right and the fallback misses. Gemini, the fallback on ELLIPSE, is right on 9.4\% of the pairs there that Decider Choice gets wrong and on 52.8\% of those it gets right.

Whichever decision model comes first, the cross-fitted cascade that matches a panel's best LLM judge saves over half of that judge's cost on most panels (Table~\ref{tab:cascade}), but on only four of nine in the whole-rubric condition (Table~\ref{tab:conditions}). With flash-tier fallbacks, no condition's Jev-first cascade gains more than 2.7 points (oracle) or 2.5 (cross-fitted) over its panel's best single judge (Table~\ref{tab:conditions}; Appendices~\ref{app:wholerubric} and~\ref{app:effort}). Higher-tier fallbacks lift a Jev-first HelpSteer2 cascade 3.7 points, but only above its condition's weak best single judge, and lift no cascade more than 1.7 points above either tier's best single judge (Appendix~\ref{app:highertier}).

\phantomsection\label{sec:juries}Pooling judges' verdicts in a \emph{jury} fares no better. In no condition, and with neither decision model, does the 95\% interval of a median jury's difference from the most accurate matched judge lie above zero (Appendix~\ref{app:shared}).

\section{Discussion}
\label{sec:discussion}

Prior work finds LLMs' errors correlated across providers~\citep{kim2025correlated,goel2025great}; we find them correlated across two kinds of judge, and they stayed so with each vendor's higher tier. The Decider, from a second vendor on another backbone, showed the same excess of judge--judge over judge--label agreement, correlated errors on six of seven ordinal panels, and small cascade gains, so these patterns are not peculiar to Jev. Two decision models are still too few to show what holds for the class, and the Decider's ELLIPSE result remains unexplained. TypeSafe warns that Jev may read scope and negation literally,\footnote{\url{https://docs.typesafe.ai/model-jaggedness/jev-1.13}.} yet LLM judges told to heed both (Appendix~\ref{app:protocol}) also placed units below the raters and repeated nearly all its confident errors, even at high effort. Literal reading by every judge, unstated conventions, shared training priors, a reluctance to grant top levels (Appendix~\ref{app:conventions}), and, on split pairs, rater disagreement (Section~\ref{sec:shared-errors}) could each contribute; our data cannot tell them apart. Practical recommendations that follow from these findings are in Appendix~\ref{app:recommendations}.

\section{Conclusion}
\label{sec:conclusion}

We compared two decision models, Jev and Perplexity's Decider, with flash-tier and higher-tier LLM judges on nine panels and replayed cascades that put each decision model first. A cascade can beat its LLM fallback only where its first stage alone is right. It therefore needs a first stage that knows when it is likely to be wrong and a fallback that is wrong somewhere else. The confidence of both decision models meets the first need on most panels. For Jev the second went unmet. Despite Jev's different design, the judges were wrong in the same places, and the LLM judges repeated almost all of Jev's most confident errors, about twice as often as independent errors would. Even oracle thresholds lifted no Jev-first cascade more than 2.7 points above the best single judge with flash-tier fallbacks, or 1.7 above either tier's best single judge with higher-tier ones. The LLM judges also repeated most of the Decider's confident errors outside ELLIPSE, and no Decider-first cascade with flash-tier fallbacks gained more than 1.7 points, largely because one of the Decider's own framings was already the best single judge on most panels. Jev is often a cheap substitute for a flash-tier LLM judge but a poor complement to one. A judge built differently need not be wrong differently, so the next decision model should be designed to get right what LLM judges get wrong, and vice versa.

\section*{Limitations}
\phantomsection
\makeatletter
\def\@currentlabel{Limitations}
\makeatother
\label{sec:limitations}

\paragraph{Observational account.} We formed our expectations about Jev on the binary panels and tested them on the ordinal panels, and our account of the judges' shortfall came afterwards and is exploratory. No experiment supplied exemplars or rewrote level descriptions. The account could be tested by rerunning ELLIPSE with exemplar essays for each level and with a grammar scale that names observable features and drops the native-like label for level 5 (Appendix~\ref{app:conventions}), and by rerunning every ordinal panel with its benchmark's own instructions to raters in place of our rubric version. Giving the same texts to untrained human readers, or giving the judges exemplars, would also distinguish missing conventions from priors that the judges share through their training data. The candidate conventions came from inspecting the judges' errors (Appendix~\ref{app:conventions}), and the cross-panel aggregations of Sections~\ref{sec:shared} and~\ref{sec:routing} were computed after the offset had been observed. We added the Decider after the Jev analyses, ran its copy through the same code, and recorded the claims to be checked on it, with their thresholds, about two and a half hours before the Decider answered any request on a study item (Appendix~\ref{app:decider}). The Decider is a fine-tune of Qwen3.8-27B with Apache-2.0 open weights, and 530,103 of its 626,033 training rows come from the tasksource collection, HelpSteer2's attribute labels among them. Its released training code labels synthetic rows with OpenAI's \texttt{gpt-5.6-sol}, a sibling tier of GPT-5.6 Luna, and its data manifest does not say how many rows are synthetic. Shared training signal is one route to correlated errors. Jev's training data are not disclosed, nor are those of the LLM judges, so shared priors cannot be ruled out.

\paragraph{Judge configuration.} The main runs' LLM judges are flash-tier models with the harness's default prompt, used zero-shot, and Luna runs at its enforced temperature (Section~\ref{sec:llm}). This matches the question of whether a decision model can replace an LLM judge in its default configuration, but it does not measure the best obtainable LLM judge. No few-shot prompt, wrapper for Jev on ordinal criteria, or LLM run without the prompt's instructions on middle levels and self-contradiction was tried, and Jev Choice was not run without the NA option. The Decider received Jev's framings and wrapper unchanged, with none tuned for it, and its documentation describes no retrieval. Each vendor's higher tier ran only with per-criterion judgments at vendor-default reasoning, and no flagship tier such as GPT-5.6 Sol, no model from a vendor other than OpenAI, Google, and DeepSeek, and no higher tier judging whole rubrics served as a fallback. For Google and DeepSeek the higher tier is an older model than the flash-tier judge it replaced, and Gemini 3.1 Pro's documented default thinking level differs from Gemini's, but it emitted about as many reasoning tokens as Gemini's main run (Appendix~\ref{app:highertier}). The judges also receive different instructions and option orders (Sections~\ref{sec:jev} and~\ref{sec:llm}). We measured the LLM judges in both judging modes, but the decision models only as designed: each receives a unit's criteria together in a single request, and we did not test whether the questions of one request influence one another's answers. Reasoning effort was requested only at medium (on all nine panels for DeepSeek, on the repeat units alone for Luna and Gemini) and at high, and only with per-criterion judgments; no LLM judge ran without reasoning or above high. Running the LLM judges without reasoning would narrow the cost ratios in either judging mode, with an unknown effect on accuracy.

\paragraph{Runs.} The main runs judged each ordinal panel once. Jev and the LLM judges are not deterministic. The Decider gave identical verdicts and probabilities on every repeated pair in two runs minutes apart, but its stability over days is untested and its vendor documents occasional differences. Across five passes over RiceChem, about 3\% of Jev's pairs flipped verdict at least once, and in a spot check of 82 pairs from the ordinal panels two Luna runs gave the same verdict on 71 (Appendix~\ref{app:protocol}). The repeat runs gave every LLM judge a second whole-rubric pass over 106 units (105 for DeepSeek), per-criterion judgments were repeated only in the effort calibration on the same units (Appendix~\ref{app:effort}) and in that spot check, and Jev was not rerun on an ordinal panel. The Decider ran on 7 October against the LLM verdicts recorded on 19 and 27 September and 2 October, with no LLM judge rerun. It has a repeat on the ordinal panels, which Jev lacks, and was not part of Jev's five-pass study. Because the whole-rubric and effort runs followed the main runs by eight days under the same undated model names, a model update or a change in routing could contribute to their differences from the main runs. DeepSeek's medium and high runs overlapped in time, so we do not compare its wall times across effort levels, and because provider prompt caching and OpenRouter's routing moved some costs, effort levels are compared on tokens and at fixed prices as well as on billed costs (Appendix~\ref{app:effort}). OpenRouter routed DeepSeek's whole-rubric run to a different mix of upstream providers than its main run and billed those requests at 1.70 times their cost at DeepSeek's list prices, against 1.34 times for the per-criterion ones (Appendix~\ref{app:wholerubric}), so routing, not only the judging mode, may lie behind DeepSeek's smaller saving. We do not compare wall times between judging modes, since on the ordinal panels the per-criterion runs also sent more requests at a time than the whole-rubric runs. The higher-tier runs came 13 days after the main runs and allowed each request 600 seconds instead of 60. OpenRouter sent most of DeepSeek V4 Pro's requests to one upstream provider; two of its requests hung, and its resumed run's wall time was not measured (Appendix~\ref{app:highertier}).

\paragraph{Samples.} Results for single criteria rest on 25 to 258 units, because the ordinal panels are 10\% or 20\% samples. Two results rest on very little data, the 8 \crit{error\_recovery} pairs that Jev Choice answered and the three USR-TC contexts that hold no fact, behind the \texttt{\_nofact} result (Appendix~\ref{app:conventions}). The confident-error sample takes only 12 errors of each decision model from each ordinal panel, whereas the band analysis of Figure~\ref{fig:confidence}b uses every error.

\paragraph{Statistics.} The main intervals resample units, not the larger sampling groups, and Section~\ref{sec:accuracy} and Appendix~\ref{app:tests} report what changes when the groups are resampled instead. The 5-point equivalence margin is post hoc (Section~\ref{sec:metrics}), and because it is absolute it is lenient where accuracy is low, as on ELLIPSE. The criteria singled out in the confidence analysis and the cut-offs of Appendix~\ref{app:offset} were chosen after inspecting the data. The offset--gap analysis of Section~\ref{sec:location} is leave-one-out, but the one-level ELLIPSE shift of Appendix~\ref{app:offset} and the USR shift of Appendix~\ref{app:conventions} were chosen on the pairs they are scored on. Under oracle thresholds cascade gains are optimistic, and the cross-fitted estimates use halves as small as 12 units. Holm corrections run within each decision model's comparisons. Cascade costs are estimates under the rules of Appendix~\ref{app:cascades}, which do not charge for what a live cascade would also defer because its first stage abstained. Only deferral on the confidence of a decision model's Choice or Noul framing was tried, and the LLM judges' token probabilities or verbalized confidence were not elicited.

\paragraph{Labels and text.} Nearly half of the ELLIPSE labels depend on the rounding rule (Section~\ref{sec:labels}). Our rubric versions depart from the benchmarks' own questions and rubrics (Appendix~\ref{app:protocol}), and we compared them with what the raters read for ELLIPSE alone. On the other six ordinal panels a convention that our texts omit may be stated in the raters' own instructions. The HelpSteer2 and LFQA labels partly reflect something other than the named criterion. HelpSteer2 \crit{correctness} tracks overall helpfulness at Pearson $r = 0.94$, and provenance alone predicts LFQA labels about as well as the judges do on exact accuracy. A judge may also have seen the items or labels of these public benchmarks in training.

\paragraph{Scope.} We tested only two decision models, from two vendors, with Jev as the reference, so nothing here is shown to hold for decision models in general. Their Decision Index scores come from classification benchmarks and may not carry over to rubric judging. Appendices~\ref{app:choicescore} to~\ref{app:confidence} study Jev's copy in depth, and several of their analyses were run for Jev alone (Appendix~\ref{app:decider}). Later requests under the same model names may return different verdicts, because no request except DeepSeek V4 Pro's and the Decider's named a fixed model version, and Gemini 3.1 Pro is a Preview (Section~\ref{sec:llm} and Appendices~\ref{app:protocol} and~\ref{app:highertier}). The ordinal panels are all English, four of the seven are dialogue, and the longest ordinal unit is 1,114 words. Prices are list or billed prices on the dates of the runs, with the Decider's read from its documentation a day earlier, and the LLM judges' cost depends on reasoning defaults that providers can change.

\section*{Ethics Statement}

This study uses public benchmarks as their authors released them and collects no new human-subject data. HealthBench's health-related conversations serve only as evaluation material. We report every judge's cost from recorded token usage and list or billed prices (Appendix~\ref{app:cost}). The authors are not incentivized by TypeSafe, which develops Jev, by Perplexity, which develops the Decider, or by their industry competitors. \ifarrreview
The anonymized supplementary material holds the verdict records of every judge, which we will release upon publication, so that the paper's numbers can be recomputed without any request to a model API. Appendix~\ref{app:release} lists the records and the values that need other inputs.
\else
We release the verdict records of every judge so that the paper's numbers can be recomputed without any request to a model API, and Appendix~\ref{app:release} lists the records and the values that need other inputs.
\fi

\section*{Generative AI Use Disclosure}
\label{sec:generative-ai-use-disclosure}
Generative AI tools assisted at different stages of preparing this paper. The corresponding author has checked their outputs and takes complete responsibility for any resulting inaccuracies.

\ifarrreview\else
\section*{Acknowledgments}
This research was developed with funding from the Defense Advanced Research Projects Agency's (DARPA) SciFy program (Agreement No. HR00112520300) and is based upon work supported in part by the Office of the Director of National Intelligence (ODNI), Intelligence Advanced Research Projects Activity (IARPA), via 56000026C0019 (the BENGAL program). The views and conclusions contained herein are those of the authors and should not be interpreted as necessarily representing the official policies, either expressed or implied, of DARPA, ODNI, IARPA, the Department of Defense, or the U.S. Government. The U.S. Government is authorized to reproduce and distribute reprints for governmental purposes notwithstanding any copyright annotation therein.
\fi

\bibliography{references}

\appendix
\counterwithin{table}{section}
\counterwithin{figure}{section}
\renewcommand{\thetable}{\thesection\arabic{table}}
\renewcommand{\thefigure}{\thesection\arabic{figure}}
\counterwithin{equation}{section}
\renewcommand{\theequation}{\thesection\arabic{equation}}
\renewcommand{\floatpagefraction}{0.8}
\renewcommand{\dblfloatpagefraction}{0.8}
\raggedbottom

\input{appendix/a_protocol}
\input{appendix/b_cost}
\input{appendix/c_tests}
\input{appendix/d_binary}
\input{appendix/e_choice_score}
\input{appendix/f_abstention}
\input{appendix/g_offset}
\input{appendix/h_conventions}
\input{appendix/i_confidence}
\input{appendix/j_shared}
\input{appendix/k_cascades}
\input{appendix/l_binary_obs}
\input{appendix/m_release}
\input{appendix/n_whole_rubric}
\input{appendix/o_effort}
\input{appendix/p_higher_tier}
\input{appendix/q_recommendations}
\input{appendix/r_decider}

\end{document}

%% file: appendix/a_protocol.tex
\section{Panels, Judges, and Protocol}
\label{app:protocol}

\begin{table*}[!tp]
\centering
\footnotesize
\setlength{\tabcolsep}{3pt}
\begin{tabular}{@{}l>{\raggedright\arraybackslash}p{2.85cm}>{\raggedright\arraybackslash}p{2.45cm}>{\raggedright\arraybackslash}p{2.55cm}rc>{\raggedright\arraybackslash}p{2.05cm}@{}}
\toprule
Panel & Population; sample & Strata & Raters per pair; label & Half-way & \makecell{Rater ref.\\exact / QWK} & Unit words: median (mean; range) \\
\midrule
RiceChem & 1{,}240 responses to 4 questions; 121 held out (10\% of each question, seed 42) & question: 32, 31, 29, 29 & released teaching-assistant label & -- & -- & -- \\
\addlinespace[2pt]
HealthBench & 200 completions from 196 conversations & theme (7) $\times$ verdict class: all MET 66, mixed 68, all UNMET 66 & 2--5 physicians (866 labels); majority & -- & -- & -- \\
\midrule
ELLIPSE & 2{,}571 test-set essays; 258 (10\%) & prompt $\times$ holistic band (106 strata) & 2 trained raters; mean & 725 & 51.4 / 0.50 & 392\newline(427.9; 46--1{,}114) \\
\addlinespace[2pt]
FED-Turn & 375 turns; 75 (20\%) & system: Meena 24, Mitsuku 26, human 25 & 5 crowd workers (4 on one pair); mean & 1 & 63.8 / 0.36 & 9\newline(12.5; 2--38) \\
\addlinespace[2pt]
FED-Dialogue & 125 dialogues; 25 (20\%) & system: Meena 8, Mitsuku 9, human 8 & 5 crowd workers (1--5 on \crit{error\_recovery}); mean & 2 & 64.9 / 0.29 & 123\newline(131.0; 42--284) \\
\addlinespace[2pt]
HelpSteer2 & 448 preference pairs (896 responses); 45 pairs, 90 responses (10\%) & preference strength: weak 52, strong 38 responses & released label: rounded mean of the 3 most-agreeing annotators & -- & -- & 215\newline(251.2; 1--905) \\
\addlinespace[2pt]
LFQA & 300 questions with 4 answers each; 30 questions, 120 answers (10\%) & none & 3 crowd workers; mean & 0 & 55.4 / 0.39 & 100\newline(104.9; 8--656) \\
\addlinespace[2pt]
USR-TC & 60 contexts with 6 responses each; 12 contexts, 72 responses (20\%) & none & 3 dialogue researchers; mean & 0 & 61.7 / 0.46 & 19.5\newline(22.9; 6--58) \\
\addlinespace[2pt]
USR-PC & 60 contexts with 5 responses each; 12 contexts, 60 responses (20\%) & none & 3 dialogue researchers; mean & 0 & 72.6 / 0.52 & 12\newline(12.9; 6--30) \\
\bottomrule
\end{tabular}
\caption{Construction of the nine panels. The sample gives sampled units and the sampled fraction. Half-way: labels whose rater mean is a half level. Rater reference (Section~\ref{sec:labels}): pooled exact accuracy (\%) and mean QWK. Unit words: whitespace words in the submission; our RiceChem records do not hold the submissions, and our HealthBench records give completion length in characters only. HealthBench's 406 pairs by theme: hedging 108, context seeking 66, health data tasks 60, communication 54, emergency referrals 50, complex responses 38, global health 30; its units by number of criteria: one 30, two 134, three 36. Criteria per panel, with the shortest and longest criterion sentence in words: RiceChem 27 (3--21), HealthBench 34 (7--481), ELLIPSE 6 (27--36), FED-Turn 7 plus 1 binary (4--13), FED-Dialogue 9 plus 1 binary (6--13), HelpSteer2 4 (15--32), LFQA 3 (14--33), and USR-TC and USR-PC 3 plus 2 binary each (11--43). The ordinal panels hold 3,778 of each matched judge's 5,003 pairs.}
\label{tab:panel-construction}
\end{table*}

\begin{table*}[!tp]
\centering
\footnotesize
\setlength{\tabcolsep}{3pt}
\begin{tabular}{@{}>{\raggedright\arraybackslash}p{2.45cm}>{\raggedright\arraybackslash}p{1.2cm}>{\raggedright\arraybackslash}p{3.8cm}>{\raggedright\arraybackslash}p{4.15cm}>{\raggedright\arraybackslash}p{2.95cm}@{}}
\toprule
Judge (criteria) & Primitive & Instruction or prompt & Options, in the order shown & Verdict \\
\midrule
bare Noul (binary panels) & Noul & \{requirement\} & none & MET if $P(\text{yes}) \geq 0.5$ \\
\addlinespace[2pt]
Jev Noul (binary panels; embedded binary criteria) & Noul & Determine whether this criterion is satisfied by the \texttt{\textasciigrave submission\textasciigrave}. Criterion: \{requirement\} & true: $D_{\text{MET}}$; false: $D_{\text{UNMET}}$ & MET if $P(\text{yes}) \geq 0.5$; cannot abstain \\
\addlinespace[2pt]
Jev Choice (binary panels) & Choice & the Jev Noul instruction & MET: $D_{\text{MET}}$; UNMET: $D_{\text{UNMET}}$; CANNOT\_ASSESS: $D_{\text{CA}}$ & most probable option \\
\addlinespace[2pt]
Jev Choice (ordinal criteria) & Choice & \{requirement\} & every option text as an option label, in scale order, the NA option last; description fields left empty & most probable option; NA is an abstention \\
\addlinespace[2pt]
Jev Score (ordinal criteria) & Score & \{requirement\} & the same option texts as level descriptions, in scale order; no NA option & expected level, rounded half to even; cannot abstain \\
\midrule
LLM judges (binary criteria) & -- & harness's default binary system prompt; user message with the criterion type, criterion, input, and submission & MET, UNMET, CANNOT\_ASSESS, defined in the system prompt with $D_{\text{MET}}$, $D_{\text{UNMET}}$, and $D_{\text{CA}}$ & JSON with the status and an explanation \\
\addlinespace[2pt]
LLM judges (ordinal criteria) & -- & harness's default multi-choice system prompt; user message with the criterion as a question, numbered options, input, and submission & option labels, the NA option included, shuffled for every pair & JSON with the option number and an explanation; NA is an abstention \\
\bottomrule
\end{tabular}
\caption{What each judge receives. \{requirement\} is the criterion text, verbatim. $D_{\text{MET}}$ = ``The thing described in the criterion IS present in the submission''; $D_{\text{UNMET}}$ = ``The thing described in the criterion IS NOT present in the submission''; $D_{\text{CA}}$ = ``Insufficient evidence to determine either way (use rarely)''. Jev's definitions are copied verbatim from the harness's binary system prompt; every criterion has positive weight, so the harness's definitions for negative criteria were never used. The wrapper of Section~\ref{sec:jev} is the Jev Noul instruction with $D_{\text{MET}}$ and $D_{\text{UNMET}}$. Bare Noul is reported in Appendices~\ref{app:binary} and~\ref{app:decider} and, where Decider bare Noul ties for the most accurate judge on HealthBench, in Table~\ref{tab:cascade} and Appendix~\ref{app:highertier}. The two LLM rows describe per-criterion judgments; for whole-rubric judgments the judge receives the same criterion texts and options together (Appendix~\ref{app:wholerubric}). The LLM judges' model identifiers, as their APIs report them, are \texttt{gpt-5.6-luna}, \texttt{gemini-3.8-flash}, and \texttt{deepseek/deepseek-v4.1-flash}, and those of the higher-tier judges (Appendix~\ref{app:highertier}) are \texttt{gpt-5.6-terra}, \texttt{gemini-3.1-pro-preview}, and \texttt{deepseek/deepseek-v4-pro-0813}. The Decider received the five rows above the rule unchanged, the first as \emph{Decider bare Noul}, and every reply reported the model that its requests named, \texttt{pplx-decider-v1.1-27b}.}
\label{tab:judge-inputs}
\end{table*}

\begin{table}[!tbp]
\centering
\footnotesize
\setlength{\tabcolsep}{2.2pt}
\begin{tabular}{@{}l l r l r@{}}
\toprule
Panel & Unit & Criteria & Scale & Pairs \\
\midrule
RiceChem & chemistry answer & 27 & binary & 819 \\
HealthBench & chatbot reply & 34 & binary & 406 \\
\midrule
ELLIPSE & learner essay & 6 & 5 levels & 1,548 \\
FED-Turn & dialogue turn & 7\,+\,1 & 3 levels & 600 \\
FED-Dialogue & dialogue & 9\,+\,1 & 3 levels & 250 \\
HelpSteer2 & assistant reply & 4 & 5 levels & 360 \\
LFQA & ELI5 answer & 3 & 3--4 levels & 360 \\
USR-TC & dialogue reply & 3\,+\,2 & 3 levels & 360 \\
USR-PC & dialogue reply & 3\,+\,2 & 3 levels & 300 \\
\bottomrule
\end{tabular}
\caption{The nine panels, binary above the rule; +$n$: binary criteria on an ordinal panel; USR-TC and USR-PC: USR Topical-Chat~\citep{gopalakrishnan2019topicalchat} and PersonaChat~\citep{zhang2018personalizing}. Sources: \citet{sonkar2024automated}, \citet{arora2025healthbench}, \citet{crossley2023english}, \citet{mehri2020unsupervised} (FED), \citet{wang2024helpsteer2}, \citet{kamoda2025quantifying}, and \citet{mehri2020usr}. Details: Table~\ref{tab:panel-construction}.}
\label{tab:panels}
\end{table}

\paragraph{Sampling.}
FED supplies two panels, FED-Turn and FED-Dialogue, and USR supplies USR-TC and USR-PC, whose conversations come from Topical-Chat~\citep{gopalakrishnan2019topicalchat} and PersonaChat~\citep{zhang2018personalizing}. LFQA's questions and human answers come from ELI5~\citep{fan2019eli5}. Each ordinal panel was drawn with seed 20260919 by proportional allocation over the strata of Table~\ref{tab:panel-construction}. Sampling groups are kept whole. The four answers to an LFQA question, the responses to one USR context, and the two responses of a HelpSteer2 preference pair enter the sample together. LFQA and the two USR panels have no strata and are simple random samples of questions or contexts. The RiceChem panel is the test part of an 80/10/10 split of each question's responses, the proportions that \citet{sonkar2024automated} used, and \citet{rao2026autorubric} evaluate AutoRubric on the same held-out set. Its criteria come from the RiceChem rubrics. HealthBench's 34 criteria are consensus criteria, which form the physician-validated subset of its criteria. They use 30 distinct texts because some criteria share a text across themes. HealthBench's completions are the validation and test sets, 100 completions each, of HealthBench-100, a sample drawn by a script distributed with AutoRubric~\citep{rao2026autorubric} and stratified by theme and by whether the labels of a completion's criteria are all MET, mixed, or all UNMET.

\paragraph{Leakage.}
The harness passes a judge only a unit's input and its submission, and a decision model's state contains only these two fields. To keep them from leaking, human labels sit in a label field and holistic scores, strata, and the identity of the system that produced a unit in a metadata field, and no judge receives either field. Nor does any judge see an option's \emph{value}, the number from 0 to 1 that the harness attaches to a level. No ordinal panel has a reference submission. HealthBench's physician-written ideal completion is removed from every judge's input, except in one ablation of Jev Noul and Decider Noul that lies outside the protocol (Appendix~\ref{app:binary}).

\paragraph{Criterion text.}
ELLIPSE calls its criteria traits, and HelpSteer2 calls them attributes. On the ordinal panels every judge receives the criterion and option strings of the rubric versions of Section~\ref{sec:panels}. These versions recast the FED and USR questions as statements, so a FED question reads ``To the average person, the response is interesting.'' HelpSteer2's level wording is paraphrased, and LFQA's level descriptions come from the rating prompt that the benchmark's authors wrote for GPT-4. On ELLIPSE the versions keep the level descriptions of the corpus's scoring rubric, typographic artifacts such as ``capitalizatio n'' and ``inappropri ately'' included, and give each trait a criterion sentence that ends by labeling level 5 ``native-like facility'', a phrase that the corpus's rubric uses only for its holistic top level. Every criterion of the ordinal panels carries the same weight.

\paragraph{Prior automatic evaluation on these benchmarks.}
Earlier automatic evaluation on the dialogue sets reports correlations with the rater mean. FED and USR introduced reference-free metrics on them~\citep{mehri2020unsupervised,mehri2020usr}, and G-Eval and LLM-Eval report LLM evaluators there~\citep{liu2023geval,lin2023llmeval}. On ELLIPSE, supervised models reach a mean QWK of 0.854 on the six analytic traits~\citep{aljuaid2025transgat}, and zero-shot GPT-4 reaches 0.307 on the holistic score~\citep{hou2025improving}, a different target. Calibration examples help GPT-4 rate short second-language essays~\citep{yancey2023rating}. Section~\ref{sec:related} cites evidence that essay graders built on LLMs score below human raters. JUDGE-BENCH, which includes both USR sets, scores ordinal ratings by correlation~\citep{bavaresco2025llms}. A correlation with the rater mean does not register a constant shift. We therefore also report exact accuracy and the offset, for four judges under one protocol.

\paragraph{Labels.}
Ratings of N/A are dropped before the mean is taken, which affects FED-Dialogue \crit{error\_recovery} (Appendix~\ref{app:abstention}) and one FED-Turn \crit{fluent} pair, whose label rests on four raters. Half-way means are kept, and Table~\ref{tab:panel-construction} counts them; outside ELLIPSE, at most two labels on any panel are half-way means. ELLIPSE labels average the two raters' individual scores rather than taking the benchmark's released score. HelpSteer2's labels are the released ones (Table~\ref{tab:panel-construction}), not the per-annotator ratings published separately. On an embedded binary criterion a tie among the raters would have removed the pair, and none occurred.

\paragraph{Rater reference.}
The rater reference of Section~\ref{sec:labels} is scored with the same metrics as the judges and pools over units with at least two raters. Each rater on a binary criterion is compared with the majority of the others, and ties are skipped. Panel exact accuracy pools the ordinal criteria in proportion to their rater comparisons, and panel QWK is mean QWK as defined in Section~\ref{sec:metrics}. The leave-one-out target of Section~\ref{sec:labels} matters most when a unit has few raters, since with three raters it is the mean of two ratings. With three raters per unit on the USR panels, the reference is 100.0\% on unanimous pairs and 34.4\% (USR-TC) or 42.0\% (USR-PC) on split pairs, so its value is set largely by the share of unanimous pairs. Of the leave-one-out \crit{factuality} targets on LFQA, 48.3\% are half-way means of two ratings and round up, whereas no three-rater label on LFQA is half-way. Label uncertainty of this kind can distort any comparison with human agreement~\citep{elangovan2025beyond}.

\paragraph{LLM judges.}
The harness routes all three LLM judges through LiteLLM. Each runs as a single judge in the harness, with a strict JSON response schema and no few-shot examples. As the harness's documentation describes, the shuffle of Section~\ref{sec:llm} draws from a generator seeded with the logged per-run seed, the submission, the criterion index, and the judge. It moves the NA option with the others, and the chosen option is mapped back to its level. Only the effort runs and the effort calibration's explicit-medium runs sent a reasoning or thinking parameter (Section~\ref{sec:llm}; Appendix~\ref{app:effort}); like the main runs, the higher-tier runs sent none (Appendix~\ref{app:highertier}). All three judges still emitted reasoning tokens in the main runs, Luna far fewer than the other two. On the ordinal panels Luna emitted 337{,}696, Gemini 2{,}556{,}073, and DeepSeek 3{,}046{,}094. Luna's count includes the first run of the spot check described below, which shared the response cache. With the cache on, twelve of Luna's embedded binary verdicts were served from that spot check's cached responses, and every other recorded verdict comes from a single pass over its panel.

\paragraph{Tier descriptions.}
{\raggedright The quotations below, accessed 27 September 2026, give the vendors' own descriptions of the flash tier's speed and price; Appendix~\ref{app:highertier} quotes the vendors on the higher tier, and Table~\ref{tab:cost} gives the wall times we measured. OpenAI called GPT-5.6 Luna ``our fastest and most affordable model'' on 30 July 2026 ({\def\UrlBreaks{\do\.\do\/\do\-}\url{https://openai.com/index/advancing-the-price-performance-frontier-with-gpt-5-6/}}), and its model page rates Luna's speed ``Fast'' and says it is ``designed for cost-sensitive, high-volume workloads'' (\url{https://developers.openai.com/api/docs/models/gpt-5.6-luna}). Google offers Gemini 3.8 Flash ``with the speed and cost efficiency of Flash'' (\url{https://ai.google.dev/gemini-api/docs/models/gemini-3.8-flash}) and on 2 September 2026 described it as ``often approaching the performance of higher-cost frontier models'' ({\def\UrlBreaks{\do\.\do\/\do\-}\url{https://blog.google/innovation-and-ai/models-and-research/gemini-models/3-8-flash-and-3-8-flash-cyber/}}). DeepSeek introduced V4.1 Flash with ``faster inference'' and ``lower API prices'' (\url{https://www.deepseek.com/en/news/deepseek-v4-1-flash/}).\par}

\paragraph{System prompts.}
The LLM judges receive the harness's default system prompts (Table~\ref{tab:judge-inputs}). The multi-choice prompt, which they receive for ordinal criteria, tells the judge to ``Pay careful attention to negation, scope, and qualifying language'' and to ``Be strict about factual accuracy but flexible about wording'', and not to retreat to a middle option when unsure (``Do not default to middle options out of uncertainty''). When a submission contradicts itself and no stance dominates on an ordered scale, the prompt asks for the lower-quality level (``select the option reflecting the weaker (lower-quality) reading''). It reserves the NA option for submissions that point to inaccessible content, cannot be judged on the question, or are too garbled to evaluate, and it warns against choosing NA when the judge is uncertain but has some evidence.

\paragraph{Routing and failures.}
OpenRouter's default routing spread DeepSeek's main-run requests on the ordinal panels over 11 upstream providers. DeepSeek returned an empty response on six ordinal pairs (four on ELLIPSE, one on HelpSteer2, and one on LFQA), which are excluded from every ordinal metric. It did the same on one RiceChem pair and two HealthBench pairs. These three are the one exception to the exclusion of judge-call failures from the denominators, because our scoring of the binary panels took the UNMET verdict that the harness recorded for each failed per-criterion request of DeepSeek and did not check the failure recorded with it. They count as verdicts in the analyses of the main runs (Table~\ref{tab:abstention}). No request to either decision model failed, and Appendix~\ref{app:decider} counts the Decider's retries. Appendix~\ref{app:wholerubric} gives the routing and the failures of the whole-rubric requests, Appendix~\ref{app:effort} those of the effort runs, and Appendix~\ref{app:highertier} those of the higher-tier runs.

\paragraph{What the decision models receive.}
The two fields of Jev's state (Section~\ref{sec:jev}) are named \texttt{input} and \texttt{submission}, matching the tags that mark the input and the submission in the LLM judges' user message (Table~\ref{tab:judge-inputs}). The input is the unit's context. It is the preceding conversation on FED-Turn and HealthBench, the conversation with the fact or persona the response is conditioned on for the USR panels, the user's prompt on HelpSteer2, the question on LFQA and RiceChem, and the title of the essay prompt on ELLIPSE. On FED-Dialogue it is the fixed sentence ``A conversation between a User and a System.'' The same code built the Decider's requests, so they carry the same state and questions as Jev's. Perplexity's API takes a request body of the same form as TypeSafe's, with fields for the model name, the state, and the named questions. Unlike the LLM judges, neither decision model receives the criterion type. Appendix~\ref{app:decider} describes the Decider's client, rate limit, retries, and repeat.

\paragraph{Run-to-run variation of Jev.}
Requests named the \texttt{jev-latest} alias, and the main runs did not log the version that answered them. A stability study on the day of the main runs logged it: five interleaved passes over the RiceChem panel and a 59-pair HealthBench subset, under all three binary framings, used 2{,}250 requests, and every one of them reports \texttt{jev-1.13.0}. On RiceChem, 2.9\% of pairs changed verdict at least once under Jev Noul and 3.3\% under Jev Choice. Under every framing, each change occurred on a pair whose mean probability of MET lay within 0.1 of 0.5 (Figure~\ref{fig:binary-confidence}b shows Jev Noul). Across passes, the standard deviation of accuracy was 0.36 points for Jev Noul on RiceChem, 0.76 points on the HealthBench subset, and at most 0.93 points for any framing. The share of pairs with identical probabilities in all five passes ranged from 4.2\% of RiceChem pairs under bare Noul to 35.8\% under Jev Choice. The median request latency was 0.18 to 0.20 seconds in every pass.

\paragraph{Repeated Luna judgments.}
In a spot check before the main runs, Luna judged 82 pairs twice, all the pairs of two units from each ordinal panel, the second time with the response cache bypassed. The two runs agreed on 71 of the 82 pairs (86.6\%). Luna's main run is a later, separate pass and was not compared with either. The second default runs of Appendix~\ref{app:effort} later repeated every LLM judge's per-criterion judgments on the 106 repeat units.

%% file: appendix/b_cost.tex
\section{Cost and Wall Time}
\label{app:cost}

\begin{table*}[!tp]
\centering
\footnotesize
\setlength{\tabcolsep}{4pt}
\begin{tabular}{@{}lrrrrrrrr@{}}
\toprule
 & & & & & & & \multicolumn{2}{c}{DeepSeek} \\
\cmidrule(l){8-9}
Panel & Pairs & \makecell[r]{Criteria\\per unit} & Jev Choice & Jev Score & Luna & Gemini & billed & estimate \\
\midrule
\multicolumn{9}{@{}l}{\emph{(a) Cost in US dollars (ratio to Jev Choice)}} \\
RiceChem & 819 & 6.8 & 0.0071 & -- & 0.284 (40$\times$) & 2.321 (326$\times$) & 0.518 (73$\times$) & 0.382 \\
HealthBench & 406 & 2.0 & 0.0129 & -- & 0.226 (18$\times$) & 1.560 (121$\times$) & 0.430 (33$\times$) & 0.349 \\
\midrule
ELLIPSE & 1{,}548 & 6.0 & 0.0239 & 0.0234 & 0.690 (29$\times$) & 9.457 (396$\times$) & 1.877 (79$\times$) & 1.436 \\
FED-Turn & 600 & 8.0 & 0.0027 & 0.0025 & 0.125 (47$\times$) & 1.636 (607$\times$) & 0.234 (87$\times$) & 0.178 \\
FED-Dialogue & 250 & 10.0 & 0.0011 & 0.0010 & 0.065 (61$\times$) & 0.825 (770$\times$) & 0.181 (169$\times$) & 0.130 \\
HelpSteer2 & 360 & 4.0 & 0.0050 & 0.0049 & 0.163 (33$\times$) & 1.450 (290$\times$) & 0.324 (65$\times$) & 0.232 \\
LFQA & 360 & 3.0 & 0.0050 & 0.0048 & 0.112 (23$\times$) & 1.266 (255$\times$) & 0.271 (55$\times$) & 0.186 \\
USR-TC & 360 & 5.0 & 0.0032 & 0.0032 & 0.103 (32$\times$) & 1.206 (373$\times$) & 0.200 (62$\times$) & 0.142 \\
USR-PC & 300 & 5.0 & 0.0025 & 0.0024 & 0.077 (31$\times$) & 0.866 (349$\times$) & 0.147 (59$\times$) & 0.107 \\
\midrule
Nine panels & 5{,}003 & -- & 0.0633 & 0.0422$^{a}$ & 1.844 (29$\times$) & 20.587 (325$\times$) & 4.183 (66$\times$) & 3.141 (50$\times$) \\
US\$ per 1{,}000 pairs & & & 0.013 & 0.011$^{a}$ & 0.37 & 4.11 & 0.84 & 0.63 \\
\midrule
\multicolumn{9}{@{}l}{\emph{(b) Wall time in seconds (ratio to Jev Choice)}} \\
RiceChem & & & 3.9 & -- & 238 (61$\times$) & 210 (54$\times$) & 1{,}988 (507$\times$) & \\
HealthBench & & & 5.3 & -- & 158 (30$\times$) & 133 (25$\times$) & 1{,}202 (227$\times$) & \\
\midrule
ELLIPSE & & & 7.4 & 7.5 & 205 (28$\times$) & 316 (43$\times$) & 1{,}643 (223$\times$) & \\
FED-Turn & & & 1.9 & 2.2 & 56 (29$\times$) & 66 (34$\times$) & 217 (112$\times$) & \\
FED-Dialogue & & & 0.8 & 0.8 & 28 (35$\times$) & 30 (38$\times$) & 203 (252$\times$) & \\
HelpSteer2 & & & 2.4 & 2.3 & 56 (24$\times$) & 55 (23$\times$) & 407 (173$\times$) & \\
LFQA & & & 3.4 & 3.5 & 46 (14$\times$) & 50 (15$\times$) & 404 (120$\times$) & \\
USR-TC & & & 2.0 & 1.8 & 40 (20$\times$) & 55 (28$\times$) & 116 (59$\times$) & \\
USR-PC & & & 1.6 & 1.6 & 33 (21$\times$) & 33 (21$\times$) & 119 (76$\times$) & \\
\midrule
Nine panels & & & 28.6 & 19.8$^{a}$ & 859 (30$\times$) & 950 (33$\times$) & 6{,}298 (220$\times$) & \\
\bottomrule
\end{tabular}
\caption{Cost and wall time per panel and judge. Concurrency: Jev 8 on every panel; Luna 16, Gemini 20, and DeepSeek 32 on the ordinal panels; every judge 8 on the binary panels. DeepSeek: billed cost (Section~\ref{sec:metrics}) and, in the last column, the harness's estimate for the same requests, with its ratio given only for the total. $^{a}$Seven ordinal panels only; Jev Score was not run on the binary panels. Jev's bare Noul and Jev Noul costs on the binary panels are in Table~\ref{tab:binary}, and judge-call failures in Table~\ref{tab:abstention}. Prices are those of the run date.}
\label{tab:cost}
\end{table*}

\paragraph{How cost is computed.}
Section~\ref{sec:metrics} gives the cost rule behind Table~\ref{tab:cost}. Each decision model's cost is the input-token count that its API reported for each request, at its price in Section~\ref{sec:jev}. The Decider's counts agree to the token across its run summaries, its records of each unit, and a log of every request. For Luna and Gemini, cost is LiteLLM's completion cost, computed for each request from the token counts the API reported, reasoning tokens included. Gemini's recorded cost is reproduced exactly by \$0.75 per million prompt tokens and \$3.75 per million completion tokens, and its prompt-token price is about 18 times Jev's. A response served from the cache carries the cost of its original request. Over the seven ordinal panels Jev Choice cost \$0.043 and Jev Score \$0.042, about the same. Appendix~\ref{app:decider} gives the Decider's costs and wall times on each panel, and the launch price in Perplexity's changelog.

\paragraph{Prices of the higher tier.}
The costs of the higher-tier runs (Appendix~\ref{app:highertier}) rest on the vendors' list prices of 2 October 2026. In US dollars per million uncached input, cached input, and output tokens, Terra and Gemini 3.1 Pro both list at \$2.00, \$0.20, and \$12.00, and DeepSeek V4 Pro at \$0.66, \$0.022, and \$1.98 at DeepSeek's off-peak rates. OpenAI also charges \$2.50 per million prompt tokens written to its prompt cache for Terra and \$0.25 for Luna, whose other prices are \$0.20, \$0.02, and \$1.20. DeepSeek V4.1 Flash lists at \$0.15, \$0.003, and \$0.60, and Gemini 3.8 Flash at the two prices given above. Priced in this way, the recorded costs of Luna, Terra, and Gemini 3.1 Pro are reproduced exactly, and Table~\ref{tab:ht-runs} uses these list prices. DeepSeek V4 Pro, like DeepSeek, is reported at its OpenRouter bill.

\paragraph{DeepSeek's billed cost.}
A panel's billed cost sums the costs that OpenRouter reported with each response. The harness's own estimate prices 4{,}747 of the 4{,}994 answered requests at DeepSeek's list prices, whichever upstream provider served them, and it is lower (Table~\ref{tab:cost}, last column). The billed cost is 1.36 times the estimate on RiceChem and 1.23 times on HealthBench. On single ordinal panels the factor runs from 1.31 to 1.46, and it is 1.34 over the seven together. The billed totals exclude the six failed requests on the ordinal panels and the three failed binary ones, which left no cached response.

\paragraph{Criteria per unit and unit length.}
Across the ordinal panels, Gemini's cost ratio rises with every increase in criteria per unit, from 255$\times$ on LFQA (3.0 criteria per unit) to 770$\times$ on FED-Dialogue (10.0). Over all nine panels it runs from 121$\times$ on HealthBench (2.0 criteria per unit) to FED-Dialogue's 770$\times$. RiceChem departs from that order. With 6.8 criteria per unit, its ratio (326$\times$) is below those of ELLIPSE and both USR panels, which have fewer. The Luna and DeepSeek ratios follow the order less closely. Among Luna's ratios, ELLIPSE's (29$\times$) is below HelpSteer2's (33$\times$), and among DeepSeek's billed ratios, HelpSteer2's (65$\times$) is above those of both USR panels (62$\times$ and 59$\times$), although HelpSteer2 has fewer criteria per unit than either. Unit length has little effect. ELLIPSE's essays are the longest ordinal units and the USR responses among the shortest (Table~\ref{tab:panel-construction}), yet with one criterion per unit more, ELLIPSE's Gemini ratio (396$\times$) is close to the USR panels' (373$\times$ and 349$\times$).

\paragraph{Output and reasoning tokens.}
For each criterion of a unit, Gemini writes an explanation and spends reasoning tokens in either judging mode, all billed at its output price, whereas a criterion adds only a question to Jev's request and nothing to Jev's output, which is free. With per-criterion judgments each criterion also puts another copy of the system prompt and of the unit's text into Gemini's prompt tokens, none of which Gemini reported as cached (Appendix~\ref{app:wholerubric}). The number of requests alone does not set the order of Gemini's ratios, since its whole-rubric ratios, at one request per unit as for Jev, mostly keep that order (Appendix~\ref{app:wholerubric}). Gemini's per-token price and its reasoning tokens widen its cost ratio. Completion tokens, which include reasoning tokens, make up 60.0\% of Gemini's per-criterion bill over the nine panels, \$6.66 of \$9.46 (70.4\%) on ELLIPSE, and 47.8\% of its \$1.64 on FED-Turn.

\paragraph{Wall time.}
Wall time was measured around each panel's evaluation and includes checkpoint writes. On the ordinal panels each judge ran in its own process. The three LLM processes ran simultaneously, each against its own provider, and the Jev Choice and Jev Score processes ran together against Jev's API. On each binary panel the LLM judges ran one after another, Luna first and DeepSeek last, and Jev's binary framings did the same in a single process. The RiceChem and HealthBench runs proceeded simultaneously and sent requests to the same providers in the same order, so their LLM wall times are not isolated measurements.

%% file: appendix/c_tests.tex
\section{Paired Comparisons and Secondary Metrics}
\label{app:tests}

Section~\ref{sec:accuracy} rests on the unit-level comparisons of Table~\ref{tab:clustered}. Table~\ref{tab:paired} adds every pair-level sign test, and Table~\ref{tab:secondary} gives secondary metrics for the ordinal panels, with rows for both decision models. The comparisons of the first two tables are Jev's, and Appendix~\ref{app:decider} gives the Decider's.

\begin{table*}[!tp]
\centering
\footnotesize
\setlength{\tabcolsep}{4pt}
\begin{tabular}{@{}l@{\hspace{6pt}}lrrlrrrl@{}}
\toprule
Panel & Judge & Units & Diff & 95\% interval & Bound & J\,:\,O & $p_{\text{unit}}$ & Class \\
\midrule
\multicolumn{9}{@{}l}{\emph{Each LLM judge against Jev Choice}} \\
\addlinespace[2pt]
RiceChem & Luna & 121 & $-3.2$ & $-5.5$, $-0.9$ & 5.1 & 40\,:\,21 & 0.020 & separated \\
 & Gemini & 121 & $-4.9$\ddg & $-7.5$, $-2.3$ & 7.0 & 46\,:\,20 & 0.002 & separated \\
 & DeepSeek & 121 & $-1.7$ & $-4.6$, $+1.2$ & 4.1 & 40\,:\,33 & 0.48 & parity \\
\addlinespace[2pt]
HealthBench & Luna & 200 & $-6.7$ & $-12.2$, $-1.0$ & 11.3 & 55\,:\,36 & 0.059 & separated \\
 & Gemini & 200 & $+2.5$ & $-1.4$, $+6.2$ & 5.6 & 22\,:\,34 & 0.14 & -- \\
 & DeepSeek & 200 & $-0.7$ & $-4.6$, $+3.1$ & 3.9 & 25\,:\,26 & 1.00 & parity \\
\addlinespace[2pt]
ELLIPSE & Luna & 258 & $+0.6$ & $-0.8$, $+2.1$ & 1.8 & 34\,:\,45 & 0.26 & parity \\
 & Gemini & 258 & $+16.4$\ddg & $+14.1$, $+18.7$ & 18.3 & 7\,:\,150 & $5 \times 10^{-36}$ & separated \\
 & DeepSeek & 258 & $+1.6$ & $-0.1$, $+3.4$ & 3.1 & 53\,:\,66 & 0.27 & parity \\
\addlinespace[2pt]
FED-Turn & Luna & 75 & $+1.0$ & $-3.3$, $+5.2$ & 4.5 & 18\,:\,29 & 0.14 & parity \\
 & Gemini & 75 & $+3.7$ & $-0.3$, $+7.8$ & 7.2 & 17\,:\,25 & 0.28 & -- \\
 & DeepSeek & 75 & $+2.2$ & $-1.7$, $+5.8$ & 5.3 & 19\,:\,30 & 0.15 & -- \\
\addlinespace[2pt]
FED-Dialogue & Luna & 25 & $-0.4$ & $-7.0$, $+6.8$ & 6.1 & 10\,:\,7 & 0.63 & -- \\
 & Gemini & 25 & $+7.9$\ddg & $+2.7$, $+13.5$ & 12.4 & 2\,:\,10 & 0.039 & separated \\
 & DeepSeek & 25 & $-1.7$ & $-6.9$, $+3.5$ & 6.0 & 10\,:\,7 & 0.63 & -- \\
\addlinespace[2pt]
HelpSteer2 & Luna & 90 & $-10.2$\ddg & $-15.8$, $-4.8$ & 14.9 & 40\,:\,14 & $5 \times 10^{-4}$ & separated \\
 & Gemini & 90 & $+5.7$ & $+0.9$, $+10.3$ & 9.5 & 13\,:\,31 & 0.010 & separated \\
 & DeepSeek & 90 & $-2.3$ & $-6.6$, $+1.7$ & 5.9 & 20\,:\,15 & 0.50 & -- \\
\addlinespace[2pt]
LFQA & Luna & 120 & $+2.8$ & $-0.6$, $+6.1$ & 5.6 & 13\,:\,22 & 0.18 & -- \\
 & Gemini & 120 & $-1.1$ & $-5.6$, $+3.1$ & 4.8 & 28\,:\,25 & 0.78 & parity \\
 & DeepSeek & 120 & $-3.7$ & $-7.6$, $0.0$ & 6.9 & 23\,:\,14 & 0.19 & -- \\
\addlinespace[2pt]
USR-TC & Luna & 72 & $+0.3$ & $-4.2$, $+4.7$ & 4.2 & 20\,:\,20 & 1.00 & parity \\
 & Gemini & 72 & $+3.1$ & $-1.1$, $+7.2$ & 6.5 & 14\,:\,24 & 0.14 & -- \\
 & DeepSeek & 72 & $-1.4$ & $-6.7$, $+3.6$ & 5.8 & 18\,:\,20 & 0.87 & -- \\
\addlinespace[2pt]
USR-PC & Luna & 60 & $+4.0$ & $+0.3$, $+7.7$ & 7.0 & 8\,:\,17 & 0.11 & separated \\
 & Gemini & 60 & $+4.3$ & $-0.3$, $+9.0$ & 8.3 & 9\,:\,19 & 0.087 & -- \\
 & DeepSeek & 60 & $-0.3$ & $-4.4$, $+4.0$ & 3.7 & 12\,:\,9 & 0.66 & parity \\
\midrule
\multicolumn{9}{@{}l}{\emph{Jev Score against Jev Choice (ordinal pairs)}} \\
\addlinespace[2pt]
ELLIPSE & Jev Score & 258 & $+1.9$\ddg & $+0.9$, $+2.9$ & 2.8 & 15\,:\,42 & $5 \times 10^{-4}$ & separated, parity \\
FED-Turn & Jev Score & 75 & $+10.5$\ddg & $+6.5$, $+14.5$ & 13.7 & 9\,:\,42 & $3 \times 10^{-6}$ & separated \\
FED-Dialogue & Jev Score & 25 & $+7.2$\ddg & $+2.0$, $+12.8$ & 11.9 & 4\,:\,13 & 0.049 & separated \\
HelpSteer2 & Jev Score & 90 & $-3.4$ & $-7.8$, $+0.9$ & 7.1 & 26\,:\,17 & 0.22 & -- \\
LFQA & Jev Score & 120 & $+0.8$ & $-2.5$, $+4.2$ & 3.6 & 11\,:\,14 & 0.69 & parity \\
USR-TC & Jev Score & 72 & $+4.2$ & $+0.9$, $+7.9$ & 7.4 & 3\,:\,11 & 0.057 & separated \\
USR-PC & Jev Score & 60 & $-1.1$ & $-5.6$, $+3.3$ & 5.0 & 9\,:\,7 & 0.80 & -- \\
\bottomrule
\end{tabular}
\caption{Paired comparisons with units resampled (Section~\ref{sec:metrics}). Diff: the second judge's accuracy minus Jev Choice's, in points, on the pairs both scored; on the ordinal panels the upper block includes the embedded binary pairs (Section~\ref{sec:jev}), and the lower block uses ordinal pairs only. 95\% interval: percentile interval over 10{,}000 resamples of units. Bound: the smallest symmetric margin, in points, that contains the 90\% interval. J\,:\,O: units on which Jev Choice is ahead against units on which the other judge is ahead, and $p_{\text{unit}}$ the exact two-sided sign test over them, with tied units set aside. Class: separated (Section~\ref{sec:metrics}), parity (Section~\ref{sec:metrics}), both, or neither (--). Parity needs a bound below 5 points, so USR-PC Jev Score, with a bound of exactly 5.0, is not at parity. \ddag: the bootstrap $p$-value survives Holm correction within its block of 27 or 7 comparisons.}
\label{tab:clustered}
\end{table*}

\begin{table*}[!tp]
\centering
\footnotesize
\setlength{\tabcolsep}{1.75pt}
\begin{tabular}[t]{@{}llrrrr@{\hspace{6pt}}r@{}}
\toprule
Panel & Judge & $n$ & Jev & LLM & $p$ & $p_{\text{Holm}}$ \\
\midrule
\multicolumn{7}{@{}l}{\emph{Jev Choice against each LLM judge}} \\
\addlinespace[2pt]
RiceChem & Luna & 819 & 64 & 38 & 0.013 & 0.30 \\
 & Gemini & 819 & 86 & 46 & $6 \times 10^{-4}$ & 0.016 \\
 & DeepSeek & 819 & 65 & 51 & 0.23 & 1.00 \\
\addlinespace[2pt]
HealthBench & Luna & 406 & 74 & 47 & 0.018 & 0.39 \\
 & Gemini & 406 & 28 & 38 & 0.27 & 1.00 \\
 & DeepSeek & 406 & 34 & 31 & 0.80 & 1.00 \\
\addlinespace[2pt]
ELLIPSE & Luna & 1{,}548 & 55 & 65 & 0.41 & 1.00 \\
 & Gemini & 1{,}548 & 30 & 284 & $5 \times 10^{-53}$ & $1 \times 10^{-51}$ \\
 & DeepSeek & 1{,}543 & 76 & 101 & 0.071 & 1.00 \\
\addlinespace[2pt]
FED-Turn & Luna & 600 & 51 & 57 & 0.63 & 1.00 \\
 & Gemini & 600 & 37 & 59 & 0.032 & 0.63 \\
 & DeepSeek & 599 & 40 & 53 & 0.21 & 1.00 \\
\addlinespace[2pt]
FED-Dialogue & Luna & 232 & 25 & 24 & 1.00 & 1.00 \\
 & Gemini & 228 & 12 & 30 & 0.008 & 0.19 \\
 & DeepSeek & 232 & 28 & 24 & 0.68 & 1.00 \\
\addlinespace[2pt]
HelpSteer2 & Luna & 352 & 68 & 32 & $4 \times 10^{-4}$ & 0.011 \\
 & Gemini & 351 & 28 & 48 & 0.029 & 0.60 \\
 & DeepSeek & 350 & 40 & 32 & 0.41 & 1.00 \\
\addlinespace[2pt]
LFQA & Luna & 358 & 17 & 27 & 0.17 & 1.00 \\
 & Gemini & 357 & 35 & 31 & 0.71 & 1.00 \\
 & DeepSeek & 354 & 30 & 17 & 0.079 & 1.00 \\
\addlinespace[2pt]
USR-TC & Luna & 360 & 36 & 37 & 1.00 & 1.00 \\
 & Gemini & 358 & 26 & 37 & 0.21 & 1.00 \\
 & DeepSeek & 359 & 37 & 32 & 0.63 & 1.00 \\
\addlinespace[2pt]
USR-PC & Luna & 299 & 17 & 29 & 0.10 & 1.00 \\
 & Gemini & 299 & 16 & 29 & 0.072 & 1.00 \\
 & DeepSeek & 299 & 23 & 22 & 1.00 & 1.00 \\
\bottomrule
\end{tabular}\hfill
\begin{tabular}[t]{@{}llrrrr@{}}
\toprule
Panel & Judge & $n$ & Jev & LLM & $p$ \\
\midrule
\multicolumn{6}{@{}l}{\emph{Jev Noul against each LLM judge (binary panels)}} \\
\addlinespace[2pt]
RiceChem & Luna & 819 & 62 & 43 & 0.078 \\
 & Gemini & 819 & 88 & 55 & 0.007 \\
 & DeepSeek & 819 & 60 & 53 & 0.57 \\
\addlinespace[2pt]
HealthBench & Luna & 406 & 73 & 45 & 0.013 \\
 & Gemini & 406 & 33 & 42 & 0.36 \\
 & DeepSeek & 406 & 39 & 35 & 0.73 \\
\addlinespace[2pt]
\midrule
\multicolumn{6}{@{}l}{\emph{Jev Score against each LLM judge (ordinal pairs)}} \\
\addlinespace[2pt]
ELLIPSE & Luna & 1{,}548 & 74 & 55 & 0.11 \\
 & Gemini & 1{,}548 & 30 & 255 & $1 \times 10^{-45}$ \\
 & DeepSeek & 1{,}543 & 92 & 88 & 0.82 \\
\addlinespace[2pt]
FED-Turn & Luna & 525 & 82 & 35 & $2 \times 10^{-5}$ \\
 & Gemini & 525 & 75 & 43 & 0.004 \\
 & DeepSeek & 524 & 75 & 38 & $6 \times 10^{-4}$ \\
\addlinespace[2pt]
FED-Dialogue & Luna & 214 & 35 & 19 & 0.040 \\
 & Gemini & 204 & 21 & 21 & 1.00 \\
 & DeepSeek & 211 & 35 & 18 & 0.027 \\
\addlinespace[2pt]
HelpSteer2 & Luna & 360 & 63 & 44 & 0.081 \\
 & Gemini & 355 & 37 & 72 & 0.001 \\
 & DeepSeek & 357 & 45 & 52 & 0.54 \\
\addlinespace[2pt]
LFQA & Luna & 359 & 22 & 29 & 0.40 \\
 & Gemini & 359 & 41 & 33 & 0.42 \\
 & DeepSeek & 354 & 33 & 17 & 0.033 \\
\addlinespace[2pt]
USR-TC & Luna & 216 & 31 & 23 & 0.34 \\
 & Gemini & 215 & 21 & 20 & 1.00 \\
 & DeepSeek & 216 & 34 & 18 & 0.036 \\
\addlinespace[2pt]
USR-PC & Luna & 180 & 15 & 25 & 0.15 \\
 & Gemini & 180 & 17 & 27 & 0.17 \\
 & DeepSeek & 180 & 20 & 25 & 0.55 \\
\bottomrule
\end{tabular}

\vspace{8pt}
\setlength{\tabcolsep}{4pt}
\begin{tabular}{@{}lrrrrrrr@{}}
\toprule
\multicolumn{8}{@{}l}{\emph{Jev Choice against Jev Score (ordinal pairs)}} \\
\addlinespace[2pt]
 & ELLIPSE & FED-Turn & FED-Dialogue & HelpSteer2 & LFQA & USR-TC & USR-PC \\
\midrule
Pairs $n$ & 1{,}548 & 525 & 208 & 352 & 358 & 216 & 179 \\
Jev Choice only right & 18 & 14 & 5 & 36 & 13 & 6 & 10 \\
Jev Score only right & 47 & 69 & 20 & 24 & 16 & 15 & 8 \\
$p$ & $4 \times 10^{-4}$ & $7 \times 10^{-10}$ & 0.004 & 0.16 & 0.71 & 0.078 & 0.81 \\
\bottomrule
\end{tabular}
\caption{Every pair-level sign test. These tests treat pairs as independent, unlike the unit-level comparisons of Jev Choice in Table~\ref{tab:clustered}. $n$ is the number of pairs both judges scored. The Jev column counts the pairs only the Jev framing got right, and the LLM column the pairs only the other judge got right; $p$ is the exact two-sided binomial test on these two counts. Left: Jev Choice against each LLM judge, with Holm-adjusted $p$ over these 27 tests. Right: Jev Noul against each LLM judge on the binary panels, and Jev Score against each LLM judge on ordinal pairs. Bottom: Jev Choice against Jev Score on ordinal pairs.}
\label{tab:paired}
\end{table*}

\begin{table*}[!t]
\centering
\scriptsize
\setlength{\tabcolsep}{5pt}
\begin{tabular}{@{}lrrrrrrr@{}}
\toprule
 & ELLIPSE & FED-Turn & FED-Dialogue & HelpSteer2 & LFQA & USR-TC & USR-PC \\
\midrule
\multicolumn{8}{l}{\emph{(a) Within-one accuracy (\%)}} \\
\quad Jev Choice & 64.3 & 90.5 & 88.0 & 81.8 & 94.7 & 99.1 & 92.7 \\
\quad Jev Score & 70.1 & 97.9 & 91.6 & 86.1 & 96.4 & 99.5 & 95.6 \\
\quad Decider Choice & 97.3 & 85.9 & 87.1 & 95.1 & 97.8 & 99.1 & 92.8 \\
\quad Decider Score & 98.0 & 93.7 & 92.9 & 96.4 & 98.9 & 99.5 & 93.3 \\
\quad Luna & 61.7 & 94.7 & 90.2 & 80.0 & 95.0 & 98.6 & 98.3 \\
\quad Gemini & 90.5 & 94.1 & 93.1 & 85.4 & 97.2 & 98.6 & 96.7 \\
\quad DeepSeek & 65.3 & 92.7 & 91.0 & 78.7 & 96.0 & 98.6 & 93.3 \\
\quad Rater reference & 98.2 & 97.1 & 97.3 & -- & 94.4 & 95.5 & 98.0 \\
\addlinespace[3pt]
\multicolumn{8}{l}{\emph{(b) Mean absolute error (levels)}} \\
\quad Jev Choice & 1.26 & 0.53 & 0.63 & 0.78 & 0.37 & 0.39 & 0.51 \\
\quad Jev Score & 1.16 & 0.35 & 0.57 & 0.73 & 0.34 & 0.34 & 0.49 \\
\quad Decider Choice & 0.56 & 0.64 & 0.72 & 0.43 & 0.33 & 0.40 & 0.49 \\
\quad Decider Score & 0.51 & 0.58 & 0.69 & 0.48 & 0.29 & 0.31 & 0.58 \\
\quad Luna & 1.29 & 0.48 & 0.63 & 0.89 & 0.34 & 0.39 & 0.41 \\
\quad Gemini & 0.80 & 0.45 & 0.50 & 0.66 & 0.36 & 0.36 & 0.43 \\
\quad DeepSeek & 1.23 & 0.48 & 0.62 & 0.85 & 0.39 & 0.43 & 0.49 \\
\addlinespace[3pt]
\multicolumn{8}{l}{\emph{(c) Offset against the unrounded rater mean (levels)}} \\
\quad Jev Choice & $-1.01$ & $-0.29$ & $-0.53$ & $-0.33$ & $-0.20$ & $+0.02$ & $-0.31$ \\
\quad Jev Score & $-0.92$ & $-0.08$ & $-0.44$ & $-0.29$ & $-0.11$ & $+0.05$ & $-0.28$ \\
\quad Decider Choice & $-0.23$ & $-0.41$ & $-0.60$ & $+0.09$ & $+0.12$ & $+0.11$ & $-0.34$ \\
\quad Decider Score & $-0.16$ & $-0.42$ & $-0.63$ & $-0.05$ & $+0.07$ & $+0.06$ & $-0.41$ \\
\quad Luna & $-1.04$ & $-0.15$ & $-0.51$ & $-0.48$ & $-0.20$ & $+0.14$ & $-0.21$ \\
\quad Gemini & $-0.54$ & $-0.07$ & $-0.33$ & $-0.26$ & $-0.11$ & $+0.02$ & $-0.34$ \\
\quad DeepSeek & $-0.93$ & $-0.10$ & $-0.48$ & $-0.39$ & $-0.14$ & $+0.18$ & $-0.28$ \\
\addlinespace[3pt]
\multicolumn{8}{l}{\emph{(d) Spearman $\rho$ of the unit aggregate with the holistic score}} \\
\quad Jev Choice & 0.70 & 0.60 & 0.76 & 0.20 & 0.51 & 0.72 & 0.77 \\
\quad Jev Score & 0.67 & 0.60 & 0.73 & 0.13 & 0.52 & 0.74 & 0.80 \\
\quad Decider Choice & 0.74 & 0.60 & 0.67 & 0.31 & 0.52 & 0.71 & 0.90 \\
\quad Decider Score & 0.73 & 0.62 & 0.75 & 0.40 & 0.62 & 0.76 & 0.87 \\
\quad Luna & 0.66 & 0.55 & 0.79 & 0.22 & 0.53 & 0.64 & 0.82 \\
\quad Gemini & 0.72 & 0.44 & 0.68 & 0.27 & 0.61 & 0.79 & 0.92 \\
\quad DeepSeek & 0.62 & 0.51 & 0.83 & 0.12 & 0.47 & 0.72 & 0.76 \\
\addlinespace[3pt]
\multicolumn{8}{l}{\emph{(e) Spearman $\rho$ with the unrounded rater mean, pooled over criteria}} \\
\quad Jev Choice & 0.42 & 0.57 & 0.56 & 0.61 & 0.52 & 0.62 & 0.59 \\
\quad Jev Score & 0.37 & 0.60 & 0.51 & 0.64 & 0.54 & 0.63 & 0.60 \\
\quad Decider Choice & 0.57 & 0.52 & 0.55 & 0.80 & 0.65 & 0.55 & 0.64 \\
\quad Decider Score & 0.54 & 0.58 & 0.58 & 0.80 & 0.65 & 0.65 & 0.62 \\
\quad Luna & 0.36 & 0.46 & 0.47 & 0.57 & 0.52 & 0.58 & 0.62 \\
\quad Gemini & 0.56 & 0.50 & 0.49 & 0.66 & 0.60 & 0.68 & 0.76 \\
\quad DeepSeek & 0.37 & 0.47 & 0.57 & 0.56 & 0.55 & 0.58 & 0.58 \\
\addlinespace[3pt]
\multicolumn{8}{l}{\emph{(f) Verdict accuracy on embedded binary criteria (\%)}} \\
\quad Jev Choice & -- & 98.7 & 80.0 & -- & -- & 75.7 & 90.8 \\
\quad Jev Score & -- & 98.7 & 80.0 & -- & -- & 75.7 & 92.5 \\
\quad Decider Choice & -- & 97.3 & 84.0 & -- & -- & 83.3 & 95.8 \\
\quad Decider Score & -- & 97.3 & 84.0 & -- & -- & 83.3 & 95.8 \\
\quad Luna & -- & 96.0 & 80.0 & -- & -- & 75.7 & 94.2 \\
\quad Gemini & -- & 97.3 & 88.0 & -- & -- & 78.3 & 95.0 \\
\quad DeepSeek & -- & 93.3 & 72.0 & -- & -- & 77.6 & 87.5 \\
\quad Rater reference & -- & 93.3 & 95.1 & -- & -- & 90.0 & 95.5 \\
\addlinespace[3pt]
\multicolumn{8}{l}{\emph{(g) Mean QWK against the rounded label}} \\
\quad Jev Choice & 0.15 & 0.42 & 0.34 & 0.29 & 0.47 & 0.57 & 0.47 \\
\quad Jev Score & 0.14 & 0.50 & 0.40 & 0.32 & 0.48 & 0.59 & 0.45 \\
\quad Decider Choice & 0.42 & 0.35 & 0.27 & 0.47 & 0.42 & 0.51 & 0.54 \\
\quad Decider Score & 0.42 & 0.37 & 0.30 & 0.47 & 0.45 & 0.58 & 0.41 \\
\quad Luna & 0.13 & 0.37 & 0.29 & 0.29 & 0.51 & 0.46 & 0.52 \\
\quad Gemini & 0.29 & 0.39 & 0.35 & 0.34 & 0.48 & 0.60 & 0.58 \\
\quad DeepSeek & 0.16 & 0.39 & 0.36 & 0.30 & 0.43 & 0.50 & 0.42 \\
\addlinespace[3pt]
\multicolumn{8}{l}{\emph{(h) Score framings' offsets against the rounded label (levels)}} \\
\quad Jev Score & $-1.15$ & $-0.16$ & $-0.48$ & $-0.29$ & $-0.09$ & $+0.03$ & $-0.34$ \\
\quad Decider Score & $-0.40$ & $-0.50$ & $-0.67$ & $-0.05$ & $+0.09$ & $+0.04$ & $-0.47$ \\
\bottomrule
\end{tabular}
\caption{Secondary metrics on the ordinal panels; the LLM judges' rows serve both decision models' copies. (a) and (b) are computed on ordinal pairs. (c) HelpSteer2's single released label serves as its rater mean, so its row repeats the offsets against the label. (d) Correlation over units between the holistic score (ELLIPSE overall score, FED and USR overall ratings, HelpSteer2 helpfulness, and LFQA acceptability) and the harness's unit aggregate, the mean of option values over a unit's criteria. (e) Predicted level and rater mean are both scaled to $[0, 1]$ within each criterion. (f) Each decision model's Choice and Score rows are its Noul verdicts, from two separate requests. (g) Per-criterion QWK averaged over each panel's ordinal criteria; Table~\ref{tab:main}b sets the matched judges' mean agreement with one another beside their mean agreement with the labels. (h) The other judges' offsets are in Table~\ref{tab:offsets}. Rater reference: as in Section~\ref{sec:labels} and, for (f), Appendix~\ref{app:protocol}.}
\label{tab:secondary}
\end{table*}

\paragraph{Comparisons by panel.}
Among its matched judges, Jev Choice ranks first on RiceChem and second on HealthBench, behind Gemini (Table~\ref{tab:main}), and is separated from Luna on both and from Gemini on RiceChem (Table~\ref{tab:clustered}). On the ordinal panels 16 of 21 comparisons are not separated; of the other five, Gemini leads on ELLIPSE, FED-Dialogue, and HelpSteer2, most on ELLIPSE (16.4 points), and Luna leads on USR-PC and trails on HelpSteer2. DeepSeek is never separated from Jev Choice.

\paragraph{Pair sets.}
Because abstentions and judge-call failures leave the denominators (Section~\ref{sec:metrics}), two judges' accuracies share a denominator only in a paired comparison. The pair set also differs between analyses, so one judge's accuracy on a panel differs between tables. Table~\ref{tab:main} counts an ordinal panel's ordinal pairs. The comparisons of Jev Choice in Table~\ref{tab:clustered} add its embedded binary pairs and keep the pairs both judges scored. Table~\ref{tab:cascade} and the cascade analyses of Appendix~\ref{app:cascades} count every pair that the decision model in the first stage answered, keeping its verdict where the cascade's fallback judge (Section~\ref{sec:routing}) abstained. The one kind of judge-call failure that stays in the denominators, DeepSeek's on the binary panels, is described in Appendix~\ref{app:protocol}.

\paragraph{Pair-level sign tests.}
For two judges, the test counts the pairs both scored that only the first got right and those that only the second got right, and applies an exact two-sided binomial test with probability 0.5 to these discordant counts, the exact McNemar test. Jev Choice is tested on the pair sets of Table~\ref{tab:clustered}. Tests involving Jev Score use ordinal pairs only, since on embedded binary pairs Jev Choice and Jev Score both answer through Jev Noul (Section~\ref{sec:jev}). Unlike the intervals of Table~\ref{tab:clustered}, which resample units because the pairs of one unit share its text, these tests ignore that clustering.

\paragraph{Unit-level analysis.}
Each comparison of Table~\ref{tab:clustered} draws its resamples from its own seeded generator. The bootstrap intervals separate as many of the 27 comparisons as the uncorrected pair-level tests do, four in each direction, and seven of the eight separations coincide. The exception is Gemini's pair-level lead on FED-Turn ($p = 0.032$), which the intervals do not separate. Of its 75 turns, 17 favor Jev Choice, 25 favor Gemini, and 33 are tied ($p_{\text{unit}} = 0.28$). The intervals separate Luna's lead on USR-PC instead, whose pair-level $p$ is 0.10. The unit sign test is significant in six comparisons, three in each direction, and after Holm correction it keeps the same three as the pair-level test. It is not significant on two comparisons with Luna that the intervals separate, Jev Choice's lead on HealthBench ($p_{\text{unit}} = 0.059$) and Luna's lead on USR-PC ($p_{\text{unit}} = 0.11$). Separation also depends on how the paired difference varies across units, so equal accuracies can differ in separation. On USR-PC, Luna and Gemini have the same exact accuracy (Table~\ref{tab:main}), and Gemini's lead over Jev Choice in Table~\ref{tab:clustered} is the larger, 4.3 points against 4.0. Gemini's interval is wider, however, and includes zero, which leaves only Luna's lead separated.

\paragraph{Single accuracies on ELLIPSE.}
Over 2{,}000 resamples of ELLIPSE's essays, the marginal 95\% interval of exact accuracy is 11.5\% to 15.3\% for Jev Choice and 27.0\% to 32.8\% for Gemini.

\paragraph{Parity and equivalence bounds.}
Parity amounts to two one-sided tests of equivalence at $\alpha = 0.05$. It holds with Luna on ELLIPSE, FED-Turn, and USR-TC; with DeepSeek on RiceChem, HealthBench, ELLIPSE, and USR-PC; and with Gemini on LFQA. On the ordinal panels, 6 of the 21 comparisons reach parity, and the other 10 that are not separated have 90\% intervals reaching 5.3 to 8.3 points from zero. The bound of Table~\ref{tab:clustered} has a median of 5.9 points and is at most 7.5 points for 21 of all 27 comparisons. Three of those 21 are also separated, so at a 7.5-point margin they would be both equivalent and separated.

\paragraph{Resampling by sampling group.}
The sampling groups are the 45 HelpSteer2 prompts, the 30 LFQA questions, and the 12 contexts of each USR panel. Resampling them in place of units widens nine of the twelve intervals with an LLM judge on these panels and changes four outcomes. Luna's lead on USR-PC is no longer separated, since its interval now runs from $-0.3$ to $+8.7$ points. Jev Choice's lead over DeepSeek on LFQA becomes separated, with an interval from $-6.5$ to $-0.6$. USR-TC Luna loses parity with a bound of 5.3 points, and LFQA Gemini with one of 5.1. That leaves eight separations, five of them Jev Choice's leads and the other three all Gemini's. With 12 to 45 groups these percentile intervals are rough.

\paragraph{Jev Score against Jev Choice.}
The unit sign test favors Jev Score on ELLIPSE, FED-Turn, and FED-Dialogue, and Holm correction keeps the first two. The bootstrap intervals also separate USR-TC (Appendix~\ref{app:choicescore}). The ELLIPSE difference of 1.9 points is both separated and at parity.

\paragraph{Multiple comparisons.}
Bootstrap $p$-values, twice the smaller share of replicates on either side of zero, are Holm-corrected within each family of comparisons, such as Jev Choice's 27 with the LLM judges. Decider Choice's 27 form a family of their own (Section~\ref{sec:metrics}), so adding the Decider changed no corrected value in Jev's copy, and Appendix~\ref{app:decider} gives the Decider's. Holm correction multiplies the $i$-th smallest of the 27 $p$-values of Jev Choice against an LLM judge by $28-i$, takes a running maximum, and caps the result at 1. Applied to the bootstrap $p$-values, it keeps four separations, two in each direction (\ddag\ in Table~\ref{tab:clustered}): Gemini's leads on ELLIPSE and on FED-Dialogue, and Jev Choice's leads over Luna on HelpSteer2 and over Gemini on RiceChem. Of the pair-level tests it keeps three, Gemini's lead on ELLIPSE and Jev Choice's leads on HelpSteer2 against Luna ($p_{\text{Holm}} = 0.011$) and on RiceChem against Gemini ($p_{\text{Holm}} = 0.016$). Gemini's FED-Dialogue lead, which remains significant when the bootstrap $p$-values are corrected, has a pair-level adjusted value of 0.19. The other 34 pair-level tests in Table~\ref{tab:paired} are reported without correction. Among them, Jev Score is significantly more accurate than an LLM judge in 7 of its 21 comparisons on ordinal pairs and significantly less accurate in 2, both against Gemini, on ELLIPSE and HelpSteer2. These 21 comparisons were not repeated at the unit level.

\paragraph{Secondary metrics.}
\emph{Within-one accuracy} (Table~\ref{tab:secondary}a) is exact accuracy with a one-level miss also counted as correct. In Jev's copy, every offset outside USR-TC is negative against the unrounded rater mean, as it is against the rounded label, so the rounding rule does not produce the sign of the offset (Table~\ref{tab:secondary}c). Each Decider framing's offset also has the same sign against both targets on every panel, although on ELLIPSE the offsets of both framings turn positive against labels rounded down (Section~\ref{sec:agreement}; Appendix~\ref{app:offset}). Jev Noul's MET rate over the embedded binary pairs (74.2\%) is close to the labels' (74.7\%), largely because two USR-TC criteria err in opposite directions, \crit{understandable} with 17 false positives and \crit{uses\_knowledge} with 16 false negatives. The unit aggregate correlates with HelpSteer2's helpfulness score at Spearman 0.12 to 0.27 for Jev's five judges, 0.31 for Decider Choice, and 0.40 for Decider Score, against 0.44 to 0.92 for every judge of either copy on the other ordinal panels (Table~\ref{tab:secondary}d), although HelpSteer2's \crit{correctness} label alone tracks helpfulness (Table~\ref{tab:conventions}).

%% file: appendix/d_binary.tex
\section{The Binary Panels}
\label{app:binary}

\begin{table*}[!tp]
\centering
\footnotesize
\setlength{\tabcolsep}{4pt}
\begin{tabular}{@{}lrrrrrrrrr@{}}
\toprule
 & Accuracy & $\kappa$ & AUROC & FPR & FNR & FP:FN & Pred.\ MET & ECE & Cost \\
 & (\%) & & & (\%) & (\%) & & (\%) & & (US\$) \\
\midrule
\multicolumn{10}{l}{\emph{RiceChem panel: 819 pairs, labeled MET rate 59.5\%}} \\
\quad bare Noul & 69.4 & 0.31 & 0.80 & 61.1 & 9.9 & 203:48 & 78.4 & 0.077 & 0.0037 \\
\quad Jev Noul & 80.1 & 0.57 & 0.86 & 34.9 & 9.7 & 116:47 & 67.9 & 0.056 & 0.0055 \\
\quad Jev Choice & 81.0 & 0.59 & 0.85 & 32.2 & 10.1 & 107:49 & 66.5 & 0.096 & 0.0071 \\
\quad Decider bare Noul & 76.6 & 0.49 & 0.85 & 45.5 & 8.4 & 151:41 & 72.9 & 0.108 & 0.0065 \\
\quad Decider Noul & 80.0 & 0.58 & 0.86 & 28.9 & 14.0 & 96:68 & 62.9 & 0.084 & 0.0070 \\
\quad Decider Choice & 80.3 & 0.59 & 0.86 & 28.6 & 13.6 & 95:66 & 63.0 & 0.087 & 0.0074 \\
\quad Luna & 77.8 & 0.54 & -- & 30.1 & 16.8 & 100:82 & 61.7 & -- & 0.284 \\
\quad Gemini & 76.1 & 0.52 & -- & 21.1 & 25.9 & 70:126 & 52.6 & -- & 2.321 \\
\quad DeepSeek & 79.2 & 0.56 & -- & 30.4 & 14.2 & 101:69 & 63.4 & -- & 0.518 \\
\addlinespace[3pt]
\multicolumn{10}{l}{\emph{HealthBench panel: 406 pairs, labeled MET rate 53.7\%}} \\
\quad bare Noul & 76.4 & 0.52 & 0.83 & 30.9 & 17.4 & 58:38 & 58.6 & 0.074 & 0.0112 \\
\quad Jev Noul & 77.3 & 0.54 & 0.84 & 31.4 & 15.1 & 59:33 & 60.1 & 0.070 & 0.0121 \\
\quad Jev Choice & 77.1 & 0.54 & 0.84 & 31.4 & 15.6 & 59:34 & 59.9 & 0.073 & 0.0129 \\
\quad Decider bare Noul & 79.6 & 0.58 & 0.89 & 33.5 & 9.2 & 63:20 & 64.3 & 0.097 & 0.0076 \\
\quad Decider Noul & 71.9 & 0.43 & 0.78 & 41.5 & 16.5 & 78:36 & 64.0 & 0.148 & 0.0079 \\
\quad Decider Choice & 75.4 & 0.50 & 0.84 & 38.8 & 12.4 & 73:27 & 65.0 & 0.139 & 0.0081 \\
\quad Luna & 70.4 & 0.42 & -- & 17.0 & 40.4 & 32:88 & 39.9 & -- & 0.226 \\
\quad Gemini & 79.6 & 0.59 & -- & 22.3 & 18.8 & 42:41 & 53.9 & -- & 1.560 \\
\quad DeepSeek & 76.4 & 0.53 & -- & 23.9 & 23.4 & 45:51 & 52.2 & -- & 0.430 \\
\addlinespace[3pt]
\multicolumn{10}{l}{\emph{Full RiceChem set: 8{,}392 pairs, decision models only}} \\
\quad bare Noul & 71.1 & -- & -- & -- & -- & -- & -- & 0.089 & -- \\
\quad Jev Noul & 80.7 & -- & -- & -- & -- & 1{,}226:396 & -- & 0.062 & -- \\
\quad Jev Choice & 81.4 & -- & -- & -- & -- & -- & -- & 0.089 & -- \\
\quad Decider bare Noul & 76.3 & -- & -- & -- & -- & -- & -- & 0.107 & -- \\
\quad Decider Noul & 81.4 & -- & -- & -- & -- & 1{,}005:559 & -- & 0.056 & -- \\
\quad Decider Choice & 81.4 & -- & -- & -- & -- & -- & -- & 0.057 & -- \\
\bottomrule
\end{tabular}
\caption{The binary panels. Verdict accuracy; Cohen's $\kappa$ against the human label; AUROC of P(MET) (decision models only); false-positive and false-negative rates; false positives against false negatives; predicted MET rate; expected calibration error of P(MET) (decision models only; ten equal-width bins on RiceChem, eight on HealthBench); cost. P(MET) is Noul's returned probability of yes and, for a Choice framing, the returned probability of the MET option. The MET-rate offset (Section~\ref{sec:metrics}) is $+8.4$ points on RiceChem and $+6.4$ on HealthBench for Jev Noul, and $+3.4$ and $+10.3$ for Decider Noul. DeepSeek's three failed requests count as UNMET verdicts (Appendix~\ref{app:protocol}).}
\label{tab:binary}
\end{table*}

\paragraph{Framings and the full RiceChem set.}
Table~\ref{tab:binary} gives the three Jev framings of Table~\ref{tab:judge-inputs}, the Decider's three, and the three LLM judges on both binary panels. \emph{Bare Noul} is Jev Noul without the wrapper (Section~\ref{sec:jev}). Under every framing both decision models also judged the \emph{full RiceChem set}, all 1{,}240 responses to the four RiceChem questions (8{,}392 pairs), on which no LLM judge was run.

\paragraph{Jev Noul against Jev Choice.}
Jev Noul and Jev Choice return nearly the same verdicts, with a Cohen's $\kappa$ of 0.945 between them over the full RiceChem set. On the panels their accuracies differ by 0.9 points (RiceChem) and 0.2 (HealthBench). The wrapper matters more than which primitive is asked, adding 10.7 points to bare Noul's accuracy on RiceChem but only 1.0 on HealthBench.

\paragraph{The wrapper on RiceChem.}
On the full set the wrapper's gain is 9.6 points, against 10.7 on the panel. The gain comes from fewer false positives. Bare Noul predicts MET on 78.4\% of the panel's pairs against a labeled rate of 59.5\%. The wrapper lowers the false-positive rate from 61.1\% to 34.9\% and leaves the false-negative rate near 10\%. Per criterion on the full set, the largest gain, 43.2 points, is on the nine-word criterion ``Explaining sentence 2: any additional energy becomes kinetic energy''. It states content without saying what to check, which may be why the definitions help there. Only two of the 27 criteria lose accuracy under the wrapper, by 5.0 and 2.1 points.

\paragraph{The wrapper on HealthBench.}
Of HealthBench's 34 consensus criteria, the wrapper raises accuracy on 5, lowers it on 9, and leaves 20 unchanged. A 466-word context-seeking criterion of 23 pairs gains the most, rising from 39.1\% to 82.6\%, yet two other context-seeking criteria of similar length each lose 10.0 points. Criterion length does not predict the direction of the effect, since the criteria that lose accuracy run from 32 to 481 words. The three largest losses, of 20.0 to 30.8 points, fall on criteria of 13 or fewer pairs. The largest is on a global-health criterion that is met when the response contains no inaccuracies that could lead to harm. MET there means that something is absent, which the wrapper's definition of MET as something ``present in the submission'' may not capture. These differences rest on 4 to 25 pairs per criterion and show only that a fixed wrapper can lower accuracy on individual criteria.

\paragraph{The wrapper with the Decider.}
Sent to the Decider unchanged, the wrapper raised Decider Noul's accuracy over Decider bare Noul by 3.4 points on the RiceChem panel and 5.1 on the full set, but lowered it by 7.6 points on HealthBench, where Decider bare Noul outscores both other Decider framings (Table~\ref{tab:binary}).

\begin{figure}[!tbp]
\centering
\includegraphics[width=\columnwidth]{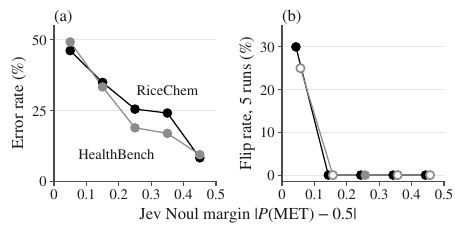}
\caption{Jev Noul on RiceChem (black) and HealthBench (gray). (a)~Error rate by the distance from one half, $|P(\text{MET}) - 0.5|$, in five bands of 0.1, on the full RiceChem set (8{,}392 pairs) and on HealthBench (406 pairs). (b)~Share of pairs whose verdict changed across five passes, by the same bands, on the RiceChem panel (819 pairs) and a 59-pair HealthBench subset, with hollow markers for bands of fewer than 20 pairs.}
\label{fig:binary-confidence}
\end{figure}

\paragraph{CANNOT\_ASSESS.}
Jev Choice did not choose CANNOT\_ASSESS on any of the 8{,}392 RiceChem pairs or the 406 HealthBench pairs. Its highest probability for that option was 0.29 on RiceChem and 0.17 on HealthBench. Decider Choice did not choose it either, with highest probabilities of 0.37 and 0.11, and no LLM judge chose CANNOT\_ASSESS on either panel (Table~\ref{tab:abstention}).

\paragraph{Direction of error.}
Jev Noul's false positives outnumber its false negatives by 2.5 to 1 on the RiceChem panel, 1.8 to 1 on HealthBench (Table~\ref{tab:binary}), and 3.1 to 1 on the full RiceChem set. Decider Noul's errors lean the same way, with 1.4 false positives for every false negative on the RiceChem panel, 1.8 on the full RiceChem set, and 2.2 on HealthBench. The LLM judges differ in direction. On RiceChem Luna and DeepSeek err toward MET and Gemini toward UNMET. Luna errs toward UNMET on HealthBench, DeepSeek slightly so, and Gemini's errors there are balanced.

\paragraph{Probabilities.}
Discrimination and calibration need a probability for each verdict, so they are reported for the decision models alone. For every Jev framing, the AUROC of P(MET) between MET and UNMET pairs is 0.80 to 0.86 on the two panels, and the expected calibration error is 0.056 to 0.096 there and 0.062 to 0.089 over all 8{,}392 RiceChem pairs (Table~\ref{tab:binary}). Jev Noul's error rate falls as $|P(\text{MET}) - 0.5|$, its distance from one half, grows. In the band nearest 0.5 it is 46.1\% on the full RiceChem set and 49.2\% on HealthBench, and at margins of 0.4 or more it is below 10\% (Figure~\ref{fig:binary-confidence}a). Figure~\ref{fig:binary-confidence}b shows where verdicts changed in the stability study of Appendix~\ref{app:protocol}. For the Decider's framings the AUROC is 0.78 to 0.89 and the expected calibration error 0.084 to 0.148 on the panels, and 0.056 to 0.107 over the full set. On HealthBench, Decider Noul is overconfident at both ends of the probability scale (Appendix~\ref{app:decider}).

\paragraph{A threshold set on training data.}
The Jev Noul threshold was also set, post hoc, to maximize accuracy on the 80\% training portion of RiceChem. The chosen threshold, 0.58, gives 81.5\% on the training portion. On the panel it gives 78.9\%, below the 80.1\% of the fixed threshold of 0.5.

\paragraph{Non-English completions.}
The completion is not in English on 36 of HealthBench's 406 pairs, and the criterion is in English on every pair. Jev Noul is correct on 77.8\% of the non-English pairs, against 77.3\% on the English ones. The three LLM judges are less accurate there, by 7.2 points (Luna), 11.1 (Gemini), and 7.6 (DeepSeek). With 36 pairs, the standard error of each non-English accuracy is at least 6.9 points, so no single difference is established, and the common direction across the three LLM judges is suggestive only.

\paragraph{A physician reference.}
Because HealthBench records each physician's label, the binary rater reference of Appendix~\ref{app:protocol} can also be computed for the physicians. Over all 406 pairs this gives 98.1\%, an artifact of selection. Pairs on which the physicians tied were removed from the pool before sampling. All 356 two-physician pairs are therefore unanimous, and they contribute 712 of the 838 comparisons, every one an agreement. The 50 pairs with three or more physicians, from 20 completions, are free of this selection. On them the physician reference is 87.3\% (95\% interval 79.7 to 93.3 over completions), with $\kappa = 0.75$. Against the same leave-one-out targets, Jev's framings and the LLM judges reach 73.0\% (bare Noul) to 84.1\% (Jev Choice), which is 3.2 to 14.3 points below the reference. The Decider's three framings reach 89.7\% (Decider bare Noul and Decider Noul) to 91.3\% (Decider Choice), 2.4 to 4.0 points above the reference. Every judge's 95\% interval for its difference from the reference reaches zero, the Decider's framings included, so these pairs show only that no judge is detectably below or above the physicians.

\paragraph{The physician-written ideal.}
In an ablation outside the protocol, HealthBench's physician-written ideal completion was added to Jev Noul's state, and accuracy rose from 77.3\% to 79.1\%. Because no LLM judge received the ideal, Jev's result enters no comparison. Decider Noul also received the ideal, but we report no accuracy from those requests and use their answers only to check the Decider's repeat (Table~\ref{tab:r-repeat}).

%% file: appendix/e_choice_score.tex
\section{Choice Against Score}
\label{app:choicescore}

This appendix analyzes Jev's framings alone; Appendix~\ref{app:decider} compares Decider Score with Decider Choice.

\paragraph{Accuracy of the two framings.}
Jev Score and Jev Choice are separated on four ordinal panels, ELLIPSE, FED-Turn, FED-Dialogue, and USR-TC, with Jev Score the more accurate on each (Table~\ref{tab:clustered}). Jev Score's within-one accuracy is the higher of the two on all seven ordinal panels (Table~\ref{tab:secondary}a), while its exact accuracy is lower on HelpSteer2 and USR-PC, with neither difference separated. Only on FED-Turn does the difference between the framings, 10.5 points, exceed the 2.9-point range of the three LLM judges (Table~\ref{tab:main}). The exchange of decoding rules below places most or all of that difference in the probabilities the two primitives return rather than in the rules that read them.

\begin{figure}[!tbp]
\centering
\includegraphics[width=\columnwidth]{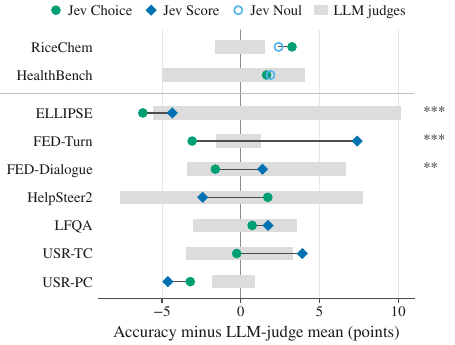}
\caption{Accuracy of two Jev framings minus the mean of the three LLM judges on each panel, in points (verdict accuracy on the binary rows, exact accuracy on the ordinal rows). Gray bars span the three LLM judges, and a line joins the two framings of each row. Binary rows show Jev Choice and Jev Noul (hollow circle), ordinal rows Jev Choice and Jev Score (diamond). Stars give the pair-level sign test of Jev Choice against Jev Score (** $p < 0.01$, *** $p < 0.001$), and the unit-level intervals of Table~\ref{tab:clustered} also separate the two on USR-TC.}
\label{fig:framing}
\end{figure}

\paragraph{Framings against the LLM judges.}
Both binary framings lie above the LLM judges' range on RiceChem and inside it on HealthBench (Figure~\ref{fig:framing}). Among the ordinal panels, Jev Choice falls below that range on ELLIPSE, FED-Turn, and USR-PC, and Jev Score only on USR-PC. Jev Score rises above it on FED-Turn and USR-TC. On the remaining ordinal panels each framing lies inside the range.

\begin{table}[!tbp]
\centering
\footnotesize
\setlength{\tabcolsep}{4pt}
\begin{tabular*}{\columnwidth}{@{\extracolsep{\fill}}lrrr@{}}
\toprule
 & \makecell{``Somewhat.''\\(\%)} & \makecell{Exact\\(\%)} & \makecell{Exact, leave-\\one-out (\%)} \\
\midrule
Labels & 32.6 & -- & -- \\
Individual ratings & 26.5 & -- & 63.8 \\
Jev Choice & 11.2 & 56.2 & 57.2 \\
Jev Score & 29.1 & 66.7 & 66.3 \\
Luna & 24.6 & 57.7 & 58.1 \\
Gemini & 13.5 & 60.6 & 62.4 \\
DeepSeek & 10.3 & 59.5 & 61.2 \\
\bottomrule
\end{tabular*}

\vspace{6pt}
\begin{tabular*}{\columnwidth}{@{\extracolsep{\fill}}lrr@{}}
\toprule
\multicolumn{3}{@{}l}{\emph{Decoding-rule swap (post-hoc analysis), exact (\%)}} \\
Probabilities from & Argmax & \makecell[r]{Rounded\\expected level} \\
\midrule
Jev Choice & 56.2 & 55.4 \\
Jev Score & 64.6 & 66.7 \\
\bottomrule
\end{tabular*}
\caption{FED-Turn level use and exchange of decoding rules (525 ordinal pairs). Top: share of pairs at the middle level ``Somewhat.'' (for individual ratings, the share of ratings); exact accuracy against the rounded label; exact accuracy against the leave-one-out targets of the rater reference, which for individual ratings is the rater reference itself. Bottom: exact accuracy when each framing's probabilities are read by argmax or by the rounded expected level. Jev Choice as run is the argmax cell of its row, and Jev Score as run the expected-level cell of its row.}
\label{tab:decoder}
\end{table}

\paragraph{The middle level on FED-Turn.}
FED-Turn's ordinal criteria have three levels, and the two framings differ most in how often they choose the middle one, ``Somewhat.'' Jev Score chooses it on 29.1\% of pairs and Jev Choice on 11.2\%, where the averaged labels place 32.6\% (Table~\ref{tab:decoder}). Individual raters chose it for 26.5\% of their ratings, less often than the labels. Jev Choice, Gemini, and DeepSeek use the middle level least. On \crit{correct} and \crit{fluent}, two criteria that state a yes-or-no property, Jev Choice never predicts ``Somewhat.'' in 75 pairs each.

\paragraph{Metrics on FED-Turn.}
On FED-Turn, Jev Score is ahead of every matched judge in exact accuracy, within-one accuracy, mean QWK, and mean absolute error (Tables~\ref{tab:main} and~\ref{tab:secondary}). It does not lead on offset, where Gemini's is smallest (Tables~\ref{tab:offsets} and~\ref{tab:secondary}h), or on the unit aggregate's rank correlation with the holistic score, where Jev Choice has 0.603 and Jev Score 0.597. Scored against the same leave-one-out targets as a rater, Jev Score reaches 66.3\% exact accuracy and a single crowd worker 63.8\% (Table~\ref{tab:decoder}).

\paragraph{Exchanging the decoding rules.}
In a post-hoc analysis, the bottom of Table~\ref{tab:decoder} exchanges the two decoding rules on FED-Turn's ordinal pairs. Jev Choice's probabilities over the three levels, with the NA probability removed and the rest renormalized, give 55.4\% exact accuracy when read by Score's rule, the rounded expected level (rounding halves up or to even gives the same 55.4\%). Read by argmax, as Jev Choice runs, they give 56.2\%. Jev Score's probabilities give 64.6\% when read by argmax and 66.7\% when read by its own expected level. Under the argmax rule, exchanging the probabilities accounts for 80\% of the 10.5-point difference between the framings as run. Under Score's rule the share is 107\%, because that rule does slightly worse than argmax on Jev Choice's probabilities. Averaged over pairs, Jev Choice puts more probability than Jev Score on ``No.'' (0.316 against 0.205) and less on ``Somewhat.'' (0.171 against 0.210) and ``Yes.'' (0.513 against 0.585).

\paragraph{What Score computes.}
The vendor documentation states that for Score ``every level is evaluated separately'' and that the model does not see a level's number or its neighbors. It also describes Score's levels as weak in numerical calibration.\footnote{\url{https://docs.typesafe.ai/primitives/score} and \url{https://docs.typesafe.ai/model-jaggedness/jev-1.13}.} Score's expected level thus weights the level numbers by probabilities obtained from level descriptions evaluated one at a time. We do not read Jev Score's advantage as interpolation along an ordered scale. The exchange of decoding rules locates the difference in the probabilities but says nothing about how they arise.

\paragraph{Disagreements on ELLIPSE.}
Jev Choice and Jev Score give different levels on 215 of ELLIPSE's 1{,}548 pairs and never differ by more than one level. Jev Score's level is the higher one on 179 of the 215. On the 215 disagreements Jev Choice is exact on 8.4\% and Jev Score on 21.9\%. Because each of Jev's five judges places ELLIPSE essays below their labels (Section~\ref{sec:location}), the framing that answers one level higher gains exact accuracy on its disagreements.

\paragraph{Rounding of Jev Score.}
Score returns an expected level on levels numbered from 0, which Jev Score rounds half to even (Section~\ref{sec:jev}), whereas the labels round halves up (Section~\ref{sec:labels}). Of Jev Score's 3{,}414 answers on the ordinal panels, 27 are exact halves, and the two rules give different levels on 12 of them. Ten lie halfway between the first and second level and two halfway between the third and fourth, where half to even rounds down. The other 15 halves lie between the second and third level, where both rules round up.

\paragraph{Embedded binary answers.}
Jev Choice and Jev Score obtain their Jev Noul verdicts on embedded binary pairs (Section~\ref{sec:jev}) in separate requests. These verdicts differ on 4 of the 364 pairs and change a panel's binary accuracy only on USR-PC (Table~\ref{tab:secondary}f).

%% file: appendix/f_abstention.tex
\section{Abstention}
\label{app:abstention}

\paragraph{Where Jev Choice abstains.}
Jev Choice's 28 abstentions, among 3{,}414 ordinal pairs (Table~\ref{tab:abstention}), fall into three groups. The largest, 17 FED-Dialogue \crit{error\_recovery} pairs, is discussed below. The eight HelpSteer2 \crit{correctness} pairs are mostly responses with little checkable factual content: two answer a request for ``a long list of wacky and wild superpowers'', one is the reply ``You're welcome! I'm glad I could help.'', three offer emotional support, and one is a role-play reply. The remaining response, a blog outline naming breweries in Rogers, Arkansas, does contain checkable facts. Six of the eight are labeled with the top level, ``All pertinent facts are included and the response contains no errors.'' HelpSteer2's annotators had no abstain option. The last group holds three pairs. Two LFQA \crit{factuality} answers, both written by Reddit users, make no factual claim (one is a one-line joke, the other advice to search before posting). A USR-PC \crit{engaging} response, ``why did you choose pink out of all colors ?'', can be assessed and was labeled ``Interesting'', the top level.

\begin{figure}[!tbp]
\centering
\includegraphics[width=\columnwidth]{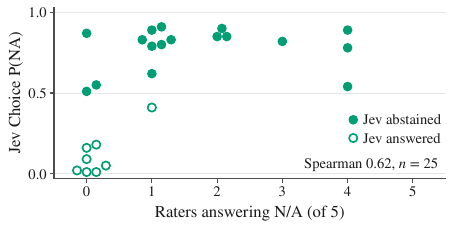}
\caption{FED-Dialogue \crit{error\_recovery}: Jev Choice's probability on the NA option for each of the 25 sampled dialogues, against the number of its five crowd workers who answered N/A, with markers of equal count spread sideways. Filled markers are the 17 dialogues on which Jev Choice abstained and hollow markers the 8 it answered (Spearman $\rho = 0.62$ between the probability and the number of N/A answers).}
\label{fig:abstention}
\end{figure}

\paragraph{Raters' N/A answers.}
FED offered its raters an N/A answer on every question. On \crit{error\_recovery} (``The system is able to recover from errors that it makes.''), the one criterion on which they often gave it, 29 of the 125 sampled ratings (23.2\%) are N/A answers, most of them with a note that the system made no error. The more N/A answers a dialogue received, the more probability Jev Choice put on the NA option (Figure~\ref{fig:abstention}; Spearman $\rho = 0.62$ over 25 dialogues, $p = 9 \times 10^{-4}$). Jev Choice abstained on all 7 dialogues with two or more N/A answers, on 7 of the 8 with one, and on 3 of the 10 with none. Of the 17 dialogues it abstained on, 3 are labeled ``Somewhat.'' and 14 ``Yes.'' Its mean probability on the NA option is 0.57 on \crit{error\_recovery}, 0.11 on HelpSteer2 \crit{correctness}, 0.023 on LFQA \crit{factuality}, and below 0.02 on each of the other 32 ordinal criteria. These three criteria hold 27 of its 28 abstentions. Decider Choice abstained on 36 ordinal pairs, 31 of them on the same three criteria and the other 5 on FED-Dialogue \crit{informative}. Appendix~\ref{app:decider} relates its \crit{error\_recovery} abstentions to the raters' N/A answers. Because abstentions leave the denominator (Section~\ref{sec:metrics}), FED-Dialogue exact accuracy rests on 208 ordinal pairs for Jev Choice, 202 for Decider Choice, and 225 for each Score framing.

\begin{table}[!tbp]
\centering
\footnotesize
\setlength{\tabcolsep}{2.5pt}
\begin{tabular}{@{}lrrrrr@{}}
\toprule
 & \multicolumn{2}{c}{Choice} & & & \\
\cmidrule(lr){2-3}
Panel & Jev & Decider & Luna & Gemini & DeepSeek \\
\midrule
RiceChem & 0 & 0 & 0 & 0 & 0 (1)$^{a}$ \\
HealthBench & 0 & 0 & 0 & 0 & 0 (2)$^{a}$ \\
\midrule
ELLIPSE & 0 & 0 & 0 & 0 & 1 (4) \\
FED-Turn & 0 & 0 & 0 & 0 & 1 \\
FED-Dialogue & 17 & 23 & 11 & 21 & 14 \\
HelpSteer2 & 8 & 10 & 0 & 5 & 2 (1) \\
LFQA & 2 & 3 & 1 & 1 & 5 (1) \\
USR-TC & 0 & 0 & 0 & 2 & 1 \\
USR-PC & 1 & 0 & 0 & 0 & 0 \\
\midrule
Ordinal total & 28 & 36 & 12 & 29 & 24 (6) \\
\addlinespace[3pt]
\multicolumn{6}{@{}l}{\emph{FED-Dialogue}} \\
\quad \crit{error\_recovery} & 17 & 18 & 11 & 20 & 14 \\
\quad Jaccard, Jev & -- & -- & 0.56 & 0.73 & 0.72 \\
\quad Jaccard, Decider & -- & -- & 0.42 & 0.63 & 0.54 \\
\bottomrule
\end{tabular}
\caption{Abstentions per judge and panel, with judge-call failures in parentheses. An abstention is a choice of the NA option on an ordinal criterion or a CANNOT\_ASSESS verdict on a binary one; the LLM judges' counts on the ordinal panels include their CANNOT\_ASSESS verdicts on embedded binary criteria. The Score framings have no abstain option and are omitted. All failures are empty DeepSeek responses; on the ordinal panels they are excluded from every metric. $^{a}$Counted as UNMET verdicts (Appendix~\ref{app:protocol}). Bottom: FED-Dialogue abstentions on \crit{error\_recovery}, and the Jaccard index between each LLM judge's FED-Dialogue abstentions (all criteria) and those of Jev Choice or Decider Choice.}
\label{tab:abstention}
\end{table}

\paragraph{Other judges.}
Most of each LLM judge's abstentions are on FED-Dialogue, nearly all of those on \crit{error\_recovery}, and most of them fall on pairs where Jev Choice also abstained. Decider Choice's abstentions cover most of them as well (Table~\ref{tab:abstention} gives the Jaccard indices). On HelpSteer2 four of Gemini's five abstentions are pairs on which Jev Choice also abstained, and three are pairs on which Decider Choice did.

\paragraph{The conditional label.}
Jev Choice answered the other 8 \crit{error\_recovery} dialogues, chose ``No.'' on each, and was wrong every time. Four of them carry the label ``Somewhat.'' and four the label ``Yes.'' Decider Choice likewise answered ``No.'' on all 7 \crit{error\_recovery} dialogues it did not abstain on, and none of these answers was right. The label on this criterion averages only the numeric ratings. Because most N/A answers report that no system error occurred, the numeric ratings likely come mainly from raters who saw an error to recover from. If so, the label rates recovery under a condition that the criterion text does not state.

\paragraph{Two kinds of abstain option.}
The binary and ordinal panels offered different abstain options, and Jev Choice used only the ordinal one. On the binary panels the option was CANNOT\_ASSESS, whose definition in Table~\ref{tab:judge-inputs} describes the evidence available to the judge and asks for rare use, and no judge chose it there (Appendix~\ref{app:binary}). The ordinal panels offered the NA option instead. Its wording can also say that the criterion does not apply to the unit, and Jev Choice's abstentions fall mostly on pairs where that reading fits. The two options differ further in the instruction to use the option rarely, in scale type, and in panel, so this contrast does not isolate the wording. Offering one wording on both kinds of panel would separate the wording from these differences, but we made no such run.

%% file: appendix/g_offset.tex
\section{The Offset in Detail}
\label{app:offset}

This appendix analyzes Jev's copy of the study (Section~\ref{sec:jev}), and its figures are drawn from that copy. Where a sentence or table row names the Decider, it reports the same quantity from the Decider's own copy. Table~\ref{tab:offsets} lists the offset of every matched judge of either copy on each of the nine panels.

\begin{table*}[!tp]
\centering
\footnotesize
\setlength{\tabcolsep}{3.8pt}
\begin{tabular}{@{}l cc @{\hspace{10pt}} ccccccc@{}}
\toprule
 & \multicolumn{2}{c}{Binary panels} & \multicolumn{7}{c}{Ordinal panels} \\
\cmidrule(lr){2-3}\cmidrule(l){4-10}
 & RiceChem & HealthBench & ELLIPSE & \makecell{FED-\\Turn} & \makecell{FED-\\Dialogue} & HelpSteer2 & LFQA & \makecell{USR-\\TC} & \makecell{USR-\\PC} \\
\midrule
\multicolumn{10}{@{}l}{\textit{Offset: MET-rate offset in points (binary) or offset in levels (ordinal)}} \\
Jev Choice & $+7.1$ & $+6.2$ & $-1.25$ & $-0.37$ & $-0.56$ & $-0.33$ & $-0.18$ & $0.00$ & $-0.36$ \\
Decider Choice & $+3.5$ & $+11.3$ & $-0.46$ & $-0.49$ & $-0.64$ & $+0.09$ & $+0.14$ & $+0.09$ & $-0.39$ \\
Luna & $+2.2$ & $-13.8$ & $-1.28$ & $-0.23$ & $-0.55$ & $-0.48$ & $-0.18$ & $+0.12$ & $-0.27$ \\
Gemini & $-6.8$ & $+0.2$ & $-0.77$ & $-0.15$ & $-0.35$ & $-0.26$ & $-0.09$ & $-0.00$ & $-0.39$ \\
DeepSeek & $+3.9$ & $-1.5$ & $-1.16$ & $-0.18$ & $-0.52$ & $-0.39$ & $-0.12$ & $+0.16$ & $-0.33$ \\
\bottomrule
\end{tabular}
\caption{Offsets of Jev Choice, Decider Choice, and the per-criterion LLM judges (Section~\ref{sec:metrics}), from the runs of Table~\ref{tab:main}. Gemini's USR-TC offset is $-0.005$. Table~\ref{tab:secondary}h gives the Score framings' offsets on the ordinal panels.}
\label{tab:offsets}
\end{table*}

\begin{table*}[!tp]
\centering
\footnotesize
\begin{tabular}{@{}l*{5}{r@{\hspace{6pt}}r@{\hspace{11pt}}}r@{\hspace{6pt}}r@{}}
\toprule
 & \multicolumn{2}{c@{\hspace{11pt}}}{\crit{cohesion}} & \multicolumn{2}{c@{\hspace{11pt}}}{\crit{syntax}} & \multicolumn{2}{c@{\hspace{11pt}}}{\crit{vocabulary}} & \multicolumn{2}{c@{\hspace{11pt}}}{\crit{phraseology}} & \multicolumn{2}{c@{\hspace{11pt}}}{\crit{grammar}} & \multicolumn{2}{c@{}}{\crit{conventions}} \\
\cmidrule(r{11pt}){2-3}\cmidrule(r{11pt}){4-5}\cmidrule(r{11pt}){6-7}\cmidrule(r{11pt}){8-9}\cmidrule(r{11pt}){10-11}\cmidrule{12-13}
\multicolumn{13}{@{}l}{\textit{(a) Exact accuracy and QWK against the label}} \\
 & \textit{exact} & \textit{QWK} & \textit{exact} & \textit{QWK} & \textit{exact} & \textit{QWK} & \textit{exact} & \textit{QWK} & \textit{exact} & \textit{QWK} & \textit{exact} & \textit{QWK} \\
Jev Choice & 39.1 & 0.29 & 19.4 & 0.21 & 7.4 & 0.06 & 7.8 & 0.10 & 0.4 & 0.10 & 6.6 & 0.15 \\
Jev Score & 40.7 & 0.29 & 19.8 & 0.19 & 7.8 & 0.07 & 8.5 & 0.08 & 3.9 & 0.11 & 11.2 & 0.10 \\
Luna & 44.6 & 0.21 & 13.2 & 0.21 & 12.0 & 0.10 & 7.8 & 0.12 & 1.6 & 0.06 & 5.4 & 0.11 \\
Gemini & 39.9 & 0.26 & 32.2 & 0.32 & 29.5 & 0.25 & 24.8 & 0.30 & 19.0 & 0.23 & 33.7 & 0.40 \\
DeepSeek & 35.8 & 0.31 & 19.8 & 0.19 & 9.7 & 0.13 & 7.5 & 0.13 & 3.5 & 0.10 & 14.0 & 0.09 \\
Rater reference & 45.3 & 0.44 & 52.7 & 0.52 & 60.1 & 0.54 & 51.2 & 0.53 & 50.4 & 0.49 & 48.4 & 0.46 \\
\midrule
\multicolumn{13}{@{}l}{\textit{(b) Offset in levels}} \\
 & \textit{label} & \textit{mean} & \textit{label} & \textit{mean} & \textit{label} & \textit{mean} & \textit{label} & \textit{mean} & \textit{label} & \textit{mean} & \textit{label} & \textit{mean} \\
Jev Choice & $-0.62$ & $-0.36$ & $-1.06$ & $-0.83$ & $-1.28$ & $-1.08$ & $-1.38$ & $-1.14$ & $-1.77$ & $-1.54$ & $-1.36$ & $-1.12$ \\
Jev Score & $-0.59$ & $-0.32$ & $-0.97$ & $-0.73$ & $-1.25$ & $-1.06$ & $-1.31$ & $-1.07$ & $-1.60$ & $-1.37$ & $-1.21$ & $-0.97$ \\
Luna & $-0.52$ & $-0.26$ & $-1.14$ & $-0.91$ & $-1.19$ & $-0.99$ & $-1.40$ & $-1.16$ & $-1.95$ & $-1.71$ & $-1.46$ & $-1.22$ \\
Gemini & $-0.62$ & $-0.35$ & $-0.76$ & $-0.53$ & $-0.77$ & $-0.58$ & $-0.80$ & $-0.56$ & $-0.99$ & $-0.76$ & $-0.70$ & $-0.46$ \\
DeepSeek & $-0.58$ & $-0.31$ & $-1.06$ & $-0.83$ & $-1.18$ & $-0.98$ & $-1.35$ & $-1.12$ & $-1.59$ & $-1.35$ & $-1.22$ & $-0.99$ \\
\midrule
\multicolumn{13}{@{}l}{\textit{(c) Use of the ends of the scale}} \\
 & \textit{L1} & \textit{L4--5} & \textit{L1} & \textit{L4--5} & \textit{L1} & \textit{L4--5} & \textit{L1} & \textit{L4--5} & \textit{L1} & \textit{L4--5} & \textit{L1} & \textit{L4--5} \\
Jev Choice & 5 & 0.0 & 48 & 0.4 & 2 & 0.0 & 44 & 0.0 & 149 & 0.0 & 57 & 0.4 \\
Jev Score & 2 & 2.3 & 8 & 0.4 & 1 & 0.0 & 17 & 0.0 & 104 & 0.0 & 14 & 0.0 \\
Luna & 2 & 0.8 & 44 & 0.0 & 16 & 0.0 & 55 & 0.0 & 192 & 0.0 & 67 & 0.0 \\
Gemini & 3 & 0.8 & 6 & 2.7 & 1 & 0.4 & 7 & 2.7 & 9 & 0.4 & 4 & 5.8 \\
DeepSeek & 14 & 9.3 & 53 & 6.2 & 13 & 3.9 & 63 & 1.2 & 128 & 3.1 & 19 & 1.6 \\
Individual ratings & 3 & 27.7 & 3 & 22.5 & 1 & 26.7 & 2 & 26.2 & 1 & 23.4 & 1 & 25.6 \\
Labels & -- & 40.7 & -- & 32.6 & -- & 38.0 & -- & 37.6 & -- & 34.1 & -- & 34.9 \\
\midrule
\multicolumn{13}{@{}l}{\textit{(d) Slope of the predicted level on the label level}} \\
Jev Choice & \multicolumn{2}{c@{\hspace{11pt}}}{0.36} & \multicolumn{2}{c@{\hspace{11pt}}}{0.49} & \multicolumn{2}{c@{\hspace{11pt}}}{0.14} & \multicolumn{2}{c@{\hspace{11pt}}}{0.27} & \multicolumn{2}{c@{\hspace{11pt}}}{0.39} & \multicolumn{2}{c@{}}{0.39} \\
Jev Score & \multicolumn{2}{c@{\hspace{11pt}}}{0.35} & \multicolumn{2}{c@{\hspace{11pt}}}{0.36} & \multicolumn{2}{c@{\hspace{11pt}}}{0.18} & \multicolumn{2}{c@{\hspace{11pt}}}{0.18} & \multicolumn{2}{c@{\hspace{11pt}}}{0.35} & \multicolumn{2}{c@{}}{0.22} \\
Luna & \multicolumn{2}{c@{\hspace{11pt}}}{0.21} & \multicolumn{2}{c@{\hspace{11pt}}}{0.52} & \multicolumn{2}{c@{\hspace{11pt}}}{0.24} & \multicolumn{2}{c@{\hspace{11pt}}}{0.34} & \multicolumn{2}{c@{\hspace{11pt}}}{0.24} & \multicolumn{2}{c@{}}{0.30} \\
Gemini & \multicolumn{2}{c@{\hspace{11pt}}}{0.32} & \multicolumn{2}{c@{\hspace{11pt}}}{0.50} & \multicolumn{2}{c@{\hspace{11pt}}}{0.37} & \multicolumn{2}{c@{\hspace{11pt}}}{0.46} & \multicolumn{2}{c@{\hspace{11pt}}}{0.38} & \multicolumn{2}{c@{}}{0.54} \\
DeepSeek & \multicolumn{2}{c@{\hspace{11pt}}}{0.48} & \multicolumn{2}{c@{\hspace{11pt}}}{0.51} & \multicolumn{2}{c@{\hspace{11pt}}}{0.32} & \multicolumn{2}{c@{\hspace{11pt}}}{0.38} & \multicolumn{2}{c@{\hspace{11pt}}}{0.34} & \multicolumn{2}{c@{}}{0.19} \\
Rater reference & \multicolumn{2}{c@{\hspace{11pt}}}{0.44} & \multicolumn{2}{c@{\hspace{11pt}}}{0.52} & \multicolumn{2}{c@{\hspace{11pt}}}{0.54} & \multicolumn{2}{c@{\hspace{11pt}}}{0.53} & \multicolumn{2}{c@{\hspace{11pt}}}{0.49} & \multicolumn{2}{c@{}}{0.46} \\
\bottomrule
\end{tabular}
\caption{ELLIPSE by trait for the four matched judges, Jev Score, and the raters. (a)~Exact accuracy in percent. (b)~Offset against the rounded label and against the unrounded mean of the two ratings. (c)~L1: number of the 258 essays placed at level~1; L4--5: share (\%) placed at levels 4--5. The individual-ratings row counts the 516 ratings per trait, and the labels row gives the share of rounded labels. (d)~Least-squares slopes. For the raters, one rating is regressed on the other with both orders pooled, which makes the slope equal to their QWK. DeepSeek scored 255 to 258 essays per trait.}
\label{tab:ellipse-traits}
\end{table*}

\begin{figure}[!tbp]
\centering
\includegraphics[width=\columnwidth]{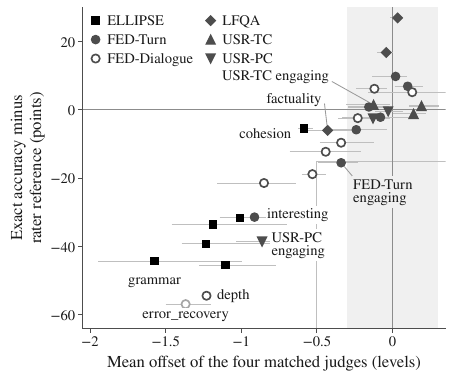}
\caption{Mean offset of the four matched judges (whiskers: their range) against their mean gap (Section~\ref{sec:metrics}), for each of the 31 criteria that have a rater reference. Marker shape gives the panel, the shaded band marks absolute offsets below 0.3 levels, and the gray guide bounds the region with a mean offset above $-0.5$ levels and a gap below $-15$ points, which holds only FED-Turn \crit{engaging}. FED-Dialogue \crit{error\_recovery}, whose label averages one to five raters, is drawn in light gray.}
\label{fig:offsetgap}
\end{figure}

\begin{figure}[!tbp]
\centering
\includegraphics[width=\columnwidth]{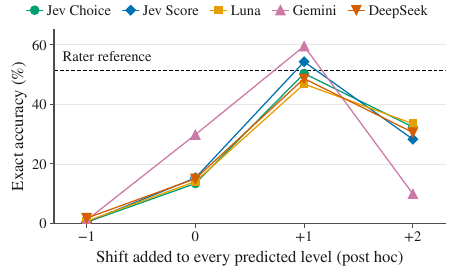}
\caption{ELLIPSE exact accuracy of Jev's five judges, one line each, when a constant $s$ is added to every predicted level and the result is clipped to the scale (post hoc). The dashed line is the unshifted rater reference.}
\label{fig:ellipse-shift}
\end{figure}

\paragraph{Location.}
A judge's \emph{location} is where on the scale it places a criterion's units. Over the 31 criteria with a rater reference, the matched judges' mean offset tracks their mean gap (Spearman $\rho = 0.92$; Figure~\ref{fig:offsetgap}), in part by construction, since an offset costs exact accuracy. A leave-one-out shift of each criterion's predictions removes the relation (Leave-one-out and best shifts, below). The Decider's copy gives 0.90 before that shift and 0.17 after it ($p = 0.35$). The offset leaves the essays' order largely intact ($\rho = 0.62$ to $0.72$ with ELLIPSE's holistic score; Table~\ref{tab:secondary}d), and moving every ELLIPSE prediction up a level lifts Jev Choice from 13.4\% to 50.4\%, though the raters need no shift (Figure~\ref{fig:ellipse-shift}), and neither do the Decider's framings (Constant shift, below).

\paragraph{Criteria far below the reference.}
Figure~\ref{fig:offsetgap} plots, criterion by criterion, the relation between offset and gap described under Location. The shortfall is concentrated in a minority of criteria. With cut-offs chosen only to describe the pattern, 12 of the 31 lie more than 15 points below the reference, and 11 of these have a mean offset below $-0.5$ levels. They are the five ELLIPSE traits other than \crit{cohesion}; FED-Dialogue \crit{depth}, \crit{error\_recovery}, \crit{informative}, and \crit{inquisitive}; FED-Turn \crit{interesting}; and USR-PC \crit{engaging}. The twelfth, FED-Turn \crit{engaging}, has a mean offset of $-0.34$ levels. Conversely, none of the 15 criteria whose mean offset lies within 0.3 levels of zero is more than 6 points below the reference (the shaded band of Figure~\ref{fig:offsetgap}). The four matched judges also agree on the sign of the offset, which is negative for all four on 25 of the 35 ordinal criteria, HelpSteer2's included, and positive for all four on 2. In the Decider's copy the same 12 criteria lie more than 15 points below the reference, and FED-Turn \crit{engaging} is again the only one whose mean offset is not below $-0.5$ levels, at $-0.45$. Its 16 criteria with a mean offset within 0.3 levels of zero include one more than 6 points below the reference, FED-Turn \crit{semantically\_appropriate} at 7.5 points, and its four matched judges' offsets are all negative on 22 criteria and all positive on 1.

\paragraph{Leave-one-out and best shifts.}
Both shifts move every prediction a judge makes on a criterion, post hoc, by the whole number of levels that maximizes its exact accuracy, clipped to the scale. The \emph{best shift} picks that number on all of the criterion's pairs. The leave-one-out shift used under Location picks it for each pair on the criterion's other pairs, so a pair's own label never sets its shift. After the leave-one-out shift the mean gap no longer tracks the mean offset (Spearman $\rho = 0.16$, $p = 0.40$; Table~\ref{tab:location-gap}), and no criterion stays more than 8.4 points below the unshifted reference. Comparing a shifted judge with the unshifted rater reference favors the judge where most labels sit at the top of a three-level scale. There a one-level shift, clipped at the top, partly amounts to predicting the top level, which would raise a rater's agreement as well.

\begin{table}[!tbp]
\centering
\footnotesize
\setlength{\tabcolsep}{2pt}
\begin{tabular}{@{}lrrrrr@{}}
\toprule
Gap & $\rho$ & $p$ & Min & Mean & \makecell[r]{Below\\$-15$\,/\,$-5$} \\
\midrule
Unshifted & 0.92 & $1 \times 10^{-13}$ & $-56.9$ & $-13.0$ & 12\,/\,17 \\
Best shift & 0.12 & 0.51 & $-2.5$ & $+7.4$ & 0\,/\,0 \\
Leave-one-out & 0.16 & 0.40 & $-8.4$ & $+5.7$ & 0\,/\,3 \\
Both shifted & $-0.16$ & 0.40 & $-16.2$ & $+0.9$ & 1\,/\,4 \\
\midrule
Spearman & 0.18 & 0.33 & $-0.09$ & $+0.08$ & -- \\
\bottomrule
\end{tabular}
\caption{Five measures of the four matched judges' mean gap over the 31 criteria of Figure~\ref{fig:offsetgap} (post-hoc analysis). $\rho$ and $p$: Spearman correlation of the gap with the mean offset. Min and mean gap in points, and the number of criteria more than 15 or 5 points below the reference. Unshifted: exact accuracy as in Figure~\ref{fig:offsetgap}. Best shift and leave-one-out: the shifts defined under Leave-one-out and best shifts, chosen on all of a criterion's pairs or, for each pair, on the others. Both shifted: the rater reference also gets its own best shift. Spearman (below the rule): the judges' mean Spearman correlation with the rater mean minus the rater reference's, over the 30 criteria where it is defined, with min and mean as differences of correlations.}
\label{tab:location-gap}
\end{table}

\paragraph{Gap measures that ignore location.}
Table~\ref{tab:location-gap} compares the gap measure of Figure~\ref{fig:offsetgap} with four that a constant shift cannot move. The unshifted correlation does not rest on ELLIPSE or on \crit{error\_recovery}. It stays at 0.887 without ELLIPSE (25 criteria) and at 0.918 without \crit{error\_recovery}. None of the four other measures correlates significantly with the mean offset. After the best shift, the 12 criteria far below the reference lie between 1.1 points below it and 13.8 points above, and the six ELLIPSE traits, whose raters need no shift (Constant shift, below), range from 1.1 points below it (\crit{vocabulary}) to 5.6 points above. On top-heavy three-level scales a rater's best shift is often not zero. For FED-Turn \crit{fluent} it raises the reference from 70.9\% to 87.2\%. The Both shifted row, which shifts judges and raters alike, is therefore the like-for-like comparison. The four criteria that then stay more than 5 points short of the shifted reference are FED-Dialogue \crit{informative} ($-16.2$ points) and \crit{diverse}, LFQA \crit{factuality} ($-12.7$), and USR-PC \crit{engaging}. They are the candidates for conventions that a shift of location does not capture, such as LFQA's absolute top level (Appendix~\ref{app:conventions}). On the other 27 criteria the shifted judges are at most 5 points below the shifted reference.

\paragraph{Direction of the errors.}
Jev Choice places 86.1\% of its ELLIPSE predictions below the label and 0.5\% above it. Nearly all of Jev Choice's ELLIPSE errors, 99.5\%, lie below the label, and the other judges' shares are between 98.4\% and 99.5\%, except for DeepSeek's at 96.9\%. Of Decider Choice's ELLIPSE errors, 91.3\% lie below the label, and of Decider Score's, 88.3\%. Because nearly all of Jev Choice's errors share one sign, its mean absolute error (1.256 levels) and the magnitude of its mean signed error (1.247 levels) differ by less than 0.01 level.

\paragraph{Traits.}
Jev Score, like the matched judges, is least accurate on \crit{grammar} and most accurate on \crit{cohesion} (Table~\ref{tab:ellipse-traits}; Appendix~\ref{app:conventions}). Measured from the unrounded rater mean, every offset is smaller in magnitude, by 0.19 to 0.27 levels, which is the amount by which the half-up rule raises each trait's labels.

\begin{table}[!tbp]
\centering
\footnotesize
\setlength{\tabcolsep}{3.5pt}
\begin{tabular}{@{}lrrrrr@{}}
\toprule
 & \multicolumn{4}{c}{Exact accuracy (\%)} & \\
\cmidrule(lr){2-5}
Judge & All & \makecell[r]{Half-\\way} & Other & \makecell[r]{Down-\\ward} & \makecell[r]{Offset vs.\\mean} \\
\midrule
Jev Choice & 13.4 & 6.8 & 19.3 & 29.1 & $-1.01$ \\
Jev Score & 15.3 & 6.6 & 23.0 & 34.8 & $-0.92$ \\
Luna & 14.1 & 7.9 & 19.6 & 28.6 & $-1.04$ \\
Gemini & 29.8 & 11.9 & 45.7 & 57.6 & $-0.54$ \\
DeepSeek & 15.0 & 9.2 & 20.2 & 29.6 & $-0.93$ \\
\midrule
Decider Choice & 47.1 & 30.9 & 61.4 & 62.3 & $-0.23$ \\
Decider Score & 51.0 & 36.1 & 64.0 & 61.8 & $-0.16$ \\
\bottomrule
\end{tabular}
\caption{ELLIPSE labels and the half-up rounding rule. Half-way labels are the 725 pairs whose two ratings average to a half level, which the rule rounds up, and other labels are the remaining 823 pairs (DeepSeek answered 1{,}543 pairs in all). The downward column is a post-hoc analysis that rounds the half-way means down instead. The last column is the offset in levels against the unrounded rater mean. The rule adds 0.23 levels to the pooled label. The rows below the rule are the Decider's framings, from its own copy.}
\label{tab:ellipse-rounding}
\end{table}

\begin{table}[!tbp]
\centering
\footnotesize
\setlength{\tabcolsep}{4pt}
\begin{tabular}{@{}lrrrr@{}}
\toprule
 & \multicolumn{2}{c}{Topical-Chat} & \multicolumn{2}{c}{PersonaChat} \\
\cmidrule(lr){2-3}\cmidrule(lr){4-5}
Judge & $s = 0$ & $s = +1$ & $s = 0$ & $s = +1$ \\
\midrule
Jev Choice & 55.6 & 47.2 & 16.9 & 69.5 \\
Jev Score & 61.1 & 43.1 & 18.3 & 76.7 \\
Luna & 59.7 & 34.7 & 26.7 & 78.3 \\
Gemini & 63.4 & 40.8 & 23.3 & 80.0 \\
DeepSeek & 62.5 & 37.5 & 23.3 & 71.7 \\
\midrule
Rater reference & 58.8 & -- & 61.1 & -- \\
\bottomrule
\end{tabular}
\caption{Post-hoc shift on USR \crit{engaging}. Exact accuracy (\%) with the recorded predictions ($s = 0$) and with every prediction moved up one level, capped at the top level ($s = +1$). Each judge scored 71 or 72 pairs on Topical-Chat and 59 or 60 on PersonaChat. The rater reference is computed on \crit{engaging} alone, as in Figure~\ref{fig:location}b, and is shown unshifted.}
\label{tab:engaging-shift}
\end{table}

\begin{figure}[!tbp]
\centering
\includegraphics[width=\columnwidth]{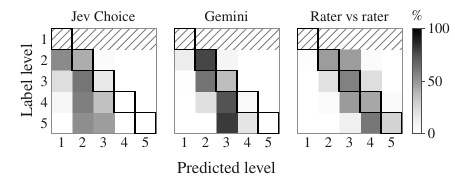}
\caption{Row-normalized ELLIPSE confusion matrices for Jev Choice, Gemini, and one rater against the other, with the diagonal outlined. Rows are the label level and columns the predicted level. For the raters, rows are one rating and columns the other rating of the same essay and trait, with both orders pooled. The level-1 rows hold fewer than 20 pairs and are omitted (hatched).}
\label{fig:ellipse-confusion}
\end{figure}

\begin{figure}[!tbp]
\centering
\includegraphics[width=\columnwidth]{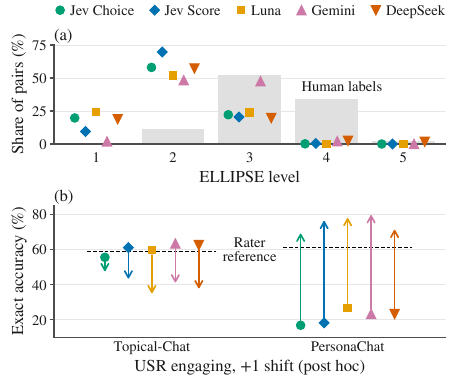}
\caption{(a) Share of ELLIPSE pairs at each level: labels (gray bars) and Jev's five judges (DeepSeek over its 1,543 answered pairs). (b) USR \crit{engaging}: exact accuracy before (marker) and after (arrowhead) a post-hoc +1 shift; dashed: each corpus's rater reference on \crit{engaging} (Table~\ref{tab:engaging-shift}).}
\label{fig:location}
\end{figure}

\paragraph{Level usage.}
Pooled over the six traits, individual ratings put 25.4\% of ratings at levels 4--5, fewer than the labels' 36.3\%, and none of Jev's five judges places more than 9.3\% of any trait's predictions there (Table~\ref{tab:ellipse-traits}c). Decider Choice places 10.7\% of its ELLIPSE predictions at these levels, and 19.4\% of its \crit{conventions} predictions. Decider Score places 10.8\%, and at most 15.9\% of one trait's (\crit{syntax}). At the bottom of the scale the pattern reverses. The rounded label is level~1 on one pair in 1{,}548, and the individual raters used level~1 on 11 of 3{,}096 ratings, whereas Jev's five judges predicted it on 30 (Gemini) to 376 (Luna) pairs, on \crit{grammar} more than on any other trait.

\paragraph{Compression.}
The judges compress the scale. Pooled over the 1{,}548 pairs, the regression slope of Table~\ref{tab:ellipse-traits}d runs from 0.29 (Jev Score) to 0.44 (Gemini), with Jev Choice at 0.37. The same regression of one rater's level on the other's, over 3{,}096 ordered pairs of ratings, gives 0.50. This slope is below one because the two raters disagree. It is not a like-for-like benchmark for the judges, whose target, the two-rater label, is less noisy than a single rating. A judge with one rater's noise would show a slope above 0.50 against the label, so a comparison with 0.50 understates the judges' compression. Per trait, the judges' slopes run from 0.14 to 0.54 and the raters' from 0.44 to 0.54 (Table~\ref{tab:ellipse-traits}d). Jev Choice's within-one accuracy is 64.3\%, and one rater is within one level of the other on 98.2\% of ratings. For labels of level~3 to level~5, Jev Choice most often predicts level~2, and Gemini most often predicts one or two levels below the label (Figure~\ref{fig:ellipse-confusion}).

\paragraph{Rank order.}
Averaged over traits, Jev Choice's Spearman correlation with the unrounded rater mean is 0.481, and one rater's correlation with the other is 0.483. By this measure Jev Choice ranks essays about as well as a single rater. As with the slopes, the comparison favors the judge.

\paragraph{Holistic band.}
The ELLIPSE sample was stratified by bands of the holistic score, with 432 pairs at 2.5 or below, 888 at 3 to 3.5, and 228 at 4 or above. In both copies every judge's offset grows more negative from band to band, from $-1.02$ levels in the lowest band to $-1.63$ in the highest for Jev Choice, from $-0.34$ to $-0.68$ for Decider Choice, and from $-0.59$ to $-1.09$ for Gemini. Compression predicts this pattern. When predictions rise by less than one level per label level, the offset grows with the label. The growth in Jev's copy is consistent with the population-norm account of Appendix~\ref{app:conventions}, but it does not by itself distinguish that account from other causes of compression.

\paragraph{Constant shift.}
A post-hoc analysis moves every prediction by $s$ levels, clipped to the scale (Figure~\ref{fig:ellipse-shift}). For each of Jev's five judges, exact accuracy peaks at $s = +1$, at 46.8\% (Luna) to 59.6\% (Gemini) against the rater reference's 51.4\%, and falls to at most 33.5\% at $s = +2$ and below 2\% at $s = -1$. Their mean QWK also peaks at $s = +1$, at 0.31 to 0.48, below the rater reference's 0.50, and Jev Choice's rises there from 0.15 to 0.34. Apart from clipping, a constant shift leaves the regression slopes above unchanged. The raters' own best shift on ELLIPSE is zero, so for these judges the shift corrects only their location. The Decider's framings, like the raters and unlike the three LLM judges, are most accurate unshifted. Decider Choice reaches 47.1\% at $s = 0$ and 45.6\% at $s = +1$, and Decider Score 51.0\% at $s = 0$ and 41.3\% at $s = +1$.

\paragraph{Rounding.}
Half-up rounding of the labels accounts for about 0.23 levels of Jev Choice's ELLIPSE offset. Table~\ref{tab:ellipse-rounding} isolates the rounding rule of Section~\ref{sec:labels}. Every judge of either copy is less accurate on the half-way labels than on the others, Jev Choice at 6.8\% against 19.3\% and Decider Choice at 30.9\% against 61.4\%. Against the unrounded rater mean, four of Jev's five judges are 0.92 to 1.04 levels low and Gemini is 0.54 levels low.

Rounding the half-way means down instead, as a post-hoc analysis, changes the order of two of Jev's five judges. Jev Choice then overtakes Luna (29.1\% against 28.6\% exact), and Gemini's lead over Jev Choice grows from 16.4 to 28.4 points. The downward rule lowers each half-way label by one level, which raises every offset by 0.47 levels (one level times the half-way share of 46.8\%). That is less than the smallest offset magnitude among Jev's five judges, Gemini's 0.77 levels, so all five stay negative. It is more than either Decider framing's, so the Decider's ELLIPSE offset changes sign with the rounding rule. Decider Choice's offset is $-0.46$ levels under the half-up rule and $+0.004$ under the downward one, and Decider Score's moves from $-0.40$ to $+0.07$.

\paragraph{USR \crit{engaging}.}
Table~\ref{tab:engaging-shift} gives each judge's exact accuracy on USR \crit{engaging} before and after the post-hoc shift of Appendix~\ref{app:conventions}, Jev Score included. On PersonaChat every shifted judge exceeds the unshifted rater reference, a comparison that favors the judges for the reason given under Leave-one-out and best shifts. The rater reference's own best shift raises it there from 61.1\% to 80.0\%. Gemini's shifted accuracy equals that value, the other four judges fall 1.7 (Luna) to 10.5 points (Jev Choice) short of it, and the four matched judges average 5.1 points below it. On Topical-Chat no shift raises the rater reference or any matched judge. The matched judges order the responses similarly on both corpora (Spearman $\rho = 0.50$ to $0.72$ with the rater mean), and part of the PersonaChat gain comes from moving predictions onto the top level, where 63.3\% of its labels sit. Holding the criterion text fixed leaves two other differences between the corpora. Topical-Chat responses are grounded in facts, which the top level's text (``presents an interesting fact'') mentions, and raters were assigned to contexts in rotation on Topical-Chat and in fixed triples on PersonaChat.

%% file: appendix/h_conventions.tex
\section{Where the Conventions Live}
\label{app:conventions}

Table~\ref{tab:conventions} lists the candidate conventions that Section~\ref{sec:conventions} introduces. They are of three kinds (a population norm on ELLIPSE, a corpus-relative standard, and annotation habits), and the raters' instructions may state some (we checked ELLIPSE only). Table~\ref{tab:criterion-texts} quotes in full the criterion texts behind its rows. Judge sets and ranges here are those of Jev's copy (Section~\ref{sec:jev}) unless the Decider is named; Table~\ref{tab:conventions-decider} gives the same rows in the Decider's copy.

\begin{table*}[!tp]
\centering
\footnotesize
\setlength{\tabcolsep}{4pt}
\renewcommand{\arraystretch}{1.12}
\begin{tabular}{@{}L{2.3cm} L{4.5cm} L{4.0cm} L{4.3cm}@{}}
\toprule
Panel, criterion & Text the judges received & What the labels also encode & Evidence \\
\midrule
ELLIPSE \crit{grammar} & ``Level 3: Some errors in grammar and usage.'' The criterion sentence adds ``level 5 = native-like facility''. & Error frequency judged against grade 8--12 English learners, learned in rater training (population norm) & Exact 0.4--19.0\% against a reference of 50.4\%; offset $-0.99$ to $-1.95$ \\
USR-PC \crit{engaging} & ``Interesting: the response is very interesting or presents an interesting fact.'' (identical on Topical-Chat) & Interest judged relative to the responses in the corpus (corpus-relative standard) & Offset $-0.73$ to $-1.03$ on PersonaChat, $-0.01$ to $-0.31$ on Topical-Chat; a post-hoc +1 shift gives 69.5--80.0\% exact, and a rater given its best shift 80.0\% \\
FED-Dialogue \crit{depth} & ``The system discusses topics in depth.'' & Depth judged by what a chatbot conversation can offer (annotation habit) & Exact 4.0\% for every matched judge (Jev Score 8.0\%) against a reference of 58.4\% \\
LFQA \crit{factuality}, top level & ``Entirely accurate: \ldots\ can be relied upon as a trusted source of information.'' & Top level granted to answers with errors that the LLM judges name, in two quoted cases (annotation habit) & Top level for 50 of 120 answers by the raters and 12 of 118 by Jev Choice; Jev Choice exact 17.4\% on 23 unanimous pairs, 46.3\% on 95 split pairs \\
HelpSteer2 \crit{correctness} & ``All pertinent facts are included and the response contains no errors.'' & Overall helpfulness, which the label tracks (annotation habit) & Label--helpfulness Pearson $r = 0.94$; judge QWK with the label 0.16--0.30 \\
USR-TC \crit{uses\_knowledge} & ``Given the fact the response is conditioned on, the response uses that fact well'' & MET when the context has no fact (annotation habit) & Raters MET on 18 of 18 \texttt{\_nofact} pairs (17 unanimous); Jev Noul UNMET on 16 \\
USR-TC \crit{understandable} & ``This is independent of whether the response is on topic or otherwise well-formed.'' & Self-contradictory responses failed, a stricter rule than the text (annotation habit) & Jev Noul MET rate 90.3\% against 68.1\%; 17.4--34.8\% correct on UNMET pairs \\
\midrule
ELLIPSE \crit{cohesion} & Level 4: ``a range of cohesive devices used appropriately such as reference and transitional words'' & none identified; the levels name observable devices & Every judge's best trait, 35.8--44.6\% exact against 45.3\%; offset $-0.52$ to $-0.62$ \\
HelpSteer2 \crit{verbosity} & ``amount of detail, relative to what the prompt asked for'' & none identified & Judge QWK 0.58--0.69, against at most 0.33 on the other three attributes \\
\bottomrule
\end{tabular}
\caption{Candidate conventions: what the labels encode beyond the criterion text of our rubric versions, quoted verbatim. The third column is our interpretation of the evidence in the fourth. Ranges span the four matched judges of Jev's copy unless stated, and every shift is a post-hoc analysis. The two rows below the rule are contrasts in which the text names something observable in the unit. Table~\ref{tab:conventions-decider} gives the Decider's copy.}
\label{tab:conventions}
\end{table*}

\begin{table*}[!tp]
\centering
\footnotesize
\setlength{\tabcolsep}{4pt}
\renewcommand{\arraystretch}{1.12}
\begin{tabular}{@{}>{\raggedright\arraybackslash}p{2.4cm}>{\raggedright\arraybackslash}p{12.9cm}@{}}
\toprule
Panel and criterion & Criterion sentence, then level descriptions from lowest to highest \\
\midrule
ELLIPSE \crit{grammar} & \textit{Criterion.} The essay demonstrates strong command of grammar and usage, per the ELLIPSE English-proficiency analytic rubric's Grammar trait (level 1 = lowest proficiency, level 5 = native-like facility). \newline \textit{Levels.} Level 1: Errors in grammar and usage throughout. $\mid$ Level 2: Many errors in grammar and usage. $\mid$ Level 3: Some errors in grammar and usage. $\mid$ Level 4: Minimal errors in grammar and usage. $\mid$ Level 5: Command of grammar and usage with few or no errors. \\
\addlinespace
USR-TC and USR-PC \crit{engaging} & \textit{Criterion.} The response is interesting rather than dull: it engages the reader (for example, by offering an opinion, a thought, or an interesting fact) rather than being generic. \newline \textit{Levels.} Dull: the response is generic and dull. $\mid$ Somewhat interesting: the response could engage the reader in the conversation (e.g., an opinion or a thought). $\mid$ Interesting: the response is very interesting or presents an interesting fact. \\
\addlinespace
FED-Dialogue \crit{depth} & \textit{Criterion.} The system discusses topics in depth. \newline \textit{Levels.} No. $\mid$ Somewhat. $\mid$ Yes. \\
\addlinespace
LFQA \crit{factuality} & \textit{Criterion.} The information provided in the answer is factually correct and aligns with well-established information. \newline \textit{Levels.} Completely inaccurate: the answer is entirely false or contains significant factual errors. It contradicts well-established information or provides misleading information. $\mid$ Partially accurate: the answer has some correct information but also contains significant inaccuracies or lacks important details. It requires additional verification or correction. $\mid$ Mostly accurate: the answer is predominantly correct and provides relevant information. However, it may still have minor inaccuracies or omissions that do not significantly impact the overall accuracy. $\mid$ Entirely accurate: the answer is entirely accurate and factually correct. It aligns with well-established information, provides clear and precise details, and can be relied upon as a trusted source of information. \\
\addlinespace
HelpSteer2 \crit{correctness} & \textit{Criterion.} The response includes all pertinent facts relevant to the prompt and contains no factual errors. \newline \textit{Levels.} Major errors and/or missing pertinent facts; the response cannot be relied on for correctness. $\mid$ Several errors or omissions of pertinent facts. $\mid$ A mix of correct and incorrect or missing content; correctness is inconsistent. $\mid$ Mostly correct and complete in its factual content, with at most minor errors or omissions. $\mid$ All pertinent facts are included and the response contains no errors. \\
\addlinespace
USR-TC \crit{uses\_knowledge} & \textit{Criterion.} Given the fact the response is conditioned on, the response uses that fact well (rather than not mentioning or referring to it at all). \newline \textit{Verdicts.} MET or UNMET (binary criterion; the LLM judges could also return CANNOT\_ASSESS). \\
\addlinespace
USR-TC \crit{understandable} & \textit{Criterion.} The response is understandable in the context of the conversation history: a reader can tell what the person is trying to say (for example, its pronouns and references resolve sensibly). This is independent of whether the response is on topic or otherwise well-formed. \newline \textit{Verdicts.} MET or UNMET (binary criterion; the LLM judges could also return CANNOT\_ASSESS). \\
\midrule
ELLIPSE \crit{cohesion} & \textit{Criterion.} The essay demonstrates strong text organization and use of cohesive devices to connect ideas across sentences and paragraphs, per the ELLIPSE English-proficiency analytic rubric's Cohesion trait (level 1 = lowest proficiency, level 5 = native-like facility). \newline \textit{Levels.} Level 1: No clear control of organization; cohesive devices not present or unsuccessfully used; presentation of ideas unclear. $\mid$ Level 2: Organization only partially developed with a lack of logical sequencing of ideas; some basic cohesive devices used but with inaccuracy or repetition. $\mid$ Level 3: Organization generally controlled; cohesive devices used but limited in type; Some repetitive, mechanical, or faulty use of cohesion use within and/or between sentences and paragraphs. $\mid$ Level 4: Organization generally well controlled; a range of cohesive devices used appropriately such as reference and transitional words and phrases to connect ideas; generally appropriate overlap of ideas $\mid$ Level 5: Text organization consistently well controlled using a variety of effective linguistic features such as reference and transitional words and phrases to connect ideas across sentences and paragraphs; appropriate overlap of ideas. \\
\addlinespace
HelpSteer2 \crit{verbosity} & \textit{Criterion.} The response's amount of detail, relative to what the prompt asked for, ranging from minimal to extensive. \newline \textit{Levels.} Minimal detail; extremely terse relative to the prompt. $\mid$ Below the level of detail one would typically expect for this prompt. $\mid$ A moderate, middling level of detail. $\mid$ Above the level of detail one would typically expect for this prompt. $\mid$ Extensive, exhaustive detail; highly elaborated relative to the prompt. \\
\bottomrule
\end{tabular}
\caption{Verbatim criterion sentences and level descriptions for the criteria of Table~\ref{tab:conventions}, with levels separated by $\mid$ and listed from lowest to highest. The NA option (Section~\ref{sec:jev}) is not listed. Typography, including capitals inside a description, is as in our rubric versions (Section~\ref{sec:panels}). The HelpSteer2 descriptions paraphrase the benchmark's guideline. The two criteria below the rule are the contrast rows of Table~\ref{tab:conventions}.}
\label{tab:criterion-texts}
\end{table*}

\begin{table*}[!t]
\centering
\footnotesize
\setlength{\tabcolsep}{4pt}
\renewcommand{\arraystretch}{1.12}
\begin{tabular}{@{}>{\raggedright\arraybackslash}p{2.5cm}>{\raggedright\arraybackslash}p{4.4cm}>{\raggedright\arraybackslash}p{8.3cm}@{}}
\toprule
Convention and criterion & Raters and judges & Quoted text \\
\midrule
Population norm; ELLIPSE \crit{grammar} & Raters: Level 2, Level 2. All five judges: Level 1 (Jev Choice confidence 0.95). & Gemini: ``The submission contains pervasive errors in grammar, usage, and sentence structure in virtually every clause [\ldots], matching Level 1.'' \\
\addlinespace
Annotation habit, absolute top level; LFQA \crit{factuality} & Two ChatGPT answers. Raters on each: Mostly, Entirely, Entirely accurate (label: Entirely accurate). All five judges on each: Partially accurate. & Answer: ``LED TVs use light emitting diodes to create images''. DeepSeek: ``In reality, LED TVs are LCD TVs with LED backlighting, not a distinct image-creating technology.'' \newline Answer: ``The SD card operates on a different protocol and doesn't require a power supply.'' Gemini: ``[\ldots] it contains major factual inaccuracies by claiming an SD card `doesn't require a power supply' and inherently avoids data loss when pulled out.'' \\
\addlinespace
Annotation habit; USR-TC \crit{uses\_knowledge} & Raters: MET, MET, MET. All five judges: UNMET. & The context ends ``Fact: \texttt{\_nofact}''. Response: ``i 've never tried one of the longest tennis matches . what do you think of the granny shot in tennis ?'' Gemini: ``No fact was provided to condition the response on (`Fact: \_nofact'), so the response does not use or reference any specified fact.'' \\
\addlinespace
Annotation habit, stricter than the text; USR-TC \crit{understandable} & Raters: MET, UNMET, UNMET (label: UNMET). All five judges: MET. & Response: ``i do n't watch it but i do n't really watch it .'' Luna: ``The response is understandable: `it' sensibly refers to Game of Thrones, and the speaker clearly says they do not watch it, despite the redundant repetition.'' \\
\bottomrule
\end{tabular}
\caption{One quoted pair for the population norm and for three annotation habits. The corpus-relative standard of USR \crit{engaging} is a property of the corpus's rating distribution and has no single-pair example (Table~\ref{tab:engaging-shift}). The explanations are quoted from the LLM judges' responses, and [\ldots] marks an omission. USR responses are quoted with the corpus's tokenization.}
\label{tab:convention-examples}
\end{table*}

\begin{table*}[!t]
\centering
\footnotesize
\setlength{\tabcolsep}{4pt}
\begin{tabular}{@{}lcrrrrrrlcrr@{}}
\toprule
 & \multicolumn{2}{c}{Criterion length} & & \multicolumn{4}{c}{Unit-length quartile} & & & \multicolumn{2}{c}{Gap to reference} \\
\cmidrule(lr){2-3}\cmidrule(lr){5-8}\cmidrule(lr){11-12}
Panel & Words & $\rho$ ($n$) & \makecell[r]{Unit\\words} & Q1 & Q2 & Q3 & Q4 & Raters & Levels & \makecell[r]{Exact\\(points)} & QWK \\
\midrule
RiceChem & 3--21 & $-0.26$ (27) & -- & -- & -- & -- & -- & TA & 2 & -- & -- \\
HealthBench & 7--481 & $-0.27$ (34) & -- & -- & -- & -- & -- & 2--5 physicians & 2 & -- & -- \\
\midrule
ELLIPSE & 27--36 & $+1.00$ (6) & 392 & 16.4 & 13.9 & 11.2 & 12.3 & 2 trained & 5 & $-37.9$ & $-0.34$ \\
FED-Turn & 4--13 & $-0.10$ (7) & 9 & 56.3 & 63.0 & 62.5 & 61.4 & 5 crowd & 3 & $-7.6$ & $+0.07$ \\
FED-Dialogue & 6--13 & $+0.30$ (9) & 123 & 37.5 & 50.0 & 67.3 & 53.0 & 5 crowd & 3 & $-16.3$ & $+0.05$ \\
HelpSteer2 & 15--32 & -- & 215 & 43.3 & 59.3 & 51.8 & 40.7 & released & 5 & -- & -- \\
LFQA & 14--33 & -- & 100 & 69.4 & 70.1 & 66.7 & 68.8 & 3 crowd & 3--4 & $+13.3$ & $+0.09$ \\
USR-TC & 11--43 & -- & 19.5 & 68.2 & 65.3 & 58.9 & 77.8 & 3 researchers & 3 & $+0.3$ & $+0.11$ \\
USR-PC & 11--43 & -- & 12 & 73.8 & 69.5 & 68.2 & 70.0 & 3 researchers & 3 & $-16.2$ & $-0.04$ \\
\bottomrule
\end{tabular}
\caption{Panel properties that might order Jev Choice's gap to the rater reference (What does not separate the panels), with that gap in the last two columns. Criterion length: range of criterion words and the Spearman correlation between a criterion's length and Jev's accuracy on it, over $n$ criteria (Jev Noul on RiceChem and HealthBench, and elsewhere Jev Choice on ordinal criteria only; the ELLIPSE value uses the criterion sentence alone). Some HealthBench criteria share a text (Appendix~\ref{app:protocol}) and so have the same length. Unit words: median whitespace words per unit. Quartiles: Jev Choice's accuracy (\%) by quartile of unit length, embedded binary pairs included. Raters: the source of the human label (TA, teaching-assistant annotation; trained essay raters; crowd workers; dialogue researchers; the released label on HelpSteer2). Gap: Jev Choice's value minus the rater reference's, in points of exact accuracy (Section~\ref{sec:metrics}) and in mean QWK. A dash marks a value that was not computed or does not exist.}
\label{tab:panel-factors}
\end{table*}

\paragraph{Natural contrasts.}
Four natural contrasts in the panels bear on the candidate conventions, and none was designed as an experiment. They vary (1) the judge, with the pairs fixed; (2) the corpus, with the criterion text fixed (USR \crit{engaging}); (3) the criterion within a rubric (Contrasts within a rubric, below); and (4) the form of the criterion, binary or ordinal (Appendix~\ref{app:binaryobs}). Only the first three hold one factor approximately fixed. In the first, the offsets share their sign across the three LLM judges in both judging modes (Table~\ref{tab:wr-downstream}). Jev Choice, which never sees the harness prompt, and Jev Score, which decodes by expected level, share it too, so neither a shared prompt nor a decoding rule is their common cause. In the second, USR \crit{engaging}, worded identically on Topical-Chat and PersonaChat, draws matched-judge offsets of $-0.73$ to $-1.03$ levels on PersonaChat, whose raters gave the top level in 59.4\% of ratings, and $-0.01$ to $-0.31$ on Topical-Chat, where they gave it in 34.7\%. A one-level shift lifts every matched judge on PersonaChat from at most 26.7\% to at least 69.5\% and lowers it on Topical-Chat (Figure~\ref{fig:location}b); we read these labels as corpus-relative (Appendix~\ref{app:offset} gives this contrast in detail).

\paragraph{Label noise.}
Label noise would scatter errors on both sides of the label and gather them on split pairs, so it cannot explain ELLIPSE, where 99.5\% of Jev Choice's errors lie below the label, LFQA \crit{factuality}, where Jev Choice misses unanimous pairs more often than split ones (an exact accuracy of 17.4\% on the 23 unanimous pairs against 46.3\% on the 95 split pairs; Table~\ref{tab:conventions}), or USR-TC's almost unanimous \texttt{\_nofact} pairs, whose context holds no fact. Beyond ELLIPSE, LFQA \crit{factuality}, and USR-TC \crit{uses\_knowledge}, the evidence that the judges place units lower than the raters do is weaker. On a three-level scale whose labels are mostly at the top, a judge that uses fewer levels than the raters shows a negative mean offset even without placing units lower. Four panel properties (criterion length, unit length, rater expertise, and scale width) do not order the panels by Jev Choice's gap (Table~\ref{tab:panel-factors}). The panel with the largest gap, ELLIPSE, is the one rated by trained raters. This is weak evidence against an account based on label quality. A shared reluctance to grant top levels and shared training priors~\citep{kim2025correlated} remain possible, and no experiment varied the text, so the conventions account is observational.

\paragraph{Population norm.}
ELLIPSE's labels may carry a population norm (Table~\ref{tab:conventions}). Its trained raters scored the essays of grade 8 to 12 English learners~\citep{crossley2023english}. A norm-referenced scale reads a level against a population~\citep{glaser1963instructional}, and training instills such expectations~\citep{weigle1994effects}. The gloss of level 5 in our criterion sentence, ``native-like facility'', and LLM essay graders' penalty for small errors~\citep{mathew2026llms} may also contribute. ELLIPSE's level descriptions are those of the corpus's English-proficiency scoring rubric. The judges received those same descriptions, so a population norm could reach the raters only through their training. A zero-shot judge has nothing but the descriptions and our criterion sentence, which ends ``(level 1 = lowest proficiency, level 5 = native-like facility)'' (Table~\ref{tab:criterion-texts}). Read against native-like writing, ``some errors'' would put a learner essay low, so the offset may reflect this gloss of level~5 as well as the missing population. These data do not distinguish a population norm from the small-error penalty of LLM essay graders. The Decider received the same texts with no exemplars, yet on ELLIPSE it comes close to the raters (The Decider, below).

The ELLIPSE example of Table~\ref{tab:convention-examples} is one of Jev Choice's confident errors (Appendix~\ref{app:shared}). Both raters put the essay at level~2 on \crit{grammar}, and all five judges put it at level~1. Gemini's explanation reports errors in nearly every clause, which it matches to the level~1 description ``Errors in grammar and usage throughout.'' The level descriptions give no base rate, so they do not say how many errors separate ``throughout'' from ``many'' in essays by grade 8--12 English learners. Rater training on that population would supply such a base rate. The judges lack it and placed this essay one level lower, as the population-norm account predicts.

\paragraph{Contrasts within a rubric.}
Within a rubric the judges, prompt, and units stay the same, so the third contrast varies only the criterion. ELLIPSE's best and worst traits differ in what their level descriptions name. Each of Jev's five judges is most accurate on \crit{cohesion}, whose level descriptions name observable devices, and least accurate on \crit{grammar}, whose levels differ only by the quantifiers ``throughout'', ``many'', ``some'', ``minimal'', and ``few or no'' (Tables~\ref{tab:conventions} and~\ref{tab:criterion-texts}). Both Decider framings are most accurate on \crit{syntax} instead, and Decider Choice is least accurate on \crit{phraseology}. The \crit{cohesion} trait still carries an offset of $-0.52$ to $-0.62$ levels, and with six traits the level descriptions are confounded with other trait differences, such as criterion length (see What does not separate the panels, below). Every matched judge, in either copy, also agrees best with the HelpSteer2 labels on \crit{verbosity}, which describes an amount of detail visible in the response.

\paragraph{Annotation habit: an absolute top level.}
LFQA's raters grant the top \crit{factuality} level far more often than Jev Choice does, to 50 of the 120 answers against 12 of the 118 that Jev Choice scored (Table~\ref{tab:conventions}). Table~\ref{tab:convention-examples} quotes two ChatGPT answers that two of the three raters placed at the top level and on which all five judges chose ``Partially accurate''. The three LLM judges name the same false claim in each answer: that LED TVs create images with light-emitting diodes, and that an SD card needs no power supply. In both cases the label departs from the text of the top level, and the judges reject the top level for the reason its text gives. HelpSteer2 shows a related pattern at a level that is neither the top nor absolute. None of Jev's five judges chose \crit{complexity} level 4 or 5 in 450 predictions, although the labels put 10 of the 90 responses at level 4. Decider Choice chose them 5 times and Decider Score 4 times.

\paragraph{Annotation habits on USR-TC.}
Three of the twelve sampled USR-TC contexts carry the literal string \texttt{\_nofact} in place of a fact, and the 18 \crit{uses\_knowledge} pairs from these contexts are the ones counted in Table~\ref{tab:conventions}. The LLM judges' explanations show that they read the string as the absence of a fact (Table~\ref{tab:convention-examples}). The \crit{understandable} criterion text excludes well-formedness from the judgment. The raters nonetheless marked self-contradictory responses such as the one quoted UNMET, and every judge is correct on at most 34.8\% of the 23 pairs labeled UNMET.

\paragraph{The USR-TC errors.}
Jev Choice makes 117 errors on USR-TC, counting its Jev Noul verdicts on the two binary criteria. Of these, 16 are the \texttt{\_nofact} pairs of \crit{uses\_knowledge} and 17 are \crit{understandable} pairs labeled UNMET. Another 63 are ordinal errors on split pairs. The remaining 21 errors (17.9\%) are 19 ordinal errors on unanimous pairs and two other binary errors. They are the only errors that fall neither on a split pair nor under a candidate convention, and those conventions were themselves identified from the judges' errors.

\paragraph{No-read baselines.}
Both no-read baselines of Section~\ref{sec:labels} are fitted on the evaluated pairs, and HelpSteer2's names each attribute's most frequent level. On two panels such a baseline is competitive on exact accuracy. The HelpSteer2 \crit{correctness} text asks for a fact check, but its label tracks overall helpfulness (Pearson $r = 0.94$ over the 90 responses; Table~\ref{tab:conventions}). HelpSteer2's constant predictor has a QWK of zero by construction. It beats each of Jev's five judges in exact accuracy, significantly in pair-level sign tests for all but Gemini ($p \le 2.1 \times 10^{-4}$; Gemini $p = 0.075$), so separations on HelpSteer2 in Jev's copy (Table~\ref{tab:clustered}) rank judges that trail a constant. Decider Choice, at 63.1\%, is above the constant without a significant difference (right alone on 45 pairs against the constant's 36, $p = 0.37$), and Decider Score, at 55.3\%, is below it, also not significantly (56 against 73, $p = 0.16$). The Decider's HelpSteer2 leads are thus over LLM judges that trail the constant. The LFQA provenance baseline is most accurate on \crit{amount\_info} (74.2\%) and \crit{formality} (73.3\%) and least on \crit{factuality} (56.7\%). Without provenance, the modal level of each criterion scores 61.1\% over the panel, against the provenance baseline's 68.1\%. In pair-level sign tests (Appendix~\ref{app:tests}), the $p$-value of each of Jev's five judges against the provenance baseline is at least 0.27, Decider Choice's is 0.83, and Decider Score's is 0.19. Every matched judge's mean QWK, 0.43 to 0.51, is above the baseline's 0.28, a difference we did not test.

\paragraph{What does not separate the panels.}
Table~\ref{tab:panel-factors} gives the four properties named under Label noise next to Jev Choice's gap to the reference in exact accuracy and in mean QWK. The two measures put Jev Choice on the same side of the reference on four panels, above it on LFQA and USR-TC and below it on ELLIPSE and USR-PC. On the two FED panels Jev Choice is below the reference in exact accuracy and above it in mean QWK, and part of this pattern may come from the rater reference's leave-one-out target (Section~\ref{sec:labels}).

Within panels, criterion length shows no consistent relation with Jev's per-criterion accuracy, and with few criteria per panel the test is weak. The correlations of Table~\ref{tab:panel-factors} are not significant on the binary panels ($p = 0.18$ on RiceChem, $0.12$ on HealthBench) or on the FED panels ($p = 0.83$ on FED-Turn, $0.43$ on FED-Dialogue). ELLIPSE is ambiguous. Its six criterion sentences share one template and differ in length by at most nine words, yet their length orders Jev Choice's per-trait accuracy perfectly (Spearman 1.00). Level-description length, which ranges from 44 to 133 words, correlates at 0.60 ($p = 0.21$). With six traits, length cannot be separated from what each trait's levels describe (Table~\ref{tab:conventions}).

By quartile of unit length, Jev Choice's accuracy on ELLIPSE falls from 16.4\% in the shortest quartile to 11.2--12.3\% in the two longest, and on FED-Dialogue it is lowest in the shortest quartile (37.5\%). Elsewhere it moves by up to 18.9 points without a consistent direction, so unit length has no common effect. The ELLIPSE offset also grows with the holistic band (Appendix~\ref{app:offset}), and this slice does not separate essay length from essay quality. Across panels, the three with the shortest units, FED-Turn, USR-PC, and USR-TC, have gaps that run from 16.2 points below the reference to 0.3 points above it.

ELLIPSE, the only ordinal panel with trained raters, has Jev Choice's largest gap (see Label noise), but among the crowd-rated panels the gap runs from $+13.3$ points on LFQA to $-16.3$ on FED-Dialogue (Table~\ref{tab:panel-factors}). Rater expertise does not order the gaps either.

Scale width changes the chance level of exact accuracy (Section~\ref{sec:metrics}), but panels with the same scale can still differ widely. USR-TC and USR-PC share their three-level scales and five byte-identical criterion texts, among them the \crit{engaging} text of the second natural contrast, and Jev Choice's gap is $+0.3$ points on USR-TC and $-16.2$ points on USR-PC.

\paragraph{The harness prompt.}
The first natural contrast excludes the harness prompt as the common cause of the offsets, but its instructions, such as its rule for self-contradicting submissions (Appendix~\ref{app:protocol}), may still add to the LLM judges' offsets. Jev Choice, which never receives them, is nonetheless offset further on ELLIPSE than Gemini and DeepSeek (Table~\ref{tab:offsets}). The paragraph on label noise above leaves open the possibility that all judges share a reluctance to grant high levels. Such a reluctance would produce a larger offset where raters use the top level more, and USR \crit{engaging} shows that pattern. The embedded binary criteria rule out only a general bias toward failing verdicts, because all judges over-grant MET on \crit{understandable}.

\begin{table*}[!t]
\centering
\footnotesize
\setlength{\tabcolsep}{4pt}
\renewcommand{\arraystretch}{1.12}
\begin{tabular}{@{}L{2.3cm} L{10.9cm} L{2.0cm}@{}}
\toprule
Panel, criterion & Evidence in the Decider's copy & As the LLM judges? \\
\midrule
ELLIPSE \crit{grammar} & Exact 1.6\% (Luna) to 40.7\% (Decider Choice) against a reference of 50.4\%, Decider Score 43.4\%; offset $-0.59$ (Decider Choice) to $-1.95$ & yes \\
USR-PC \crit{engaging} & Offset $-0.73$ to $-1.10$ on PersonaChat (both Decider framings $-1.10$), $-0.01$ to $-0.33$ on Topical-Chat; a post-hoc +1 shift gives 70.0--80.0\% exact on PersonaChat, lifting Decider Choice from 11.7\%, and lowers Decider Choice on Topical-Chat from 54.2\% to 45.8\% & yes \\
FED-Dialogue \crit{depth} & Exact 4.0\% for every matched judge and for Decider Score, against a reference of 58.4\% & yes \\
LFQA \crit{factuality}, top level & Top level for 70 of the 117 answers Decider Choice scored (offset $+0.21$), for 43 of 120 by Decider Score, and for 50 of 120 by the raters; Decider Choice exact 70.8\% on 24 unanimous pairs, 41.9\% on 93 split pairs & no, higher than the raters \\
HelpSteer2 \crit{correctness} & QWK with the label 0.49 for Decider Choice (80 pairs; offset $+0.41$) and for Decider Score, 0.18--0.30 for the LLM judges & no, higher than the label \\
USR-TC \crit{uses\_knowledge} & Decider Noul MET on 16 of the 18 \texttt{\_nofact} pairs; each LLM judge UNMET on 12 or 13 & no, as the raters \\
USR-TC \crit{understandable} & Decider Noul MET rate 97.2\% against 68.1\%; correct on 2 of the 23 UNMET pairs & yes \\
\midrule
ELLIPSE \crit{cohesion} & Exact 35.8--53.1\% against 45.3\%, Decider Choice the highest; offset $-0.38$ to $-0.62$; both Decider framings most accurate on \crit{syntax} (Decider Choice 57.8\%, Decider Score 61.2\%) & -- \\
HelpSteer2 \crit{verbosity} & QWK 0.58--0.67, each matched judge's best attribute; at most 0.58 on the other three attributes (Decider Score, \crit{complexity}) & -- \\
\bottomrule
\end{tabular}
\caption{The rows of Table~\ref{tab:conventions} in the Decider's copy, where the matched judges are Decider Choice and the three LLM judges and ranges span them unless stated. Last column: whether the Decider departs from the labels in the direction the LLM judges do. On USR-TC's binary criteria both Decider framings use Decider Noul (Section~\ref{sec:jev}). Shifts are post hoc.}
\label{tab:conventions-decider}
\end{table*}

\paragraph{The Decider.}
Four of the seven candidate rows go the same way in the Decider's copy and three do not (Table~\ref{tab:conventions-decider}). Decider Choice grants LFQA's top \crit{factuality} level to more answers than the raters did, and it is more accurate on that criterion's unanimous pairs than on its split ones, as label noise would predict. It places HelpSteer2 responses above their \crit{correctness} label on average, and both Decider framings agree with that label more closely than any LLM judge, at a QWK of 0.49 against 0.18 to 0.30. The Decider's data manifest lists HelpSteer2's attribute labels among its training sources (Appendix~\ref{app:decider}), so the model had been trained on HelpSteer2's rating scheme and labels, whether or not our sampled responses were among those rows, and on that panel it was not judging the raters' conventions zero-shot. On the \texttt{\_nofact} pairs, Decider Noul answers MET, as the raters did, on 16 of 18. On ELLIPSE, Decider Score's exact accuracy comes within half a point of the rater reference, and both Decider framings, like the raters, do best unshifted (Appendix~\ref{app:offset}). No source in that manifest names ELLIPSE or the essay corpus it samples, and nothing disclosed accounts for this result. Of the three observations set against label noise above, only ELLIPSE's holds in the Decider's copy, where Decider Choice errs below the label on 91.3\% of its wrong pairs, and even there the sign of its offset depends on how the labels are rounded.

\paragraph{What the contrasts suggest.}
How close a judge comes to the raters appears to depend on what the raters knew beyond the text the judges read. In the contrasts within a rubric, ELLIPSE \crit{cohesion} and HelpSteer2 \crit{verbosity}, the judges come closest where a level description points to something a reader can see in the unit. Where, on our reading, a level's meaning depends on a population, a corpus, or an annotation habit, Jev and every LLM judge, the best included, fall short in the same direction, and the Decider does so on four of the seven rows. If every judge reads the criterion text as written, agreement with labels on an ordinal rubric measures in part whether a judge was given the raters' conventions, such as reading ELLIPSE essays against the population of learner essays they come from. Section~\ref{sec:discussion} counts such literal reading among several possible contributors to this shared shortfall, together with shared training priors and a reluctance to grant top levels, and our data cannot tell these contributors apart.

%% file: appendix/i_confidence.tex
\section{Confidence}
\label{app:confidence}

\paragraph{The returned field.}
The vendor describes the confidence field that Choice and Score return as a number computed from how the probabilities are spread but gives no exact formula for either primitive. Its interactive demo approximates the field for three options by $(3 p_{\max} - 1)/2$.\footnote{\url{https://docs.typesafe.ai/confidence}.} For Jev Choice the field matches the general form $(K p_{\max} - 1)/(K - 1)$, where $p_{\max}$ is the largest returned probability and $K$ counts every option, the NA option included; it is 0 for a uniform distribution and 1 when one option receives all the probability. The largest deviation from this formula is 0.0225 over Jev Choice's 3{,}414 answers on ordinal pairs, with a mean of 0.006, and 0.015 over the 8{,}392 binary answers on the full RiceChem set (Appendix~\ref{app:binary}), where $K = 3$. These deviations are larger than two-decimal rounding of the returned values would produce, so the formula approximates the field closely without describing it exactly. The formula does not describe Jev Score's confidence (median deviation 0.10, maximum 0.46), which is used as returned. Noul returns no confidence field, and the Noul confidence of Section~\ref{sec:jev} is the same normalization with $K = 2$. The embedded binary pairs of the ordinal panels enter every Jev Choice analysis with this Noul confidence. Decider Choice's field follows the formula to floating-point precision on its 3{,}414 ordinal answers, and Decider Score's, which does not, is also used as returned (Appendix~\ref{app:decider}).

\begin{table*}[p]
\centering
\footnotesize
\setlength{\tabcolsep}{4pt}
\begin{tabular*}{\textwidth}{@{\extracolsep{\fill}}lrrrrr@{}}
\toprule
 & \multicolumn{4}{c}{Confidence band} & \\
\cmidrule(lr){2-5}
Panel & [0, 0.25) & [0.25, 0.5) & [0.5, 0.75) & [0.75, 1] & AUROC \\
\midrule
\multicolumn{6}{@{}l}{\textit{(a) Jev Choice error rate, \% (pairs)}} \\
ELLIPSE & 100.0 (1) & 85.8 (654) & 87.1 (697) & 87.2 (196) & 0.489 \\
FED-Turn & 52.8 (36) & 62.8 (137) & 45.5 (121) & 23.2 (306) & 0.698 \\
FED-Dialogue & 60.0 (10) & 58.7 (63) & 46.6 (58) & 41.2 (102) & 0.569 \\
HelpSteer2 & 88.9 (27) & 54.5 (132) & 48.1 (79) & 41.2 (114) & 0.639 \\
LFQA & 100.0 (1) & 43.8 (73) & 49.6 (113) & 13.5 (171) & 0.703 \\
USR-TC & 39.1 (23) & 47.2 (89) & 36.6 (101) & 19.7 (147) & 0.638 \\
USR-PC & 52.2 (23) & 41.9 (62) & 32.1 (81) & 18.8 (133) & 0.634 \\
\quad \crit{engaging} only & -- & 81.8 (11) & 85.7 (21) & 81.5 (27) & 0.523 \\
\quad without \crit{engaging} & 52.2 (23) & 33.3 (51) & 13.3 (60) & 2.8 (106) & -- \\
\midrule
\multicolumn{6}{@{}l}{\textit{(b) Jev Score error rate by its own confidence, \% (pairs)}} \\
ELLIPSE & -- & 60.0 (5) & 82.2 (1,066) & 90.6 (477) & \\
FED-Turn & 48.0 (102) & 44.9 (136) & 30.8 (104) & 13.2 (258) & \\
FED-Dialogue & 66.7 (42) & 59.4 (69) & 53.6 (56) & 18.1 (83) & \\
HelpSteer2 & 85.7 (42) & 83.0 (53) & 50.3 (143) & 39.3 (122) & \\
LFQA & 66.7 (6) & 50.0 (62) & 42.5 (113) & 14.5 (179) & \\
USR-TC & 33.3 (39) & 40.6 (96) & 33.7 (86) & 19.4 (139) & \\
USR-PC & 51.6 (31) & 45.1 (71) & 24.4 (78) & 19.2 (120) & \\
\midrule
\multicolumn{6}{@{}l}{\textit{(c) Repeated errors, observed\,/\,independence baseline, \% (Jev Choice errors); last column, all errors}} \\
ELLIPSE & 33.3\,/\,67.9 (1) & 55.7\,/\,49.1 (561) & 64.0\,/\,55.8 (607) & 80.5\,/\,63.7 (171) & 62.6\,/\,54.0 \\
FED-Turn & 24.6\,/\,24.8 (19) & 48.4\,/\,33.9 (86) & 55.5\,/\,39.1 (55) & 85.9\,/\,43.7 (71) & 59.7\,/\,37.4 \\
FED-Dialogue & 5.6\,/\,11.3 (6) & 52.3\,/\,32.2 (37) & 62.0\,/\,46.5 (27) & 85.7\,/\,58.3 (42) & 64.8\,/\,44.4 \\
HelpSteer2 & 27.8\,/\,13.1 (24) & 44.6\,/\,22.3 (72) & 65.8\,/\,26.6 (38) & 89.4\,/\,41.1 (47) & 58.5\,/\,26.9 \\
LFQA & 0.0\,/\,38.9 (1) & 52.1\,/\,28.2 (32) & 67.1\,/\,31.9 (56) & 92.8\,/\,43.4 (23) & 67.5\,/\,33.3 \\
USR-TC & 63.0\,/\,53.4 (9) & 48.4\,/\,35.2 (42) & 66.4\,/\,36.9 (37) & 87.2\,/\,33.6 (29) & 64.8\,/\,36.8 \\
USR-PC & 27.8\,/\,15.6 (12) & 32.1\,/\,24.8 (26) & 70.5\,/\,48.7 (26) & 85.3\,/\,54.6 (25) & 57.7\,/\,38.9 \\
\midrule
\multicolumn{6}{@{}l}{\textit{(d) Jev Choice accuracy on its most confident pairs, \%}} \\
\textit{Share of pairs} & \textit{top 25\%} & \textit{top 50\%} & \textit{top 75\%} & \textit{all} & \\
\cmidrule(lr){2-5}
ELLIPSE & 14.5 & 13.2 & 12.3 & 13.4 & \\
FED-Turn & 78.0 & 76.7 & 69.1 & 61.5 & \\
FED-Dialogue & 55.2 & 57.8 & 54.3 & 51.9 & \\
HelpSteer2 & 60.2 & 56.8 & 54.5 & 48.6 & \\
LFQA & 88.9 & 84.4 & 73.9 & 68.7 & \\
USR-TC & 82.2 & 78.3 & 70.7 & 67.5 & \\
USR-PC & 80.0 & 79.3 & 74.6 & 70.2 & \\
\bottomrule
\end{tabular*}

\vspace{0.9em}

\begin{tabular}{@{}>{\raggedright\arraybackslash}p{2.0cm}>{\raggedright\arraybackslash}p{\dimexpr\textwidth-2.0cm-2\tabcolsep\relax}@{}}
\toprule
\multicolumn{2}{@{}l}{\textit{(e) Jev Choice AUROC by criterion}} \\
ELLIPSE & \crit{cohesion} 0.502, \crit{syntax} 0.562, \crit{vocabulary} 0.615, \crit{phraseology} 0.404, \crit{grammar} 0.000$^{*}$, \crit{conventions} 0.227 \\
FED-Turn & \crit{interesting} 0.392, \crit{engaging} 0.565, \crit{specific} 0.605, \crit{relevant} 0.742, \crit{correct} 0.625, \crit{semantically\_appropriate} 0.576, \crit{fluent} 0.806, \crit{understandable} 0.980$^{*}$ \\
FED-Dialogue & \crit{coherent} 0.611, \crit{diverse} 0.699, \crit{depth} 0.542$^{*}$, \crit{likeable} 0.931, \crit{understanding} 0.901, \crit{flexible} 0.670, \crit{informative} 0.335, \crit{inquisitive} 0.562, \crit{consistent} 0.790 \\
HelpSteer2 & \crit{correctness} 0.756, \crit{coherence} 0.640, \crit{complexity} 0.569, \crit{verbosity} 0.492 \\
LFQA & \crit{factuality} 0.413, \crit{formality} 0.667, \crit{amount\_info} 0.689 \\
USR-TC & \crit{understandable} 0.767, \crit{natural} 0.622, \crit{maintains\_context} 0.629, \crit{engaging} 0.565, \crit{uses\_knowledge} 0.692 \\
USR-PC & \crit{understandable} 0.884$^{*}$, \crit{natural} 0.766, \crit{maintains\_context} 0.796, \crit{engaging} 0.523, \crit{uses\_knowledge} 0.895 \\
\bottomrule
\end{tabular}
\caption{Jev's confidence on the ordinal panels. (a)~Jev Choice's error rate by confidence band, with the number of pairs in parentheses, and the AUROC of confidence for Jev Choice's correctness; USR-PC is also split by criterion. (b)~Jev Score's error rate by Jev Score's own confidence. (c)~Repeated errors as a share of the LLM verdicts on Jev Choice's wrong pairs, then the independence baseline, both as defined in Section~\ref{sec:shared-errors}, with the number of Jev Choice errors in parentheses. The baseline is undefined for three verdicts each on ELLIPSE and HelpSteer2, whose combination of criterion and label occurs on no other pair. (d)~Jev Choice's accuracy on its most confident pairs, with ties broken by unit and criterion. (e)~AUROC for each criterion with at least 25 answered pairs and at least one pair of each outcome; $^{*}$ marks criteria with fewer than five pairs in the smaller outcome, whose AUROC rests on too few pairs to interpret. Embedded binary pairs are included in (a) to (e).}
\label{tab:confidence}
\end{table*}

\begin{figure}[!tbp]
\centering
\includegraphics[width=\columnwidth]{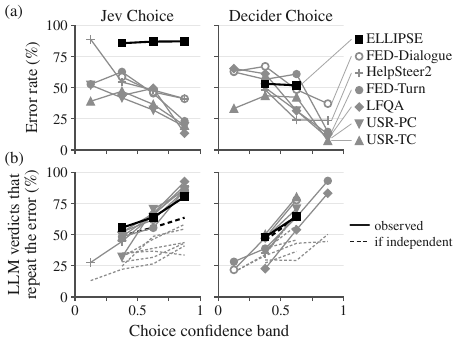}
\caption{Confidence band of Jev Choice (left) and Decider Choice (right) against (a) its error rate and (b) the share of LLM verdicts on its wrong pairs that repeat its answer (solid; dashed: independence baseline), one line per ordinal panel, ELLIPSE in black. Bands with under 20 pairs (a) or errors (b) are omitted.}
\label{fig:confidence}
\end{figure}

\paragraph{Panels.}
Outside ELLIPSE, Jev Choice's error rate falls between the 0.25--0.5 band and the top band on every panel, by 13.3 to 39.6 points (Table~\ref{tab:confidence}a), and the three largest falls are on FED-Turn, LFQA, and USR-TC. On ELLIPSE, where the error rate is flat across bands (Section~\ref{sec:confidence}; Figure~\ref{fig:confidence}a), confidence is also unrelated to the size of the error (Spearman $-0.03$ between confidence and absolute level error, $p = 0.23$, over 1{,}548 pairs). The lowest band is small on every panel and holds a single pair on ELLIPSE and on LFQA.

\paragraph{Criteria.}
The criterion-level AUROCs of Table~\ref{tab:confidence}e show where confidence fails inside a panel whose errors it otherwise ranks. The criteria named below were singled out after inspecting these values. The AUROC is below 0.45 on five criteria with at least five pairs in each outcome: ELLIPSE \crit{phraseology} and \crit{conventions}, FED-Turn \crit{interesting}, FED-Dialogue \crit{informative}, and LFQA \crit{factuality}. All five carry a negative offset for Jev Choice, from $-0.58$ levels on \crit{factuality} to $-1.38$ on \crit{phraseology}, which suggests that confidence fails where Jev Choice errs mostly in one direction. ELLIPSE \crit{vocabulary} is an exception, where Jev Choice's offset is $-1.28$ levels yet the AUROC is 0.62. LFQA \crit{factuality}, with an AUROC of 0.41 against 0.70 for its panel, also concentrates the panel's errors. Jev Choice is wrong on 59.3\% of its 118 pairs, against 18.3\% and 16.7\% on the panel's other two criteria. On USR-PC \crit{engaging} (AUROC 0.52), 83.1\% of Jev Choice's verdicts on the 59 pairs are wrong at a mean confidence of 0.72, and the error rate is flat over the three occupied bands. Without \crit{engaging}, USR-PC's error rate falls steeply with confidence (Table~\ref{tab:confidence}a). Confidence ranks errors well on HelpSteer2 \crit{correctness} (AUROC 0.76), a criterion whose labels follow overall helpfulness (Table~\ref{tab:conventions}).

\paragraph{Jev Score.}
Jev Score's error rate is lower in the top band than in the lowest on six panels and rises between the two upper bands on ELLIPSE (Table~\ref{tab:confidence}b). On FED-Turn's ordinal pairs alone, without the embedded binary pairs that Table~\ref{tab:confidence}d includes, Jev Score's most confident quarter is 86.3\% exact against 66.7\% overall, and Jev Choice's 71.0\% against 56.2\%.

\paragraph{Most confident pairs.}
Table~\ref{tab:confidence}d restricts Jev Choice to the 25\%, 50\%, and 75\% of pairs on which it is most confident. At 25\% coverage, accuracy is 9.8 to 20.2 points above the full-coverage value on five panels, 3.2 points above on FED-Dialogue, and 1.0 point above on ELLIPSE. These accuracies include the embedded binary pairs, so on the FED and USR panels the full-coverage column differs from Table~\ref{tab:main}.

\paragraph{The Decider's confidence.}
Decider Choice's confidence, which Table~\ref{tab:confidence} does not show, ranks its errors on the six panels other than ELLIPSE, with an AUROC of 0.72 on FED-Turn, 0.60 on FED-Dialogue, 0.70 on HelpSteer2, 0.78 on LFQA, 0.69 on USR-TC, and 0.75 on USR-PC. On ELLIPSE its AUROC is 0.52 and its error rate is flat (Figure~\ref{fig:confidence}a, right). It is 53.2\% in the 0.25--0.5 band and 51.9\% in the 0.5--0.75 band, which together hold all but two of its pairs, and no pair reaches the top band. Table~\ref{tab:r-confidence}a gives its error rate in every band, and Appendix~\ref{app:decider} the intervals of the ELLIPSE AUROC and of FED-Dialogue's, which includes 0.5.

\paragraph{Binary panels.}
We computed no calibration error for the ordinal panels. On the binary panels, the ECE column of Table~\ref{tab:binary} and Figure~\ref{fig:binary-confidence}a give the calibration of P(MET) (Appendix~\ref{app:binary}).

%% file: appendix/j_shared.tex
\section{Repeated Errors and Juries}
\label{app:shared}

\begin{table}[!tbp]
\centering
\footnotesize
\setlength{\tabcolsep}{3pt}
\begin{tabular}{@{}lcrrrr@{}}
\toprule
Panel & Conf. & \makecell[r]{LLM\\repeats} & \makecell[r]{Ex-\\pected} & \makecell[r]{Same\\answer} & \makecell[r]{Unan.\\raters} \\
\midrule
ELLIPSE & 0.89--0.95 & 33 & 17.9 & 10 & 6 \\
FED-Turn & 0.98--1.00 & 35 & 15.3 & 11 & 0 \\
FED-Dialogue & 0.97--0.99 & 36 & 27.6 & 12 & 0 \\
HelpSteer2 & 0.90--0.99 & 35 & 16.6 & 11 & -- \\
LFQA & 0.94--1.00 & 36 & 18.5 & 12 & 0 \\
USR-TC & 0.89--0.99 & 32 & 11.8 & 9 & 3 \\
USR-PC & 0.92--1.00 & 35 & 19.0 & 11 & 1 \\
\midrule
Total & & 242 & 126.7 & 76 & \\
\midrule
HealthBench & -- & 32 & 10.0 & -- & 11 \\
\bottomrule
\end{tabular}
\caption{The confident-error sample (Section~\ref{sec:shared-errors}). Conf.: range of Jev Choice's confidence over the 12 errors of a panel. LLM repeats: verdicts of Luna, Gemini, and DeepSeek that equal Jev's wrong answer, out of 36 per panel, and Expected: the number the independence baseline predicts. Same answer: errors on which all five judges give Jev Choice's answer, out of 12. Unan.\ raters: errors whose raters all gave the same level, out of 12 (undefined for HelpSteer2, which releases a single label). HealthBench: the separately sampled errors described below, with physician unanimity.}
\label{tab:confident-errors}
\end{table}

\begin{table*}[!tbp]
\centering
\footnotesize
\setlength{\tabcolsep}{5pt}
\begin{tabular}{@{}llrrlrrr@{}}
\toprule
 & \multicolumn{2}{c}{Most accurate matched judge} & \multicolumn{2}{c}{Median of three LLM judges} & \multicolumn{2}{c}{Median of four judges} & \\
\cmidrule(lr){2-3}\cmidrule(lr){4-5}\cmidrule(lr){6-7}
Panel & Judge & Acc. & $\Delta$ & 95\% interval & $\Delta$ lower & $\Delta$ upper & All wrong \\
\midrule
RiceChem & Jev Choice & 81.0 & $-2.8$ & $-5.4$, $-0.1$ & $-2.3$ & $0.0$ & -- \\
HealthBench & Gemini & 79.6 & $-2.2$ & $-4.7$, $+0.2$ & $-2.2$ & $-0.7$ & -- \\
\midrule
ELLIPSE & Gemini & 29.9 & $-12.8$ & $-14.8$, $-10.7$ & $-17.7$ & $-10.8$ & 64.0 \\
FED-Turn & Gemini & 65.3 & $-0.2$ & $-3.0$, $+2.5$ & $-2.7$ & $+0.8$ & 22.7 \\
FED-Dialogue & Gemini & 60.5 & $-3.1$ & $-7.4$, $+1.7$ & $-8.8$ & $0.0$ & 30.5 \\
HelpSteer2 & Gemini & 54.4 & $-8.0$ & $-12.1$, $-4.0$ & $-9.5$ & $-0.6$ & 31.5 \\
LFQA & Luna & 71.5 & $-2.0$ & $-5.1$, $+1.1$ & $-2.8$ & $-0.6$ & 17.8 \\
USR-TC & Gemini & 70.9 & $0.0$ & $-2.5$, $+2.5$ & $-0.6$ & $-1.1$ & 17.2 \\
USR-PC & Gemini & 74.6 & $-0.7$ & $-2.7$, $+1.3$ & $-1.3$ & $+1.0$ & 24.0 \\
\bottomrule
\end{tabular}
\caption{Median juries on the pairs all four matched judges scored with a verdict (post-hoc analysis). Most accurate matched judge: as defined under Juries, below, with its accuracy (\%). $\Delta$: jury accuracy minus that judge's, in points, and for the three-LLM median the 95\% interval of that difference over resampled units. All wrong: share (\%) of the ordinal pairs on which all four matched judges are wrong, so that picking a correct judge on every pair where one exists would reach 100 minus this value.}
\label{tab:juries}
\end{table*}

\paragraph{Sampling rule.}
The pool is every pair Jev Choice answered wrongly, abstentions excluded, with confidence as in Appendix~\ref{app:confidence}. Pairs are sorted by confidence, highest first, with ties broken by unit identifier and then by criterion name. The sample holds 84 errors, 83 on ordinal pairs and one on an embedded binary pair, a \texttt{\_nofact} pair of USR-TC \crit{uses\_knowledge}. On four panels (ELLIPSE, FED-Turn, HelpSteer2, and USR-TC) one further error shares the twelfth-highest confidence value and falls outside the sample. Selecting the tied errors in the order the pairs are stored gives the same LLM repeat counts on every panel.

\paragraph{Repeated errors.}
Luna repeats 81 of the 84 errors, Gemini 80, and DeepSeek 81 (Table~\ref{tab:confident-errors} gives the panels). Under the independence baseline the three would repeat 126.7 of their 252 verdicts. The expected count per panel is highest on FED-Dialogue, 27.6 of 36, and lowest on USR-TC, 11.8. Jev Score gives Jev Choice's answer on all 84, but it is the same model asked a different question, so it is not counted as corroboration.

\paragraph{The Decider's sample.}
Decider Choice's 84 confident errors, drawn by the same rule (Table~\ref{tab:r-confidence}c), get 218 repeats among the LLM judges' 252 verdicts, against 120.7 expected. On each panel but ELLIPSE the LLM judges repeat 30 to 36 of their 36 verdicts, where the baseline expects 13.7 to 20.4. On ELLIPSE they repeat 16 against 19.5 expected, and the count stays at or below the expected one in every condition of Table~\ref{tab:conditions}, down to 9 against 16.7 with higher-tier judges. Appendix~\ref{app:decider} describes these 12 errors and the LLM verdicts on them. Decider Noul's HealthBench sample, drawn like Jev Noul's below, gets 16 repeats against 8.9 expected, 13 of them on its six false positives and 3 on its six false negatives.

\paragraph{Composition.}
In all 12 ELLIPSE errors Jev Choice put the essay at level~1 (eight on \crit{grammar}, three on \crit{syntax}, and one on \crit{conventions}), and all five judges gave that level on 10 of them. Nine of these labels are level~2 and three are level~3. The two raters agreed on six, each time at level~2, and the other six labels are half-way means that the half-up rule rounded up, from ratings of 1 and 2 on three errors and of 2 and 3 on the other three. Every FED-Dialogue error is a ``No.'' prediction shared by all five judges, eight of them on \crit{depth}, and 11 of the 12 FED-Turn errors are ``No.'' predictions too. The five judges also agree on each of the 12 LFQA errors, ten on \crit{formality} and two on \crit{amount\_info}. On USR-PC the errors are 12 \crit{engaging} pairs on which Jev Choice chose ``Dull'' against a label of ``Somewhat interesting'' (nine) or ``Interesting'' (three). One of them has unanimous raters, and so do three USR-TC errors, the \texttt{\_nofact} pair among them. Five HelpSteer2 \crit{coherence} errors lie one level above the label, where all five judges chose the top level, ``Fully clear and internally consistent throughout.''

\paragraph{Split and unanimous pairs.}
One reading is that the LLM judges repeat these errors because Jev's confidence measures how clearly the criterion text, applied to the unit, points to one answer, so LLM judges given the same text reach that answer as well and all miss a label that departs from it. A second is that the errors sit on pairs whose raters disagreed, and that some of the apparent repeats reflect that disagreement. The last column of Table~\ref{tab:confident-errors} leaves the second reading open for most of the sample, since on the six panels with several raters per pair only 10 of the 72 errors have unanimous raters, and none of the FED or LFQA errors has. That reading cannot apply to those ten, the USR-TC \texttt{\_nofact} pair among them, or to HealthBench, whose physicians were unanimous on 11 of its 12 sampled errors. In the Decider's sample 5 of the 72 errors on those panels have unanimous raters, and its 12 HealthBench errors all have unanimous physicians.

\paragraph{When every matched judge is wrong.}
Table~\ref{tab:juries} also gives, over every ordinal pair the four matched judges all scored, the share on which all four are wrong. In Jev's copy it is 17.2\% to 31.5\% on every ordinal panel but ELLIPSE, where it is 64.0\%, so choosing a correct judge for each ELLIPSE pair would reach 36.0\%. Gemini, the best of Jev's matched judges there, reaches 29.8\% exact accuracy (29.9\% on the pairs all four scored), against 51.4\% for the rater reference (Table~\ref{tab:main}; Section~\ref{sec:agreement}). Of the panels with a rater reference, ELLIPSE is also the only one on which choosing a correct judge wherever one exists falls short of that reference. In the Decider's copy the share is 45.8\% on ELLIPSE and 11.9\% to 30.3\% elsewhere, and choosing a correct judge exceeds the reference on every such panel, ELLIPSE included (Table~\ref{tab:r-cascades}c).

\paragraph{Repetition in every band.}
The sample covers only the most confident errors. For all of Jev Choice's errors, Table~\ref{tab:confidence}c gives the shares of repeated errors by band, which Figure~\ref{fig:confidence}b plots, next to their independence baseline. The baseline rises with confidence on several panels, from 49.1\% in the 0.25--0.5 band to 63.7\% in the top band on ELLIPSE and from 24.8\% to 54.6\% on USR-PC. Part of the rise in Figure~\ref{fig:confidence}b therefore follows from each LLM judge's own tendencies given the criterion and the label. The excess over the baseline is nonetheless largest in the top band on every ordinal panel (Table~\ref{tab:confidence}c). Two variants leave the rise with confidence intact. Restricting the count to ordinal pairs changes the top band only on USR-TC, to 83.3\%. It lowers USR-TC's 0.5--0.75 band from 66.4\% to 58.3\%, because the LLM judges also give Jev's answer on its embedded binary errors there, and moves no other band of Figure~\ref{fig:confidence}b by more than 4.0 points. Counting LLM abstentions and judge-call failures as non-repeats lowers no band by more than 1.9 points. Table~\ref{tab:r-confidence}b gives the same bands for Decider Choice's errors.

\paragraph{HealthBench.}
HealthBench's confident errors were selected by a different rule and are reported separately. They are Jev Noul's six false positives with the highest probability of yes, from 0.91 to 0.96, and the six false negatives with the lowest, from 0.08 to 0.12. The three LLM judges give Jev's answer in 32 of their 36 verdicts on them, against 10.0 that the independence baseline expects (Table~\ref{tab:confident-errors}): Luna in 11, Gemini in 10, and DeepSeek in 11. Over all 92 of Jev Noul's HealthBench errors the three LLM judges repeat 55.8\% of their verdicts, against a baseline of 27.5\%. For Jev Choice's errors the repeat share is 58.4\% and the baseline 27.1\%.

\paragraph{Juries.}
In Jev's copy, the three LLM judges' median level (their majority on binary pairs) never beats the most accurate matched judge (chosen on the same pairs, which favors it) and trails it by up to 12.8 points (ELLIPSE), significantly so on ELLIPSE, HelpSteer2, and RiceChem (95\% intervals below zero). Jev Choice as a fourth juror never helps significantly, and the median gives Jev's wrong answer on 82 of the 84 confident errors (Table~\ref{tab:juries}). No median jury's interval lies above that judge in the other conditions (Tables~\ref{tab:wr-downstream}, \ref{tab:ef-downstream}, and~\ref{tab:ht-downstream}). The most accurate matched judge of Table~\ref{tab:juries} is the best single judge of the cascades (Section~\ref{sec:cascades}) on eight panels; on FED-Turn that role goes to Jev Score, which is not a matched judge. A median of four levels is not unique when the four judges split evenly, which happens on 7.4\% to 22.9\% of a panel's pairs. We therefore score the four-judge jury with both the lower and the upper median, so that the tie rule favors no judge (Table~\ref{tab:juries}). The upper median is the most accurate of the three juries on every panel except USR-TC. Its best result, on USR-PC, is 1.0 point above the most accurate matched judge, with an interval of $-2.3$ to $+4.3$ points. Restricted to ordinal pairs, the three-LLM median is above that judge only on USR-TC, by 1.4 points (interval $-2.8$ to $+5.1$). Of the two confident errors on which the three-LLM median does not give Jev's answer, it gives the label on a USR-TC pair and a third level on an ELLIPSE pair. The four-judge lower median is right on none of the 84 confident errors and the upper median on one.

\paragraph{Juries in the Decider's copy.}
Table~\ref{tab:r-cascades}c repeats Table~\ref{tab:juries} for the Decider's copy, with Decider Choice as the fourth juror. The three-LLM median falls significantly short of the most accurate matched judge only on ELLIPSE and HelpSteer2, where that judge is Decider Choice. In none of the four conditions is a median jury significantly more accurate than that judge, although in the higher tier the USR-TC upper median's interval starts at exactly zero (0.0 to $+5.9$). The three-LLM median returns the Decider's wrong answer on 73 of its 84 confident errors.

%% file: appendix/k_cascades.tex
\section{Cascades}
\label{app:cascades}

\begin{table*}[!tp]
\centering
\footnotesize
\setlength{\tabcolsep}{3pt}
\begin{tabular}{@{}llrrrrrrrrrrr@{}}
\toprule
 & & & & \multicolumn{3}{c}{Accuracy (\%)} & & \multicolumn{3}{c}{Cost} & & \\
\cmidrule(lr){5-7}\cmidrule(lr){9-11}
Panel & Fallback & $\tau$ & \makecell[r]{Deferred\\(\%)} & Cascade & Jev & Fallback & \makecell[r]{vs.\\fallback} & \makecell[r]{Cascade\\(\$)} & \makecell[r]{Fallback\\(\$)} & \makecell[r]{Ratio\\(\%)} & \makecell[r]{Reach\\(\%)} & \makecell[r]{$\tau = 0.75$\\(\%)} \\
\midrule
RiceChem & Luna & 0.00 & 0.0 & 80.1 & 80.1 & 77.8 & $+2.3$ & 0.006 & 0.284 & 2.0 & 0.0 & -- \\
 & \textbf{Gemini} & 0.12 & 12.1 & 80.6 & 80.1 & 76.1 & $+4.5$ & 0.286 & 2.321 & 12.3 & 0.0 & -- \\
 & DeepSeek & 0.12 & 12.1 & 80.5 & 80.1 & 79.2 & $+1.2$ & 0.052 & 0.382 & 13.5 & 0.0 & -- \\
\addlinespace
HealthBench & Luna & 0.16 & 22.2 & 80.0 & 77.3 & 70.4 & $+9.6$ & 0.062 & 0.226 & 27.5 & 0.0 & -- \\
 & \textbf{Gemini} & 0.22 & 33.5 & 81.5 & 77.3 & 79.6 & $+2.0$ & 0.535 & 1.560 & 34.3 & 16.7 & -- \\
 & DeepSeek & 0.12 & 16.7 & 80.3 & 77.3 & 76.4 & $+3.9$ & 0.071 & 0.349 & 20.2 & 0.0 & -- \\
\midrule
ELLIPSE & Luna & 0.75 & 87.3 & 14.1 & 13.4 & 14.1 & $+0.1$ & 0.626 & 0.690 & 90.8 & 80.6 & 14.1 \\
 & \textbf{Gemini} & 0.95 & 99.9 & 29.8 & 13.4 & 29.8 & $0.0$ & 9.469 & 9.457 & 100.1 & 99.9 & 28.0 \\
 & DeepSeek & 0.60 & 66.4 & 15.2 & 13.4 & 15.1 & $+0.1$ & 1.270 & 1.877 & 67.7 & 66.4 & 14.7 \\
\addlinespace
FED-Turn & Luna & 0.80 & 55.7 & 63.2 & 61.5 & 62.5 & $+0.7$ & 0.072 & 0.125 & 57.8 & 38.5 & 62.5 \\
 & \textbf{Gemini} & 0.75 & 49.0 & 65.5 & 61.5 & 65.2 & $+0.3$ & 0.804 & 1.636 & 49.2 & 49.0 & 65.5 \\
 & DeepSeek & 0.75 & 49.0 & 64.3 & 61.5 & 63.7 & $+0.7$ & 0.118 & 0.234 & 50.1 & 25.3 & 64.3 \\
\addlinespace
FED-Dialogue & Luna & 0.55 & 38.6 & 54.9 & 51.9 & 51.5 & $+3.4$ & 0.026 & 0.065 & 40.3 & 0.0 & 53.2 \\
 & \textbf{Gemini} & 1.00 & 99.6 & 59.7 & 51.9 & 59.7 & $0.0$ & 0.822 & 0.825 & 99.7 & 99.6 & 57.9 \\
 & DeepSeek & 0.25 & 4.3 & 53.2 & 51.9 & 50.2 & $+3.0$ & 0.009 & 0.181 & 4.9 & 0.0 & 51.1 \\
\addlinespace
HelpSteer2 & Luna & 0.30 & 14.5 & 51.1 & 48.6 & 38.4 & $+12.8$ & 0.029 & 0.163 & 17.6 & 0.0 & 41.8 \\
 & \textbf{Gemini} & 0.50 & 45.2 & 54.8 & 48.6 & 54.3 & $+0.6$ & 0.660 & 1.450 & 45.5 & 30.1 & 54.8 \\
 & DeepSeek & 0.30 & 14.5 & 50.9 & 48.6 & 46.3 & $+4.5$ & 0.052 & 0.324 & 16.0 & 0.0 & 47.4 \\
\addlinespace
LFQA & \textbf{Luna} & 0.75 & 52.2 & 72.1 & 68.7 & 71.5 & $+0.6$ & 0.063 & 0.112 & 56.7 & 34.4 & 72.1 \\
 & Gemini & 0.00 & 0.0 & 68.7 & 68.7 & 67.6 & $+1.1$ & 0.005 & 1.266 & 0.4 & 0.0 & 67.9 \\
 & DeepSeek & 0.00 & 0.0 & 68.7 & 68.7 & 65.1 & $+3.6$ & 0.005 & 0.271 & 1.8 & 0.0 & 65.9 \\
\addlinespace
USR-TC & Luna & 0.90 & 80.8 & 68.6 & 67.5 & 67.8 & $+0.8$ & 0.086 & 0.103 & 84.0 & 38.1 & 68.3 \\
 & \textbf{Gemini} & 0.75 & 59.2 & 70.8 & 67.5 & 70.6 & $+0.3$ & 0.717 & 1.206 & 59.4 & 59.2 & 70.8 \\
 & DeepSeek & 0.00 & 0.0 & 67.5 & 67.5 & 66.1 & $+1.4$ & 0.003 & 0.200 & 1.6 & 0.0 & 65.8 \\
\addlinespace
USR-PC & Luna & 0.55 & 32.4 & 74.9 & 70.2 & 74.2 & $+0.7$ & 0.027 & 0.077 & 35.7 & 28.4 & 73.6 \\
 & \textbf{Gemini} & 0.50 & 28.4 & 75.9 & 70.2 & 74.6 & $+1.3$ & 0.249 & 0.866 & 28.7 & 10.0 & 75.9 \\
 & DeepSeek & 0.35 & 13.0 & 72.9 & 70.2 & 69.9 & $+3.0$ & 0.022 & 0.147 & 14.7 & 0.0 & 70.2 \\
\bottomrule
\end{tabular}
\caption{Ledger of the Jev-first cascades for every panel and fallback judge (post-hoc analysis with oracle thresholds). Deferral follows the \emph{Threshold rules} below. Accuracies are over the cascade's pair set, embedded binary pairs included. vs.\ fallback: cascade minus fallback alone, in points, computed from unrounded accuracies. Cost: Jev's cost plus the deferred share of the fallback's panel cost, a linear estimate, with DeepSeek priced at the harness's estimate for the binary panels and at its billed cost for the ordinal panels; ratio: the cascade's cost divided by the fallback's. Reach: the smallest deferred share at which the cascade matches its fallback alone; the paragraph \emph{Reaching the fallback} below gives the cost of these matches for each panel's best LLM judge. $\tau = 0.75$: cascade accuracy at that fixed threshold on the ordinal panels. Bold: the fallback of each panel's best cascade.}
\label{tab:cascade-ledger}
\end{table*}

\begin{table*}[!tp]
\centering
\footnotesize
\setlength{\tabcolsep}{3.3pt}
\begin{tabular}{@{}llrrrrrrrrrr@{}}
\toprule
 & & & \multicolumn{4}{c}{Kept pairs} & & \multicolumn{4}{c}{Deferred pairs} \\
\cmidrule(lr){4-7}\cmidrule(lr){9-12}
Panel & Fallback & $\tau$ & $n$ & \makecell[r]{Jev\\(\%)} & \makecell[r]{Fallback\\(\%)} & J\,:\,F ($p$) & & $n$ & \makecell[r]{Jev\\(\%)} & \makecell[r]{Fallback\\(\%)} & J\,:\,F ($p$) \\
\midrule
RiceChem & Gemini & 0.12 & 720 & 83.2 & 78.1 & 63\,:\,26 ($1\times10^{-4}$) & & 99 & 57.6 & 61.6 & 25\,:\,29 (0.68) \\
HealthBench & Gemini & 0.22 & 270 & 86.3 & 83.3 & 18\,:\,10 (0.18) & & 136 & 59.6 & 72.1 & 15\,:\,32 (0.019) \\
\midrule
ELLIPSE & Gemini & 0.95 & 2 & 0.0 & 0.0 & 0\,:\,0 & & 1,546 & 13.5 & 29.9 & 30\,:\,284 ($5\times10^{-53}$) \\
FED-Turn & Gemini & 0.75 & 306 & 76.8 & 76.1 & 10\,:\,8 (0.81) & & 294 & 45.6 & 53.7 & 27\,:\,51 (0.009) \\
FED-Dialogue & Gemini & 1.00 & 1 & 100.0 & 100.0 & 0\,:\,0 & & 232 & 51.7 & 59.5 & 12\,:\,30 (0.008) \\
HelpSteer2 & Gemini & 0.50 & 193 & 56.0 & 54.9 & 14\,:\,12 (0.85) & & 159 & 39.6 & 53.5 & 14\,:\,36 (0.003) \\
LFQA & Luna & 0.75 & 171 & 86.5 & 85.4 & 2\,:\,0 (0.50) & & 187 & 52.4 & 58.8 & 15\,:\,27 (0.088) \\
USR-TC & Gemini & 0.75 & 147 & 80.3 & 79.6 & 2\,:\,1 (1.00) & & 213 & 58.7 & 64.3 & 24\,:\,36 (0.16) \\
USR-PC & Gemini & 0.50 & 214 & 76.2 & 74.3 & 10\,:\,6 (0.45) & & 85 & 55.3 & 75.3 & 6\,:\,23 (0.002) \\
\midrule
HelpSteer2 & Gemini & 0.30 & 301 & 54.5 & 56.1 & 27\,:\,32 (0.60) & & 51 & 13.7 & 43.1 & 1\,:\,16 ($3\times10^{-4}$) \\
\bottomrule
\end{tabular}
\caption{Kept and deferred pairs of each panel's best Jev-first cascade (post-hoc analysis); Table~\ref{tab:cascade-ledger} marks its fallback and oracle threshold. J\,:\,F counts the pairs only Jev got right against the pairs only the fallback got right, with the $p$-value of a two-sided sign test. The last row repeats HelpSteer2 at $\tau = 0.30$.}
\label{tab:cascade-split}
\end{table*}

\begin{figure*}[!tp]
\centering
\includegraphics[width=\textwidth]{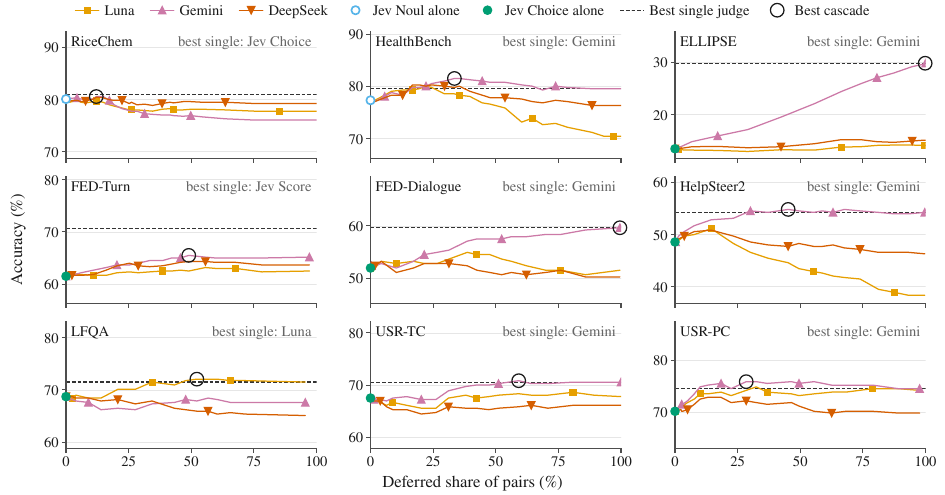}
\caption{Accuracy of the Jev-first cascades against the share of pairs deferred to each fallback judge (post-hoc analysis), one line per fallback. The marker at a deferred share of zero is Jev alone, in the framing that Section~\ref{sec:cascades} uses for the panel. The dashed line is the best single judge of Table~\ref{tab:cascade} with Jev first, named in each panel, and the black ring marks the panel's best cascade (bold in Table~\ref{tab:cascade-ledger}). Every panel spans the same range of accuracy points.}
\label{fig:cascade-curves}
\end{figure*}

\paragraph{Threshold rules.}
The replay uses the cascade and the framings of Section~\ref{sec:cascades}. On the ordinal panels the cascade threshold $\tau$ runs from 0 to 1 in steps of 0.05. On the binary panels a pair is deferred when its distance from one half, $|P(\text{yes}) - 0.5|$, which is half the Noul confidence of Section~\ref{sec:jev}, is below $\tau$, and there $\tau$ runs from 0 to 0.5 in steps of 0.02. The pair set is that of Appendix~\ref{app:tests}. Where the fallback's request failed, the cascade also keeps Jev's verdict. For each fallback, the reported oracle threshold is the $\tau$ with the highest accuracy, with ties going to the lower deferred share. A panel's best cascade is the one whose fallback, at its oracle threshold, gives the highest accuracy (bold in Table~\ref{tab:cascade-ledger}). With Jev first, its fallback is Gemini on every panel but LFQA, where it is Luna, and its deferred share runs from 12.1\% of the pairs on RiceChem to 99.9\% on ELLIPSE. Appendix~\ref{app:highertier} repeats the replay with each vendor's higher tier as fallback, gives the terms of Equation~\eqref{eq:identity} for each such fallback at its own oracle threshold on HelpSteer2 and HealthBench, and also measures those cascades against each panel's best single judge of either tier.

\paragraph{Decider-first cascades.}
With the Decider in Jev's place, these rules give the Decider-first cascades of Table~\ref{tab:cascade}, which put Decider Choice first on the ordinal panels and Decider Noul on the binary ones. The tables and figure of this appendix describe the Jev-first cascades, and Table~\ref{tab:r-cascades} gives the Decider-first ones, with their fallbacks, deferred shares, kept and deferred pairs, halvings, and held-out results.

\begin{table}[!tbp]
\centering
\footnotesize
\setlength{\tabcolsep}{4pt}
\begin{tabular}{@{}lrr@{}}
\toprule
Panel & \makecell[r]{Halvings with\\a gain (\%)} & \makecell[r]{Held-out\\difference} \\
\midrule
RiceChem & 6 & $+1.2$ \\
HealthBench & 80 & $-0.5$ \\
\midrule
ELLIPSE & 0 & $-0.1$ \\
FED-Turn & 0 & $-0.7$ \\
FED-Dialogue & 40 & $-2.1$ \\
HelpSteer2 & 4 & $-1.8$ \\
LFQA & 52 & $-0.7$ \\
USR-TC & 30 & $-1.1$ \\
USR-PC & 76 & $-0.2$ \\
\bottomrule
\end{tabular}
\caption{Cross-fitted Jev-first cascades of Table~\ref{tab:cascade} (post-hoc analysis). Halvings with a gain: share of the 50 halvings whose cascade beats the best single judge. Held-out difference: the cascade that matches the panel's best LLM judge on one half, minus that judge, in points of accuracy on the other half, averaged over the halves; Table~\ref{tab:cascade} gives its cost.}
\label{tab:cascade-halvings}
\end{table}

\paragraph{Cross-fitted thresholds.}
Each of the 50 halvings of Table~\ref{tab:cascade} splits a panel's units into two halves with a seeded permutation. Each half in turn makes four choices on its own pairs: (1) a threshold for each fallback, (2) the best cascade's fallback and threshold under the rule above, (3) the best single judge, and (4) the smallest threshold at which a cascade with that half's best LLM judge as fallback matches that judge. The other half is scored with these choices, and a halving pools its two scored halves. The same code reproduces Tables~\ref{tab:cascade} and~\ref{tab:cascade-ledger} when it makes the oracle-threshold choices. Table~\ref{tab:cascade-halvings} gives each panel's share of halvings with a gain and the held-out result of choice (4). With Jev first, cross-fitted gains over the fallback alone stay large only where the fallback is far less accurate than the panel's best LLM judge, 12.2 points over Luna on HelpSteer2 and 8.3 over Luna on HealthBench. With each panel's best LLM judge as fallback they run from $-2.1$ to $+1.3$ points. Jev Score is the best single judge in 99 of the 100 FED-Turn halves. On the six panels on which oracle thresholds match the best LLM judge for 10\% to 59\% of that judge's cost (see \emph{Reaching the fallback}), the cross-fitted cascade matches that judge on the held-out half in only 46\% to 64\% of halves. When the fallback is the best LLM judge, cross-fitted Jev-first cascades cost 16\% to 48\% as much as that judge on these six panels for 0.2 to 1.8 points less held-out accuracy, and 6.6\% on RiceChem, where Jev by itself beats it, for 1.2 points more (Tables~\ref{tab:cascade} and~\ref{tab:cascade-halvings}).

\paragraph{The identity.}
A cascade and its fallback agree on the deferred pairs, so
\begin{equation}
a_{\text{cascade}} - a_{\text{fallback}} = s_{\text{kept}} \bigl(a^{\text{kept}}_{\text{Jev}} - a^{\text{kept}}_{\text{fallback}}\bigr),
\label{eq:identity}
\end{equation}
with $a$ denoting accuracy and $s_{\text{kept}}$ the kept share. Let $a^{\text{def}}_{\text{fallback}}$ be the fallback's accuracy on the deferred pairs. On ordinal panels the fallback's accuracy counts Jev's verdict wherever the fallback abstained. The accuracies of the cascade and of its fallback are
\begin{align*}
a_{\text{cascade}} &= s_{\text{kept}}\,a^{\text{kept}}_{\text{Jev}} + (1-s_{\text{kept}})\,a^{\text{def}}_{\text{fallback}},\\
a_{\text{fallback}} &= s_{\text{kept}}\,a^{\text{kept}}_{\text{fallback}} + (1-s_{\text{kept}})\,a^{\text{def}}_{\text{fallback}},
\end{align*}
and their difference is Equation~\eqref{eq:identity}, in which $a^{\text{def}}_{\text{fallback}}$ cancels. Equation~\eqref{eq:identity} thus bounds a cascade's gain over its fallback by Jev's lead over the fallback on the kept pairs, and the fallback's lead on deferred pairs, which lifts a cascade above Jev, never lifts it above the fallback. Recomputing both sides from the recorded verdicts gives equal values at the reported $\tau$ on all nine panels.

\paragraph{Kept and deferred pairs.}
Table~\ref{tab:cascade-split} applies the identity to each panel's best Jev-first cascade. Against that cascade's fallback, Jev alone is right on only 12 to 37 pairs of an ordinal panel (Table~\ref{tab:paired}), and on kept pairs the two are mostly right or wrong together. The fallback leads Jev on the deferred pairs of every panel, significantly on six in two-sided sign tests (all but RiceChem, LFQA, and USR-TC). The five ordinal panels where Jev keeps more than two pairs are also the five on which the best cascade beats its fallback. On their kept pairs Jev leads the fallback by 0.7 to 1.9 points, and no lead is significant ($p \geq 0.45$). Jev's confidence keeps few of the pairs on which Jev is right and the fallback wrong. It keeps none of them on ELLIPSE and FED-Dialogue, a minority on FED-Turn, LFQA, and USR-TC, and half on HelpSteer2, whereas on USR-PC, the ordinal panel with the largest gain, 10 of the 16 such pairs are kept. Correlated errors make the cascades cheap. Where the fallback would mostly repeat Jev's errors, keeping Jev's confident verdicts costs little accuracy, which is why most cross-fitted cascades avoid most of the best LLM judge's cost (Table~\ref{tab:cascade}).

RiceChem is the one panel where Jev leads significantly on its kept pairs (63 to 26, $p = 1 \times 10^{-4}$). The best RiceChem cascade defers to Gemini, which Jev Noul outscores, so the cascade gains 4.5 points over Gemini. It still falls below Jev Choice alone. HealthBench's 2.0-point gain over Gemini rests on a kept-pair lead of 18 to 10 ($p = 0.18$).

To illustrate the identity, the last row of Table~\ref{tab:cascade-split} repeats HelpSteer2 with Gemini at $\tau = 0.30$, the oracle threshold on this panel with Luna or DeepSeek as fallback. Gemini wins the deferred pairs 16 to 1 ($p = 3 \times 10^{-4}$), an advantage that drops out of the cascade's difference from Gemini because the cascade gives Gemini's own verdicts there. On the kept pairs Jev is behind Gemini (54.5\% against 56.1\%), so the cascade (52.8\%) ends below Gemini alone (54.3\%).

\paragraph{Fixed threshold and flat curves.}
The last column of Table~\ref{tab:cascade-ledger} gives cascade accuracy at a fixed $\tau = 0.75$ on every ordinal panel. For the best cascade's fallback this is already the oracle threshold on FED-Turn, LFQA, and USR-TC, and it matches the oracle accuracy on HelpSteer2 and USR-PC, whose oracle threshold is 0.50. The fixed threshold falls below Gemini alone on ELLIPSE and FED-Dialogue. Near their maxima the curves are flat (Figure~\ref{fig:cascade-curves}). With Gemini as fallback, accuracy stays within 1.4 points of the oracle threshold's for every $\tau$ from 0.55 up on FED-Turn, from 0.30 up on USR-PC, and from 0.40 up on HelpSteer2.

\paragraph{Reaching the fallback.}
With the best LLM judge of each panel as fallback and oracle thresholds, the reach of Table~\ref{tab:cascade-ledger} stays below 60\% on HealthBench and on the five ordinal panels other than ELLIPSE and FED-Dialogue. Those two need nearly every pair deferred, because Gemini's lead over Jev Choice is separated on both and, on ELLIPSE, Jev Choice's confidence does not rank its errors. As a share of that judge's cost, with Jev's own cost added to the deferred share, these matches cost 17.5\% on HealthBench, 49.2\% on FED-Turn, 30.5\% on HelpSteer2, 38.8\% on LFQA, 59.4\% on USR-TC, and 10.3\% on USR-PC, against 100.1\% on ELLIPSE and 99.7\% on FED-Dialogue. Where Jev alone is at least as accurate as a fallback, as for all three RiceChem fallbacks, the reach is zero, so on RiceChem Jev alone already matches the best LLM judge, DeepSeek. On FED-Turn, Jev Score alone reaches 70.7\% on the cascade pair set, above every cascade, at about 1/650 of Gemini's cost.

\paragraph{Criterion routing.}
LFQA's cascade does not reduce to sending one criterion to the fallback. Routing every \crit{factuality} pair to Luna and keeping Jev Choice's verdicts elsewhere gives 69.8\%, below both the confidence cascade (72.1\%) and Luna alone (71.5\%). Routing any one or two of the three criteria to Luna gives at most 70.9\%, reached with \crit{factuality} and \crit{amount\_info} together.

\paragraph{Cost.}
The linear cost estimate of Table~\ref{tab:cascade-ledger} assumes that a deferred pair costs the fallback's average per pair. Deferred pairs may be longer or shorter than average and need more or fewer reasoning tokens.

%% file: appendix/l_binary_obs.tex
\section{Binary-Panel Expectations Re-examined}
\label{app:binaryobs}

Several expectations formed on the two binary panels did not hold on the ordinal panels, and others held only in narrower forms (Table~\ref{tab:binary-observations}). The direction of Jev's errors, for instance, reverses between the two kinds of panel (see the last paragraph of this appendix), and Jev Choice, which never chose CANNOT\_ASSESS on the binary panels, abstained on 28 ordinal pairs. Besides the scale, the two binary panels share checklist criteria (Section~\ref{sec:panels}), expert labels, and one wording of the abstain option. Because the ordinal panels also change the domain, the options Jev receives, and Jev's instruction, we attribute no failure to scale type alone.

\begin{table*}[!tp]
\centering
\footnotesize
\setlength{\tabcolsep}{4pt}
\renewcommand{\arraystretch}{1.0}
\begin{tabular}{@{}>{\raggedright\arraybackslash}p{4.1cm}>{\raggedright\arraybackslash}p{2.2cm}>{\raggedright\arraybackslash}p{8.9cm}@{}}
\toprule
Expectation formed on the binary panels & Status on the ordinal panels & Evidence \\
\midrule
Jev costs two orders of magnitude less than an LLM judge. & holds in narrower form & Nine-panel cost ratios are 29$\times$ (Luna), 66$\times$ (DeepSeek), and 325$\times$ (Gemini), and single-panel ratios span 18$\times$ to 770$\times$ (Table~\ref{tab:cost}). A factor of 100 or more holds over the nine panels only for Gemini, and on single panels also for DeepSeek on FED-Dialogue (169$\times$). \\
\addlinespace[2pt]
No LLM judge is significantly more accurate than Jev. & fails & Unit-level intervals separate Gemini's leads over Jev Choice on ELLIPSE, FED-Dialogue, and HelpSteer2 and Luna's lead on USR-PC, and the ELLIPSE and FED-Dialogue leads survive Holm correction over 27 comparisons. \\
\addlinespace[2pt]
Jev over-grants MET. & reversed & On ordinal criteria 31 of the 35 judge-by-panel offsets are negative, including Jev Choice's on six of seven panels. On embedded binary criteria the sign varies with the criterion, as on the two USR-TC criteria discussed at the end of this appendix. \\
\addlinespace[2pt]
Jev's confidence predicts which of its verdicts are wrong. & holds in narrower form & AUROC 0.57 to 0.70 on six ordinal panels and 0.49 on ELLIPSE, with failures on single criteria such as LFQA \crit{factuality} (0.41) and USR-PC \crit{engaging} (0.52) (Appendix~\ref{app:confidence}). \\
\addlinespace[2pt]
The primitive used to ask Jev does not change its verdicts. & fails & Unit-level intervals separate Jev Score ahead of Jev Choice on ELLIPSE, FED-Turn, FED-Dialogue, and USR-TC, by up to 10.5 points (Table~\ref{tab:clustered}). Jev Noul was not run on ordinal criteria, and the two binary framings agree at $\kappa = 0.945$ (Appendix~\ref{app:binary}). \\
\addlinespace[2pt]
Jev never abstains. & fails & Jev Choice abstained 28 times, 17 on FED-Dialogue \crit{error\_recovery}, where P(NA) follows the raters' N/A count (Appendix~\ref{app:abstention}). It never chose the epistemic CANNOT\_ASSESS option in 8,798 binary pairs. \\
\addlinespace[2pt]
Jev's most confident errors are mostly label errors. & undetermined & Of the 72 sampled errors from multi-rater panels, 10 have unanimous labels, six of them on ELLIPSE, and unanimous labels cannot be read as label noise. The USR-TC \texttt{\_nofact} labels, one of which is in the sample, are unanimous on 17 of 18 (Table~\ref{tab:conventions}; Appendix~\ref{app:shared}). The other 62 errors, every FED and LFQA one among them, sit on split pairs, where the recorded verdicts cannot tell a wrong label from a contested one. On HealthBench the physicians were unanimous on 11 of Jev Noul's 12 confident errors. \\
\addlinespace[2pt]
Other judges repeat Jev's confident errors. & holds & The three LLM judges repeat 242 of 252 verdicts on the ordinal panels and 32 of 36 on HealthBench, against 126.7 and 10.0 under the independence baseline (Section~\ref{sec:shared-errors}; Appendix~\ref{app:shared}). \\
\addlinespace[2pt]
Criterion length does not predict Jev's per-criterion accuracy. & holds & Spearman $-0.10$ on FED-Turn (7 criteria) and $+0.30$ on FED-Dialogue (9), against $-0.26$ and $-0.27$ on RiceChem and HealthBench. The test is weak, with few criteria and narrow length ranges, and ELLIPSE is ambiguous (Appendix~\ref{app:conventions}). \\
\addlinespace[2pt]
A second judge helps where Jev's uncertainty can be resolved. & holds in narrower form & In the post-hoc cascades the fallback beats Jev on every panel's deferred pairs, yet the best cascade on an ordinal panel exceeds the best single judge's accuracy by at most 1.3 points with oracle thresholds, and by at most 0.8 with cross-fitted ones. On five ordinal panels an oracle-threshold cascade reaches the best LLM judge for 10--59\% of that judge's cost, and ELLIPSE and FED-Dialogue need at least 99.6\% deferral (Appendix~\ref{app:cascades}). \\
\addlinespace[2pt]
The verdict-definition wrapper matters. & not examined & No framing for ordinal criteria used a wrapper (Section~\ref{sec:jev}). Its binary-panel gains, $+10.7$ and $+1.0$ points, are in Appendix~\ref{app:binary}. \\
\addlinespace[2pt]
Jev is stable across runs. & not examined & Jev was re-run only on the binary panels, where changes were confined to pairs near the threshold (Appendix~\ref{app:protocol}). \\
\addlinespace[2pt]
Jev has no non-English penalty. & not examined & All seven ordinal panels are in English, and the binary-panel check rests on 36 non-English HealthBench pairs (Appendix~\ref{app:binary}). \\
\bottomrule
\end{tabular}
\caption{Each expectation from the binary panels and its status on the seven ordinal panels. \emph{holds}: the ordinal panels agree. \emph{holds in narrower form}: they agree within the scope given in the evidence. \emph{reversed}: they show the opposite direction. \emph{fails}: they contradict it. \emph{undetermined}: the ordinal panels as run do not decide it. \emph{not examined}: the ordinal panels as run could not test it. Cascade results are post-hoc analysis.}
\label{tab:binary-observations}
\end{table*}

The expectations in the first column of Table~\ref{tab:binary-observations} are those we formed on the binary panels and wrote down before any judge was run on the ordinal panels (see \ref{sec:limitations}). They concern Jev alone. The Decider, added after these analyses, played no part in forming them, and the claims we checked on it were recorded separately before it had judged any study item (Appendix~\ref{app:decider}). The six status labels, and the assignment of each expectation to one of them, were decided after the ordinal panels had been analyzed. The evidence column gives the statistic behind each status, and a reader can apply a stricter or a looser rule to it. Since the two kinds of panel differ in several ways at once, a status records whether an expectation carried over and leaves open which difference decided the outcome.

\paragraph{Binary and ordinal criteria.}
The fourth natural contrast of Appendix~\ref{app:conventions} varies the form of the criterion, although it also changes the panel, the domain, and the label source. For Jev, moving from binary to ordinal criteria reverses the direction of the error. Jev over-grants MET on both binary panels (Appendix~\ref{app:binary}), on which the LLM judges' MET-rate offsets have no common sign (Table~\ref{tab:offsets}). On ordinal criteria Jev's offsets are negative outside USR-TC. The two embedded binary criteria of USR-TC split in direction. Every judge over-grants MET on \crit{understandable} (Jev Noul's MET rate is 90.3\% against 68.1\% labeled) and under-grants it on \crit{uses\_knowledge} (37.5\% against 58.3\%). On the same responses, then, the direction of every judge's error follows the criterion, so Jev's reversal need not come from the change of form itself.

%% file: appendix/m_release.tex
\section{Release and Derived Statistics}
\label{app:release}

\paragraph{Release.}
\ifarrreview The release, provided for review as anonymized supplementary material and to be made public upon publication, contains\else The release contains\fi{} the code that runs every judge, the frozen panel samples, and one verdict record per judge and pair. A panel sample holds each unit as the judges received it, the criterion texts (with level descriptions on the ordinal panels), the label, and the individual rater scores behind it. A verdict record gives the judge's level or verdict, whether it abstained, and any judge-call failure. For a decision model it also holds the returned probabilities and confidence, and for the LLM judges the explanation each one returned. The per-panel summaries that the runners wrote, which record cost and wall time, accompany the verdict records.

\paragraph{Scripts.}
A driver script regenerates the paper's numbers and figures from these files, sending no request to any model. It runs per-panel analysis scripts and then the analysis script, which computes the values that the paper quotes from the records and the per-panel outputs. The analysis script writes them to a JSON file and a readable summary, which names for each value the file and key it is read from, or the rule that computes it. Two figure scripts draw every figure at its printed size from these outputs. One draws each copy of the study's figures, and the other draws Figures~\ref{fig:parity} and~\ref{fig:confidence}, which set the two decision models side by side; Figure~\ref{fig:whole-rubric} also reads the judging-mode comparison. The four comparison scripts described below likewise send no model request, and each records the source of every value it writes.

The analysis script checks itself against the per-panel analysis outputs. Its recomputed discordant counts for the 21 ordinal-panel tests between Jev Choice and the LLM judges and the 7 between Jev Choice and Jev Score, its confidence-band tables for the seven ordinal panels, and its 21 cascade curves on the ordinal panels all equal those that the per-panel analysis computed. The paired comparisons, group resampling, juries, cross-fitted cascades, and physician reference draw from one fixed seed, with a separate generator for each quantity. The code behind them recomputes pair counts, shares of repeated errors, rater references, oracle-threshold cascades, and physician comparisons, and each equals the value computed without resampling. The ELLIPSE marginal intervals of Appendix~\ref{app:tests} use a seed of their own and one generator shared by the judges.

\paragraph{Inputs beyond the panel records.}
Some values need records other than the panel verdicts and the per-panel summaries. The LLM judges' cached responses hold the per-response costs that sum to DeepSeek's billed cost (Appendix~\ref{app:cost}), the reasoning-token counts, the upstream providers that served DeepSeek, and the reasoning effort that each request of the effort runs carried. They also show which twelve Luna verdicts were served from the cache (Appendix~\ref{app:protocol}).\ifarrreview{} In the supplementary material each cache is replaced by an extract that keeps only the fields the scripts read, and the request parameters and per-request latencies are not among them.\fi{} Jev's runs outside the panels (the full RiceChem set, the five-pass stability study, and the HealthBench ideal-completion ablation; Appendices~\ref{app:protocol} and~\ref{app:binary}) have records of their own. So do Luna's two spot-check runs.

\paragraph{Whole-rubric runs.}
The whole-rubric runs of Appendix~\ref{app:wholerubric} form a second results tree in the release, with the same layout as the first. Its Jev runs, panel samples, and rater files are links into the per-criterion tree, so only the three LLM judges' verdict records, run summaries, and cached responses are new. The repeat runs form a third tree, which also keeps DeepSeek's failed first attempt on ELLIPSE. The release lists the units of the repeat runs and of the smoke test. A setup script for these runs builds the second tree, draws the two sets of units, and reads each judge's logged seed from the per-criterion records. A launch script runs the three judges on the smoke-test units, the full panels, and the repeat units, and a comparison script reads the three trees and writes the judging-mode comparison, like the analysis script, as a JSON file with a readable summary. A single setting tells the scripts which results tree to read. With the setting pointed at the second tree, the scripts that summarize each run, the analysis script, and the figure script run unchanged on the whole-rubric verdicts, and with it unset they reproduce the per-criterion outputs exactly.

\paragraph{Effort runs.}
Four more trees, laid out like the first, hold the effort and calibration runs of Appendix~\ref{app:effort}, one tree each for the high runs, DeepSeek's medium run, the calibration's explicit-medium runs, and the calibration's second default runs, the last two on the repeat units. Each tree records the effort its requests named, and each calibration tree also records which units it covers. As in the second tree, links into the per-criterion tree supply Jev's runs, the panel samples, and the rater files, and the medium tree also links Luna's and Gemini's main runs as their medium runs. The medium tree's response caches contain DeepSeek's responses only, so the values read from caches, the reasoning and cached prompt tokens, are absent for those two linked runs and are reported as not applicable rather than as zero. When a setting names an effort, the code that configures the LLM judges passes it to the harness, which adds it to every request. Without that setting the requests are unchanged. Another launch script runs the three judges for each tree, and the setup script builds these trees too, links included. A second comparison script reads all the trees and compares the runs, the calibration, accuracy, and the paper's statistics across conditions. With the setting pointed at the high or the medium tree, the analysis and figure scripts run unchanged on that condition, and these reruns supply the agreement and offset counts that Table~\ref{tab:ef-downstream}c gives for the medium and high conditions.

\paragraph{Higher-tier runs.}
The higher-tier runs of Appendix~\ref{app:highertier} fill a further tree, again of the same layout and linked into the per-criterion tree for Jev's verdicts, the panels, and the rater files. The tree records which model replaced each flash-tier judge, the seeds read from that judge's records, and the settings of the runs. A log in the tree gives the start, end, and exit status of every session of every run, with a note on each of the two stopped sessions, and the summed session times are formed from it. A third launch script runs the three higher-tier judges, and a third comparison script reads this tree, the per-criterion tree, and the high tree. It compares every higher-tier judge with its flash sibling, as run and at high effort, on accuracy, repeated errors, cascades, and juries, the cascades also against the per-panel best single judge of either tier. For each run, the pilot included, it records the tokens, the costs (as recorded, as billed, and at list prices), the wall times, and their ratios to Jev Choice and to the flash sibling. The analysis and figure scripts need no change to cover the higher-tier judges once pointed at this tree.

\paragraph{Decider runs.}
The Decider's runs of Appendix~\ref{app:decider} have a results tree of their own, laid out like the first. They keep the names of Jev's runs, so that the analysis, figure, and comparison scripts read them unchanged, and in this tree such a name means the Decider in that framing. Each run's record names the grader that sent the Decider's requests, and no run in the tree links into one of Jev's. A short declaration in the tree gives the model identifier, the price, the documented rate limit, and the client's settings; it was frozen after the runs and before any analysis read it. The requests went through a small client that sends the request bodies built for Jev to Perplexity's endpoint and keeps the request rate under the documented limit. It retries, within a fixed budget, attempts that hit the rate limit, time out, or fail on the server's side, and it logs every attempt with its send time, wait in the queue, latency, status, reported model, and input tokens. A summary step turns these request logs into each run's counts of requests, attempts, retries, and rate-limit responses, together with its latencies, the share of time spent waiting for the rate limit, and its dates. Four more Decider trees pair the Decider's runs, linked from this tree, with the LLM judges' runs of the whole-rubric, high, medium, and higher-tier trees, and a sixth holds its repeat of the 106 units.\ifarrreview{} A seventh holds the smoke run and the log of the requests sent right after it.\fi{} A script pairs each repeated request with its first sending by the request body and compares the two answers. Pointed at a Decider tree, the analysis, figure, and comparison scripts produce the Decider's copy of the statistics they compute, and their JSON output keeps Jev's names as keys, which there stand for the Decider. A fourth comparison script reads both models' trees and sets their values side by side, key by key, with each model's run details, and it computes no difference, ratio, or test between the two models. It also checks that every value depending only on the LLM judges is identical in the two copies and that Jev's column reproduces Table~\ref{tab:conditions}.

\paragraph{Derived statistics.}
Differences between reported values are computed before rounding, and stated bounds such as ``at most'' compare rounded values. The analysis script derives the following statistics from the recorded verdicts, labels, and summaries, and marks the post-hoc analyses among them.\par
\begin{itemize}[leftmargin=*,nosep,before=\raggedright]
\item The paired comparisons with units resampled: bootstrap intervals, equivalence bounds, sign tests over units, Holm correction, and the resampling of sampling groups (Table~\ref{tab:clustered}).
\item Pair-level sign tests for Jev Choice, Jev Noul, and Jev Score, with the Holm correction of Jev Choice's tests (Table~\ref{tab:paired}).
\item The paired accuracy differences and per-panel cost ratios of Figure~\ref{fig:parity}, and DeepSeek's billed cost per panel.
\item Judge--judge and judge--label QWK (Table~\ref{tab:main}), and the all-wrong shares (Table~\ref{tab:juries}).
\item The per-criterion aggregates of Figure~\ref{fig:offsetgap}: the Spearman correlation over 31 criteria, the counts of criteria beyond the 15-point cut-off and of criteria with near-zero offsets, the sign agreement of the four offsets on 35 criteria, and the gap measures that ignore location (post hoc; Table~\ref{tab:location-gap}).
\item MET-rate offsets and Cohen's $\kappa$ on the binary panels, and MET rates on the embedded binary criteria.
\item For ELLIPSE, level usage, the sign structure of the errors, slopes, offsets by trait and by holistic band, accuracy on half-way labels, the post-hoc downward tie-break and constant-shift curves, and the marginal bootstrap intervals of Appendix~\ref{app:tests}.
\item The USR \crit{engaging} shift (post hoc) and rank correlations, the \texttt{\_nofact} counts, the decomposition of Jev Choice's USR-TC errors, and the rater reference split by rater unanimity.
\item Each judge's use of ``Somewhat.'' on FED-Turn, the exchange of decoding rules there (post hoc), and the number of Jev Score's exact halves (Appendix~\ref{app:choicescore}).
\item The relation between P(NA) and the raters' N/A answers on FED-Dialogue \crit{error\_recovery}, and the overlap of the judges' abstentions (Appendix~\ref{app:abstention}).
\item The fit of the confidence formula, AUROC per panel and per criterion, accuracy on the most confident pairs, repeated errors by band with their independence baseline, the confident-error sample, and the median juries (post hoc; Appendices~\ref{app:confidence} and~\ref{app:shared}).
\item For the cascades (post hoc), the oracle-threshold and cross-fitted results, the kept and deferred splits with their sign tests, accuracy at a fixed threshold, the deferred share needed to reach each fallback, cost estimates, and the LFQA criterion-routing check (Appendix~\ref{app:cascades}).
\item The HealthBench physician reference (Appendix~\ref{app:binary}).
\item The no-read baselines and their sign tests, criterion-length correlations on FED and ELLIPSE, accuracy by unit-length quartile, and the agreement between Luna's two spot-check runs.
\end{itemize}

\paragraph{Sums over panels.}
Several counts in the paper add per-panel values: the 31 negative offsets (of 35, five judges on seven panels), the 242 repeated verdicts (of 252, three LLM judges and 12 errors on each of seven panels), the 28 abstentions (of 3,414 ordinal pairs on seven panels), and the 8 separated, 8 at-parity, and 11 inconclusive comparisons (of the 27 pairings of Jev Choice and an LLM judge on the nine panels). The nine-panel costs and wall times are sums of the per-panel values, with DeepSeek at its billed cost. The class tallies, confident-error totals, and nine-panel sums of Appendices~\ref{app:effort}, \ref{app:highertier}, and~\ref{app:decider} are formed the same way.

\paragraph{What the scripts do not compute.}
Some statements in the paper are not produced by the analysis script. Values published by the source benchmarks, such as ELLIPSE's grade band and rater design, are cited to their papers. Facts about the decision models, such as their prices and the description of their primitives, come from the vendors' documentation, the Decider's backbone and training data from its public model card, and both models' Decision Index scores from the index itself. Other values were read or computed directly from the recorded per-pair results, rater files, or run logs:\par
\begin{itemize}[leftmargin=*,nosep,before=\raggedright]
\item the HealthBench strata of Table~\ref{tab:panel-construction}, the FED-Turn \crit{fluent} pair with four raters, the absence of ties among the embedded binary labels, and the twelve Luna verdicts served from the spot check's cached responses (Appendix~\ref{app:protocol});
\item the rater-reference rows for within-one and embedded binary accuracy in Table~\ref{tab:secondary};
\item the per-criterion wrapper effects on HealthBench (Appendix~\ref{app:binary});
\item the FED-Turn mean probabilities per level, the embedded binary disagreements between the two framings for ordinal criteria, the levels at which Jev Score's exact halves fall, and the agreement of the two rounding rules on Jev Choice's probabilities (Appendix~\ref{app:choicescore});
\item the labels of the \crit{error\_recovery} dialogues that Jev Choice abstained on (Appendix~\ref{app:abstention}) and the ratings behind the ELLIPSE confident errors (Appendix~\ref{app:shared});
\item for the Decider, its ELLIPSE offset against labels rounded down (Section~\ref{sec:agreement}), the LLM judges' verdicts on Decider Choice's ELLIPSE errors (Sections~\ref{sec:shared-errors} and~\ref{sec:cascades}), and the offsets of Decider Choice and the LLM judges on HelpSteer2 \crit{correctness} (Appendices~\ref{app:conventions} and~\ref{app:decider});
\item from the cached responses, the request latencies of Table~\ref{tab:ht-runs}, and in Appendices~\ref{app:wholerubric} and~\ref{app:highertier} the counts of prompt-cache hits and writes, the costs at uncached or list prices and their ratios to the bills, and the replies costed above off-peak rates, together with the system-prompt token counts of Appendix~\ref{app:wholerubric} (OpenAI's o200k\_base tokenizer);
\item in Appendix~\ref{app:decider}, the intervals over resampled units of Decider Choice's ELLIPSE and FED-Dialogue AUROCs and of its three positive offsets, a rerun with 200 fresh seeds of the bootstrap that compares it with Gemini on HealthBench, the repeats on a confident-error sample drawn from its ordinal pairs alone, the judges' verdicts on one HealthBench hedging criterion, and the mean probability and MET rate of Decider Noul's HealthBench answers in the highest and lowest probability bins;
\item from the Decider's request logs, the part of its Choice run's wall time that the rate limit accounts for (Section~\ref{sec:price}) and, in Appendix~\ref{app:decider}, what its rate-limit responses reported, the input tokens of the three attempts that timed out and the time that one of them added to the Choice run, the waiting share of the nine-panel Choice run as a whole, the replies that preceded the run and their timestamps, the delays of repeated requests, the answers to duplicate requests outside the repeat, and the request identifiers and latencies behind the check for a response cache.
\end{itemize}
The readings of individual pairs in Appendices~\ref{app:abstention}, \ref{app:conventions}, and~\ref{app:shared} quote the records they read and are not measurements.

%% file: appendix/n_whole_rubric.tex
\section{Whole-Rubric Judgments}
\label{app:wholerubric}

\begin{figure*}[!tp]
\centering
\includegraphics[width=\textwidth]{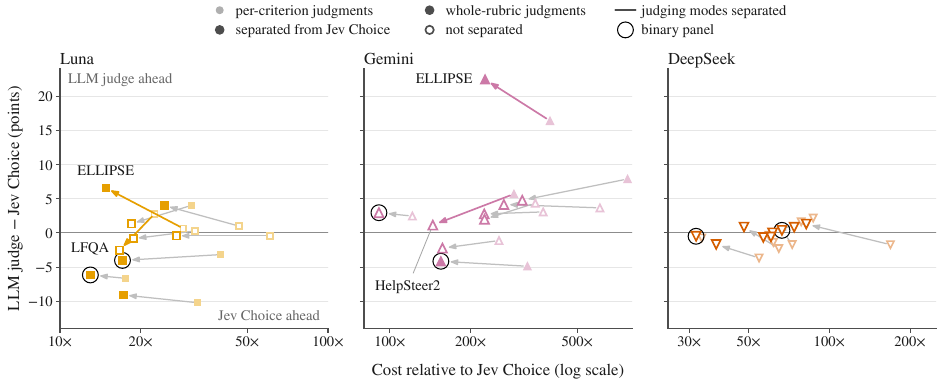}
\caption{Figure~\ref{fig:parity}a redrawn for both judging modes of Section~\ref{sec:llm}, one plot per LLM judge. Each logarithmic cost axis spans one decade, so equal cost ratios give arrows of equal horizontal length in every plot. An arrow runs from a judge's point on one panel with per-criterion judgments (small, pale marker) to its point with whole-rubric judgments (full-size marker). At either end a filled marker marks a comparison separated from Jev Choice in that judging mode and a hollow one a comparison that is not; rings mark the binary panels. The four colored, labeled arrows are the judge--panel combinations whose accuracy is separated between the two judging modes (Table~\ref{tab:wr-accuracy}); the other arrows are gray. Arrows too short for a head are drawn as plain lines or hidden under their marker.}
\label{fig:whole-rubric}
\end{figure*}

This appendix describes how the whole-rubric runs and the repeat runs of Section~\ref{sec:llm} were made and reports the whole-rubric results that Table~\ref{tab:conditions} condenses into one column. Figure~\ref{fig:whole-rubric} and the tables below also support the whole-rubric remarks of Sections~\ref{sec:shared} and~\ref{sec:routing}. Every per-criterion value in these tables is that of the main runs, and Jev's verdicts, costs, and times are the same in both modes. The tables give Jev's copy of each analysis. The Decider's verdicts do not change with the mode either, and Appendix~\ref{app:decider} gives its copy of this condition beside the other conditions of Table~\ref{tab:conditions}.

\paragraph{Prompt and reply.}
In the whole-rubric runs the harness sends one request per unit. The system prompt states the task and quotes as guides those system prompts of Table~\ref{tab:judge-inputs}, binary or multi-choice, that match the kinds of criterion the unit carries. It ends with the reply format: a JSON object with one judgment per criterion identifier, each holding a verdict or an option number and an explanation. The user message lists the criteria first, each tagged with its identifier and worded as in a per-criterion request, options included, and then gives the input and the submission once. Because the whole-rubric runs reused the model names and each judge's logged seed, and the option shuffle was keyed as in Appendix~\ref{app:protocol}, every criterion's options appeared in their recorded order. The units, model identifiers, temperatures, reasoning defaults, and the harness's retries were those of the main runs, the response cache was on, and requests timed out after 600 seconds.

\paragraph{Failed and incomplete replies.}
The harness reads a reply one criterion at a time. A judgment that is missing, duplicated, or unusable fails only its own criterion, as a parse failure, and the pair is excluded from every denominator like any other failed pair. A request that fails, or whose reply is not a readable JSON object, fails every pair of its unit, and the unit then has no verdict record at all, so none of its pairs can stand as UNMET (compare Appendix~\ref{app:protocol}). All of Luna's and Gemini's replies were complete. DeepSeek's five failed requests fell on RiceChem (two units, 14 pairs), HealthBench (one unit, 2 pairs), LFQA (one unit, 3 pairs), and USR-TC (one unit, 5 pairs). The four on RiceChem, HealthBench, and LFQA ran long and then returned nothing, and the USR-TC reply was cut short inside its JSON. DeepSeek's two single-criterion failures, both on ELLIPSE, were a missing and a duplicated judgment. The judges abstained about as often in either mode (Table~\ref{tab:wr-runs}), except DeepSeek on USR-TC, whose abstentions rose from 1 to 16.

\paragraph{Repeat runs.}
Each of the twelve runs, one per ordinal panel, one per RiceChem question, and one for HealthBench, contributed $\max(1, \lceil 0.1\,n \rceil)$ of its $n$ units, drawn without replacement by a generator seeded with a CRC-32 checksum of the run's name. All three judges repeated the same units: 14 on RiceChem, 20 on HealthBench, 26 on ELLIPSE, 8 on FED-Turn, 3 on FED-Dialogue, 9 on HelpSteer2, 12 on LFQA, 8 on USR-TC, and 6 on USR-PC, 529 pairs in all. The repeat runs started from an empty response cache and kept every other setting of the whole-rubric runs. DeepSeek's first repeat attempt on ELLIPSE failed on all 26 units with local network errors, before any request reached the model and with nothing billed. We reran those units with the same settings and keep the failed attempt in the release. One DeepSeek repeat unit on FED-Dialogue hit the 600-second timeout and was left failed, so DeepSeek's repeat covers 105 units and 519 pairs. A smoke test of three units per run (36 units) preceded the runs and enters no analysis.

\begin{table*}[!tp]
\centering
\footnotesize
\setlength{\tabcolsep}{4pt}
\begin{tabular}{@{}lrrrrrr@{}}
\toprule
 & \multicolumn{2}{c}{Luna} & \multicolumn{2}{c}{Gemini} & \multicolumn{2}{c}{DeepSeek} \\
\cmidrule(lr){2-3}\cmidrule(lr){4-5}\cmidrule(l){6-7}
 & Per-criterion & Whole-rubric & Per-criterion & Whole-rubric & Per-criterion & Whole-rubric \\
\midrule
\multicolumn{7}{@{}l}{\emph{(a) Nine panels}} \\
Requests & 5{,}003 & 1{,}021 & 5{,}003 & 1{,}021 & 5{,}003 & 1{,}021 \\
Failed requests & 0 & 0 & 0 & 0 & 9 & 5 \\
Failed pairs & 0 & 0 & 0 & 0 & 9 & 26 \\
Abstentions & 12 & 19 & 29 & 25 & 24 & 36 \\
Prompt tokens & 12{,}063{,}678 & 3{,}884{,}372 & 10{,}972{,}689 & 3{,}703{,}999 & 11{,}241{,}762 & 3{,}667{,}224 \\
Completion tokens & 866{,}600 & 572{,}284 & 3{,}295{,}232 & 2{,}331{,}522 & 4{,}254{,}957 & 2{,}935{,}010 \\
\quad of which reasoning & 502{,}270 & 252{,}742 & 2{,}981{,}135 & 1{,}957{,}145 & 3{,}946{,}419 & 2{,}604{,}817 \\
Cost (US\$) & 1.84 (29$\times$) & 1.01 (16$\times$) & 20.59 (325$\times$) & 11.52 (182$\times$) & 4.18 (66$\times$) & 3.40 (54$\times$) \\
\quad harness estimate & & & & & 3.14 (50$\times$) & 0.91 (14$\times$) \\
Wall time (s) & 859 (30$\times$) & 802 (28$\times$) & 950 (33$\times$) & 1{,}288 (45$\times$) & 6{,}298 (220$\times$) & 10{,}026 (350$\times$) \\
Requests at a time & 16\,/\,8 & 8\,/\,8 & 20\,/\,8 & 8\,/\,8 & 32\,/\,8 & 8\,/\,8 \\
\bottomrule
\end{tabular}

\medskip
\setlength{\tabcolsep}{3pt}
\begin{tabular}{@{}lrrrrrrrr@{}}
\toprule
 & & \multicolumn{4}{c}{Cost in US dollars} & \multicolumn{3}{c}{Wall time in seconds} \\
\cmidrule(lr){3-6}\cmidrule(l){7-9}
 & & & & \multicolumn{2}{c}{DeepSeek} & & & \\
\cmidrule(lr){5-6}
Panel & Units & Luna & Gemini & billed & estimate & Luna & Gemini & DeepSeek \\
\midrule
\multicolumn{9}{@{}l}{\emph{(b) Whole-rubric judgments by panel}} \\
RiceChem & 121 & 0.122 (17$\times$) & 1.104 (155$\times$) & 0.474 (67$\times$) & 0.116 & 144 (37$\times$) & 147 (38$\times$) & 907 (231$\times$) \\
HealthBench & 200 & 0.167 (13$\times$) & 1.171 (91$\times$) & 0.410 (32$\times$) & 0.110 & 151 (28$\times$) & 110 (21$\times$) & 2{,}610 (493$\times$) \\
\midrule
ELLIPSE & 258 & 0.354 (15$\times$) & 5.410 (226$\times$) & 1.467 (61$\times$) & 0.403 & 220 (30$\times$) & 617 (84$\times$) & 3{,}727 (506$\times$) \\
FED-Turn & 75 & 0.066 (25$\times$) & 0.717 (266$\times$) & 0.199 (74$\times$) & 0.052 & 47 (24$\times$) & 69 (36$\times$) & 803 (414$\times$) \\
FED-Dialogue & 25 & 0.029 (27$\times$) & 0.334 (312$\times$) & 0.088 (82$\times$) & 0.031 & 25 (30$\times$) & 46 (58$\times$) & 317 (394$\times$) \\
HelpSteer2 & 90 & 0.086 (17$\times$) & 0.721 (144$\times$) & 0.240 (48$\times$) & 0.061 & 69 (29$\times$) & 77 (33$\times$) & 374 (159$\times$) \\
LFQA & 120 & 0.083 (17$\times$) & 0.779 (157$\times$) & 0.188 (38$\times$) & 0.053 & 69 (20$\times$) & 91 (27$\times$) & 671 (199$\times$) \\
USR-TC & 72 & 0.061 (19$\times$) & 0.727 (225$\times$) & 0.196 (61$\times$) & 0.048 & 44 (22$\times$) & 69 (35$\times$) & 372 (190$\times$) \\
USR-PC & 60 & 0.046 (18$\times$) & 0.559 (225$\times$) & 0.141 (57$\times$) & 0.035 & 34 (22$\times$) & 60 (38$\times$) & 245 (156$\times$) \\
\midrule
Nine panels & 1{,}021 & 1.014 (16$\times$) & 11.521 (182$\times$) & 3.402 (54$\times$) & 0.908 (14$\times$) & 802 (28$\times$) & 1{,}288 (45$\times$) & 10{,}026 (350$\times$) \\
\bottomrule
\end{tabular}
\caption{The LLM judges' runs in both judging modes. (a) Nine-panel totals. Requests: those the harness issued; LiteLLM's internal retries are not recorded. Failed pairs: pairs whose judgment failed, which in the per-criterion mode are the pairs of DeepSeek's nine failed requests (three of them recorded as UNMET on the binary panels) and in the whole-rubric mode the 24 pairs of its five failed requests plus its two single-criterion failures on ELLIPSE. Abstentions: CANNOT\_ASSESS and NA answers, failures excluded. Completion tokens include the reasoning tokens, which are read from the cached responses. Cost: DeepSeek at its billed cost; the row below gives what the harness itself estimated for those requests. Parentheses give the ratio to Jev Choice (Table~\ref{tab:cost}). Requests at a time: ordinal\,/\,binary panels; Jev sent 8 on every panel. (b) Whole-rubric judgments by panel, one request per unit, with ratios to Jev Choice's run on that panel; DeepSeek's two costs as in Table~\ref{tab:cost}.}
\label{tab:wr-runs}
\end{table*}

\paragraph{Cost, tokens, and time.}
Whole-rubric judgments bring the requests of each LLM judge down from 5,003 to 1,021 (Table~\ref{tab:wr-runs}). Prompt tokens fall only to about a third, because a whole-rubric request, with its longer system prompt and every criterion of the unit (see \emph{Prompt and reply}), carries about 1.6 times the prompt tokens of a per-criterion one. Completion tokens fall by 29\% to 34\%, mostly through reasoning tokens, which drop by about half for Luna and a third for Gemini and DeepSeek. The rest of the completion tokens, which in either mode carry a verdict and an explanation for each criterion, fall by 12\% for Luna and rise by 19\% for Gemini and 7\% for DeepSeek. On the ordinal panels the cached whole-rubric replies hold 162{,}354 reasoning tokens for Luna, 1{,}635{,}010 for Gemini, and 1{,}939{,}805 for DeepSeek, against the per-criterion counts of Appendix~\ref{app:protocol}. Cost, computed for both modes by the rule of Section~\ref{sec:metrics}, falls by 45\% (Luna), 44\% (Gemini), and 19\% (DeepSeek). With 4.9 times as many requests, per-criterion judgments cost Luna 1.82 times as much as whole-rubric judgments, Gemini 1.79 times, and DeepSeek, by its bills, 1.23 times.

Like its per-criterion cost ratio, Gemini's whole-rubric cost ratio is smallest on HealthBench (91$\times$, 2.0 criteria per unit) and largest on FED-Dialogue (312$\times$, 10), but it drops from LFQA (157$\times$, 3) to HelpSteer2 (144$\times$, 4), and RiceChem (155$\times$, 6.8) sits below every ordinal panel with five or more criteria per unit. With whole-rubric judgments, a unit's median duration across the ordinal panels is 4.0 to 6.6 seconds for Luna, 4.9 to 15.7 for Gemini, and 6.5 to 30.1 for DeepSeek. The 90th percentile of DeepSeek's unit durations reaches 127.1 seconds on ELLIPSE and 160.2 on FED-Dialogue. Table~\ref{tab:wr-runs} gives wall times for both modes, but they do not measure the judging mode (see \ref{sec:limitations}). The whole-rubric runs sent as many requests at a time as Jev did. Whole-rubric Luna, Gemini, and DeepSeek cost 16, 182, and 54 times as much as Jev Choice and, at that concurrency, took 28, 45, and 350 times as long (Table~\ref{tab:wr-runs}a). Equal concurrency leaves one difference between the runs that favors the LLM judges in these time ratios. Jev Choice and Jev Score shared one API, while each LLM judge had its own provider.

\paragraph{Why cost falls less than requests.}
Completion tokens, which fall least, make up most of each judge's cost. At the list prices of Appendix~\ref{app:cost}, which reproduce the recorded costs of Luna and Gemini, they account for 56\% of Luna's per-criterion cost and 68\% of its whole-rubric cost, and for 60\% and 76\% of Gemini's. OpenRouter splits the cost of each DeepSeek response between prompt and completion, and by its bills completion tokens took 79\% of DeepSeek's per-criterion cost and 88\% of its whole-rubric cost.

Prompt caching lowered the price of the system prompt that every per-criterion request repeats. The harness opens each per-criterion request with a fixed system prompt, one for binary and one for multi-choice criteria (Table~\ref{tab:judge-inputs}), of 1{,}629 and 1{,}642 tokens as OpenAI's o200k\_base tokenizer counts them. A provider's prompt cache can serve such an opening, at a lower price, to later requests that begin with it. Luna's provider served 79.7\% of its per-criterion prompt tokens from its prompt cache and DeepSeek's providers 67.5\% (Table~\ref{tab:ef-runs}), against 72.0\% and 57.6\% of their whole-rubric prompt tokens. Gemini reported no cached tokens in either mode.

On all but three of Luna's per-criterion requests, the prompt cache served no more than the opening that the requests for one kind of criterion share. Of the per-criterion replies in Luna's response cache, 1{,}588 of the 1{,}589 for binary criteria report exactly 1{,}979 cached prompt tokens, and 3{,}480 of the 3{,}484 for multi-choice criteria exactly 1{,}895 (counts that include the 70 spot-check replies of Appendix~\ref{app:protocol} that the main run did not reuse). These counts, a few hundred tokens above the length of the system prompt, did not change with the unit or with its number of criteria, which on the binary panels ran from one to eight. Because each request gives its criterion before the input and the submission, it shared little more than that opening with the requests for the unit's other criteria, and all but three of Luna's per-criterion requests paid for the unit's text in full. OpenAI billed more than 99\% of Luna's uncached prompt tokens in either mode as writes to its prompt cache, at \$0.25 per million tokens instead of the uncached \$0.20 (Appendix~\ref{app:cost}), so Luna paid more for the unit's text than it would have without the cache. Whole-rubric requests also open with a fixed system prompt for each mix of criterion kinds (see \emph{Prompt and reply}), and on 985 of Luna's 1{,}021 whole-rubric requests the prompt cache served a constant opening of 2{,}305, 2{,}324, or 4{,}054 tokens.

Had Luna paid its uncached input price for every prompt token, its per-criterion run would have cost \$3.45 instead of \$1.84 and its whole-rubric run \$1.46 instead of \$1.01, so that per-criterion judgments would have cost 2.36 times as much as whole-rubric ones instead of 1.82 times. At DeepSeek's list prices, its per-criterion run would have cost \$3.12 with its cached tokens and \$4.24 without them, and its whole-rubric run \$2.00 and \$2.31, which puts the per-criterion run at 1.56 times the whole-rubric run's cost with the prompt cache and 1.83 times without it. Caching thus brought the cost of per-criterion judgments closer to that of whole-rubric ones for Luna and DeepSeek. Even so, without the prompt cache the 4.9-fold rise in requests would have raised Luna's cost only 2.36-fold, and Gemini's cost, with no cached tokens, rose 1.79-fold, close to Luna's 1.82. We think the gap between requests and cost comes mostly from the completion tokens, which a judge spends on every criterion in either mode, and otherwise from the longer whole-rubric prompt and, for Luna and DeepSeek, from the prompt cache.

On panels with more criteria per unit, per-criterion judgments tend to cost more relative to whole-rubric ones, though far less than in proportion to their extra requests. On single panels, per-criterion judgments cost Luna 1.35 (HealthBench and LFQA) to 2.33 times (RiceChem) as much as whole-rubric judgments and Gemini 1.33 (HealthBench) to 2.47 times (FED-Dialogue), while sending 2.0 to 10.0 times as many requests (Tables~\ref{tab:cost} and~\ref{tab:wr-runs}b). By DeepSeek's bills, they cost it 1.02 (USR-TC) to 2.06 times (FED-Dialogue) as much, a range that may also reflect the routing described below.

\paragraph{DeepSeek's bill in the two modes.}
To split DeepSeek's billed cost by panel, each cached whole-rubric reply is matched to its panel through the explanations it shares with the verdict records, and every successful reply found a match. For the whole-rubric requests the billed cost is 3.7 times the harness's estimate over the nine panels, and 2.8 (FED-Dialogue) to 4.1 times (USR-TC) on single panels, against 1.3 times for the per-criterion requests over the nine panels (Appendix~\ref{app:cost}). The two estimates are not comparable, because LiteLLM's price map changed between the runs, as it did for the effort runs (Appendix~\ref{app:effort}). The per-criterion estimate prices tokens close to DeepSeek's list prices, whereas the whole-rubric estimate prices completion tokens at about \$0.29 per million, under half the list price of \$0.60, so the harness-estimate row of Table~\ref{tab:wr-runs}a overstates DeepSeek's saving. Measured against its cost at list prices, DeepSeek's bill is 1.34 times as high for the per-criterion requests and 1.70 times for the whole-rubric ones, and at list prices whole-rubric judgments would have saved it 36\%, less than Luna and Gemini saved, which fits the larger share of DeepSeek's cost that lies in completion tokens. OpenRouter's routing may account for part of DeepSeek's smaller saving as billed. Its 698 answered whole-rubric requests on the ordinal panels (two more failed) came from 15 upstream providers, chiefly Parasail (256 requests), Together (224), and CoreWeave (129), whereas Morph and DeepInfra together had served just over half of its per-criterion requests there.

\begin{table*}[!tp]
\centering
\footnotesize
\setlength{\tabcolsep}{3pt}
\begin{tabular}{@{}l@{\hspace{6pt}}lrrrrlrrrrl@{}}
\toprule
 & & & & \multicolumn{2}{c}{Accuracy (\%)} & & & & & & \\
\cmidrule(lr){5-6}
Panel & Judge & Units & Pairs & \makecell[r]{Per-\\criterion} & \makecell[r]{Whole-\\rubric} & Diff & 95\% interval & Bound & P\,:\,W & $p_{\text{unit}}$ & Class \\
\midrule
RiceChem & Luna & 121 & 819 & 77.8 & 76.9 & $-0.9$ & $-3.4$, $+1.7$ & 3.0 & 33\,:\,30 & 0.80 & parity \\
 & Gemini & 121 & 819 & 76.1 & 76.8 & $+0.7$ & $-1.1$, $+2.6$ & 2.3 & 20\,:\,27 & 0.38 & parity \\
 & DeepSeek & 119 & 805 & 79.0 & 81.2 & $+2.2$ & $-0.4$, $+4.9$ & 4.5 & 24\,:\,38 & 0.098 & parity \\
\addlinespace[2pt]
HealthBench & Luna & 200 & 406 & 70.4 & 70.9 & $+0.5$ & $-2.9$, $+3.9$ & 3.3 & 20\,:\,22 & 0.88 & parity \\
 & Gemini & 200 & 406 & 79.6 & 80.0 & $+0.5$ & $-1.3$, $+2.3$ & 2.0 & 6\,:\,8 & 0.79 & parity \\
 & DeepSeek & 199 & 404 & 76.2 & 76.5 & $+0.2$ & $-3.1$, $+3.5$ & 3.0 & 22\,:\,22 & 1.00 & parity \\
\addlinespace[2pt]
ELLIPSE & Luna & 258 & 1{,}548 & 14.1 & 20.0 & $+5.9$\ddg & $+3.8$, $+8.1$ & 7.8 & 39\,:\,84 & $6 \times 10^{-5}$ & separated \\
 & Gemini & 258 & 1{,}548 & 29.8 & 35.9 & $+6.1$\ddg & $+4.1$, $+8.1$ & 7.8 & 26\,:\,88 & $5 \times 10^{-9}$ & separated \\
 & DeepSeek & 258 & 1{,}536 & 15.0 & 13.5 & $-1.5$ & $-3.2$, $+0.3$ & 2.9 & 70\,:\,52 & 0.12 & parity \\
\addlinespace[2pt]
FED-Turn & Luna & 75 & 600 & 62.5 & 65.5 & $+3.0$ & $-0.5$, $+6.5$ & 5.8 & 17\,:\,31 & 0.059 & -- \\
 & Gemini & 75 & 600 & 65.2 & 65.7 & $+0.5$ & $-2.2$, $+3.0$ & 2.7 & 17\,:\,20 & 0.74 & parity \\
 & DeepSeek & 75 & 599 & 63.8 & 62.4 & $-1.3$ & $-4.3$, $+1.5$ & 3.8 & 23\,:\,21 & 0.88 & parity \\
\addlinespace[2pt]
FED-Dialogue & Luna & 25 & 233 & 51.5 & 51.5 & $0.0$ & $-4.7$, $+5.1$ & 4.3 & 9\,:\,8 & 1.00 & parity \\
 & Gemini & 25 & 228 & 60.5 & 57.9 & $-2.6$ & $-7.0$, $+1.7$ & 6.5 & 7\,:\,4 & 0.55 & -- \\
 & DeepSeek & 25 & 234 & 50.0 & 53.0 & $+3.0$ & $-2.6$, $+8.9$ & 8.0 & 7\,:\,9 & 0.80 & -- \\
\addlinespace[2pt]
HelpSteer2 & Luna & 90 & 360 & 39.2 & 39.2 & $0.0$ & $-5.0$, $+4.7$ & 4.2 & 29\,:\,30 & 1.00 & parity \\
 & Gemini & 90 & 355 & 54.6 & 50.1 & $-4.5$ & $-8.2$, $-0.8$ & 7.6 & 22\,:\,11 & 0.080 & separated \\
 & DeepSeek & 90 & 357 & 46.8 & 49.3 & $+2.5$ & $-2.0$, $+7.0$ & 6.2 & 16\,:\,24 & 0.27 & -- \\
\addlinespace[2pt]
LFQA & Luna & 120 & 359 & 71.6 & 66.3 & $-5.3$\scm & $-8.6$, $-2.0$ & 8.1 & 26\,:\,7 & 0.001 & separated \\
 & Gemini & 120 & 358 & 67.6 & 66.2 & $-1.4$ & $-4.5$, $+1.7$ & 3.9 & 16\,:\,11 & 0.44 & parity \\
 & DeepSeek & 119 & 351 & 65.2 & 67.8 & $+2.6$ & $-1.7$, $+6.5$ & 5.9 & 16\,:\,25 & 0.21 & -- \\
\addlinespace[2pt]
USR-TC & Luna & 72 & 357 & 68.3 & 67.2 & $-1.1$ & $-5.6$, $+3.9$ & 5.0 & 23\,:\,14 & 0.19 & -- \\
 & Gemini & 72 & 358 & 70.7 & 70.7 & $0.0$ & $-2.8$, $+2.8$ & 2.5 & 10\,:\,10 & 1.00 & parity \\
 & DeepSeek & 71 & 339 & 68.4 & 69.9 & $+1.5$ & $-2.4$, $+5.3$ & 4.7 & 14\,:\,17 & 0.72 & parity \\
\addlinespace[2pt]
USR-PC & Luna & 60 & 300 & 74.0 & 71.3 & $-2.7$ & $-6.0$, $+0.7$ & 5.7 & 18\,:\,10 & 0.18 & -- \\
 & Gemini & 60 & 300 & 74.3 & 72.0 & $-2.3$ & $-5.0$, $0.0$ & 4.3 & 11\,:\,4 & 0.12 & parity \\
 & DeepSeek & 60 & 300 & 69.7 & 69.3 & $-0.3$ & $-4.7$, $+4.0$ & 4.0 & 15\,:\,13 & 0.85 & parity \\
\midrule
Nine panels & Luna & 1{,}021 & 4{,}982 & & & $+1.5$ & $+0.4$, $+2.6$ & 2.4 & 214\,:\,236 & 0.32 & \\
 & Gemini & 1{,}021 & 4{,}972 & & & $+1.4$ & $+0.5$, $+2.3$ & 2.2 & 135\,:\,183 & 0.008 & \\
 & DeepSeek & 1{,}016 & 4{,}925 & & & $+0.3$ & $-0.7$, $+1.4$ & 1.2 & 207\,:\,221 & 0.53 & \\
\addlinespace[2pt]
Seven ordinal & Luna & 700 & 3{,}757 & & & $+2.1$ & $+0.8$, $+3.4$ & 3.2 & 161\,:\,184 & 0.24 & \\
 & Gemini & 700 & 3{,}747 & & & $+1.7$ & $+0.5$, $+2.8$ & 2.6 & 109\,:\,148 & 0.018 & \\
 & DeepSeek & 698 & 3{,}716 & & & $-0.1$ & $-1.3$, $+1.1$ & 1.1 & 161\,:\,161 & 1.00 & \\
\addlinespace[2pt]
All but ELLIPSE & Luna & 763 & 3{,}434 & & & $-0.5$ & $-1.9$, $+0.8$ & 1.7 & 175\,:\,152 & 0.22 & \\
 & Gemini & 763 & 3{,}424 & & & $-0.7$ & $-1.6$, $+0.3$ & 1.5 & 109\,:\,95 & 0.36 & \\
 & DeepSeek & 758 & 3{,}389 & & & $+1.2$ & $-0.1$, $+2.5$ & 2.3 & 137\,:\,169 & 0.076 & \\
\bottomrule
\end{tabular}
\caption{Each LLM judge's whole-rubric run against its per-criterion run, on the pairs both runs scored, with units resampled as in Table~\ref{tab:clustered}. Accuracy is verdict accuracy on the binary panels and, on the ordinal panels, the share of correct verdicts over the ordinal and embedded binary pairs together. Diff: whole-rubric minus per-criterion accuracy, in points. P\,:\,W: units where the per-criterion run leads against units where the whole-rubric run leads, and $p_{\text{unit}}$ their sign test. Bound and Class as in Table~\ref{tab:clustered}. \ddag: significant once the bootstrap $p$-values of the 27 rows above the rule are Holm-corrected; \S: significant after that correction only in the sign test over units. The bottom rows pool each judge's pairs over several panels and lie outside the Holm correction.}
\label{tab:wr-accuracy}
\end{table*}

\paragraph{Accuracy by judging mode.}
Table~\ref{tab:wr-accuracy} sets each judge's two runs side by side on their common pairs, using the paired statistics of Section~\ref{sec:metrics} with one Holm family of 27 judge--panel comparisons. Four of these changes are separated, 16 at parity, and 7 inconclusive. Of the 27 differences, 13 are positive, 3 are exactly zero, and 11 are negative. The mode raised Luna's ELLIPSE accuracy by 5.9 points and Gemini's by 6.1, the only two of the 27 per-panel changes significant once their bootstrap $p$-values are corrected. Besides these ELLIPSE gains, whose 95\% intervals run from 3.8 to 8.1 and from 4.1 to 8.1 points, the separated changes are two losses, Luna's of 5.3 points on LFQA and Gemini's of 4.5 on HelpSteer2. None of DeepSeek's changes is separated. Holm correction of the sign test over units keeps Luna's LFQA loss as well as the two ELLIPSE gains, although the bootstrap $p$-value for that loss exceeds 0.05 after correction ($0.070$). Gemini's HelpSteer2 loss has a bootstrap $p$-value of 0.021, which the correction raises to 0.50.

Across the nine panels together, Luna's pooled gain is 1.5 points and Gemini's 1.4, but both pooled differences are negative once ELLIPSE's 1,548 pairs are left out, and their intervals then include zero. Luna's ELLIPSE offset moved toward zero, from $-1.28$ to $-1.05$ levels, and Gemini's from $-0.77$ to $-0.67$, whereas DeepSeek's moved from $-1.16$ to $-1.22$ levels, with an accuracy change at parity. On a panel with embedded binary pairs the pooled measure of Table~\ref{tab:wr-accuracy} can differ in sign from the measure of Table~\ref{tab:main}, which counts ordinal pairs only; on USR-TC, for instance, Luna's difference is $-1.1$ points pooled and $+1.4$ on the ordinal pairs. The largest changes in mean QWK are Luna's fall on LFQA, from 0.51 to 0.39, and DeepSeek's rise on USR-PC, from 0.42 to 0.51. Luna's ELLIPSE gain did not extend to every trait. On \crit{cohesion} its exact accuracy fell from 44.6\% to 32.6\%, while \crit{syntax} rose from 13.2\% to 30.6\%, \crit{vocabulary} from 12.0\% to 26.0\%, and \crit{phraseology} from 7.8\% to 19.4\%.

\begin{table*}[!tp]
\centering
\footnotesize
\setlength{\tabcolsep}{4pt}
\begin{tabular}{@{}l@{\hspace{6pt}}lrrlrrrl@{\hspace{10pt}}rl@{}}
\toprule
 & & \multicolumn{7}{c}{Whole-rubric LLM judge} & \multicolumn{2}{c}{Per-criterion} \\
\cmidrule(lr){3-9}\cmidrule(l){10-11}
Panel & Judge & Units & Diff & 95\% interval & Bound & J\,:\,O & $p_{\text{unit}}$ & Class & Diff & Class \\
\midrule
RiceChem & Luna & 121 & $-4.0$ & $-6.9$, $-1.4$ & 6.4 & 38\,:\,21 & 0.036 & separated & $-3.2$ & separated \\
 & Gemini & 121 & $-4.2$ & $-6.9$, $-1.4$ & 6.5 & 40\,:\,22 & 0.030 & separated & $-4.9$\ddg & separated \\
 & DeepSeek & 119 & $+0.4$ & $-2.0$, $+2.7$ & 2.4 & 21\,:\,28 & 0.39 & parity & $-1.7$ & parity \\
\addlinespace[2pt]
HealthBench & Luna & 200 & $-6.2$ & $-11.8$, $-0.5$ & 10.8 & 45\,:\,32 & 0.17 & separated & $-6.7$ & separated \\
 & Gemini & 200 & $+3.0$ & $-0.7$, $+6.6$ & 6.0 & 19\,:\,33 & 0.070 & -- & $+2.5$ & -- \\
 & DeepSeek & 199 & $-0.5$ & $-4.6$, $+3.6$ & 3.9 & 29\,:\,29 & 1.00 & parity & $-0.7$ & parity \\
\addlinespace[2pt]
ELLIPSE & Luna & 258 & $+6.6$\ddg & $+4.5$, $+8.7$ & 8.3 & 30\,:\,87 & $1 \times 10^{-7}$ & separated & $+0.6$ & parity \\
 & Gemini & 258 & $+22.5$\ddg & $+19.7$, $+25.3$ & 24.8 & 7\,:\,166 & $1 \times 10^{-40}$ & separated & $+16.4$\ddg & separated \\
 & DeepSeek & 258 & $0.0$ & $-1.4$, $+1.5$ & 1.2 & 49\,:\,48 & 1.00 & parity & $+1.6$ & parity \\
\addlinespace[2pt]
FED-Turn & Luna & 75 & $+4.0$ & $+0.2$, $+7.8$ & 7.3 & 19\,:\,34 & 0.053 & separated & $+1.0$ & parity \\
 & Gemini & 75 & $+4.2$ & $-0.3$, $+8.7$ & 8.0 & 22\,:\,31 & 0.27 & -- & $+3.7$ & -- \\
 & DeepSeek & 75 & $+0.8$ & $-2.8$, $+4.5$ & 3.8 & 23\,:\,28 & 0.58 & parity & $+2.2$ & -- \\
\addlinespace[2pt]
FED-Dialogue & Luna & 25 & $-0.4$ & $-8.7$, $+8.1$ & 7.4 & 10\,:\,8 & 0.81 & -- & $-0.4$ & -- \\
 & Gemini & 25 & $+4.8$ & $0.0$, $+10.0$ & 9.2 & 4\,:\,9 & 0.27 & -- & $+7.9$\ddg & separated \\
 & DeepSeek & 25 & $+1.3$ & $-5.2$, $+7.6$ & 6.5 & 8\,:\,11 & 0.65 & -- & $-1.7$ & -- \\
\addlinespace[2pt]
HelpSteer2 & Luna & 90 & $-9.1$ & $-15.1$, $-3.1$ & 14.0 & 37\,:\,16 & 0.005 & separated & $-10.2$\ddg & separated \\
 & Gemini & 90 & $+1.1$ & $-4.3$, $+6.5$ & 5.6 & 18\,:\,29 & 0.14 & -- & $+5.7$ & separated \\
 & DeepSeek & 90 & $+0.9$ & $-4.3$, $+5.7$ & 4.9 & 22\,:\,23 & 1.00 & parity & $-2.3$ & -- \\
\addlinespace[2pt]
LFQA & Luna & 120 & $-2.5$ & $-6.4$, $+1.1$ & 5.6 & 25\,:\,16 & 0.21 & -- & $+2.8$ & -- \\
 & Gemini & 120 & $-2.2$ & $-7.0$, $+2.5$ & 6.2 & 30\,:\,26 & 0.69 & -- & $-1.1$ & parity \\
 & DeepSeek & 119 & $-1.7$ & $-5.9$, $+2.3$ & 5.1 & 23\,:\,19 & 0.64 & -- & $-3.7$ & -- \\
\addlinespace[2pt]
USR-TC & Luna & 72 & $-0.8$ & $-5.6$, $+3.6$ & 4.8 & 21\,:\,19 & 0.87 & parity & $+0.3$ & parity \\
 & Gemini & 72 & $+2.8$ & $-0.8$, $+6.4$ & 5.8 & 12\,:\,22 & 0.12 & -- & $+3.1$ & -- \\
 & DeepSeek & 71 & $-0.6$ & $-5.0$, $+3.6$ & 4.2 & 15\,:\,16 & 1.00 & parity & $-1.4$ & -- \\
\addlinespace[2pt]
USR-PC & Luna & 60 & $+1.3$ & $-2.7$, $+5.4$ & 4.7 & 14\,:\,14 & 1.00 & parity & $+4.0$ & separated \\
 & Gemini & 60 & $+2.0$ & $-2.3$, $+6.4$ & 5.7 & 11\,:\,15 & 0.56 & -- & $+4.3$ & -- \\
 & DeepSeek & 60 & $-0.7$ & $-5.4$, $+3.7$ & 4.4 & 14\,:\,11 & 0.69 & parity & $-0.3$ & parity \\
\bottomrule
\end{tabular}
\caption{Jev Choice against each whole-rubric LLM judge, in the layout of Table~\ref{tab:clustered}. Diff: accuracy of the LLM judge less that of Jev Choice (points) over the pairs that both scored, embedded binary pairs included on the ordinal panels. J\,:\,O: units with Jev Choice ahead against units with the LLM judge ahead. \ddag: the separation survives Holm correction (bootstrap $p$, 27 comparisons). The last two columns repeat the per-criterion difference and class of Table~\ref{tab:clustered}.}
\label{tab:wr-paired}
\end{table*}

\paragraph{Jev Choice against the whole-rubric judges.}
Between the judging modes, 5 of Jev Choice's 27 comparisons change separation (Figure~\ref{fig:whole-rubric}; Table~\ref{tab:wr-paired}). Luna gains a separated lead over Jev Choice on ELLIPSE and FED-Turn, and three leads of the LLM judges lose their separation: Luna's on USR-PC and Gemini's on FED-Dialogue and HelpSteer2. Jev Choice is left with three separated deficits, Luna's two new leads among them (Table~\ref{tab:conditions}), and correcting the bootstrap $p$-values within this mode's 27 comparisons leaves only Luna's and Gemini's ELLIPSE leads significant. DeepSeek is not separated from Jev Choice in either mode. On RiceChem its whole-rubric verdict accuracy of 81.2\% puts it ahead of Jev Choice's 81.0\% ($+0.4$ points on the pairs both scored, at parity), which drops Jev Choice to second place there. On LFQA, Jev Choice becomes the most accurate matched judge. In both modes Gemini is the most accurate matched judge on the six other ordinal panels, tied with Luna on USR-PC with per-criterion judgments. On the ordinal panels, the ten inconclusive comparisons have 90\% intervals that reach 5.1 to 9.2 points from zero. Resampling sampling groups instead of units changes one outcome, against four with per-criterion judgments (Appendix~\ref{app:tests}). Luna on USR-TC loses parity, its bound rising from 4.8 to 5.3 points.

In pair-level sign tests eight comparisons are significant before correction, four in each direction. Jev Choice leads Luna and Gemini on RiceChem and leads Luna on HealthBench and HelpSteer2; Luna and Gemini lead it on ELLIPSE and FED-Turn. After Holm correction of these sign tests, Luna's and Gemini's ELLIPSE leads and Jev Choice's HelpSteer2 lead over Luna remain. As with per-criterion judgments, Gemini's FED-Turn lead is significant pair by pair but not separated. HelpSteer2's constant predictor is now significantly more accurate than each of Jev's five judges in pair-level sign tests, Gemini included ($p = 0.002$). Neither Decider framing, whose verdicts come from its single run, differs significantly from it (Decider Choice $p = 0.37$, Decider Score $p = 0.16$).

\begin{table*}[!tp]
\centering
\footnotesize
\setlength{\tabcolsep}{2.8pt}
\begin{tabular}{@{}lr@{\hspace{8pt}}rrrr@{\hspace{8pt}}rrrr@{\hspace{8pt}}rrrr@{}}
\toprule
 & & \multicolumn{4}{c}{Luna} & \multicolumn{4}{c}{Gemini} & \multicolumn{4}{c}{DeepSeek} \\
\cmidrule(lr){3-6}\cmidrule(lr){7-10}\cmidrule(l){11-14}
 & & \multicolumn{1}{c}{All} & \multicolumn{3}{c}{Repeat units} & \multicolumn{1}{c}{All} & \multicolumn{3}{c}{Repeat units} & \multicolumn{1}{c}{All} & \multicolumn{3}{c}{Repeat units} \\
\cmidrule(lr){3-3}\cmidrule(lr){4-6}\cmidrule(lr){7-7}\cmidrule(lr){8-10}\cmidrule(lr){11-11}\cmidrule(l){12-14}
Panel & $n$ & W\,=\,P & R\,=\,W & W\,=\,P & R\,=\,P & W\,=\,P & R\,=\,W & W\,=\,P & R\,=\,P & W\,=\,P & R\,=\,W & W\,=\,P & R\,=\,P \\
\midrule
RiceChem & 95\,/\,14 & 88.2 & 94.7 & 84.2 & 85.3 & 93.4 & 94.7 & 91.6 & 92.6 & 86.3 & 95.8 & 85.3 & 85.3 \\
HealthBench & 42\,/\,20 & 88.2 & 90.5 & 88.1 & 83.3 & 96.6 & 100.0 & 100.0 & 100.0 & 87.6 & 88.1 & 85.7 & 78.6 \\
\midrule
ELLIPSE & 156\,/\,26 & 61.0 & 82.1 & 62.2 & 64.1 & 82.9 & 87.8 & 81.4 & 84.6 & 67.8 & 82.7 & 65.4 & 65.4 \\
FED-Turn & 64\,/\,8 & 80.2 & 81.2 & 76.6 & 76.6 & 89.2 & 98.4 & 96.9 & 95.3 & 81.8 & 90.6 & 81.2 & 81.2 \\
FED-Dialogue & 30\,/\,3 & 75.2 & 93.3 & 86.7 & 86.7 & 88.4 & 100.0 & 90.0 & 90.0 & 74.4 & 90.0$^{a}$ & 85.0$^{a}$ & 95.0$^{a}$ \\
HelpSteer2 & 36\,/\,9 & 66.9 & 80.6 & 75.0 & 72.2 & 83.9 & 91.7 & 83.3 & 86.1 & 71.0 & 83.3 & 75.0 & 83.3 \\
LFQA & 36\,/\,12 & 85.8 & 83.3 & 94.4 & 83.3 & 90.3 & 97.2 & 88.9 & 91.7 & 80.1 & 77.8 & 80.6 & 80.6 \\
USR-TC & 40\,/\,8 & 80.0 & 87.5 & 77.5 & 85.0 & 90.3 & 82.5 & 82.5 & 85.0 & 79.4 & 90.0 & 72.5 & 77.5 \\
USR-PC & 30\,/\,6 & 86.0 & 86.7 & 90.0 & 76.7 & 92.0 & 90.0 & 86.7 & 96.7 & 84.7 & 96.7 & 90.0 & 93.3 \\
\midrule
Nine panels & 529\,/\,106 & 75.7 & 86.2 & 77.1 & 76.4 & 88.4 & 92.6 & 88.1 & 90.2 & 77.4 & 87.9$^{a}$ & 77.1$^{a}$ & 78.0$^{a}$ \\
\quad mean QWK & & 0.62 & 0.71 & 0.62 & 0.60 & 0.83 & 0.91 & 0.74 & 0.76 & 0.60 & 0.71 & 0.62 & 0.62 \\
\quad Cohen's $\kappa$ & & 0.78 & 0.89 & 0.72 & 0.72 & 0.89 & 0.91 & 0.86 & 0.92 & 0.74 & 0.89 & 0.72 & 0.68 \\
\quad cross-mode excess & & & \multicolumn{3}{c}{$-9.5$ ($-12.7$, $-6.1$)} & & \multicolumn{3}{c}{$-3.5$ ($-6.4$, $-0.7$)} & & \multicolumn{3}{c}{$-10.3$ ($-14.1$, $-6.5$)} \\
\bottomrule
\end{tabular}
\caption{Share of pairs (\%) on which two runs give the identical verdict: the same level, the same MET or UNMET verdict, or an abstention in both. P: the per-criterion run; W: the whole-rubric run; R: the repeat run. All: W against P over every pair of the panel (5,003 pairs per judge; 4,968 for DeepSeek). Repeat units: the three comparisons on the repeat pairs, whose number of pairs and units is $n$. Pairs touched by a failed request in either run are set aside. $^{a}$DeepSeek's repeat lacks one FED-Dialogue unit, leaving it 20 pairs on 2 units there and 519 pairs on 105 units in all. Mean QWK: over the ordinal criteria, each criterion's QWK between the two runs; Cohen's $\kappa$: over the binary pairs. Cross-mode excess: the mean of the W\,=\,P and R\,=\,P shares minus the R\,=\,W share, in points, with a 95\% interval from resampling units; it would be zero in expectation if the judging mode changed verdicts no more than a rerun does.}
\label{tab:wr-agreement}
\end{table*}

\paragraph{Agreement between runs.}
Over all pairs, the whole-rubric run gives a different verdict from the per-criterion run on 24.3\% of Luna's pairs, 11.6\% of Gemini's, and 22.6\% of DeepSeek's, the complements of the W\,=\,P column of Table~\ref{tab:wr-agreement}. On the repeat units, rerunning the identical whole-rubric requests changes 13.8\%, 7.4\%, and 12.1\% of their verdicts (the R\,=\,W column). There, the two comparisons across modes (the whole-rubric run and the repeat run, each against the per-criterion run) agree less often, on average, than the two whole-rubric runs agree with each other. The shortfall is 9.5 points for Luna, 3.5 for Gemini, and 10.3 for DeepSeek, each with a 95\% interval that excludes zero (the cross-mode excess of Table~\ref{tab:wr-agreement}). The repeat pairs are few on single panels, 20 to 156, so Table~\ref{tab:wr-agreement} gives per-panel shares without intervals, and our comparison of the modes rests on the pooled rows. The cross-mode comparison spans eight days (see \ref{sec:limitations}). Identical verdicts between the modes are rarest on ELLIPSE for every judge, 61.0\% of the pairs for Luna, 82.9\% for Gemini, and 67.8\% for DeepSeek. In 6 of the 27 judge--panel cells the mean of the two cross-mode shares matches or exceeds the share between the two whole-rubric runs (Luna and DeepSeek on LFQA; Gemini on HealthBench, USR-TC, and USR-PC; DeepSeek on FED-Dialogue), each on 42 or fewer repeat pairs. For comparison, Luna's two per-criterion runs in the spot check of Appendix~\ref{app:protocol} agreed on 86.6\% of the 82 pairs of the ordinal panels, and its two whole-rubric runs on 83.7\% of the 392 repeat pairs on those panels.

\begin{table*}[!tp]
\centering
\scriptsize
\setlength{\tabcolsep}{2.8pt}
\begin{tabular}{@{}l@{\hspace{6pt}}cc@{\hspace{6pt}}rr@{\hspace{6pt}}rr@{\hspace{6pt}}rr@{\hspace{6pt}}rr@{\hspace{6pt}}rr@{\hspace{6pt}}rr@{}}
\toprule
 & \multicolumn{2}{c}{\makecell{Judge--judge /\\judge--label QWK}} & \multicolumn{2}{c}{\makecell{Repeated errors,\\excess (points)}} & \multicolumn{2}{c}{\makecell{Confident errors\\repeated}} & \multicolumn{2}{c}{\makecell{Oracle cascade\\vs.\ best}} & \multicolumn{2}{c}{\makecell{Cross-fitted\\vs.\ best}} & \multicolumn{2}{c}{\makecell{Cross-fitted\\cost (\%)}} & \multicolumn{2}{c}{\makecell{Jury\\vs.\ best}} \\
\cmidrule(lr){2-3}\cmidrule(lr){4-5}\cmidrule(lr){6-7}\cmidrule(lr){8-9}\cmidrule(lr){10-11}\cmidrule(lr){12-13}\cmidrule(l){14-15}
Panel & P & W & P & W & P & W & P & W & P & W & P & W & P & W \\
\midrule
\multicolumn{15}{@{}l}{\emph{(a) By panel}} \\
RiceChem & -- & -- & -- & -- & -- & -- & $-0.4$ & $+0.4$ & $-1.0$ & $+0.1$ & 6.6 & 57.0 & $-2.8$\dg & $-1.7$\dg \\
HealthBench & -- & -- & $+29.0^{b}$ & $+30.6^{b}$ & 32/36$^{b}$ & 32/36$^{b}$ & $+2.0$ & $+2.0$ & $+1.5$ & $+1.7$ & 15.6 & 28.3 & $-2.2$ & $-2.5$ \\
\midrule
ELLIPSE & 0.38\,/\,0.18 & 0.42\,/\,0.21 & $+8.6$ & $+6.8$ & 33/36 & 33/36 & $0.0$ & $0.0$ & $-0.1$ & $-0.1$ & 99.8 & 100.4 & $-12.8$\dg & $-14.7$\dg \\
FED-Turn & 0.57\,/\,0.39 & 0.60\,/\,0.40 & $+22.3$ & $+19.1$ & 35/36 & 35/36 & $-5.2$ & $-5.0$ & $-7.3$ & $-6.8$ & 45.5 & 99.1 & $-0.2$ & $-0.5$ \\
FED-Dialogue & 0.61\,/\,0.35 & 0.59\,/\,0.33 & $+20.4$ & $+25.7$ & 36/36 & 36/36 & $0.0$ & $-1.7$ & $-0.9$ & $-3.0$ & 77.9 & 93.4 & $-3.1$ & $-0.9$ \\
HelpSteer2 & 0.66\,/\,0.31 & 0.64\,/\,0.30 & $+32.0$ & $+30.8$ & 35/36 & 35/36 & $+0.6$ & $+2.6$ & $-0.9$ & $+2.5$ & 35.6 & 36.6 & $-8.0$\dg & $-1.4$ \\
LFQA & 0.70\,/\,0.46 & 0.62\,/\,0.45 & $+34.2$ & $+29.7$ & 36/36 & 36/36 & $+0.6$ & $-0.6$ & $+0.6$ & $-1.1$ & 39.0 & 23.4 & $-2.0$ & $-0.6$ \\
USR-TC & 0.65\,/\,0.53 & 0.65\,/\,0.56 & $+28.0$ & $+25.6$ & 32/36 & 32/35 & $+0.3$ & $+1.1$ & $-0.7$ & $-0.1$ & 47.5 & 73.0 & $0.0$ & $-0.6$ \\
USR-PC & 0.61\,/\,0.50 & 0.69\,/\,0.50 & $+18.8$ & $+22.7$ & 35/36 & 36/36 & $+1.3$ & $+1.7$ & $+0.8$ & $+1.6$ & 31.3 & 44.6 & $-0.7$ & $-1.7$ \\
\bottomrule
\end{tabular}

\medskip
\footnotesize
\setlength{\tabcolsep}{3pt}
\begin{tabular}{@{}>{\raggedright\arraybackslash}p{9.6cm}>{\raggedright\arraybackslash}p{2.9cm}>{\raggedright\arraybackslash}p{2.9cm}@{}}
\toprule
 & Per-criterion & Whole-rubric \\
\midrule
\multicolumn{3}{@{}l}{\emph{(b) Pooled statistics}} \\
Least-agreeing judge pair above the best judge--label QWK (panels) & 5 of 7 & 6 of 7 \\
ELLIPSE: all four matched judges wrong; best-of-four selection (\%) & 64.0; 36.0 & 58.7; 41.3 \\
Negative panel offsets, five judges on seven ordinal panels & 31 of 35 & 31 of 35 \\
Ordinal panels with every matched judge's offset negative & 6 of 7 & 6 of 7 \\
ELLIPSE offsets of the matched judges, range (levels) & $-1.28$ to $-0.77$ & $-1.25$ to $-0.67$ \\
Offset--gap Spearman $\rho$, 31 criteria (Figure~\ref{fig:offsetgap}) & 0.924 & 0.918 \\
Criteria more than 15 points below the reference; of these, offset below $-0.5$ & 12; 11 & 12; 11 \\
Criteria with offset within 0.3 levels of zero; their largest shortfall (points) & 15; 5.8 & 15; 4.8 \\
Criteria (of 35) with all four offsets negative; positive & 25; 2 & 23; 3 \\
Leave-one-out shift: Spearman $\rho$ of gap and offset ($p$) & 0.16 (0.40) & 0.09 (0.63) \\
Leave-one-out shift: worst shortfall from the unshifted reference (points) & 8.4 & 10.4 \\
Both shifted: criteria more than 5 points below; worst shortfall (points) & 4; 16.2 & 5; 17.2 \\
Confident errors: LLM verdicts repeating Jev's wrong answer & 242 of 252 (96.0\%) & 243 of 251 (96.8\%) \\
Confident errors: independence baseline (\%) & 50.3 & 50.7 \\
Top confidence band: repeated share; independence baseline (\%) & 80.5--92.8; 33.6--63.7 & 72.9--93.3; 38.5--62.8 \\
Largest gain over the best single judge, cross-fitted thresholds (points) & $+1.5$ (HealthBench) & $+2.5$ (HelpSteer2) \\
Largest gain over the best single judge, oracle threshold (points) & $+2.0$ (HealthBench) & $+2.6$ (HelpSteer2) \\
Panels where the cross-fitted cascade trails the best single judge & 6 of 9 & 5 of 9 \\
Confident errors: three-LLM median gives Jev's answer; the label & 82 of 84; 1 & 83 of 83; 0 \\
Four-judge upper median: largest gain (points) & $+1.0$ (USR-PC) & $+2.8$ (LFQA) \\
\bottomrule
\end{tabular}
\caption{The analyses of Sections~\ref{sec:shared} and~\ref{sec:routing} with per-criterion (P) and whole-rubric (W) LLM judges; Jev's verdicts are the same in both. (a) Judge--judge and judge--label QWK as in Table~\ref{tab:main}. Repeated errors: among the LLM judges' verdicts on pairs that Jev Choice got wrong, the share giving its answer, less the independence baseline (Section~\ref{sec:shared-errors}). Confident errors repeated: LLM verdicts on the confident-error sample that give Jev's answer, out of those given; one DeepSeek verdict is missing on USR-TC with whole-rubric judgments. $^{b}$HealthBench: Jev Noul's errors (Appendix~\ref{app:shared}). Cascade columns, with Jev first, as in Table~\ref{tab:cascade}: the oracle-threshold and cross-fitted cascade minus the best single judge, in points, and the cross-fitted cascade's cost for reaching the best LLM judge's accuracy, as a share of that judge's own cost; a whole-rubric fallback is charged per unit. Jury: the three-LLM median minus the most accurate matched judge, in points (Section~\ref{sec:juries}); \dag: its 95\% interval lies below zero. (b) Pooled statistics of Sections~\ref{sec:agreement} to~\ref{sec:juries} and Appendix~\ref{app:offset}. The ELLIPSE offset range runs from Luna to Gemini with per-criterion judgments and from Jev Choice to Gemini with whole-rubric judgments. Both worst shortfalls fall on FED-Dialogue \crit{informative} in either mode. The four-judge upper median's gain is over the most accurate matched judge.}
\label{tab:wr-downstream}
\end{table*}

\paragraph{Repeated errors and cascades.}
Across all of Jev Choice's errors on each ordinal panel, the whole-rubric LLM judges repeat its answer more often than the independence baseline predicts (Table~\ref{tab:wr-downstream}a). Except on FED-Dialogue, the share of repeated errors exceeds the baseline most in Jev's highest confidence band. On FED-Dialogue, among bands that hold at least 20 of Jev's errors, it exceeds the baseline by 26.1 points between confidence 0.25 and 0.5, 26.7 between 0.5 and 0.75, and 24.5 in the top band.

Equation~\eqref{eq:identity} holds for every whole-rubric cascade, so where an ordinal cascade beats its whole-rubric fallback, Jev beats that fallback on the kept pairs, by 3.0 points on HelpSteer2, 2.5 on LFQA, 2.1 on USR-TC, and 1.9 on USR-PC. These kept-pair leads on HelpSteer2, LFQA, and USR-TC are larger than the 0.7 to 1.9 points Jev holds over the per-criterion fallbacks (Appendix~\ref{app:cascades}). On HelpSteer2, where the whole-rubric cascades' oracle and cross-fitted gains over the best single judge both peak (Table~\ref{tab:wr-downstream}a), the cross-fitted gain of 2.5 points is positive in all 50 halvings. Gemini's loss of 4.5 points on HelpSteer2, from which these gains arise, is separated but does not survive Holm correction (Table~\ref{tab:wr-accuracy}). The best HelpSteer2 cascade defers 14.5\% of the pairs but touches 46.7\% of the units. Among its kept pairs, those that only Jev gets right outnumber those that only Gemini gets right by 35 to 26 (pair-level sign test, $p = 0.31$), and its accuracy still trails that of HelpSteer2's constant predictor.

Billing a whole-rubric fallback a full request for each unit that has a deferred pair erodes the saving, most where a unit carries many criteria. At the threshold that matches Gemini on USR-TC, the deferred 38.1\% of the pairs fall on 81.9\% of the units. The cross-fitted cascade that matches the best LLM judge costs 99.1\% of that judge's cost on FED-Turn instead of 45.5\%, 57.0\% on RiceChem instead of 6.6\%, and 93.4\% on FED-Dialogue instead of 77.9\%, though the fallback's change in accuracy also moves these shares. ELLIPSE had no saving to lose. Only on four panels does such a cascade stay below half of the judge's cost, at 23\% to 45\%, with held-out accuracy from 0.6 points below the judge to 2.4 above. Also charging the units whose only deferral is a Jev Choice abstention, as a live cascade would, raises the cost of matching Gemini on HelpSteer2 at the oracle threshold from 7.5\% to 15.6\% of Gemini's own cost. On the binary panels every fallback, DeepSeek included, is charged the harness's own cost estimate, as in the per-criterion binary cascades; on the ordinal panels DeepSeek is charged its billed cost per unit.

\paragraph{Juries.}
A median of the three whole-rubric LLM judges is less accurate than the most accurate matched judge on every panel, by up to 14.7 points (ELLIPSE), and its 95\% interval lies below zero on RiceChem and ELLIPSE (Table~\ref{tab:wr-downstream}a). With Jev Choice as a fourth juror, no gain has an interval that excludes zero; the largest, 2.8 points on LFQA, has an interval from 0.0 to 5.9. On all 83 confident errors that every judge scored, the three-LLM median picks Jev's wrong answer.

\paragraph{The HealthBench physician reference.}
On the 50 HealthBench pairs with three or more physicians (Appendix~\ref{app:binary}), whole-rubric Luna reaches 62.7\% against the leave-one-out targets, 24.6 points below the physician reference, with a 95\% interval of $-45.7$ to $-5.3$ points. It is the only judge in either mode whose interval excludes zero. The intervals of whole-rubric Gemini (86.5\%, 0.8 points below) and DeepSeek (81.7\%, 5.6 below) include zero, and Jev's values are unchanged. These pairs come from 20 completions, so the result is fragile.

%% file: appendix/o_effort.tex
\section{Reasoning Effort}
\label{app:effort}

This appendix describes the effort runs and the calibration of Section~\ref{sec:llm} and reports what high effort changed. Its tables hold the evidence for the high-effort column of Table~\ref{tab:conditions} and for the effort remarks of Sections~\ref{sec:shared} and~\ref{sec:routing}. Throughout, Jev's verdicts, costs, and times come from the main runs, as do Luna's and Gemini's values in the medium condition. Appendix~\ref{app:decider} sets out the Decider's copy of the high condition, and what follows is Jev's.

\paragraph{Requests.}
The reasoning defaults of the main runs stated in Section~\ref{sec:llm} are those the vendors document.\footnote{\raggedright Vendor documentation, accessed 27 September 2026: \url{https://developers.openai.com/api/docs/models/gpt-5.6-luna} (Luna), \url{https://ai.google.dev/gemini-api/docs/latest-model} (Gemini), and \url{https://api-docs.deepseek.com/guides/thinking_mode/} (DeepSeek).} In the effort runs, and in the calibration runs that requested medium, every request named a reasoning effort, medium or high, in LiteLLM's \texttt{reasoning\_effort} parameter. LiteLLM, in version 1.101.0, passes the value unchanged to OpenAI as Luna's reasoning effort, turns it into Gemini's thinking level of the same name, with thought summaries requested, and forwards it to OpenRouter as the reasoning effort of each DeepSeek request. The request parameters stored with every cached response of these runs carry the level, and those of the main runs carry none. For DeepSeek we know only which level was requested. Everything else repeated the main runs: the units, the model identifiers, the temperatures, the logged seeds (so every pair's options appeared in their recorded order), and the requests in flight (16 for Luna, 20 for Gemini, and 32 for DeepSeek on the ordinal panels, and 8 on the binary panels). Each run began with an empty response cache, and a request could run for 600 seconds before it timed out, against 60 seconds in the main runs.

The effort and calibration runs were all made on 27 September 2026, and DeepSeek's medium run overlapped its high run in time. On RiceChem, two of DeepSeek's high requests received no reply and were not ended by the timeout. Each time we stopped the runner and restarted it from its checkpoint, which completed the pending units, so every pair has a verdict record. The run's summary, however, covers only its last session, and DeepSeek's high wall time was therefore not measured.

\begin{table*}[!tp]
\centering
\footnotesize
\setlength{\tabcolsep}{4.5pt}
\begin{tabular}{@{}lrrrrrrr@{}}
\toprule
 & \multicolumn{2}{c}{Luna} & \multicolumn{2}{c}{Gemini} & \multicolumn{3}{c}{DeepSeek} \\
\cmidrule(lr){2-3}\cmidrule(lr){4-5}\cmidrule(l){6-8}
 & Main & High & Main & High & Main & Medium & High \\
\midrule
\multicolumn{8}{@{}l}{\emph{(a) Nine panels}} \\
Failed requests & 0 & 1 & 0 & 0 & 9 & 6 & 17 \\
\quad recorded as UNMET & & & & & 3 & 3 & 8 \\
Abstentions & 12 & 14 & 29 & 29 & 24 & 17 & 21 \\
Prompt tokens & 12{,}063{,}678 & 12{,}061{,}045 & 10{,}972{,}689 & 10{,}972{,}689 & 11{,}241{,}762 & 10{,}870{,}560 & 10{,}942{,}101 \\
\quad cached (\%) & 79.7 & 80.3 & -- & -- & 67.5 & 76.0 & 74.6 \\
Completion tokens & 866{,}600 & 1{,}259{,}767 & 3{,}295{,}232 & 9{,}210{,}990 & 4{,}254{,}957 & 3{,}498{,}689 & 4{,}948{,}960 \\
\quad of which reasoning & 502{,}270 & 891{,}451 & 2{,}981{,}135 & 8{,}879{,}115 & 3{,}946{,}419 & 3{,}156{,}795 & 4{,}603{,}531 \\
Cost (US\$) & 1.84 & 2.30 & 20.59 & 42.77 & 4.18 & 4.44 & 5.90 \\
\quad ratio to Jev Choice & 29$\times$ & 36$\times$ & 325$\times$ & 675$\times$ & 66$\times$ & 70$\times$ & 93$\times$ \\
Wall time (s) & 859 & 1{,}237 & 950 & 3{,}361 & 6{,}298 & -- & -- \\
\quad ratio to Jev Choice & 30$\times$ & 43$\times$ & 33$\times$ & 117$\times$ & 220$\times$ & -- & -- \\
\midrule
\multicolumn{8}{@{}l}{\emph{(b) Ratio to the run in the column to the left}} \\
Completion tokens & & 1.45 & & 2.80 & & 0.82 & 1.41 \\
Reasoning tokens & & 1.77 & & 2.98 & & 0.80 & 1.46 \\
Cost at fixed prices & & 1.25 & & 2.08 & & 0.85 & 1.32 \\
Cost as billed & & 1.25 & & 2.08 & & -- & 1.33 \\
\bottomrule
\end{tabular}
\caption{The LLM judges' per-criterion runs, summed over the nine panels; every run recorded one request per pair, 5{,}003 in all, although DeepSeek's high run sent some of them again after its restarts (see text). Main: the main runs of 19 September; Medium and High: the effort runs of 27 September (Section~\ref{sec:llm}). Failed requests: requests without a usable reply, of which the harness recorded DeepSeek's on the binary panels as UNMET verdicts (next row). Abstentions: answers of CANNOT\_ASSESS or NA, not counting failures. Cached: the share of prompt tokens that the provider served from its prompt cache; Gemini reported none. Reasoning tokens, counted within the completion tokens, come from the cached responses; unlike Appendix~\ref{app:protocol}, Luna's main-run count leaves out the 70 spot-check responses that shared its cache. Cost: Luna and Gemini at list prices and DeepSeek at its billed cost, with ratios to Jev Choice's cost and wall time over the nine panels (Table~\ref{tab:cost}); the wall-time ratios carry the concurrency caveat of Section~\ref{sec:price}. DeepSeek's wall time is given for its main run only, because that of its high run was not measured and DeepSeek's wall times are not compared across effort levels. (b) For Luna and Gemini the high run over the main run; for DeepSeek the medium run over the main run and the high run over the medium run, which was made the same day. Fixed prices: one price table per judge for every run (see text). DeepSeek's billed-cost ratio to its main run is left out because OpenRouter routed the runs of the two dates to different upstream providers, so the difference between its billed main and high costs does not measure effort.}
\label{tab:ef-runs}
\end{table*}

\paragraph{Failures and costs.}
Table~\ref{tab:ef-runs} sums each run over the nine panels. Luna's one failed request at high effort, on ELLIPSE, was an infrastructure failure, and its pair is excluded like any failed ordinal pair. Of DeepSeek's 17 failed requests at high, one returned a reply that could not be parsed, and the other 16 fell in the same unclassified category as the empty responses of its main run (Appendix~\ref{app:protocol}). On the binary panels the harness again recorded DeepSeek's failed requests as UNMET verdicts. Tables~\ref{tab:ef-accuracy} and~\ref{tab:ef-downstream} keep them as verdicts, as Appendix~\ref{app:protocol} describes for the main runs. Setting them aside instead would change no class or Holm result in Table~\ref{tab:ef-accuracy}. Failed and hung requests left no usage record, so any charge for them is missing from the costs, as it is for the main runs.

As in Section~\ref{sec:metrics}, costs are list prices for Luna and Gemini, which the harness applied alike in every run, and OpenRouter's billed cost for DeepSeek. The harness's own estimates for DeepSeek do not compare across runs, because LiteLLM loads its price map from the network each time it starts and the map changed between runs. Besides effort, prompt caching affects the costs. Providers bill the prompt tokens they serve from their prompt cache at a lower price, and a cache serves the opening that a request shares with recent ones, such as the system prompt with which every request of a run begins (Appendix~\ref{app:wholerubric}). Luna's cached share of prompt tokens barely moved (79.7\% in its main run, 80.3\% at high), Gemini reported no cached tokens, and DeepSeek's share rose from 67.5\% to 76.0\% at medium and 74.6\% at high. Routing also affects DeepSeek's costs, and OpenRouter routed DeepSeek differently on the two dates. It sent the main run's requests mostly to DeepInfra (1{,}316), Morph (1{,}311), and Wafer (714), and those of both effort runs mostly to Parasail (2{,}039 at high and 2{,}111 at medium) and Together (1{,}286 and 1{,}269). We therefore compare effort levels on tokens and on costs at fixed prices. These price each run's uncached prompt, cached prompt, and completion tokens with one table per judge, which holds the list prices for Luna and Gemini and, for DeepSeek, the mean effective prices of its main run's billed responses. At those prices DeepSeek's high run would have cost 1.12 times its main run and its medium run 0.85 times. Measured against the medium condition, high effort raised the cost by 25\% for Luna, 108\% for Gemini, and 33\% for DeepSeek as billed (Table~\ref{tab:ef-runs}b). With completion tokens up 41\% to 180\% over that condition, the LLM judges in the high condition cost 36 to 675 times as much as Jev Choice (Table~\ref{tab:ef-runs}a).

DeepSeek's medium run used 0.82 times the main run's completion tokens and 0.80 times its reasoning tokens, and its high run 1.16 and 1.17 times, against high-over-main reasoning-token ratios of 1.8 for Luna and 3.0 for Gemini. DeepSeek's high run used 1.5 times the reasoning tokens of its medium run, which suggests that its default was already close to high. Luna's and Gemini's high runs took 1.44 and 3.54 times as long as their main runs with the same requests in flight, though on another day and with the longer timeout.

\begin{table}[!tp]
\centering
\footnotesize
\setlength{\tabcolsep}{5.2pt}
\begin{tabular}{@{}lrrr@{}}
\toprule
 & Luna & Gemini & DeepSeek \\
\midrule
\multicolumn{4}{@{}l}{\emph{(a) Pairs with the main run's verdict (\%)}} \\
Second default & 83.6 & 97.0 & 80.4 \\
Explicit medium & 86.0 & 96.2 & 79.6 \\
High & 83.7 & 94.0 & 78.9 \\
\midrule
\multicolumn{4}{@{}l}{\emph{(b) Difference from the second default run (points)}} \\
Explicit medium & $+2.5$ & $-0.8$ & $-0.8$ \\
\quad interval & $-1.1$, $+6.0$ & $-2.3$, $+0.7$ & $-3.6$, $+2.0$ \\
\quad bound, class & 5.4, -- & 2.0, parity & 3.2, parity \\
High & $+0.2$ & $-3.0$ & $-1.5$ \\
\quad interval & $-2.9$, $+3.2$ & $-5.2$, $-0.9$ & $-4.6$, $+1.3$ \\
\quad bound, class & 2.7, parity & 4.8, both & 4.1, parity \\
\midrule
\multicolumn{4}{@{}l}{\emph{(c) Completion tokens per unit, ratio to the main run}} \\
Second default & 1.01 & 1.03 & 1.05 \\
Explicit medium & 0.97 & 0.99 & 0.88 \\
High & 1.39 & 2.26 & 1.12 \\
\bottomrule
\end{tabular}
\caption{Calibration on the 106 repeat units (529 pairs): each LLM judge's second default run, explicit-medium run, and high run against its main run. (a) Identical verdicts as in Table~\ref{tab:wr-agreement}. DeepSeek's shares leave out the pairs of failed requests (see text), 525 pairs on 105 units in the medium comparisons and 526 in the high ones. (b) The run's share in (a) minus the second default run's, with its 95\% interval from resampling units, and bound and class as in Table~\ref{tab:clustered}; both: separated and at parity. (c) Median over units of a unit's completion tokens divided by those in the main run; DeepSeek's medians are over the 103, 105, and 106 units without a failed request in either run.}
\label{tab:ef-calibration}
\end{table}

\paragraph{Calibration.}
The calibration ran each LLM judge twice more on the 106 repeat units of Section~\ref{sec:llm}, once at medium effort (the \emph{explicit-medium run}) and once with no effort requested. The latter repeated the main run's requests eight days later, and we call it the \emph{second default run}. The high runs cover these units too. Table~\ref{tab:ef-calibration} compares each of the three runs with the main run in two ways. The first is the share of pairs on which a run returns the main run's verdict, less the second default run's share on the same pairs. If requesting an effort changed nothing, a run that requested one and the second default run, both made on the same day, eight days after the main run, would be exchangeable with each other, and this difference would be zero in expectation. Pairs that a failed request touched in any of the runs compared are set aside, which removes four of DeepSeek's pairs from its medium comparisons and three from its high ones, in both cases including the single pair of a HealthBench unit whose second default run failed as a whole. The second comparison divides a run's completion tokens summed over each unit's requests by the main run's, and the table gives the median of these ratios over units. For Luna and Gemini each unit's prompt-token count was also identical in all four runs, which confirms that the runs sent the same prompts.

Luna's explicit-medium run used almost the same completion tokens per unit as its main run (median ratio 0.97, against 1.01 for its second default run), and it gave the main run's verdict 2.5 points more often than the second default run did, although on 529 pairs that difference is too uncertain to show parity. Gemini's explicit-medium run matched its main run on both counts. Reading the two judges' main runs as medium thus rests on agreement and tokens for Gemini and chiefly on tokens for Luna. DeepSeek's explicit-medium run agreed with its main run at parity with the second default run but spent fewer completion tokens per unit (median ratio 0.88, against 1.05), while its high run came closer to the second default run (1.12). High effort raised the median ratio to 1.39 for Luna and 2.26 for Gemini. Luna's high verdicts still agreed with its main run at parity with its second default run, whereas Gemini's agreed 3.0 points less often, a separated difference that stays within the 5-point margin.

DeepSeek's baseline also reflects eight days of changes on the serving side. Although its requests repeated those of the main run byte for byte, its second default run's prompt-token count matched the main run's on only 5 of 103 units. Its explicit-medium and high runs, made the same day as its second default run, return that run's verdict on 86.7\% and 86.3\% of the pairs, against 80.4\% between the main run and the second default run. Each of DeepSeek's effort runs is thus compared with a baseline that spans the same eight days as its own comparison with the main run.

\begin{table*}[!tp]
\centering
\scriptsize
\setlength{\tabcolsep}{3.7pt}
\begin{tabular}{@{}l@{\hspace{6pt}}rrl@{\hspace{8pt}}rrl@{\hspace{8pt}}rrl@{\hspace{8pt}}rrl@{}}
\toprule
 & \multicolumn{9}{c}{High minus main run} & \multicolumn{3}{c}{Medium minus main run} \\
\cmidrule(lr){2-10}\cmidrule(l){11-13}
 & \multicolumn{3}{c}{Luna} & \multicolumn{3}{c}{Gemini} & \multicolumn{3}{c}{DeepSeek} & \multicolumn{3}{c}{DeepSeek} \\
\cmidrule(lr){2-4}\cmidrule(lr){5-7}\cmidrule(lr){8-10}\cmidrule(l){11-13}
Panel & Diff & 95\% interval & Class & Diff & 95\% interval & Class & Diff & 95\% interval & Class & Diff & 95\% interval & Class \\
\midrule
\multicolumn{13}{@{}l}{\emph{(a) Each judge's accuracy}} \\
RiceChem & $+1.5$ & $-0.2$, $+3.2$ & parity & $+0.5$ & $-0.7$, $+1.7$ & parity & $-0.6$ & $-2.7$, $+1.5$ & parity & $-1.6$ & $-3.9$, $+0.7$ & parity \\
HealthBench & $+0.7$ & $-2.2$, $+3.8$ & parity & $-0.7$ & $-2.7$, $+1.0$ & parity & $+0.5$ & $-2.3$, $+3.3$ & parity & $+2.0$ & $-1.0$, $+5.0$ & parity \\
ELLIPSE & $+0.1$ & $-1.3$, $+1.4$ & parity & $+2.5$\ddg & $+1.2$, $+3.7$ & both & $-1.9$ & $-3.6$, $-0.3$ & both & $-2.3$\dg & $-3.9$, $-0.7$ & both \\
FED-Turn & $-2.5$ & $-5.2$, $+0.2$ & parity & $-0.5$ & $-2.3$, $+1.3$ & parity & $-1.7$ & $-4.2$, $+0.8$ & parity & $-2.5$ & $-5.0$, $0.0$ & parity \\
FED-Dialogue & $-1.7$ & $-5.1$, $+1.7$ & parity & $-0.4$ & $-3.0$, $+1.7$ & parity & $-0.4$ & $-5.3$, $+4.7$ & parity & $-3.0$ & $-7.6$, $+1.3$ & -- \\
HelpSteer2 & $+1.9$ & $-1.9$, $+5.8$ & -- & $-2.3$ & $-5.1$, $+0.3$ & parity & $-1.1$ & $-5.4$, $+3.1$ & parity & $-2.0$ & $-6.7$, $+2.8$ & -- \\
LFQA & $-3.1$ & $-6.1$, $-0.3$ & separated & $+1.1$ & $-1.7$, $+3.9$ & parity & $+3.7$ & $-0.3$, $+7.4$ & -- & $+1.1$ & $-2.3$, $+4.5$ & parity \\
USR-TC & $+1.1$ & $-2.2$, $+4.4$ & parity & $-0.8$ & $-3.1$, $+1.1$ & parity & $+5.0$ & $+0.6$, $+9.7$ & separated & $+6.1$\dg & $+1.9$, $+10.6$ & separated \\
USR-PC & $+0.3$ & $-3.3$, $+4.0$ & parity & $-0.7$ & $-3.0$, $+1.7$ & parity & $-0.3$ & $-3.7$, $+3.0$ & parity & $+2.7$ & $-0.3$, $+5.7$ & -- \\
\addlinespace[2pt]
Nine panels & $-0.0$ & $-0.9$, $+0.8$ & parity & $+0.5$ & $-0.1$, $+1.2$ & parity & $-0.4$ & $-1.3$, $+0.6$ & parity & $-0.7$ & $-1.7$, $+0.2$ & parity \\
Seven ordinal & $-0.5$ & $-1.4$, $+0.5$ & parity & $+0.7$ & $-0.1$, $+1.4$ & parity & $-0.4$ & $-1.5$, $+0.8$ & parity & $-0.8$ & $-1.9$, $+0.3$ & parity \\
All but ELLIPSE & $-0.1$ & $-1.1$, $+1.0$ & parity & $-0.4$ & $-1.0$, $+0.4$ & parity & $+0.4$ & $-0.8$, $+1.5$ & parity & $0.0$ & $-1.2$, $+1.2$ & parity \\
\bottomrule
\end{tabular}

\medskip
\footnotesize
\setlength{\tabcolsep}{3.6pt}
\begin{tabular}{@{}l@{\hspace{6pt}}rlrl@{\hspace{10pt}}rlrl@{\hspace{10pt}}rlrl@{}}
\toprule
 & \multicolumn{4}{c}{Luna} & \multicolumn{4}{c}{Gemini} & \multicolumn{4}{c}{DeepSeek} \\
\cmidrule(lr){2-5}\cmidrule(lr){6-9}\cmidrule(l){10-13}
 & \multicolumn{2}{c}{High} & \multicolumn{2}{c}{Main} & \multicolumn{2}{c}{High} & \multicolumn{2}{c}{Main} & \multicolumn{2}{c}{High} & \multicolumn{2}{c}{Main} \\
\cmidrule(lr){2-3}\cmidrule(lr){4-5}\cmidrule(lr){6-7}\cmidrule(lr){8-9}\cmidrule(lr){10-11}\cmidrule(l){12-13}
Panel & Diff & Class & Diff & Class & Diff & Class & Diff & Class & Diff & Class & Diff & Class \\
\midrule
\multicolumn{13}{@{}l}{\emph{(b) Against Jev Choice}} \\
RiceChem & $-1.7$ & parity & $-3.2$ & separated & $-4.4$ & separated & $-4.9$ & separated & $-2.3$ & parity & $-1.7$ & parity \\
HealthBench & $-5.9$ & separated & $-6.7$ & separated & $+1.7$ & parity & $+2.5$ & -- & $-0.2$ & parity & $-0.7$ & parity \\
ELLIPSE & $+0.7$ & parity & $+0.6$ & parity & $+18.9$ & separated & $+16.4$ & separated & $-0.3$ & parity & $+1.6$ & parity \\
FED-Turn & $-1.5$ & -- & $+1.0$ & parity & $+3.2$ & -- & $+3.7$ & -- & $+0.5$ & parity & $+2.2$ & -- \\
FED-Dialogue & $-2.1$ & -- & $-0.4$ & -- & $+7.4$ & separated & $+7.9$ & separated & $-2.2$ & -- & $-1.7$ & -- \\
HelpSteer2 & $-7.7$ & separated & $-10.2$ & separated & $+3.1$ & -- & $+5.7$ & separated & $-3.4$ & -- & $-2.3$ & -- \\
LFQA & $-0.3$ & parity & $+2.8$ & -- & $+0.3$ & parity & $-1.1$ & parity & $0.0$ & parity & $-3.7$ & -- \\
USR-TC & $+1.4$ & -- & $+0.3$ & parity & $+2.2$ & -- & $+3.1$ & -- & $+3.6$ & -- & $-1.4$ & -- \\
USR-PC & $+4.3$ & -- & $+4.0$ & separated & $+3.7$ & -- & $+4.3$ & -- & $-0.7$ & parity & $-0.3$ & parity \\
\bottomrule
\end{tabular}
\caption{Accuracy by reasoning effort, with units resampled as in Table~\ref{tab:clustered}. (a) Each LLM judge's high run minus its main run, and DeepSeek's medium run minus its main run, in points, over the pairs scored in both runs and measured as in Table~\ref{tab:wr-accuracy}; DeepSeek's failed requests on the binary panels count as UNMET verdicts. Class as in Table~\ref{tab:clustered}, with both marking a difference that is separated and at parity. \ddag: significant after Holm correction within the 27 high-effort comparisons, for the bootstrap $p$-values and for both sign tests; \dag: significant after correction within DeepSeek's nine medium comparisons, in pair-level sign tests only. The last three rows pool panels and lie outside both corrections. (b) Each LLM judge against Jev Choice at high effort and in the main runs: the LLM judge's accuracy less Jev Choice's, in points, and class, as in Table~\ref{tab:clustered}.}
\label{tab:ef-accuracy}
\end{table*}

\paragraph{Accuracy.}
Table~\ref{tab:ef-accuracy}a sets each judge's high run, and also DeepSeek's medium run, against the same judge's main run over the pairs that both runs scored. The intervals and classes are those of Section~\ref{sec:metrics}, and for Holm correction the 27 high-effort comparisons form one family and DeepSeek's nine medium comparisons another. Of the 27 high-effort differences, 12 are positive and 15 negative. Twenty-three are at parity, two of them also separated, two are separated without parity, and two are inconclusive. The four separated differences are Gemini's gain and DeepSeek's loss on ELLIPSE, Luna's loss on LFQA, and DeepSeek's gain on USR-TC, and correction leaves only Gemini's ELLIPSE gain of 2.5 points significant, in the bootstrap and in both sign tests. At high effort each judge's accuracy pooled over the nine panels moved by at most 0.5 points. Without ELLIPSE, Gemini's pooled difference turns negative. Among DeepSeek's medium comparisons, its ELLIPSE loss and USR-TC gain stay significant after correction in pair-level sign tests but not in the bootstrap or the sign test over units.

Against Jev Choice, the high runs change the class of nine of the 27 comparisons (Table~\ref{tab:ef-accuracy}b). Three separations lapse (Jev Choice's lead over Luna on RiceChem, Gemini's lead on HelpSteer2, and Luna's lead on USR-PC), and no comparison becomes separated, so Jev Choice keeps five of its eight separations, none reversed. Of Jev Choice's three separated leads at high, one is over Gemini (RiceChem) and two over Luna (HealthBench and HelpSteer2), and Gemini keeps its two (ELLIPSE and FED-Dialogue). In the medium condition two comparisons become separated, both with DeepSeek, which was never separated from Jev Choice in the main runs. Jev Choice leads on RiceChem by 3.3 points and DeepSeek on USR-TC by 4.7, which brings the medium condition to 10 separated comparisons of the 27. Neither stays separated at high.

\begin{table*}[!tp]
\centering
\footnotesize
\setlength{\tabcolsep}{3.4pt}
\begin{tabular}{@{}l@{\hspace{6pt}}lrlrr@{\hspace{8pt}}rrr@{\hspace{8pt}}rrr@{\hspace{8pt}}r@{}}
\toprule
 & \multicolumn{2}{c}{Best single, H} & \multicolumn{3}{c}{Oracle cascade, H} & \multicolumn{3}{c}{Oracle vs.\ best} & \multicolumn{3}{c}{Cross-fitted vs.\ best} & \makecell[r]{Halvings\\$>0$, H} \\
\cmidrule(lr){2-3}\cmidrule(lr){4-6}\cmidrule(lr){7-9}\cmidrule(lr){10-12}
Panel & Judge & Acc. & Fallback & Deferred & Acc. & M & Med & H & M & Med & H & \\
\midrule
\multicolumn{13}{@{}l}{\emph{(a) Cascades}} \\
RiceChem & Jev Choice & 81.0 & DeepSeek & 14.8 & 81.2 & $-0.4$ & $+0.2$ & $+0.2$ & $-1.0$ & $-0.6$ & $-0.4$ & 14 \\
HealthBench & Gemini & 78.8 & Gemini & 33.5 & 81.5 & $+2.0$ & $+2.0$ & $+2.7$ & $+1.5$ & $+1.8$ & $+2.4$ & 88 \\
\midrule
ELLIPSE & Gemini & 32.3 & Gemini & 99.9 & 32.3 & $0.0$ & $0.0$ & $0.0$ & $-0.1$ & $-0.1$ & $-0.1$ & 0 \\
FED-Turn & Jev Score & 70.7 & Gemini & 97.2 & 64.7 & $-5.2$ & $-5.2$ & $-6.0$ & $-7.3$ & $-7.0$ & $-7.7$ & 0 \\
FED-Dialogue & Gemini & 59.2 & Gemini & 85.8 & 59.2 & $0.0$ & $0.0$ & $0.0$ & $-0.9$ & $-0.7$ & $+0.2$ & 50 \\
HelpSteer2 & Gemini & 51.7 & Gemini & 30.1 & 53.4 & $+0.6$ & $+0.6$ & $+1.7$ & $-0.9$ & $-0.9$ & $+1.0$ & 74 \\
LFQA & Jev Score & 69.6 & Luna$^{a}$ & 34.4 & 70.9 & $+0.6$ & $+0.6$ & $+1.4$ & $+0.6$ & $+0.6$ & $+1.8$ & 98 \\
USR-TC & DeepSeek & 71.1 & DeepSeek & 64.2 & 71.4 & $+0.3$ & $0.0$ & $+0.3$ & $-0.7$ & $+0.1$ & $+0.2$ & 52 \\
USR-PC & Luna & 74.6 & Gemini & 28.4 & 75.6 & $+1.3$ & $+1.3$ & $+1.0$ & $+0.8$ & $+0.8$ & $-0.0$ & 36 \\
\bottomrule
\end{tabular}

\medskip
\setlength{\tabcolsep}{4pt}
\begin{tabular}{@{}l@{\hspace{6pt}}lrrr@{\hspace{10pt}}rrr@{\hspace{10pt}}rrr@{}}
\toprule
 & \multicolumn{4}{c}{\makecell{Matching the best\\LLM judge, H}} & \multicolumn{3}{c}{\makecell{Repeated errors,\\excess (points)}} & \multicolumn{3}{c}{\makecell{Jury\\vs.\ best (points)}} \\
\cmidrule(lr){2-5}\cmidrule(lr){6-8}\cmidrule(l){9-11}
Panel & Judge & \makecell[r]{Oracle\\cost} & \makecell[r]{Cross-\\fitted cost} & \makecell[r]{Held-\\out} & M & Med & H & M & Med & H \\
\midrule
\multicolumn{11}{@{}l}{\emph{(b) By panel}} \\
RiceChem & Luna & -- & 5.7 & $+1.3$ & -- & -- & -- & $-2.8$\dg & $-2.9$\dg & $-1.8$ \\
HealthBench & Gemini & 8.9 & 11.1 & $-0.1$ & -- & -- & -- & $-2.2$ & $-1.2$ & $-1.7$ \\
\midrule
ELLIPSE & Gemini & 100.0 & 99.7 & $-0.1$ & $+8.6$ & $+9.1$ & $+8.0$ & $-12.8$\dg & $-13.9$\dg & $-16.2$\dg \\
FED-Turn & Gemini & 97.2 & 48.2 & $-1.7$ & $+22.3$ & $+23.7$ & $+24.5$ & $-0.2$ & $-2.0$ & $-2.0$ \\
FED-Dialogue & Gemini & 85.9 & 68.7 & $-2.0$ & $+20.4$ & $+20.6$ & $+20.3$ & $-3.1$ & $-5.3$\dg & $-4.4$ \\
HelpSteer2 & Gemini & 7.8 & 21.8 & $-1.4$ & $+32.0$ & $+31.8$ & $+31.6$ & $-8.0$\dg & $-6.9$\dg & $-4.9$\dg \\
LFQA & Gemini & 48.0 & 24.8 & $+0.7$ & $+34.2$ & $+33.7$ & $+30.6$ & $-2.0$ & $-1.4$ & $-0.3$ \\
USR-TC & DeepSeek & 52.3 & 50.5 & $-0.5$ & $+28.0$ & $+26.9$ & $+28.6$ & $0.0$ & $+0.8$ & $+0.3$ \\
USR-PC & Luna & 21.1 & 34.1 & $-0.6$ & $+18.8$ & $+18.7$ & $+20.0$ & $-0.7$ & $+0.7$ & $-2.0$ \\
\bottomrule
\end{tabular}

\medskip
\setlength{\tabcolsep}{3pt}
\begin{tabular}{@{}>{\raggedright\arraybackslash}p{9.0cm}>{\raggedright\arraybackslash}p{2.1cm}>{\raggedright\arraybackslash}p{2.1cm}>{\raggedright\arraybackslash}p{2.1cm}@{}}
\toprule
 & M & Med & H \\
\midrule
\multicolumn{4}{@{}l}{\emph{(c) Pooled statistics}} \\
Jev Choice against an LLM judge (27 comparisons): separated, parity, inconclusive & 8, 8, 11 & 10, 7, 10 & 5, 11, 11 \\
Ordinal panels with judge--judge above judge--label QWK, of 7 & 7 & 7 & 7 \\
Least-agreeing judge pair above the best judge--label QWK, panels of 7 & 5 & 5 & 6 \\
Negative panel offsets, five judges on seven ordinal panels, of 35 & 31 & 31 & 30 \\
Ordinal panels with every matched judge's offset negative, of 7 & 6 & 6 & 6 \\
Confident errors: LLM verdicts repeating Jev's answer, of 252 & 242 (96.0\%) & 245 (97.2\%) & 245 (97.2\%) \\
Confident errors: independence baseline (\%) & 50.3 & 50.4 & 51.2 \\
Confident errors: repeats by Luna, Gemini, and DeepSeek, of 84 each & 81, 80, 81 & 81, 80, 84 & 82, 80, 83 \\
Confident errors: three-LLM median gives Jev's answer, of 84 & 82 & 84 & 84 \\
Four-judge upper median: largest gain over the most accurate matched judge (points) & $+1.0$ (USR-PC) & $+2.3$ (USR-PC) & $+1.7$ (LFQA) \\
\bottomrule
\end{tabular}
\caption{The analyses of Sections~\ref{sec:shared} and~\ref{sec:routing} in the main runs (M), the medium condition (Med), and the high condition (H). Jev's verdicts are the same in all three. (a) Jev-first cascades as in Tables~\ref{tab:cascade} and~\ref{tab:cascade-ledger}: at high, the best single judge and, for the cascade with an oracle threshold, its fallback, deferred share (\%), and accuracy (\%); in each condition, the gain of that condition's best oracle-threshold cascade over its best single judge, and the cross-fitted gain, in points; and the share (\%) of the 50 halvings with a positive cross-fitted gain at high. $^{a}$Luna and DeepSeek tie as fallbacks (see text). (b) Matching the best LLM judge, with that judge chosen at high: the cost (\%) of the oracle-threshold cascade (as under Reaching the fallback, Appendix~\ref{app:cascades}) and of the cross-fitted cascade (as in Table~\ref{tab:cascade}), and the cross-fitted held-out difference (points). Repeated errors: the share of the LLM judges' verdicts on Jev Choice's wrong pairs that give its answer, minus the independence baseline (Section~\ref{sec:shared-errors}), on the ordinal panels. Jury: accuracy of the three-LLM median less that of the most accurate matched judge (Section~\ref{sec:juries}), with \dag\ where the 95\% interval excludes zero on the negative side. (c) Jev Choice's 27 comparisons with an LLM judge by class; judge--judge against judge--label QWK and the offsets of the five judges, as in Tables~\ref{tab:main}b, \ref{tab:offsets}, and~\ref{tab:secondary}h, on the seven ordinal panels; the confident-error sample (Section~\ref{sec:shared-errors}); and juries.}
\label{tab:ef-downstream}
\end{table*}

\paragraph{Shared errors, cascades, and juries.}
Table~\ref{tab:ef-downstream} repeats the analyses of Sections~\ref{sec:shared} and~\ref{sec:routing} for each condition. In all three conditions judge--label agreement stays below judge--judge agreement on all seven ordinal panels, and every offset of Jev's five judges outside USR-TC is negative, so the shortfall of Section~\ref{sec:agreement} persists. On USR-TC, Gemini's offset of $-0.005$ levels in the main runs becomes $+0.02$ at high. At high the least-agreeing judge pair also exceeds the best judge--label QWK on USR-TC and USR-PC, but no longer on ELLIPSE, where it falls short by less than 0.01. Counting every error of Jev Choice, the excess of repeated errors over the baseline moves by less than 4 points on each ordinal panel, most on LFQA. The confident-error sample is the same in every condition, since it depends on Jev's verdicts alone. On four panels its cut-off splits errors of tied confidence (Appendix~\ref{app:shared}).

With oracle thresholds the cascade's accuracy at high fell on five panels, rose on three, and stayed the same on HealthBench. Its gain over the best single judge grew most on HelpSteer2, LFQA, and HealthBench, each time because the best single judge lost more accuracy than the cascade, whereas on RiceChem the cascade itself improved, from 80.6\% to 81.2\%, with DeepSeek in place of Gemini as its fallback. On HealthBench the cascade with Gemini as fallback is right on 331 of 406 pairs in both conditions, and Gemini alone on 323 and then 320. Among its kept pairs, 15 are right for Jev alone and 4 for Gemini alone at high ($p = 0.019$, pair-level sign test), where the main runs had 18 and 10 ($p = 0.18$). Eleven of Gemini's verdicts on kept pairs changed, 7 from right to wrong and 4 from wrong to right. On the deferred pairs at high, Gemini alone is right on 33 and Jev alone on 16 ($p = 0.021$). With cross-fitted thresholds the HealthBench cascade's held-out accuracy fell from 80.0\% to 79.6\% and the best single judge's from 78.5\% to 77.3\%. Gemini was the fallback chosen in 53 of the 100 halves at high, against 71 in the main runs. The cross-fitted gain had already risen to 1.8 points in the medium condition, where only DeepSeek's verdicts differ from the main runs.

On LFQA, Luna's accuracy fell from 71.5\% to 68.4\%, which leaves Jev Score (69.6\%) the best single judge, and the cascade (70.9\%) is 1.4 points above it. Luna and DeepSeek tie as its fallback, each giving 254 correct verdicts on the 358 pairs at $\tau = 0.6$ with 123 pairs deferred. Jev alone is right on 12 kept pairs and Luna alone on 3 ($p = 0.035$), or DeepSeek alone on 5 against 13 for Jev ($p = 0.096$). In the main runs' cascade the counts had been 2 against 0 on its 171 kept pairs ($p = 0.50$). On HelpSteer2 the cascade fell from 54.8\% to 53.4\% at $\tau = 0.4$, but Gemini alone fell from 54.3\% to 51.7\%. Jev alone is right on 25 of its kept pairs and Gemini alone on 19 ($p = 0.45$), against 14 and 12 in the main runs' cascade ($p = 0.85$). As for the main runs (Table~\ref{tab:cascade-split}), these $p$-values are nominal, uncorrected for the number of panels, and computed at oracle thresholds. FED-Turn's best cascade defers 97.2\% of its pairs and still trails Jev Score by 6.0 points.

When the best LLM judge of the high condition is the fallback to be matched, the cross-fitted cascade costs less than half of that judge's cost on six panels, 5.7\% to 48.2\%, with held-out accuracy from 1.7 points below that judge to 1.3 above; in the main runs it did so on seven panels (Jev first in Table~\ref{tab:cascade}).

The three-LLM median trails the most accurate matched judge on eight panels at high, with 95\% intervals below zero on ELLIPSE and HelpSteer2, and is ahead of that judge on USR-TC alone, by one pair (0.3 points). In the medium condition it exceeds that judge on USR-TC and USR-PC, by 0.8 and 0.7 points. In no condition does the three-LLM median or the four-judge upper median have a 95\% interval above zero against that judge. At medium and high effort the three-LLM median agrees with Jev's wrong answer on every one of the 84 confident errors.

%% file: appendix/p_higher_tier.tex
\section{Higher-Tier Judges}
\label{app:highertier}

Here we set out how the higher-tier runs of Section~\ref{sec:llm} were made, what they cost, how long they took, and how accurate they were. Table~\ref{tab:conditions} summarizes them in its higher-tier column, and Sections~\ref{sec:shared-errors} and~\ref{sec:cascades} draw on the tables below. Jev's and the flash-tier judges' values all come from the main runs, and those of the high condition from Appendix~\ref{app:effort}. Apart from the most accurate judges of the Decider's copy, named in the accuracy paragraph below, the tables and comparisons here are Jev's copy; the rest of the Decider's copy of this condition is in Appendix~\ref{app:decider}.

\paragraph{Requests.}
The higher-tier judges' replies name the models \texttt{gpt-5.6-terra}, \texttt{gemini-3.1-pro-preview}, and \texttt{deepseek/deepseek-v4-pro-0813}, and no reply names another. Section~\ref{sec:llm} pairs each with the flash-tier judge of the same vendor, but none is necessarily that vendor's most capable or most recent model.\footnote{\raggedright Vendor documentation, accessed 2 October 2026: \url{https://developers.openai.com/api/docs/models/gpt-5.6-terra}, \url{https://developers.openai.com/api/docs/changelog}, and \url{https://developers.openai.com/api/docs/guides/reasoning} (Terra), \url{https://ai.google.dev/gemini-api/docs/models}, \url{https://ai.google.dev/gemini-api/docs/changelog}, and \url{https://ai.google.dev/gemini-api/docs/generate-content/thinking} (Gemini 3.1 Pro), and \url{https://api-docs.deepseek.com/quick_start/pricing}, \url{https://api-docs.deepseek.com/news/news260910}, and \url{https://api-docs.deepseek.com/guides/thinking_mode/} (DeepSeek V4 Pro).} OpenAI released GPT-5.6 Terra together with Luna on 9 July 2026 as the GPT-5.6 model that ``balances intelligence and cost'', and it names GPT-5.6 Sol, not Terra, as the GPT-5.6 flagship. Google lists Gemini 3.1 Pro, the only Pro model of its 3.x line, as a Preview, released on 19 February 2026, whereas Gemini 3.8 Flash became generally available on 2 September 2026. The DeepSeek V4 Pro release we used (0813) predates V4.1 Flash, released on 10 September, and DeepSeek reports that ``Tests by multiple parties put V4.1-Flash ahead of V4-Pro''.

No request named a reasoning effort or a thinking level, as the request parameters stored with all 5{,}003 cached responses of each judge confirm, so the vendors' defaults applied. OpenAI documents medium for GPT-5.6 models. For Gemini 3.1 Pro, whose thinking cannot be switched off, Google documents high, one level above the default of Gemini's main run (Section~\ref{sec:llm}). Gemini 3.1 Pro nonetheless emitted 1.01 times the reasoning tokens of that run and about a third of those of Gemini's high run (Table~\ref{tab:ef-runs}). DeepSeek documents high for its own API, and through OpenRouter the default is not documented, but DeepSeek V4 Pro emitted as many reasoning tokens as DeepSeek did (Table~\ref{tab:ht-runs}b).

From its flash sibling each higher-tier judge kept the temperature, which every request carried (1.0 for Terra, as for Luna, and 0 for the other two); the logged seeds, so that every pair's options appeared in the order recorded for that sibling; and the number of requests in flight (16 for Terra, 20 for Gemini 3.1 Pro, and 32 for DeepSeek V4 Pro on each ordinal panel, against 8 on each binary one). Requests timed out after 600 seconds, ten times the limit of the main runs, and went through LiteLLM 1.101.0. The three judges ran in parallel, starting at 13:20 UTC on 2 October 2026, and the full runs ended 5{,}907 seconds after they began. A pilot on 36 units (211 requests per judge) with the same seeds, made just before, had no failure and enters no analysis. OpenRouter did not pin DeepSeek V4 Pro to a provider, and 13 upstream providers served its 5{,}003 replies, Baidu 3{,}518 of them (70\%), Ionstream 708, and Wafer 307. DeepSeek's own API served none.

\begin{table*}[!tp]
\centering
\footnotesize
\setlength{\tabcolsep}{6pt}
\begin{tabular}{@{}lrrr@{}}
\toprule
 & Terra & Gemini 3.1 Pro & DeepSeek V4 Pro \\
\midrule
\multicolumn{4}{@{}l}{\emph{(a) Full runs, nine panels}} \\
Requests (failed) & 5{,}003 (0) & 5{,}003 (0) & 5{,}003 (0) \\
Abstentions & 19 & 45 & 25 \\
Prompt tokens & 12{,}063{,}678 & 10{,}972{,}689 & 12{,}563{,}791 \\
\quad cached (\%) & 80.3 & 0.1 & 71.8 \\
Completion tokens & 571{,}757 & 3{,}364{,}733 & 4{,}299{,}548 \\
\quad of which reasoning & 224{,}235 & 3{,}004{,}487 & 3{,}931{,}241 \\
Cost recorded by the harness (US\$) & 14.73 & 62.30 & 11.06 \\
Cost at list prices (US\$) & 14.73 & 62.30 & 11.05 \\
Cost as billed (US\$) & 14.73 & 62.30 & 4.27 \\
\quad ratio to Jev Choice & 233$\times$ & 984$\times$ & 67$\times$ \\
Wall time, run summaries (s) & 990 & 4{,}645 & -- \\
\quad ratio to Jev Choice & 35$\times$ & 162$\times$ & -- \\
Wall time, summed sessions (s) & 1{,}006 & 4{,}658 & 5{,}826$^{a}$ \\
Request latency, mean\,/\,median\,/\,maximum (s) & 2.4\,/\,2.0\,/\,31.5 & 12.1\,/\,7.5\,/\,584.3 & 16.5\,/\,8.7\,/\,332.3 \\
Unit duration, mean\,/\,median\,/\,maximum (s) & 11.3\,/\,9.9\,/\,39.3 & 58.5\,/\,35.6\,/\,878.3 & 76.4\,/\,49.4\,/\,372.7 \\
Requests at a time, ordinal\,/\,binary panels & 16\,/\,8 & 20\,/\,8 & 32\,/\,8 \\
\midrule
\multicolumn{4}{@{}l}{\emph{(b) Ratio to the flash sibling (Luna, Gemini, and DeepSeek in the main runs)}} \\
Completion tokens & 0.66 & 1.02 & 1.01 \\
Reasoning tokens & 0.45 & 1.01 & 1.00 \\
Cost at list prices & 7.99 & 3.03 & 3.54 \\
Cost as recorded & 7.99 & 3.03 & 3.52 \\
Cost as billed & 7.99 & 3.03 & 1.02 \\
Wall time, run summaries & 1.15 & 4.89 & -- \\
Mean request latency & 1.17 & 4.31 & 1.16 \\
\midrule
\multicolumn{4}{@{}l}{\emph{(c) Pilot, 36 units and 211 requests per judge}} \\
Cost as billed (US\$) & 0.549 & 2.006 & 0.136 \\
Cost as recorded (US\$) & 0.549 & 2.006 & 0.463 \\
Wall time (s) & 98 & 176 & 806 \\
\bottomrule
\end{tabular}
\caption{Cost and time of the higher-tier runs. (a) The full runs, per-criterion judgments on the nine panels. Abstentions: CANNOT\_ASSESS or NA answers (see text). Cached: prompt tokens served from the provider's prompt cache, as a share of all prompt tokens. Reasoning tokens are part of the completion tokens and are read from the cached responses. Costs: the harness's record; the same tokens at the vendors' list prices (Appendix~\ref{app:cost}), DeepSeek's off-peak prices for DeepSeek V4 Pro; and the bill, which for DeepSeek V4 Pro sums the costs OpenRouter returned. Ratios to Jev Choice's cost (\$0.0633) and wall time (28.6 seconds) over the nine panels (Table~\ref{tab:cost}) use the billed cost and the run summaries; for the concurrency behind the time ratios see Section~\ref{sec:price}. Summed sessions: the time between start and end of every session of a run, added up. $^{a}$Stopped and resumed twice; the sum includes two stalls and bounds the wall time at 204 times Jev Choice's (178 times without the stalls; see text). Latency: per request, from the cached responses; unit duration: the time a unit took to be judged, for resumed units in their finishing session. (b) Each higher-tier judge over its flash sibling; DeepSeek V4 Pro's wall time is not compared with DeepSeek's (see text). (c) The pilot (see text); its judges ran in parallel, 806 seconds in all.}
\label{tab:ht-runs}
\end{table*}

\paragraph{Failures and costs.}
Every request of the three judges returned a usable reply, and no unit failed (Table~\ref{tab:ht-runs}a). The judges abstained with NA on 19 (Terra), 32 (Gemini 3.1 Pro), and 23 (DeepSeek V4 Pro) ordinal pairs, and with CANNOT\_ASSESS on 13 (Gemini 3.1 Pro) and 2 (DeepSeek V4 Pro) embedded binary pairs, all on USR-TC; none abstained on the binary panels. As with DeepSeek's high run, two of DeepSeek V4 Pro's requests, one on an ELLIPSE unit and one on a FED-Turn unit, hung without a reply. After its runner had written no verdict record for about six minutes (373 and 358 seconds), we stopped it each time and resumed the run from its checkpoint; the resumed sessions completed the pending units without duplicating any. DeepSeek V4 Pro therefore judged the seven ordinal panels in three sessions, and its run summaries, which cover only the last session, do not measure its wall time over the nine panels. Added up, the sessions took 5{,}826 seconds, stalls included, and about 5{,}095 seconds without the stalls. Four of its panels have measured times. RiceChem (1{,}733 seconds) and HealthBench (942) each ran as a session of its own, and USR-TC (144) and USR-PC (125) were judged wholly in the last session. FED-Dialogue, LFQA, and HelpSteer2 finished within the second session, which no run summary covers, and only ELLIPSE and FED-Turn straddle a stopped session. Every other run went through without a stop. Terra and Gemini 3.1 Pro took 35 and 162 times as long as Jev Choice, at the main runs' concurrency (Table~\ref{tab:ht-runs}a). DeepSeek V4 Pro's times are not compared with DeepSeek's, whose main run had a 60-second timeout, failed nine requests, and was routed to a different mix of providers. The time ratios to the flash siblings in Table~\ref{tab:ht-runs}b compare runs that kept the requests in flight but were made 13 days apart.

Terra cost 233 times as much as Jev Choice, Gemini 3.1 Pro 984 times, and DeepSeek V4 Pro 67 times as billed (Table~\ref{tab:ht-runs}a). The costs of Terra and Gemini 3.1 Pro are their list prices, which the harness recorded and which we take as billed. OpenAI's prompt cache served 80.3\% of Terra's prompt tokens, and nearly all the rest, 2{,}359{,}194 tokens, were written to that cache, which OpenAI prices at \$2.50 per million, 1.25 times its input rate. Gemini 3.1 Pro reported cached tokens on 31 replies only. DeepSeek V4 Pro's bill, \$4.27, is 0.39 times its cost at DeepSeek's off-peak list prices, \$11.05 (the runs fell outside DeepSeek's peak hours), because OpenRouter sent most requests to Baidu. The harness recorded \$0.016 more than the list price, from 50 replies costed above the off-peak rates, a difference we could not trace. Priced at its recorded cost, DeepSeek V4 Pro comes to 175 times Jev Choice's cost instead of 67 times, and to 3.52 times DeepSeek's recorded cost instead of 1.02 times its bill. DeepSeek's own bill exceeded the harness's estimate (Appendix~\ref{app:cost}) and DeepSeek V4 Pro's fell below it, so the billed ratio between the two DeepSeek judges reflects routing more than tier, and the ratio at list prices is the better guide.

In tokens, Terra emitted 0.66 times Luna's completion tokens and 0.45 times its reasoning tokens, yet cost 7.99 times as much. The other two higher-tier judges emitted about as many completion and reasoning tokens as their siblings, and DeepSeek V4 Pro read 1.12 times as many prompt tokens. Completion tokens make up 76.3\% of Gemini 3.1 Pro's \$31.41 on ELLIPSE. Over the nine panels, Gemini 3.1 Pro's mean request took 4.31 times as long as Gemini's, and its slowest, 584.3 seconds, came within 16 seconds of the timeout. Requests in flight when DeepSeek V4 Pro's runner was stopped, and the two failed attempts behind Gemini 3.1 Pro requests that LiteLLM retried, left no usage record, so their charges, if any, are missing here as well. With the pilot, the higher-tier study thus cost at least \$83.99 as billed, against \$91.11 as recorded and \$91.09 at list prices.

\begin{table*}[!tp]
\centering
\scriptsize
\setlength{\tabcolsep}{3.4pt}
\begin{tabular}{@{}l@{\hspace{6pt}}rrrr@{\hspace{9pt}}rrrr@{\hspace{9pt}}rrrr@{\hspace{9pt}}r@{}}
\toprule
 & \multicolumn{4}{c}{Terra minus Luna} & \multicolumn{4}{c}{Gemini 3.1 Pro minus Gemini} & \multicolumn{4}{c}{DeepSeek V4 Pro minus DeepSeek} & \makecell[r]{Gemini 3.1 Pro\\minus Gemini, H} \\
\cmidrule(lr){2-5}\cmidrule(lr){6-9}\cmidrule(lr){10-13}\cmidrule(l){14-14}
Panel & Acc. & Diff & F\,:\,T & $p$ & Acc. & Diff & F\,:\,T & $p$ & Acc. & Diff & F\,:\,T & $p$ & Diff \\
\midrule
\multicolumn{14}{@{}l}{\emph{(a) Each higher-tier judge against its flash sibling}} \\
RiceChem & 79.4 & $+1.6$ & 37\,:\,50 & 0.20 & 77.4 & $+1.3$ & 18\,:\,29 & 0.14 & 79.7 & $+0.5$ & 51\,:\,55 & 0.77 & $+0.9$ \\
HealthBench & 71.4 & $+1.0$ & 32\,:\,36 & 0.72 & 79.6 & $0.0$ & 14\,:\,14 & 1.00 & 77.3 & $+1.0$ & 24\,:\,28 & 0.68 & $+0.7$ \\
\midrule
ELLIPSE & 19.6 & $+5.7$\ddg & 60\,:\,148 & $9 \times 10^{-10}$ & 27.6 & $-2.3$ & 123\,:\,88 & 0.019 & 22.6 & $+7.5$\ddg & 87\,:\,202 & $1 \times 10^{-11}$ & $-4.7$\ddg \\
FED-Turn & 61.1 & $+3.0$ & 42\,:\,60 & 0.092 & 57.0 & $-3.8$ & 39\,:\,16 & 0.003 & 60.8 & $+1.5$ & 28\,:\,37 & 0.32 & $-3.3$ \\
FED-Dialogue & 43.7 & $-2.5$ & 28\,:\,22 & 0.48 & 51.9 & $-3.9$ & 12\,:\,3 & 0.035 & 53.1 & $+6.4$ & 18\,:\,33 & 0.049 & $-3.5$ \\
HelpSteer2 & 43.3 & $+4.2$ & 33\,:\,48 & 0.12 & 48.9 & $-5.4$ & 36\,:\,17 & 0.013 & 46.4 & $-0.3$ & 40\,:\,39 & 1.00 & $-2.9$ \\
LFQA & 68.2 & $-3.4$ & 24\,:\,12 & 0.065 & 67.5 & $0.0$ & 18\,:\,18 & 1.00 & 68.3 & $+3.1$ & 24\,:\,35 & 0.19 & $-1.4$ \\
USR-TC & 63.4 & $-1.4$ & 34\,:\,29 & 0.61 & 58.1 & $-3.8$ & 28\,:\,15 & 0.066 & 62.0 & $+1.4$ & 33\,:\,38 & 0.64 & $-4.6$ \\
USR-PC & 54.4 & $-3.7$ & 24\,:\,13 & 0.099 & 60.0 & $-0.7$ & 8\,:\,6 & 0.79 & 58.9 & $+1.7$ & 21\,:\,26 & 0.56 & $0.0$ \\
\addlinespace[2pt]
Nine panels & 52.2 & $+2.1$ & \multicolumn{2}{l}{$+1.0$, $+3.2$} & 55.4 & $-1.8$ & \multicolumn{2}{l}{$-2.8$, $-0.9$} & 54.3 & $+3.4$ & \multicolumn{2}{l}{$+2.2$, $+4.5$} & $-2.5$ \\
Seven ordinal & 44.2 & $+2.3$ & & & 48.0 & $-2.7$ & & & 46.3 & $+4.3$ & & & \\
All but ELLIPSE & & $+0.5$ & \multicolumn{2}{l}{$-0.9$, $+1.9$} & & $-1.6$ & \multicolumn{2}{l}{$-2.7$, $-0.6$} & & $+1.5$ & \multicolumn{2}{l}{$+0.2$, $+2.9$} & $-1.4$ \\
\bottomrule
\end{tabular}

\medskip
\footnotesize
\setlength{\tabcolsep}{4pt}
\begin{tabular}{@{}>{\raggedright\arraybackslash}p{4.3cm}>{\raggedright\arraybackslash}p{3.5cm}>{\raggedright\arraybackslash}p{3.5cm}>{\raggedright\arraybackslash}p{3.5cm}@{}}
\toprule
 & Terra & Gemini 3.1 Pro & DeepSeek V4 Pro \\
\midrule
\multicolumn{4}{@{}l}{\emph{(b) Against the same vendor's judge at high effort, and against Jev Choice}} \\
Minus the high run, nine panels & $+2.1$ ($+1.0$, $+3.2$) & $-2.5$ ($-3.5$, $-1.5$) & $+3.7$ ($+2.7$, $+4.8$) \\
Minus the high run, all but ELLIPSE & $+0.6$ ($p = 0.44$) & $-1.4$ ($-2.5$, $-0.4$) & $+1.2$ ($p = 0.0498$) \\
Panels whose difference from the high run survives correction & ELLIPSE $+5.6$ (69\,:\,156)\textsuperscript{\ddag}; FED-Turn $+5.5$ (36\,:\,69)\textsuperscript{\dag} & ELLIPSE $-4.7$ (150\,:\,77)\textsuperscript{\ddag} & ELLIPSE $+9.3$ (56\,:\,199)\textsuperscript{\ddag} \\
Significant pair-level sign tests against Jev Choice (Jev\,:\,LLM) & HealthBench, Jev ahead (65\,:\,42); ELLIPSE (67\,:\,165)\textsuperscript{\ddag} and FED-Turn, Terra ahead & RiceChem, Jev ahead (71\,:\,42); ELLIPSE, Gemini 3.1 Pro ahead (24\,:\,243)\textsuperscript{\ddag} & ELLIPSE (59\,:\,199)\textsuperscript{\ddag} and FED-Turn, DeepSeek V4 Pro ahead \\
\bottomrule
\end{tabular}
\caption{Accuracy of the higher-tier judges. (a) Acc.: the higher-tier judge's accuracy (\%), measured as in Table~\ref{tab:main}; the pooled rows divide its correct verdicts by its scored pairs over the panels pooled. Diff: the higher-tier judge's accuracy less its flash sibling's, in points, over the pairs scored by both; intervals come from resampling units (Table~\ref{tab:clustered}). F\,:\,T: pairs only the flash sibling got right against pairs only the higher-tier judge got right, and $p$ the two-sided sign test on these counts. \ddag: survives Holm correction over the 27 comparisons in the pair-level and unit-level sign tests and for the bootstrap $p$-value alike. The pooled rows give the difference and its 95\% interval and lie outside the correction. Gemini 3.1 Pro minus Gemini, H: against Gemini's high run, which shares Gemini 3.1 Pro's documented thinking level (see text). (b) Differences from the same vendor's high run, in points, with the 95\% interval or the bootstrap $p$-value; the comparisons that survive Holm correction within their 27, with \ddag\ for all three tests and \dag\ for the pair-level test and the bootstrap only, and pairs only the high run got right against pairs only the higher-tier judge got right; and every pair-level sign test against Jev Choice with $p < 0.05$, uncorrected, with \ddag\ where it survives correction within the 27.}
\label{tab:ht-accuracy}
\end{table*}

\paragraph{Accuracy.}
The higher tier is not reliably more accurate. Each higher-tier judge is compared with its flash sibling in Table~\ref{tab:ht-accuracy}a, on the pairs both scored, and the 27 per-panel comparisons form one family for Holm correction. The 27 differences run from $-5.4$ points (Gemini 3.1 Pro on HelpSteer2) to $+7.5$ (DeepSeek V4 Pro on ELLIPSE), with a median of $+0.5$, and 14 of them are positive, 2 zero, and 11 negative. By the classes of Section~\ref{sec:metrics}, 10 are at parity only, 6 separated only, 1 both separated and at parity, and 10 inconclusive. Uncorrected pair-level sign tests favor the higher tier in three comparisons (Terra and DeepSeek V4 Pro on ELLIPSE, and DeepSeek V4 Pro on FED-Dialogue) and the flash sibling in four, all four with Gemini ahead of Gemini 3.1 Pro (ELLIPSE, FED-Turn, FED-Dialogue, and HelpSteer2). After correction only the two ELLIPSE gains remain, Terra's ($p_{\text{Holm}} = 2 \times 10^{-8}$) and DeepSeek V4 Pro's ($3 \times 10^{-10}$), in all three tests. Over all nine panels together, Terra's gain over Luna (2.1 points) and DeepSeek V4 Pro's over DeepSeek (3.4) are both separated, but without ELLIPSE they shrink to 0.5 points ($p = 0.52$) and 1.5 ($p = 0.028$). Gemini 3.1 Pro loses 1.8 points to Gemini, and 1.6 without ELLIPSE ($p = 0.002$). Against Gemini's high run, whose requested level is Gemini 3.1 Pro's documented default, Gemini 3.1 Pro loses 2.5 points pooled. It trails that run on six panels, ties it on USR-PC, and leads only on the binary panels, and its ELLIPSE deficit of 4.7 points survives correction in all three tests. Against the high condition Terra also gains on FED-Turn after correction, but only in the pair-level test and the bootstrap (Table~\ref{tab:ht-accuracy}b). Pooled over all their scored pairs, the higher-tier judges are right on 52.2\% (Terra), 55.4\% (Gemini 3.1 Pro), and 54.3\% (DeepSeek V4 Pro) of them, against 50.1\%, 57.3\%, and 51.0\% for their siblings and 51.1\% for Jev Choice.

Among Jev's framings and the judges of the higher-tier condition, Gemini 3.1 Pro is the most accurate on HealthBench, ELLIPSE, HelpSteer2, and USR-PC; DeepSeek V4 Pro on FED-Dialogue; Jev Choice on RiceChem; and Jev Score on FED-Turn, LFQA, and USR-TC. Across Jev's framings and every LLM judge of the main runs, the high condition, and the higher tier, however, no higher-tier judge is the most accurate on any panel. On HealthBench, Gemini 3.1 Pro only ties Gemini (79.6\%). Elsewhere the most accurate judge is Gemini on FED-Dialogue (56.9\%) and HelpSteer2 (54.6\%), Gemini at high effort on ELLIPSE (32.3\%) and USR-TC (67.6\%), Luna on LFQA (71.6\%), Luna and Gemini jointly on USR-PC (60.6\%), Jev Choice on RiceChem (81.0\%), and Jev Score on FED-Turn (66.7\%). In the Decider's copy of the higher-tier condition, the most accurate judges are Decider Choice (RiceChem and HelpSteer2), Decider Score (ELLIPSE, LFQA, and USR-TC), Terra (FED-Turn), DeepSeek V4 Pro (FED-Dialogue), and Gemini 3.1 Pro (USR-PC), while on HealthBench Gemini 3.1 Pro ties Decider bare Noul. When the Decider's framings replace Jev's in the pool of all three conditions, Terra is the most accurate judge on FED-Turn (61.1\%), and Gemini, Gemini 3.1 Pro, and Decider bare Noul tie on HealthBench (79.6\%). Jev Choice's 27 comparisons with the higher-tier judges fall into 6 separated, 9 at parity, and 12 inconclusive (Table~\ref{tab:ht-downstream}b). Its two separated leads are both on binary panels, and its four separated deficits are the three ELLIPSE comparisons and Terra on FED-Turn, whose lead's interval starts only 0.17 points above zero. Correction, with any of the three tests, keeps only the three ELLIPSE deficits.

\begin{table}[!tp]
\centering
\footnotesize
\setlength{\tabcolsep}{3pt}
\begin{tabular}{@{}lrrrrr@{}}
\toprule
 & \multicolumn{2}{c}{Oracle vs.\ best} & \multicolumn{2}{c}{\makecell{Cross-fitted\\vs.\ best}} & \makecell[r]{Cross-\\fitted} \\
\cmidrule(lr){2-3}\cmidrule(lr){4-5}
Panel & Cond. & Either & Cond. & Either & cost (\%) \\
\midrule
\multicolumn{6}{@{}l}{\emph{(a) Cascades with higher-tier fallbacks}} \\
RiceChem & $0.0$ & $0.0$ & $-0.7$ & $-0.6$ & 5.6 \\
HealthBench & $+1.7$ & $+1.7$ & $+1.5$ & $+1.3$ & 18.9 \\
\midrule
ELLIPSE & $0.0$ & $-2.3$ & $-0.1$ & $-2.3$ & 97.8 \\
FED-Turn & $-4.5$ & $-4.5$ & $-5.8$ & $-5.8$ & 46.8 \\
FED-Dialogue & $-1.7$ & $-3.0$ & $-1.7$ & $-2.8$ & 53.1 \\
HelpSteer2 & $+3.7$ & $-1.4$ & $+3.7$ & $-3.0$ & 12.4 \\
LFQA & $+0.6$ & $-1.4$ & $+0.3$ & $-2.6$ & 17.0 \\
USR-TC & $-1.4$ & $-1.9$ & $-3.0$ & $-3.3$ & 28.5 \\
USR-PC & $+1.0$ & $+0.3$ & $+0.5$ & $+0.1$ & 18.5 \\
\bottomrule
\end{tabular}

\medskip
\setlength{\tabcolsep}{3pt}
\begin{tabular}{@{}lrrrr@{}}
\toprule
Fallback & $\tau$ & \makecell[r]{Kept\\(\%)} & \makecell[r]{Jev $-$ F,\\kept} & \makecell[r]{Cascade\\$-$ F} \\
\midrule
\multicolumn{5}{@{}l}{\emph{(b) Each fallback at its oracle threshold}} \\
\multicolumn{5}{@{}l}{HelpSteer2} \\
\quad Terra & 0.30 & 85.5 & $+9.6$ & $+8.2$ \\
\quad Gemini 3.1 Pro & 0.30 & 85.5 & $+3.0$ & $+2.6$ \\
\quad DeepSeek V4 Pro & 0.30 & 85.5 & $+7.6$ & $+6.5$ \\
\multicolumn{5}{@{}l}{HealthBench} \\
\quad Terra & 0.12 & 83.3 & $+9.8$ & $+8.1$ \\
\quad Gemini 3.1 Pro & 0.22 & 66.5 & $+1.9$ & $+1.2$ \\
\quad DeepSeek V4 Pro & 0.12 & 83.3 & $+4.7$ & $+3.9$ \\
\bottomrule
\end{tabular}
\caption{Jev-first cascades with higher-tier fallbacks (post-hoc analysis), replayed as in Section~\ref{sec:cascades} and Appendix~\ref{app:cascades}. (a) Gain of the best cascade over the best single judge, in points, with an oracle threshold and cross-fitted thresholds. Cond.: over the best single judge of the higher-tier condition, chosen among Jev's framings and the higher-tier judges; Either: over either tier's best single judge (Table~\ref{tab:conditions}). Cross-fitted cost: what a cross-fitted cascade that reaches the condition's best LLM judge costs, relative to that judge's own cost (\%), as in Table~\ref{tab:cascade}. (b) Each higher-tier judge as the fixed fallback: kept share of pairs, Jev's accuracy minus the fallback's on the kept pairs, and the cascade's accuracy minus the fallback's, in points, which Equation~\eqref{eq:identity} relates. Table~\ref{tab:cascade-ledger} gives the flash-tier fallbacks.}
\label{tab:ht-cascades}
\end{table}

\begin{table*}[!tp]
\centering
\footnotesize
\setlength{\tabcolsep}{4pt}
\begin{tabular}{@{}lrrlrrrr@{}}
\toprule
 & \makecell[r]{Confident errors\\repeated} & \multicolumn{6}{c}{Juries} \\
\cmidrule(l){3-8}
Panel & & Pairs & \makecell[l]{Most accurate matched\\judge (\%)} & \makecell[r]{Three-LLM\\median $\Delta$} & 95\% interval & \makecell[r]{Upper\\median $\Delta$} & \makecell[r]{Lower\\median $\Delta$} \\
\midrule
\multicolumn{8}{@{}l}{\emph{(a) By panel, higher-tier condition}} \\
RiceChem & -- & 819 & Jev Choice (81.0) & $-1.0$ & $-3.1$, $+1.2$ & $+0.4$ & $-0.6$ \\
HealthBench & 32/36$^{b}$ & 406 & Gemini 3.1 Pro (79.6) & $-0.2$ & $-2.7$, $+2.0$ & $+1.2$ & $-1.0$ \\
\midrule
ELLIPSE & 34/36 & 1{,}534 & Gemini 3.1 Pro (27.4) & $-4.4$\dg & $-6.0$, $-2.9$ & $-3.2$ & $-11.6$ \\
FED-Turn & 36/36 & 600 & Terra (65.5) & $0.0$ & $-2.0$, $+2.0$ & $+0.8$ & $-3.2$ \\
FED-Dialogue & 34/36 & 231 & DeepSeek V4 Pro (57.1) & $-0.4$ & $-5.2$, $+4.4$ & $+2.2$ & $-5.6$ \\
HelpSteer2 & 34/36 & 351 & Gemini 3.1 Pro (49.3) & $-2.6$ & $-6.9$, $+1.7$ & $+2.3$ & $-5.1$ \\
LFQA & 36/36 & 354 & Jev Choice (68.9) & $+0.3$ & $-3.7$, $+4.0$ & $+0.6$ & $+1.7$ \\
USR-TC & 30/35 & 345 & Jev Choice (69.6) & $+1.2$ & $-3.5$, $+5.6$ & $+0.9$ & $+1.2$ \\
USR-PC & 34/36 & 299 & Gemini 3.1 Pro (73.9) & $-1.0$ & $-3.4$, $+1.7$ & $+1.7$ & $-2.0$ \\
\bottomrule
\end{tabular}

\medskip
\setlength{\tabcolsep}{3pt}
\begin{tabular}{@{}>{\raggedright\arraybackslash}p{8.6cm}>{\raggedright\arraybackslash}p{2.2cm}>{\raggedright\arraybackslash}p{2.2cm}>{\raggedright\arraybackslash}p{2.2cm}@{}}
\toprule
 & M & H & T \\
\midrule
\multicolumn{4}{@{}l}{\emph{(b) Pooled statistics}} \\
Jev Choice against an LLM judge (27 comparisons): separated, parity, inconclusive & 8, 8, 11 & 5, 11, 11 & 6, 9, 12 \\
Jev Choice's separated leads (on binary panels); separated deficits & 4 (3); 4 & 3 (2); 2 & 2 (2); 4 \\
Ordinal panels with judge--judge above judge--label QWK, of 7 & 7 & 7 & 7 \\
Least-agreeing judge pair above the best judge--label QWK, panels of 7 & 5 & 6 & 5 \\
Ordinal panels with every matched judge's offset negative, of 7 & 6 & 6 & 6 \\
Confident errors: LLM verdicts repeating Jev's answer & 242 of 252 (96.0\%) & 245 of 252 (97.2\%) & 238 of 251 (94.8\%) \\
Confident errors: independence baseline (\%) & 50.3 & 51.2 & 52.0 \\
Confident errors with HealthBench's 12 (96 in all): repeats by the OpenAI, Google, and DeepSeek judge & 92, 90, 92 & 92, 92, 95 & 91, 91, 88 \\
Largest cascade gain over the condition's best single judge, cross-fitted\,/\,oracle (points) & 1.5\,/\,2.0 & 2.4\,/\,2.7 & 3.7\,/\,3.7 \\
Largest cascade gain over the best single judge of either tier, cross-fitted\,/\,oracle (points) & 1.7\,/\,2.0 & 1.5\,/\,2.0 & 1.3\,/\,1.7 \\
Panels where the cross-fitted cascade matching the best LLM judge costs under half of it & 7 & 6 & 7 \\
Ordinal panels with the three-LLM median below the most accurate matched judge; of these, with its 95\% interval below zero & 6; 2 & 6; 2 & 4; 1 \\
Ordinal panels with the four-judge upper median below that judge & 4 & 2 & 1 \\
Three-LLM median minus that judge, range over the ordinal panels (points) & $-12.8$ to $0.0$ & $-16.2$ to $+0.3$ & $-4.4$ to $+1.2$ \\
\bottomrule
\end{tabular}
\caption{Correlated errors, cascades, and juries (Sections~\ref{sec:shared} and~\ref{sec:routing}) with higher-tier LLM judges (T), beside the main runs (M) and the high condition (H), with Jev's verdicts unchanged across the three. (a) Confident errors repeated: higher-tier verdicts on the confident-error sample (Section~\ref{sec:shared-errors}) that give Jev Choice's answer, out of those given; on USR-TC one Gemini 3.1 Pro answer is CANNOT\_ASSESS. $^{b}$HealthBench: Jev Noul's errors (Appendix~\ref{app:shared}). Juries as in Table~\ref{tab:juries}: the most accurate of the four matched judges on the pairs all four scored, with its accuracy, and each median jury's accuracy less that judge's (points), with the three-LLM median's 95\% interval over resampled units; \dag: the interval lies below zero. (b) As in Table~\ref{tab:ef-downstream}c. The OpenAI, Google, and DeepSeek judges are the flash-tier judges in M and H and the higher-tier judges in T; the 96 confident errors add Jev Noul's 12 HealthBench errors to the 84. Cascade rows as in Table~\ref{tab:ht-cascades}, each with its condition's fallbacks; the either-tier best single judge comes from the same candidates in all three conditions. Jury rows cover the seven ordinal panels; in none of the three conditions does a median jury's 95\% interval lie wholly above zero on any panel.}
\label{tab:ht-downstream}
\end{table*}

\paragraph{Shared errors, cascades, and juries.}
The higher tier leaves the shortfall of Section~\ref{sec:agreement} in place (Table~\ref{tab:ht-downstream}b). Judge--judge agreement exceeds judge--label agreement on all seven ordinal panels (0.375 against 0.189 on ELLIPSE and 0.596 against 0.518 on USR-TC), and 31 of the 35 offsets of the five judges on the ordinal panels are negative. On USR-TC, Jev Choice's offset is 0.00 levels and Terra's $+0.13$, and Gemini 3.1 Pro is the only judge with a negative offset there. On the confident-error sample the higher-tier judges give Jev Choice's answer in 238 of their 251 verdicts, against 130.4 that the independence baseline expects. Terra repeats 79 of the 84 errors (44.5 expected), Gemini 3.1 Pro 82 of its 83 verdicts (43.6), and DeepSeek V4 Pro 77 of 84 (42.3). Gemini 3.1 Pro's missing verdict is a genuine CANNOT\_ASSESS on a USR-TC \crit{uses\_knowledge} pair that Jev Choice judged UNMET at confidence 0.92 against a label of MET. On HealthBench the higher-tier judges give Jev Noul's wrong answer in 32 of 36 verdicts on its sampled errors (Terra 12, Gemini 3.1 Pro 9, and DeepSeek V4 Pro 11), where 9.2 are expected. Adding Jev Noul's sampled HealthBench errors to the confident-error sample gives 96 errors. On these, no vendor's higher tier repeats more than four fewer than its flash sibling, and none significantly fewer. Terra repeats 91 and Luna 92, Gemini 3.1 Pro 91 and Gemini 90, and DeepSeek V4 Pro 88 and DeepSeek 92 ($p = 0.29$, sign test on the errors only one of them repeats). Compared with DeepSeek's high run, which repeats 95, DeepSeek V4 Pro's 88 gives a nominal $p = 0.039$, uncorrected for the other comparisons. Pooled over the seven ordinal panels, the share of repeated errors rises with Jev Choice's confidence band as in the main runs, from 26.0\% in the lowest band to 83.0\% in the highest (main runs: 29.2\% to 84.5\%).

The best higher-tier cascade gains 3.7 points over the condition's best single judge on HelpSteer2, with an oracle threshold and with cross-fitted thresholds alike, as that judge is weak there (Table~\ref{tab:ht-cascades}a). With DeepSeek V4 Pro as fallback at $\tau = 0.30$, it defers 14.5\% of the pairs and reaches 52.8\%. Gemini 3.1 Pro, the condition's best single judge, reaches 49.1\% on the cascade pair set, and Gemini 54.3\% on the same pairs, which the cascade trails by 1.4 points. Measured against the best single judge of either tier (Table~\ref{tab:conditions}), the cascade thus loses 1.4 points with its oracle threshold and 3.0 with cross-fitted ones. That judge is, on every panel, the one Table~\ref{tab:cascade} names best with Jev first, and on HealthBench Gemini 3.1 Pro ties it exactly. There the best higher-tier cascade, Jev Noul with DeepSeek V4 Pro at $\tau = 0.12$ and 16.7\% of pairs deferred, reaches 81.3\% against 79.6\%. Its gain, 1.7 points with an oracle threshold and 1.3 with cross-fitted ones, is the largest under the either-tier yardstick. Under this yardstick, cross-fitted cascades fall short on seven of the nine panels. Where single judges tie on a training half, these cross-fitted gains depend on the order in which the tied judges are considered, by up to 0.4 points on four panels, though the largest gain under this yardstick, with oracle or cross-fitted thresholds, does not move. Because each half picks its best single judge on its own units, a wider pool of candidates can also raise a cross-fitted gain, as it does for the main runs' cascades (1.7 against 1.5 points; Table~\ref{tab:ht-downstream}b). On ELLIPSE the best cascade defers 98.5\% of the pairs to Gemini 3.1 Pro and equals it, and only DeepSeek V4 Pro's curve peaks at full deferral. On RiceChem the cascade with DeepSeek V4 Pro ties Jev Choice exactly (81.0\%).

Equation~\eqref{eq:identity} holds for every higher-tier cascade, and Table~\ref{tab:ht-cascades}b gives its terms for each fallback on HelpSteer2 and HealthBench. On HelpSteer2 at $\tau = 0.30$, Jev keeps 85.5\% of the pairs and leads DeepSeek V4 Pro on them by 7.6 points, which puts the cascade 6.5 points above that fallback. The gain thus comes from keeping Jev's verdicts on pairs where this fallback does worse. Correcting Jev's errors instead would require a fallback whose errors do not coincide with Jev's. Gemini 3.1 Pro, the most accurate higher-tier judge on HelpSteer2, trails Jev by 3.0 points on the kept pairs, and on HealthBench Jev's kept-pair lead of 1.9 points over Gemini 3.1 Pro yields a cascade only 1.2 points above it. Pooled over the nine panels, cross-fitted cascades are 1.9 points more accurate with Terra than with Luna as fallback, 3.1 more with DeepSeek V4 Pro than with DeepSeek, and 1.9 less with Gemini 3.1 Pro than with Gemini (against the high condition, $+2.0$, $+3.2$, and $-2.5$). Matching the condition's best LLM judge, a cross-fitted cascade costs less than half as much as that judge on seven panels, from 5.6\% of its cost (RiceChem) to 46.8\% (FED-Turn), and an oracle-threshold cascade does so on six.

Against the most accurate matched judge, the three-LLM median falls short on four of the seven ordinal panels, significantly only on ELLIPSE ($-4.4$ points), and ties Terra on FED-Turn (Table~\ref{tab:ht-downstream}a). It is ahead of that judge on LFQA by one pair of 354 (0.3 points) and on USR-TC by four of 345 (1.2 points), both panels on which Jev Choice is that judge, and neither lead is significant. None of the three median juries has an interval entirely above zero on any panel; the closest lower ends are $-2.0$ points for the three-LLM median on FED-Turn and $-1.1$ for each four-judge median, the upper one on RiceChem and the lower one on LFQA. Jev's wrong answer is also the three-LLM median's answer on 82 of the 83 confident errors with a higher-tier verdict from every judge, and the label on the remaining one; the four-judge upper median is right once, and the lower median never.

%% file: appendix/q_recommendations.tex
\section{Practical Recommendations}
\label{app:recommendations}

Table~\ref{tab:implications} turns the results into recommendations, each marked with the status of its evidence. Where the evidence names a decision model, it comes from that model's copy of the analyses (Section~\ref{sec:jev}; Appendix~\ref{app:decider}), and a recommendation that names Jev rests on Jev's copy alone.

\begin{table*}[!tp]
\centering
\footnotesize
\setlength{\tabcolsep}{4pt}
\renewcommand{\arraystretch}{1.0}
\begin{tabular}{@{}L{5.3cm} L{8.53cm} L{1.6cm}@{}}
\toprule
Recommendation & Evidence & Status \\
\midrule
Consider a decision model in place of a flash-tier LLM judge on binary criteria like those of RiceChem and HealthBench & Jev Choice most accurate matched judge on RiceChem and second to Gemini on HealthBench, the LLM judges costing 18 (Luna, HealthBench) to 326 (Gemini, RiceChem) times as much there; Decider Choice first on RiceChem and third on HealthBench, behind Gemini and DeepSeek, the LLM judges costing 28 (Luna, HealthBench) to 312 (Gemini, RiceChem) times as much & measured (two panels) \\
\addlinespace[1.5pt]
On ordinal criteria, expect a decision model's Choice framing to be within 5 points of a flash-tier LLM judge in some comparisons and behind the best LLM judge on some panels & Jev Choice: 6 of 21 comparisons at parity and 10 inconclusive; Gemini ahead on three panels, Luna on one, and Jev Choice ahead of Luna on one. Decider Choice: 3 at parity, 7 inconclusive, and 11 separated; ahead of all three LLM judges on ELLIPSE and HelpSteer2, behind all three on FED-Turn, and behind Luna and Gemini on FED-Dialogue & measured \\
\addlinespace[1.5pt]
Consider Jev Score for ordinal criteria when abstention is not needed, and report Jev Choice for matched comparisons & Jev Score separated ahead of Jev Choice on four panels and never separated behind it; it cannot abstain. Decider Score separated ahead of Decider Choice on two panels and behind it on two & measured \\
\addlinespace[1.5pt]
Give Jev the verdict definitions (the wrapper) on binary criteria & +10.7 (RiceChem) and +1.0 points (HealthBench) over bare Noul; for the Decider +3.4 on RiceChem but $-7.6$ on HealthBench & measured \\
\addlinespace[1.5pt]
Offer an NA option on every ordinal criterion & 28 abstentions by Jev Choice and 36 by Decider Choice in 3,414 pairs; P(NA) tracks human N/A answers (Spearman $\rho = 0.62$ for Jev, 0.65 for the Decider) & observed \\
\addlinespace[1.5pt]
Check confidence per criterion before deferring on it & Jev Choice's confidence is uninformative on ELLIPSE and has an AUROC of 0.41 on LFQA \crit{factuality}; Decider Choice's has an AUROC of 0.52 on ELLIPSE and 0.43 on USR-PC \crit{engaging} & observed \\
\addlinespace[1.5pt]
Use a cascade with a decision model first to lower cost, and expect little gain in accuracy & Jev first: largest cross-fitted gain +1.5 points (HealthBench; +2.4 at high effort; with higher-tier fallbacks +3.7 on HelpSteer2 over that condition's weak best single judge, but +1.3 on HealthBench over either tier's best single judge); on six panels, 16--48\% of the best LLM judge's cost for at most 1.8 points less accuracy, and on RiceChem 6.6\% of its cost for 1.2 points more. Decider first: largest cross-fitted gain +1.2 (LFQA; +0.8 at high effort; with higher-tier fallbacks +1.4 on FED-Dialogue, and +0.4 on HealthBench over either tier's best single judge); on LFQA, USR-TC, and USR-PC, 18--28\% of the best LLM judge's cost for at most 1.0 point less, and where the Decider alone beats that judge (ELLIPSE, HelpSteer2, and RiceChem), 0.3--5.3\% of its cost for 1.3--17.2 points more & measured (post hoc, cross-fitted) \\
\addlinespace[1.5pt]
Do not expect a jury of these judges to beat the most accurate matched judge & In both copies the median of the LLM judges is below or tied with that judge on all nine panels, and it gives Jev's wrong answer on 82 of 84 confident errors and the Decider's on 73 of 84 & measured (post hoc) \\
\addlinespace[1.5pt]
Report the offset beside agreement, and judge--judge beside judge--label agreement & Tables~\ref{tab:main} and~\ref{tab:offsets}; Figure~\ref{fig:offsetgap} & observed \\
\addlinespace[1.5pt]
Validate a replacement judge on identical pairs, with enough pairs to be decisive and the judging mode held fixed & Whole-rubric judgments changed the separation of 5 of Jev Choice's 27 comparisons (Appendix~\ref{app:wholerubric}) and of 3 of Decider Choice's & measured \\
\addlinespace[1.5pt]
On norm-referenced scales, supply exemplars & Jev Choice's per-trait ELLIPSE offsets $-0.62$ to $-1.77$ levels. The Decider's copy does not support this row, since both Decider framings do best on ELLIPSE with no shift and Decider Choice's per-trait offsets ($-0.26$ to $-0.62$ against our half-up labels) depend on the rounding rule (Section~\ref{sec:conventions}; Appendix~\ref{app:offset}) & not tested \\
\addlinespace[1.5pt]
Where labeled pairs exist, correct each criterion's offset & Worst remaining gap after the leave-one-out shift of Section~\ref{sec:location}, $-8.4$ points in Jev's copy and $-9.65$ in the Decider's; when judges and raters both get their best shift, $-16.2$ in Jev's and $-16.45$ in the Decider's (Appendix~\ref{app:offset}) & measured (post hoc) \\
\addlinespace[1.5pt]
Avoid absolute top levels unless meant literally & LFQA \crit{factuality} (Table~\ref{tab:conventions}), whose top level Jev Choice chose less often than the raters; reversed for Decider Choice, which chose it more often (see below) & not tested \\
\addlinespace[1.5pt]
Avoid three-level scales for yes-or-no properties & Jev Choice on FED-Turn's middle level (Appendix~\ref{app:choicescore}) & not tested \\
\bottomrule
\end{tabular}
\caption{Recommendations and the status of their evidence. \emph{Measured}: a paired comparison or an exact computation from the recorded verdicts supports the recommendation on these panels, validated on held-out units only where marked cross-fitted. \emph{Observed}: the data show the pattern without a test of the recommendation. \emph{Not tested}: no experiment addressed it. The evidence is from the per-criterion main runs except where a row names another condition; Appendix~\ref{app:wholerubric} gives what whole-rubric judgments change, Appendix~\ref{app:effort} what high reasoning effort changes, and Appendix~\ref{app:highertier} what higher-tier judges change.}
\label{tab:implications}
\end{table*}

\paragraph{Rubric wording and offset correction.}
The rubric-writing rows of Table~\ref{tab:implications} rest on observations that we did not test by intervention. The judges read top levels worded as absolutes (``Entirely accurate'', ``Fully clear and internally consistent throughout'') differently from the raters, in both directions. Jev Choice seldom used LFQA's absolute top \crit{factuality} level, which the raters often chose (Appendix~\ref{app:conventions}), whereas on five of its confident errors all five judges chose HelpSteer2's absolute \crit{coherence} top level where the label was one level lower (Appendix~\ref{app:shared}). Decider Choice went the other way on LFQA, choosing the top \crit{factuality} level for 70 of the 117 answers it rated, where the labels hold it for 50 of 120. Jev Choice never predicted the middle level on FED-Turn \crit{correct} and \crit{fluent}, whose texts state yes-or-no properties (Appendix~\ref{app:choicescore}). One global correction would not fit every trait, since Jev Choice's per-trait ELLIPSE offsets in Table~\ref{tab:implications} span more than a level.

\paragraph{Evaluating rubric judges.}
Beyond the rows of Table~\ref{tab:implications}, evaluations of rubric judges should separate unanimous from split pairs before interpreting errors, report each judge's framing and decoding rule (and, for an LLM judge, its judging mode), and, when attributing a property to one judge, compare several judges on identical pairs. For combined judges, such as cascades and juries, evaluations should also report the share of repeated errors against an independence baseline (Section~\ref{sec:shared-errors}) alongside any gain in accuracy. Before trusting a cascade, one can make the same comparison on the labeled pairs that set its threshold~\citep{jung2025trust}, to see how far its judges err alike. Correlated errors may also explain why our juries of diverse judges~\citep{verga2024replacing} did not help (Section~\ref{sec:juries}).

%% file: appendix/r_decider.tex
\section{A Second Decision Model: Perplexity's Decider}
\label{app:decider}

This appendix records how Perplexity's Decider was run and reports its copy of the study. The Decider judged the same 5,003 pairs as Jev with the framings and wrapper of Table~\ref{tab:judge-inputs} unchanged, and its answers passed through the same analysis code. Its copy reuses the recorded LLM verdicts, and a check of the 39,709 values that depend on the LLM judges alone found each one identical in the two copies. Every value below belongs to the Decider's copy, with the Decider's matched judges, five judges, and confidence (Section~\ref{sec:jev}), and none is set against Jev's. Jev remains the reference. We designed the study and chose its analyses on Jev, added the Decider when those analyses were done, and checked on it a list of claims fixed before its first answer on a study item (Table~\ref{tab:r-claims}).

\paragraph{Model and documentation.}
Each request asked for the model \texttt{pplx-decider-v1.1-27b}, and each of the 6,283 replies of the full run and the 266 of the repeat named it. Perplexity launched the API in October 2026 and released version 1.1 on 5 October. Its API documentation\footnote{\raggedright\url{https://docs.perplexity.ai/api-reference/decisions-post}, \url{https://docs.perplexity.ai/docs/getting-started/pricing}, \url{https://docs.perplexity.ai/docs/resources/changelog}, and the model card \url{https://huggingface.co/perplexity-ai/pplx-decider-v1.1-27b}, all accessed 6 October 2026; the quickstart is cited in Section~\ref{sec:jev}.} lists that identifier and one other, \texttt{pplx-decider-v1-27b}, and no alias. A request holds the model name, the record of Section~\ref{sec:jev}, and named questions of the three primitives. The documentation describes no web search or retrieval. Before the run, the code that builds Jev's requests built every body offline with the Decider's model name and checked it against the documented limits, and every body sent matched one built in that check. The manifest of every run names the Decider, and no run reads an answer of Jev's. Perplexity calls the model a decision model and mentions Jev only as a column of the benchmark tables on its model cards. The word ``reproduction'' comes from the community Decision Index of Section~\ref{sec:jev}, which tracks open reproductions of Jev and describes itself as unofficial and unaffiliated with TypeSafe. We quote only the index's edition 0.3, whose data and our reading of them both date from 6 October 2026. It ranks the Decider's version 1.1 first of 112 entries with a score of 62.75 and gives Jev 60.11, a score it displays as tied for second. We do not quote the score of an earlier edition that Perplexity's model card reports from its own run.

\paragraph{Lineage.}
The model card gives Qwen3.8-27B as the base model, fine-tuned in full, with a decision readout in place of the layer that generates text. It credits version 1.1's gains mainly to removing the causal mask from the full-attention layers and partly to added training data. Its data manifest lists the sources behind the counts given in the Limitations. Among them are HelpSteer2's labels for all four attributes that our HelpSteer2 panel asks about, and rows from HelpSteer and HelpSteer3, but the manifest does not say whether the responses of our panel are among the HelpSteer2 rows. No source it lists names ELLIPSE or its essays, FED, USR, LFQA, RiceChem, or HealthBench, and the rows it calls original are not itemized. Beyond reading these source names, we did not check the training rows against the panels' items.

\paragraph{Runs.}
The full run lasted from 04:21:22 to 04:36:45 UTC on 7 October 2026. Because the rate limit applies to the whole organization, its four steps ran one after another. The ordinal panels went first under Decider Choice and then under Decider Score. RiceChem followed under its three binary framings, for the panel and for the full set of 1,240 responses (Appendix~\ref{app:binary}), and HealthBench came last with four request sets, its three framings and the ablation with the physician-written ideal, whose answers serve here only to check the repeat. The repeat began four seconds after the full run ended and finished at 04:38:04. Each step kept 8 requests in flight and spaced them at no more than 9 per second, under the documented 10, and no one-second window held more than 10 sends. An attempt timed out after 30 seconds, and a request could make up to 7 attempts within 180 seconds, waiting as long as a rate-limit response asked.

Of the full run's 6,724 attempts, 438 met a rate-limit response and 3 timed out. Every request was answered, so no unit, criterion, or framing has a judge-call failure. Most rate-limit responses, 393 of them and all but one on RiceChem, gave no waiting time and reported that a maximum number of requests had been reached, which fits the token limit that the documentation mentions without a value. The other 45 asked for a one-second wait. The repeat met 13 rate-limit responses and no timeout. Two probe requests on a synthetic one-sentence submission preceded everything. A smoke run of 30 requests, one unit of each ordinal panel, of each RiceChem question, and of HealthBench, under each framing, then checked the answer types, the probabilities, and the response headers. With the probes and a burst of requests at the client's rate, it returned 92 replies, which enter none of the reported results or costs except the check of duplicate requests below.

\begin{table*}[!tp]
\centering
\footnotesize
\setlength{\tabcolsep}{4.5pt}
\begin{tabular}{@{}lrrrrrrrr@{}}
\toprule
 & & \multicolumn{4}{c}{Decider} & & & \\
\cmidrule(lr){3-6}
Panel & Pairs & Choice & Score & Noul & bare Noul & Luna & Gemini & DeepSeek \\
\midrule
\multicolumn{9}{@{}l}{\emph{(a) Cost in US dollars; LLM judges as a multiple of Decider Choice}} \\
RiceChem & 819 & 0.0074 & -- & 0.0070 & 0.0065 & 38$\times$ & 312$\times$ & 70$\times$ \\
HealthBench & 406 & 0.0081 & -- & 0.0079 & 0.0076 & 28$\times$ & 193$\times$ & 53$\times$ \\
\midrule
ELLIPSE & 1{,}548 & 0.0241 & 0.0238 & -- & -- & 29$\times$ & 393$\times$ & 78$\times$ \\
FED-Turn & 600 & 0.0031 & 0.0030 & -- & -- & 40$\times$ & 522$\times$ & 75$\times$ \\
FED-Dialogue & 250 & 0.0016 & 0.0015 & -- & -- & 42$\times$ & 530$\times$ & 116$\times$ \\
HelpSteer2 & 360 & 0.0053 & 0.0052 & -- & -- & 31$\times$ & 276$\times$ & 62$\times$ \\
LFQA & 360 & 0.0028 & 0.0027 & -- & -- & 40$\times$ & 458$\times$ & 98$\times$ \\
USR-TC & 360 & 0.0032 & 0.0031 & -- & -- & 33$\times$ & 381$\times$ & 63$\times$ \\
USR-PC & 300 & 0.0022 & 0.0021 & -- & -- & 35$\times$ & 399$\times$ & 68$\times$ \\
\midrule
Nine panels & 5{,}003 & 0.0577 & 0.0416$^{a}$ & -- & -- & 32$\times$ & 357$\times$ & 73$\times$ \\
\midrule
\multicolumn{9}{@{}l}{\emph{(b) Wall time in seconds; LLM judges as a multiple of Decider Choice}} \\
RiceChem & & 20.1 & -- & 17.4 & 14.5 & 11.8$\times$ & 10.5$\times$ & 99.1$\times$ \\
HealthBench & & 48.8 & -- & 34.5 & 44.4 & 3.2$\times$ & 2.7$\times$ & 24.6$\times$ \\
\midrule
ELLIPSE & & 29.0 & 29.2 & -- & -- & 7.0$\times$ & 10.9$\times$ & 56.6$\times$ \\
FED-Turn & & 8.6 & 8.5 & -- & -- & 6.5$\times$ & 7.7$\times$ & 25.1$\times$ \\
FED-Dialogue & & 3.1 & 3.2 & -- & -- & 9.1$\times$ & 9.9$\times$ & 66.1$\times$ \\
HelpSteer2 & & 10.3 & 10.2 & -- & -- & 5.5$\times$ & 5.4$\times$ & 39.7$\times$ \\
LFQA & & 13.5 & 13.8 & -- & -- & 3.4$\times$ & 3.7$\times$ & 30.0$\times$ \\
USR-TC & & 8.1 & 8.2 & -- & -- & 4.9$\times$ & 6.8$\times$ & 14.3$\times$ \\
USR-PC & & 7.0 & 6.8 & -- & -- & 4.6$\times$ & 4.7$\times$ & 16.9$\times$ \\
\midrule
Nine panels & & 148.6 & 80.0$^{a}$ & -- & -- & 5.8$\times$ & 6.4$\times$ & 42.4$\times$ \\
\bottomrule
\end{tabular}

\medskip
\setlength{\tabcolsep}{3pt}
\begin{tabular}{@{}lrrrrrrrrrr@{}}
\toprule
 & Requests & Attempts & \makecell[r]{Rate\\limited} & Timeouts & Failed & \makecell[r]{Input\\tokens} & \makecell[r]{Cost\\(US\$)} & \makecell[r]{Limiter\\wait (\%)} & \makecell[r]{Latency\\(s)} & \makecell[r]{Session\\(s)} \\
\midrule
\multicolumn{11}{@{}l}{\emph{(c) Requests by step}} \\
Ordinal panels, Choice & 700 & 706 & 6 & 0 & 0 & 2{,}106{,}748 & 0.0421 & 69.2 & 0.24 & 82 \\
Ordinal panels, Score & 700 & 706 & 6 & 0 & 0 & 2{,}079{,}436 & 0.0416 & 66.7 & 0.26 & 83 \\
RiceChem$^{b}$ & 4{,}083 & 4{,}503 & 418 & 2 & 0 & 11{,}719{,}695 & 0.2344 & 39.0 & 0.43 & 585 \\
HealthBench$^{c}$ & 800 & 809 & 8 & 1 & 0 & 1{,}710{,}932 & 0.0342 & 41.6 & 0.32 & 172 \\
\addlinespace[2pt]
Full run & 6{,}283 & 6{,}724 & 438 & 3 & 0 & 17{,}616{,}811 & 0.3523 & 45.2 & 0.39 & 922 \\
Repeat & 266 & 279 & 13 & 0 & 0 & 711{,}083 & 0.0142 & 36.2 & 0.41 & 74 \\
\bottomrule
\end{tabular}
\caption{The Decider's runs. (a) and (b) Each Decider framing per panel, laid out as Table~\ref{tab:cost}, which also gives the LLM judges' costs and times; each multiple divides an LLM judge's value by Decider Choice's on the same panel, with DeepSeek at its billed cost. Each Decider step ran alone, with 8 requests in flight; Table~\ref{tab:cost} gives the LLM judges' concurrency. Noul is the framed Noul of Table~\ref{tab:judge-inputs}. $^{a}$Seven ordinal panels. (c) The request log of each step. Requests: one per unit and framing; attempts: every request sent, retries included; failed: requests left without an answer. Cost: input tokens at \$0.02 per million. Limiter wait: time that attempts spent waiting for the client's rate limiter, as a share of that wait plus their latency. Rate limited: attempts answered with a rate-limit response. Latency: median over answered requests. Session: start to end of the step. $^{b}$Three framings on the 121 units of the panel and the 1,240 of the full set. $^{c}$Three framings and the request set with the physician-written ideal (Appendix~\ref{app:binary}). The repeat covers the 106 repeat units.}
\label{tab:r-runs}
\end{table*}

\paragraph{Price, cost, and time.}
Decider Choice cost \$0.058 over the nine panels and took 149 seconds; Luna, Gemini, and DeepSeek cost 32, 357, and 73 times as much and took 5.8, 6.4, and 42 times as long (Table~\ref{tab:r-runs}). In every condition the LLM judges' nine-panel totals came to at least 17.6 times Decider Choice's cost and, where measured, at least 5.4 times its wall time, whole-rubric Luna setting both minima (Table~\ref{tab:r-conditions}). The cost rule of Section~\ref{sec:metrics} prices the Decider as its quickstart, API reference, and pricing page did on 6 October 2026. The changelog's launch entry alone gives another price, \$0.04 per million input tokens, and no page calls either price introductory. At the launch price each multiple of Decider Choice's cost would halve (Table~\ref{tab:r-conditions}). Luna's per-criterion multiple would fall from 32.0 to 16.0, and whole-rubric Luna's from 17.6 to 8.8. The full run and the repeat used 18.3 million input tokens and cost \$0.367 at the documented price, or \$0.733 at the launch price. That total leaves out the 92 replies that preceded the run and three attempts that timed out, which carried 6,176 input tokens. Most of the Decider Choice run's wall time was set by the rate limit. At 9 requests per second, the 1,044 attempts of the nine-panel Choice run need 116 seconds, 78\% of its 148.6, and they spent 58\% of their time in the client's limiter. On the ordinal panels alone, its 706 attempts at that rate need 98.5\% of the 79.7 seconds they took. One HealthBench request of that run timed out after 30 seconds and was answered on its retry, which added about 22 seconds to that run's time in Table~\ref{tab:r-runs}b, and we keep the time as measured. Across the full run an answered request took a median of 0.39 seconds, with a 90th percentile of 0.56.

\paragraph{Confidence and abstentions.}
The documentation says only that the confidence returned by Choice and Score is not the top probability. For Decider Choice it equals $(K p_{\max} - 1)/(K - 1)$ of Appendix~\ref{app:confidence} on all 3,414 ordinal answers, up to floating-point rounding. Decider Score's confidence departs from that form (median deviation 0.14) and is used as returned, and Decider Noul's comes from $P(\text{yes})$ as in Section~\ref{sec:jev}. Appendix~\ref{app:abstention} breaks down Decider Choice's 36 abstentions, 18 of which fall on FED-Dialogue \crit{error\_recovery}. On that criterion its probability on the NA option rises with the number of raters who answered N/A (Spearman 0.65 over 25 dialogues, $p = 5 \times 10^{-4}$), and it abstained on all 7 dialogues that drew at least two N/A answers. It never chose CANNOT\_ASSESS on a binary panel, the full RiceChem set included.

\begin{table*}[!tp]
\centering
\footnotesize
\setlength{\tabcolsep}{3.5pt}
\begin{tabular}{@{}l@{\hspace{5pt}}lrr@{\hspace{11pt}}lrrrlrr@{}}
\toprule
 & & \multicolumn{7}{c}{Units resampled} & \multicolumn{2}{c}{Pairs} \\
\cmidrule(lr){3-9}\cmidrule(l){10-11}
Panel & Judge & Units & Diff & 95\% interval & Bound & D\,:\,O & $p_{\text{unit}}$ & Class & D\,:\,O & $p$ \\
\midrule
\multicolumn{11}{@{}l}{\emph{Each LLM judge against Decider Choice}} \\
\addlinespace[2pt]
RiceChem & Luna & 121 & $-2.6$ & $-5.0$, $-0.1$ & 4.6 & 36\,:\,20 & 0.044 & \makecell[l]{separated,\\parity} & 63\,:\,42 & 0.050 \\
 & Gemini & 121 & $-4.3$\mk{**} & $-6.6$, $-2.0$ & 6.2 & 45\,:\,21 & 0.004 & separated & 74\,:\,39 & 0.001 \\
 & DeepSeek & 121 & $-1.1$ & $-3.7$, $+1.5$ & 3.3 & 32\,:\,26 & 0.51 & parity & 60\,:\,51 & 0.45 \\
\addlinespace[2pt]
HealthBench & Luna & 200 & $-4.9$ & $-11.1$, $+1.0$ & 10.0 & 53\,:\,42 & 0.30 & -- & 86\,:\,66 & 0.12 \\
 & Gemini & 200 & $+4.2$ & $0.0$, $+8.3$ & 7.6 & 20\,:\,38 & 0.025 & -- & 33\,:\,50 & 0.078 \\
 & DeepSeek & 200 & $+1.0$ & $-3.2$, $+5.3$ & 4.6 & 27\,:\,36 & 0.31 & parity & 40\,:\,44 & 0.74 \\
\addlinespace[2pt]
ELLIPSE & Luna & 258 & $-33.0$\mk{**} & $-36.0$, $-30.0$ & 35.5 & 215\,:\,6 & $9 \times 10^{-56}$ & separated & 552\,:\,41 & $2 \times 10^{-115}$ \\
 & Gemini & 258 & $-17.2$\mk{**} & $-20.0$, $-14.6$ & 19.5 & 167\,:\,22 & $9 \times 10^{-29}$ & separated & 344\,:\,77 & $2 \times 10^{-41}$ \\
 & DeepSeek & 258 & $-32.1$\mk{**} & $-35.1$, $-29.1$ & 34.7 & 206\,:\,8 & $8 \times 10^{-51}$ & separated & 550\,:\,54 & $2 \times 10^{-104}$ \\
\addlinespace[2pt]
FED-Turn & Luna & 75 & $+6.5$\mk{**} & $+2.3$, $+10.5$ & 9.8 & 14\,:\,37 & 0.002 & separated & 43\,:\,82 & $6 \times 10^{-4}$ \\
 & Gemini & 75 & $+9.2$\mk{**} & $+5.0$, $+13.7$ & 12.8 & 12\,:\,34 & 0.002 & separated & 22\,:\,77 & $2 \times 10^{-8}$ \\
 & DeepSeek & 75 & $+7.7$\mk{**} & $+4.3$, $+11.0$ & 10.4 & 11\,:\,36 & $3 \times 10^{-4}$ & separated & 22\,:\,68 & $1 \times 10^{-6}$ \\
\addlinespace[2pt]
FED-Dialogue & Luna & 25 & $+7.1$ & $+0.5$, $+13.8$ & 12.6 & 6\,:\,12 & 0.24 & separated & 26\,:\,42 & 0.068 \\
 & Gemini & 25 & $+14.3$\mk{**} & $+8.3$, $+20.4$ & 19.4 & 2\,:\,16 & 0.001 & separated & 12\,:\,44 & $2 \times 10^{-5}$ \\
 & DeepSeek & 25 & $+4.0$ & $-1.8$, $+10.1$ & 9.1 & 8\,:\,11 & 0.65 & -- & 21\,:\,30 & 0.26 \\
\addlinespace[2pt]
HelpSteer2 & Luna & 90 & $-24.0$\mk{**} & $-30.6$, $-17.4$ & 29.5 & 51\,:\,8 & $9 \times 10^{-9}$ & separated & 110\,:\,26 & $2 \times 10^{-13}$ \\
 & Gemini & 90 & $-8.0$\mk{**} & $-13.2$, $-3.1$ & 12.3 & 33\,:\,16 & 0.021 & separated & 56\,:\,28 & 0.003 \\
 & DeepSeek & 90 & $-15.9$\mk{**} & $-21.9$, $-10.0$ & 20.9 & 44\,:\,11 & $9 \times 10^{-6}$ & separated & 86\,:\,31 & $4 \times 10^{-7}$ \\
\addlinespace[2pt]
LFQA & Luna & 120 & $+2.0$ & $-4.2$, $+8.1$ & 7.0 & 42\,:\,43 & 1.00 & -- & 51\,:\,58 & 0.57 \\
 & Gemini & 120 & $-2.0$ & $-7.0$, $+3.1$ & 6.1 & 33\,:\,24 & 0.29 & -- & 41\,:\,34 & 0.49 \\
 & DeepSeek & 120 & $-4.5$ & $-9.8$, $+0.6$ & 8.9 & 42\,:\,26 & 0.068 & -- & 52\,:\,36 & 0.11 \\
\addlinespace[2pt]
USR-TC & Luna & 72 & $-1.9$ & $-6.9$, $+3.1$ & 6.1 & 23\,:\,18 & 0.53 & -- & 43\,:\,36 & 0.50 \\
 & Gemini & 72 & $+1.1$ & $-3.1$, $+5.6$ & 4.8 & 19\,:\,21 & 0.87 & parity & 33\,:\,37 & 0.72 \\
 & DeepSeek & 72 & $-3.3$ & $-8.4$, $+1.4$ & 7.5 & 16\,:\,14 & 0.86 & -- & 43\,:\,31 & 0.20 \\
\addlinespace[2pt]
USR-PC & Luna & 60 & $+1.0$ & $-3.0$, $+4.7$ & 4.3 & 13\,:\,14 & 1.00 & parity & 20\,:\,23 & 0.76 \\
 & Gemini & 60 & $+1.3$ & $-2.7$, $+5.3$ & 4.7 & 11\,:\,16 & 0.44 & parity & 15\,:\,19 & 0.61 \\
 & DeepSeek & 60 & $-3.3$ & $-7.3$, $+0.7$ & 6.7 & 18\,:\,10 & 0.18 & -- & 25\,:\,15 & 0.15 \\
\midrule
\multicolumn{11}{@{}l}{\emph{Decider Score against Decider Choice (ordinal pairs)}} \\
\addlinespace[2pt]
ELLIPSE & Score & 258 & $+3.9$\mk{**} & $+2.3$, $+5.5$ & 5.2 & 31\,:\,77 & $1 \times 10^{-5}$ & separated & 45\,:\,105 & $1 \times 10^{-6}$ \\
FED-Turn & Score & 75 & $-2.1$ & $-6.7$, $+2.5$ & 5.9 & 31\,:\,28 & 0.79 & -- & 61\,:\,50 & 0.34 \\
FED-Dialogue & Score & 25 & $-0.5$ & $-5.9$, $+5.0$ & 5.0 & 9\,:\,7 & 0.80 & -- & 14\,:\,13 & 1.00 \\
HelpSteer2 & Score & 90 & $-7.1$\mk{**} & $-11.7$, $-2.6$ & 11.0 & 29\,:\,12 & 0.012 & separated & 43\,:\,18 & 0.002 \\
LFQA & Score & 120 & $+2.5$ & $-0.6$, $+5.3$ & 5.0 & 10\,:\,19 & 0.14 & -- & 10\,:\,19 & 0.14 \\
USR-TC & Score & 72 & $+8.8$\mk{**} & $+4.2$, $+13.4$ & 13.0 & 3\,:\,19 & $9 \times 10^{-4}$ & separated & 7\,:\,26 & 0.001 \\
USR-PC & Score & 60 & $-8.9$\mk{**} & $-15.0$, $-3.3$ & 13.9 & 18\,:\,5 & 0.011 & separated & 22\,:\,6 & 0.004 \\
\bottomrule
\end{tabular}
\caption{The Decider's paired comparisons, with the columns of Table~\ref{tab:clustered} on the left and, on the right, the pair-level sign test of Table~\ref{tab:paired} on the pairs both judges scored. Diff: the second judge's accuracy minus Decider Choice's, in points. D\,:\,O: units where Decider Choice leads against units where the other judge does (left), and pairs that Decider Choice alone got right against those the other judge alone got right (right). **: separated, and kept by Holm correction of the bootstrap $p$-values over the Decider's 27 comparisons, or over the seven of the lower block. Holm correction of the 27 pair-level tests keeps the same comparisons except HelpSteer2 Gemini. HealthBench Gemini's interval ends at exactly zero.}
\label{tab:r-paired}
\end{table*}

\paragraph{Paired comparisons.}
Decider Choice differs significantly from the LLM judges in 13 of the 27 comparisons (one, RiceChem Luna, also at parity) and reaches parity in 5 more (Figure~\ref{fig:parity}b; Table~\ref{tab:r-paired}). How firm are these 13 separations? Holm correction within the 27 comparisons keeps 11 (seven Decider leads and four deficits) and drops RiceChem Luna and FED-Dialogue Luna. Resampling HelpSteer2's prompts, LFQA's questions, and USR's contexts in place of units keeps all 13. The sign test over units also finds 13 significant comparisons, but it trades FED-Dialogue Luna for HealthBench Gemini (20 units against 38), a deficit on a binary panel. Gemini's 4.2-point lead on HealthBench falls just short of separation. With the seed of our analysis code, its 95\% interval ends exactly at zero. Rerun with 200 fresh seeds, the bootstrap separated it in 31.5\% of them, and each such run would count 14 separated comparisons out of 27, one of them a deficit on a binary panel. The pair-level sign tests find 7 significant Decider leads and 4 significant deficits. Over all 27 comparisons the equivalence bound has a median of 8.9 points and is within 7.5 points for 11.

The separations follow the panels more than the LLM judges. Nine of the 13 fall on ELLIPSE, HelpSteer2, and FED-Turn, one against each LLM judge, and the other four on RiceChem and FED-Dialogue, against Luna and Gemini, so DeepSeek is separated from Decider Choice three times. Decider Choice leads on RiceChem, ELLIPSE, and HelpSteer2 (by 8 to 33 points on the last two) and trails on FED-Turn and FED-Dialogue. Among the matched judges, it is the most accurate on the three panels where it leads, second on LFQA, third on HealthBench, USR-TC, and USR-PC (tied there with DeepSeek), and fourth on FED-Turn and FED-Dialogue. Gemini leads the matched judges, or ties for the lead, on FED-Turn, FED-Dialogue, USR-TC, and USR-PC, and Luna leads on LFQA.

Decider Score is separated ahead of Decider Choice on two ordinal panels and behind it on two, and is the most accurate of the Decider's five judges on ELLIPSE, LFQA, and USR-TC. In 7 of its 21 pair-level sign tests against an LLM judge on ordinal pairs, it is significantly ahead, and in 7 it is significantly behind. Its ELLIPSE lead over Decider Choice falls short of parity, with a bound of 5.2 points. Neither Decider framing differs significantly from a no-read baseline, as Appendix~\ref{app:conventions} shows for HelpSteer2's constant predictor. Against LFQA's provenance baseline, Decider Choice is right alone on 44 pairs and the baseline on 41 ($p = 0.83$), and Decider Score on 48 against 35 ($p = 0.19$).

\paragraph{Offsets and agreement.}
Decider Score's offsets are negative on five of the seven ordinal panels (Table~\ref{tab:secondary}h), so 28 of the 35 offsets of the Decider's five judges are negative. The three positive offsets of Decider Choice have 95\% intervals over resampled units of $+0.003$ to $+0.167$ levels on HelpSteer2, $+0.062$ to $+0.213$ on LFQA, and $0.000$ to $+0.181$ on USR-TC. The first two owe much to single criteria, HelpSteer2 \crit{correctness} ($+0.41$) and LFQA \crit{factuality} ($+0.21$), on which all three LLM judges have negative offsets. For both Decider framings the offset largest in size is FED-Dialogue's, not ELLIPSE's. Over the 35 ordinal criteria, all four matched judges' offsets are negative on 22 and positive on one. Decider Choice's own QWK with the labels exceeds its mean QWK with the three LLM judges on ELLIPSE (0.42 against 0.35) and HelpSteer2 (0.47 against 0.46), although across all four matched judges agreement with one another still exceeds agreement with the labels on both panels (Table~\ref{tab:main}b). In the main runs even the matched judges' weakest pairwise agreement exceeds their best agreement with the labels on two ordinal panels (Table~\ref{tab:r-conditions}).

\begin{table*}[!tp]
\centering
\footnotesize
\setlength{\tabcolsep}{4pt}
\begin{tabular*}{\textwidth}{@{\extracolsep{\fill}}lrrrrr@{}}
\toprule
 & \multicolumn{4}{c}{Confidence band of Decider Choice} & \\
\cmidrule(lr){2-5}
Panel & [0, 0.25) & [0.25, 0.5) & [0.5, 0.75) & [0.75, 1] & AUROC \\
\midrule
\multicolumn{6}{@{}l}{\textit{(a) Decider Choice error rate, \% (pairs)}} \\
ELLIPSE & 50.0 (2) & 53.2 (1{,}211) & 51.9 (335) & -- & 0.520 \\
FED-Turn & 62.7 (75) & 56.7 (178) & 61.0 (141) & 14.6 (206) & 0.718 \\
FED-Dialogue & 63.0 (46) & 67.2 (64) & 48.5 (66) & 37.3 (51) & 0.602 \\
HelpSteer2 & 77.8 (18) & 48.3 (145) & 24.1 (166) & 23.8 (21) & 0.704 \\
LFQA & 65.4 (26) & 61.2 (67) & 31.5 (92) & 12.8 (172) & 0.784 \\
USR-TC & 33.3 (30) & 43.6 (117) & 42.4 (92) & 7.4 (121) & 0.689 \\
USR-PC & 40.0 (15) & 50.0 (76) & 31.6 (76) & 9.8 (133) & 0.751 \\
\midrule
\multicolumn{6}{@{}l}{\textit{(b) Repeated errors, observed\,/\,independence baseline, \% (Decider Choice errors); last column, all errors}} \\
ELLIPSE & 0.0\,/\,1.0 (1) & 48.2\,/\,45.7 (644) & 64.7\,/\,63.8 (174) & -- & 51.6\,/\,49.4 (819) \\
FED-Turn & 28.4\,/\,20.1 (47) & 38.9\,/\,33.2 (101) & 64.2\,/\,43.1 (86) & 93.3\,/\,44.6 (30) & 51.5\,/\,35.4 (264) \\
FED-Dialogue & 21.8\,/\,19.9 (29) & 47.6\,/\,34.1 (43) & 77.9\,/\,66.6 (32) & 89.3\,/\,48.8 (19) & 55.8\,/\,41.4 (123) \\
HelpSteer2 & 23.8\,/\,17.0 (14) & 35.1\,/\,27.3 (70) & 67.5\,/\,36.6 (40) & 86.7\,/\,40.4 (5) & 46.0\,/\,29.6 (129) \\
LFQA & 30.0\,/\,25.0 (17) & 22.8\,/\,29.8 (41) & 54.0\,/\,29.4 (29) & 83.3\,/\,50.3 (22) & 44.5\,/\,33.1 (109) \\
USR-TC & 50.0\,/\,42.3 (10) & 50.3\,/\,39.2 (51) & 80.3\,/\,45.8 (39) & 100.0\,/\,58.3 (9) & 65.1\,/\,43.4 (109) \\
USR-PC & 44.4\,/\,16.5 (6) & 47.4\,/\,38.0 (38) & 70.8\,/\,57.4 (24) & 100.0\,/\,50.6 (13) & 62.6\,/\,44.2 (81) \\
\bottomrule
\end{tabular*}

\vspace{0.9em}
\setlength{\tabcolsep}{4pt}
\begin{tabular}{@{}lcrcrrrr@{}}
\toprule
\multicolumn{8}{@{}l}{\textit{(c) The confident-error sample}} \\
Panel & Conf. & \makecell[r]{LLM\\repeats} & \makecell{Luna\,/\,Gemini\\/\,DeepSeek} & \makecell[r]{Ex-\\pected} & \makecell[r]{Same\\answer} & \makecell[r]{Binary\\pairs} & \makecell[r]{Unan.\\raters} \\
\midrule
ELLIPSE & 0.63--0.69 & 16 & 1\,/\,11\,/\,4 & 19.5 & 1 & 0 & 0 \\
FED-Turn & 0.85--0.98 & 35 & 12\,/\,11\,/\,12 & 14.7 & 11 & 0 & 0 \\
FED-Dialogue & 0.79--0.93 & 34 & 12\,/\,10\,/\,12 & 14.9 & 10 & 2 & 0 \\
HelpSteer2 & 0.68--0.79 & 30 & 9\,/\,11\,/\,10 & 13.7 & 8 & 0 & -- \\
LFQA & 0.82--0.96 & 32 & 11\,/\,12\,/\,9 & 20.4 & 9 & 0 & 1 \\
USR-TC & 0.73--0.94 & 35 & 11\,/\,12\,/\,12 & 19.7 & 11 & 4 & 1 \\
USR-PC & 0.77--0.93 & 36 & 12\,/\,12\,/\,12 & 17.7 & 12 & 2 & 3 \\
\midrule
Total & & 218 & 68\,/\,79\,/\,71 & 120.7 & 62 & 8 & \\
\midrule
HealthBench & -- & 16 & 2\,/\,5\,/\,9 & 8.9 & -- & -- & 12 \\
\bottomrule
\end{tabular}
\caption{The Decider's confidence and the errors it shares with the LLM judges, with the definitions of Tables~\ref{tab:confidence} and~\ref{tab:confident-errors}. (a) No ELLIPSE pair reaches confidence 0.75. (b) The ELLIPSE top band holds no error. (c) Conf.: confidence range of Decider Choice's 12 sampled errors on the panel; LLM repeats out of 36, also given per judge; Expected: under the independence baseline; Same answer: sampled errors where every one of the Decider's five judges gives its answer, out of 12; Binary pairs: embedded binary pairs among the 12; Unan.\ raters: as in Table~\ref{tab:confident-errors}. HealthBench: the six false positives of Decider Noul with the highest $P(\text{MET})$ and its six false negatives with the lowest, whose physicians were unanimous on all 12.}
\label{tab:r-confidence}
\end{table*}

\paragraph{Confidence and repeated errors.}
The AUROC of 0.52 on ELLIPSE has a 95\% interval of 0.487 to 0.550 over resampled units, and FED-Dialogue's 0.60, from 227 pairs, an interval of 0.488 to 0.716 that also contains one half (Table~\ref{tab:r-confidence}a). On LFQA \crit{factuality} the AUROC is 0.60 over 117 pairs, and on USR-PC \crit{engaging} it is 0.43, where Decider Choice is wrong on 88.3\% of 60 pairs.

Over all errors, every ordinal panel has more repeated errors than the baseline predicts, by 11.4 to 21.7 points on six panels and by 2.1 on ELLIPSE. Wherever the top band holds errors, its repeat share and its excess over the baseline are the largest of the panel, 83.3\% to 100\% against 40.4\% to 58.3\%, but only FED-Turn and LFQA have the 20 top-band errors that Figure~\ref{fig:confidence}b needs. Two bands fall at or below the baseline, ELLIPSE's lowest, with one error, and LFQA's band from 0.25 to 0.5. Treating abstentions and judge-call failures as verdicts that do not repeat moves no band down by more than 1.6 points. The confident-error sample holds 76 ordinal and 8 embedded binary pairs, and Decider Score gives Decider Choice's answer on all 84. The LLM judges repeat 23 of their 24 verdicts on the binary pairs, and when the 12 errors of each panel are drawn from ordinal pairs alone they still repeat 85.3\% of their verdicts against 49.0\% expected. Of the 72 sampled errors on the six multi-rater panels, 67 fall on split pairs and 5 on pairs whose raters were unanimous.

On ELLIPSE the 12 sampled errors are nine \crit{grammar} pairs placed at level~2 against a label of 3, one placed at 2 against 4, and two \crit{cohesion} pairs placed at 3 against 4. The LLM judges are right on none of their 36 verdicts. Of the 20 verdicts below even the Decider's level, Luna gives 11, DeepSeek 8, and Gemini 1, and 19 of Luna's and DeepSeek's 24 verdicts sit one level under the Decider's. Over all 819 ELLIPSE errors of Decider Choice, 1,266 LLM verdicts repeat its answer, 942 lie further below the label, and 172 are correct. Luna is exact on 5.0\% of the pairs where Decider Choice is wrong and on 24.3\% of those where it is right, Gemini on 9.4\% and 52.8\%, and DeepSeek on 6.6\% and 24.5\%.

On HealthBench, Decider Noul's 114 errors are 78 false positives and 36 false negatives, and the sampled ones sit at $P(\text{MET})$ of 0.983 to 0.988 and 0.0011 to 0.0028. Most of the LLM judges' repeats fall on the false positives (Appendix~\ref{app:shared}). Five of the six false negatives fall on one hedging criterion whose 7 pairs are all labeled MET. The Decider says UNMET on 6 of them, and Luna and Gemini are right on 6 of the 7. Over all errors the LLM judges repeat 43.6\% of their verdicts on Decider Noul's and 46.7\% on Decider Choice's, against a baseline of 25.6\% for each. Decider Noul's HealthBench probabilities are overconfident at both ends. Its 155 pairs with $P(\text{MET})$ of at least 0.875 average 0.952 against a MET rate of 0.781, and its 77 pairs below 0.125 average 0.041 against 0.195.

\begin{table*}[!tp]
\centering
\footnotesize
\setlength{\tabcolsep}{3.6pt}
\begin{tabular}{@{}llrrrrrrrrr@{}}
\toprule
 & & & & \multicolumn{3}{c}{Accuracy (\%)} & & & & \\
\cmidrule(lr){5-7}
Panel & Fallback & $\tau$ & \makecell[r]{Deferred\\(\%)} & Cascade & Decider & Fallback & \makecell[r]{vs.\\fallback} & \makecell[r]{Reach\\(\%)} & \makecell[r]{Halvings with\\a gain (\%)} & \makecell[r]{Held-out\\difference} \\
\midrule
\multicolumn{11}{@{}l}{\emph{(a) Each panel's best cascade (oracle threshold) and the cross-fitted cascades}} \\
RiceChem & DeepSeek & 0.06 & 5.9 & 81.1 & 80.0 & 79.2 & $+1.8$ & 0.0 & 62 & $+1.3$ \\
HealthBench & Gemini & 0.50 & 100.0 & 79.6 & 71.9 & 79.6 & $0.0$ & 100.0 & 80 & $-0.3$ \\
\midrule
ELLIPSE & Gemini & 0.30 & 1.2 & 47.3 & 47.1 & 29.8 & $+17.4$ & 0.0 & 0 & $+17.2$ \\
FED-Turn & Gemini & 0.80 & 71.3 & 65.3 & 56.0 & 65.2 & $+0.2$ & 65.7 & 22 & $-0.4$ \\
FED-Dialogue & Gemini & 0.90 & 97.8 & 59.9 & 45.8 & 59.9 & $0.0$ & 97.8 & 0 & $-2.0$ \\
HelpSteer2 & Gemini & 0.20 & 2.3 & 64.0 & 63.1 & 55.1 & $+8.9$ & 0.0 & 40 & $+8.0$ \\
LFQA & Luna & 0.65 & 38.7 & 73.7 & 69.5 & 71.4 & $+2.2$ & 21.3 & 78 & $-1.0$ \\
USR-TC & Gemini & 0.55 & 47.2 & 71.1 & 69.7 & 70.8 & $+0.3$ & 40.8 & 0 & $-0.6$ \\
USR-PC & Luna & 0.55 & 36.0 & 75.3 & 73.0 & 74.0 & $+1.3$ & 30.3 & 76 & $-0.4$ \\
\bottomrule
\end{tabular}

\medskip
\setlength{\tabcolsep}{4pt}
\begin{tabular}{@{}lrrrlrrrrl@{}}
\toprule
 & \multicolumn{4}{c}{Kept pairs} & & \multicolumn{4}{c}{Deferred pairs} \\
\cmidrule(lr){2-5}\cmidrule(l){7-10}
Panel & $n$ & \makecell[r]{Decider\\(\%)} & \makecell[r]{Fallback\\(\%)} & D\,:\,F ($p$) & & $n$ & \makecell[r]{Decider\\(\%)} & \makecell[r]{Fallback\\(\%)} & D\,:\,F ($p$) \\
\midrule
\multicolumn{10}{@{}l}{\emph{(b) Kept and deferred pairs of the best cascade}} \\
RiceChem & 771 & 81.7 & 79.8 & 53\,:\,38 (0.14) & & 48 & 52.1 & 70.8 & 8\,:\,17 (0.11) \\
HealthBench & 0 & -- & -- & -- & & 406 & 71.9 & 79.6 & 32\,:\,63 (0.002) \\
\midrule
ELLIPSE & 1{,}530 & 47.4 & 29.7 & 342\,:\,72 ($4 \times 10^{-43}$) & & 18 & 22.2 & 38.9 & 2\,:\,5 (0.45) \\
FED-Turn & 172 & 90.1 & 89.5 & 2\,:\,1 (1.00) & & 428 & 42.3 & 55.4 & 20\,:\,76 ($7 \times 10^{-9}$) \\
FED-Dialogue & 5 & 80.0 & 80.0 & 0\,:\,0 & & 222 & 45.0 & 59.5 & 12\,:\,44 ($2 \times 10^{-5}$) \\
HelpSteer2 & 342 & 64.6 & 55.6 & 56\,:\,25 ($8 \times 10^{-4}$) & & 8 & 0.0 & 37.5 & 0\,:\,3 (0.25) \\
LFQA & 219 & 86.3 & 82.6 & 19\,:\,11 (0.20) & & 138 & 42.8 & 53.6 & 32\,:\,47 (0.11) \\
USR-TC & 190 & 80.5 & 80.0 & 5\,:\,4 (1.00) & & 170 & 57.6 & 60.6 & 28\,:\,33 (0.61) \\
USR-PC & 192 & 84.9 & 82.8 & 8\,:\,4 (0.39) & & 108 & 51.9 & 58.3 & 12\,:\,19 (0.28) \\
\bottomrule
\end{tabular}

\medskip
\setlength{\tabcolsep}{3.5pt}
\begin{tabular}{@{}llrrlrrlr@{}}
\toprule
 & \multicolumn{2}{c}{Most accurate matched judge} & \multicolumn{2}{c}{Median of three LLM judges} & \multicolumn{3}{c}{Median of four judges} & \\
\cmidrule(lr){2-3}\cmidrule(lr){4-5}\cmidrule(lr){6-8}
Panel & Judge & Acc. & $\Delta$ & 95\% interval & $\Delta$ lower & $\Delta$ upper & 95\% interval & All wrong \\
\midrule
\multicolumn{9}{@{}l}{\emph{(c) Median juries}} \\
RiceChem & Decider Choice & 80.3 & $-2.2$ & $-4.5$, $+0.2$ & $-1.6$ & $-0.1$ & $-2.0$, $+1.7$ & -- \\
HealthBench & Gemini & 79.6 & $-2.2$ & $-4.7$, $+0.2$ & $-3.2$ & $+0.7$ & $-1.5$, $+3.1$ & -- \\
\midrule
ELLIPSE & Decider Choice & 47.2 & $-30.1$ & $-33.1$, $-27.0$ & $-30.3$ & $-14.6$ & $-17.1$, $-12.3$ & 45.8 \\
FED-Turn & Gemini & 65.3 & $-0.2$ & $-3.0$, $+2.5$ & $-4.8$ & $+0.3$ & $-2.5$, $+3.0$ & 23.3 \\
FED-Dialogue & Gemini & 60.5 & $-3.1$ & $-7.6$, $+1.8$ & $-8.5$ & $-1.3$ & $-5.8$, $+3.2$ & 30.3 \\
HelpSteer2 & Decider Choice & 63.2 & $-16.2$ & $-21.9$, $-10.7$ & $-15.1$ & $-3.8$ & $-8.5$, $+0.9$ & 22.6 \\
LFQA & Luna & 71.5 & $-2.0$ & $-5.1$, $+1.1$ & $-2.5$ & $-0.8$ & $-5.3$, $+3.7$ & 11.9 \\
USR-TC & Gemini & 70.9 & $0.0$ & $-2.5$, $+2.5$ & $+0.3$ & $-0.3$ & $-3.7$, $+3.1$ & 17.7 \\
USR-PC & Gemini & 74.3 & $-0.7$ & $-2.7$, $+1.3$ & $-2.0$ & $+2.0$ & $-1.0$, $+5.0$ & 23.9 \\
\bottomrule
\end{tabular}
\caption{Decider-first cascades and juries (post-hoc analyses), by the rules of Appendix~\ref{app:cascades} and Table~\ref{tab:juries}. (a) The ledger columns of Table~\ref{tab:cascade-ledger}, for the fallback that each panel's best cascade uses, whose first stage is Decider Noul on the binary panels and Decider Choice on the ordinal ones; Decider: that first stage alone. The last two columns belong to the cross-fitted cascades of Table~\ref{tab:cascade}, defined as in Table~\ref{tab:cascade-halvings}. At HealthBench's oracle threshold every pair is deferred. (b) As Table~\ref{tab:cascade-split}, with D\,:\,F counting pairs that only the Decider or only the fallback got right. (c) As Table~\ref{tab:juries}, the four-judge jury including Decider Choice, with an interval for the upper median as well.}
\label{tab:r-cascades}
\end{table*}

\paragraph{Cascades.}
The best Decider-first cascades defer to Gemini on six panels, to Luna on LFQA and USR-PC, and to DeepSeek on RiceChem, and they defer from 1.2\% of the pairs on ELLIPSE to all of them on HealthBench (Table~\ref{tab:r-cascades}a). On HealthBench the protocol's first stage, Decider Noul, is the Decider's least accurate binary framing (Table~\ref{tab:binary}). Recomputed from the verdicts, Equation~\eqref{eq:identity} holds at every reported threshold, and on ELLIPSE the cascade's lead over Gemini, 17.4 points, is the kept share of 0.988 times the Decider's 17.7-point lead on the kept pairs. Cross-fitted Decider-first cascades trail the best single judge on five panels (Table~\ref{tab:cascade}), and the largest oracle gain, 1.7 points on LFQA, is not significant in a kept-pair sign test ($p = 0.20$). The Decider's lead on kept pairs is significant only on ELLIPSE and HelpSteer2, the two ordinal panels with Decider Choice at the top of the matched judges, and the fallback is ahead on the deferred pairs of all nine panels, significantly on FED-Turn, FED-Dialogue, and HealthBench (Table~\ref{tab:r-cascades}b). Counting kept and deferred pairs together, the Decider alone is right on 344 ELLIPSE pairs against Gemini's 77, on 56 HelpSteer2 pairs against 28, and on 12 to 51 pairs of each other ordinal panel against 23 to 77.

With an oracle threshold, a cascade matches the best LLM judge without deferring any pair on RiceChem, ELLIPSE, and HelpSteer2, but must defer nearly every pair on FED-Dialogue and every pair on HealthBench. Cross-fitted, the cascade that matches that judge on one half beats it on the other half of ELLIPSE by 17.2 points at 0.3\% of its cost, of HelpSteer2 by 8.0 points at 0.4\%, and of RiceChem by 1.3 points at 5.3\%. On LFQA, USR-TC, and USR-PC it costs 17.9\% to 28.1\% of that judge for 0.4 to 1.0 points less, and on FED-Turn, FED-Dialogue, and HealthBench 66.3\% to 97.3\% for 0.3 to 2.0 points less (Table~\ref{tab:cascade}).

\paragraph{Juries.}
Appendix~\ref{app:shared} discusses the juries of Table~\ref{tab:r-cascades}c. Picking a correct matched judge wherever one exists would reach 54.2\% on ELLIPSE, against Decider Choice's 47.1\% exact accuracy and its 47.2\% on the pairs all four scored. On ELLIPSE and HelpSteer2, where the three-LLM median is significantly less accurate than Decider Choice, it also trails the most accurate LLM judge, by 12.8 and 8.1 points. With Decider Choice as a fourth juror, the upper median comes out above the most accurate matched judge on HealthBench, FED-Turn, and USR-PC, by at most 2.0 points, with no interval above zero. The four-judge median is not unique on 7.8\% (RiceChem) to 38.6\% (ELLIPSE) of the pairs.

\begin{table*}[!tp]
\centering
\footnotesize
\setlength{\tabcolsep}{3.5pt}
\renewcommand{\arraystretch}{1.1}
\begin{tabular}{@{}>{\raggedright\arraybackslash}p{3.0cm}>{\raggedright\arraybackslash}p{3.0cm}>{\raggedright\arraybackslash}p{3.0cm}>{\raggedright\arraybackslash}p{3.0cm}>{\raggedright\arraybackslash}p{3.0cm}@{}}
\toprule
 & Per-criterion & Whole-rubric & High effort & Higher tier \\
\midrule
\multicolumn{5}{@{}l}{\emph{RQ1}} \\
Cost ($\times$ Choice) & 32.0\,/\,357.0\,/\,72.5 & 17.6\,/\,199.8\,/\,59.0 & 39.9\,/\,741.7\,/\,102.2 & 255.4\,/\,1{,}080.3\,/\,74.0 \\
\quad at the launch price & 16.0\,/\,178.5\,/\,36.3 & 8.8\,/\,99.9\,/\,29.5 & 19.9\,/\,370.8\,/\,51.1 & 127.7\,/\,540.2\,/\,37.0 \\
Wall time ($\times$ Choice) & 5.8\,/\,6.4\,/\,42.4 & 5.4\,/\,8.7\,/\,67.5 & 8.3\,/\,22.6\,/\,-- & 6.7\,/\,31.3\,/\,-- \\
Sep.\,/\,parity\,/\,incon. & 13 (1)\,/\,5\,/\,9 & 14 (0)\,/\,5\,/\,8 & 10 (0)\,/\,5\,/\,12 & 13 (1)\,/\,4\,/\,10 \\
Holm: leads, deficits & 7, 4 & 6, 4 & 7, 3 & 6, 5 \\
\addlinespace[2pt]
Leads & RiceChem L, G; ELLIPSE, HelpSteer2 all & as per-criterion & RiceChem G; ELLIPSE, HelpSteer2 all & RiceChem G; ELLIPSE, HelpSteer2 all \\
\addlinespace[2pt]
Deficits & FED-Turn all; FED-Dialogue L, G & HealthBench G; FED-Turn all; FED-Dialogue G, D & FED-Turn G, D; FED-Dialogue G & HealthBench G; FED-Turn all; FED-Dialogue G, D \\
\addlinespace[2pt]
\multicolumn{5}{@{}l}{\emph{RQ2}} \\
Judge--judge $>$ label & 7 & 7 & 7 & 7 \\
\quad weakest pair $>$ label & 2 & 3 & 4 & 4 \\
Offsets all negative & 4 & 4 & 4 & 4 \\
Confident repeats (\%) & 86.5 (47.9) & 88.1 (48.9) & 86.9 (47.3) & 79.8 (47.8) \\
\quad verdicts repeating & 218 of 252 & 222 of 252 & 218 of 251 & 201 of 252 \\
\quad ELLIPSE, of 36 & 16 (19.5) & 19 (19.8) & 17 (18.3) & 9 (16.7) \\
\quad HealthBench, of 36 & 16 (8.9) & 13 (7.4) & 15 (6.9) & 15 (7.8) \\
\addlinespace[2pt]
\multicolumn{5}{@{}l}{\emph{RQ3}} \\
Gain, cross-fitted\,/\,oracle & 1.18\,/\,1.68, both LFQA & 0.79 (HealthBench)\,/\,1.68 (LFQA) & 0.75 (LFQA)\,/\,1.14 (HelpSteer2) & 1.41\,/\,1.76, both FED-Dialogue \\
\addlinespace[2pt]
Either tier & 1.33\,/\,1.68, both LFQA & -- & 0.82 (LFQA)\,/\,1.14 (HelpSteer2) & 0.42 (HealthBench)\,/\,1.43 (HelpSteer2) \\
\addlinespace[2pt]
Cost under half & 6, all but FED-Turn, FED-Dialogue, and HealthBench & 4, ELLIPSE, HelpSteer2, LFQA, and USR-PC & 6, as per-criterion & 6, as per-criterion \\
\bottomrule
\end{tabular}
\caption{The Decider's rows of Table~\ref{tab:conditions} with their components. Triples, and L, G, and D: Luna, Gemini, and DeepSeek in the flash-tier conditions and Terra, Gemini 3.1 Pro, and DeepSeek V4 Pro in the higher tier, DeepSeek's judges at their billed cost. Launch price: the price in Perplexity's changelog, given in the paragraph \emph{Price, cost, and time}. Sep.: separated comparisons, with those also at parity in parentheses; incon.: inconclusive. Holm: separations still significant after Holm correction, applied to the bootstrap $p$-values of the condition's 27 comparisons. Leads and deficits: separations by panel and LLM judge; ``all'' means all three LLM judges. Weakest pair: ordinal panels on which the matched judges' least-agreeing pair exceeds the best judge--label QWK. Confident repeats: share of LLM verdicts repeating Decider Choice's confident errors (baseline share), with their count below; for ELLIPSE and for HealthBench as in Table~\ref{tab:r-confidence}c, with the expected count in parentheses. Gains in points, with their panels.}
\label{tab:r-conditions}
\end{table*}

\paragraph{Conditions.}
Table~\ref{tab:r-conditions} spells out the Decider's rows of Table~\ref{tab:conditions}. With the Decider's verdicts fixed, the LLM judges stay far costlier and slower throughout, and 10 to 14 of Decider Choice's 27 comparisons are separated in each condition, most of them Decider leads. The Decider's leads fall on RiceChem, ELLIPSE, and HelpSteer2 in every condition, and its deficits on FED-Turn and FED-Dialogue, joined on HealthBench by Gemini in the whole-rubric condition and by Gemini 3.1 Pro in the higher tier. The sign test over units finds that HealthBench deficit in all four conditions. Between the two judging modes 3 of the 27 classes change between separated and not, and high effort keeps 10 of the 13 per-criterion separations, reversing none and adding none. The Decider's matched judges agree more closely among themselves than with the labels on every ordinal panel, though every matched offset is negative on only four of these panels. Whatever the condition, the LLM judges also repeat Decider Choice's confident errors far above baseline, and the confident-error share falls short of the 80\% fixed before the run (Table~\ref{tab:r-claims}) only in the higher tier, where 202 repeated verdicts rather than 201 would have reached it. On ELLIPSE the share never exceeds the baseline. No Decider-first cascade gains as much as 2 points. With flash-tier fallbacks the largest gain over a panel's best single judge in any condition is 1.7 points (oracle) or 1.2 (cross-fitted), and with higher-tier fallbacks the largest over the best single judge from either tier is 1.4 or 0.4. In every condition the panels with a cross-fitted cascade under half the cost include ELLIPSE and HelpSteer2. In the main runs Decider Choice by itself reaches the best LLM judge's accuracy on both, and every condition's matching cascade beats that judge on the held-out half.

\begin{table}[!tbp]
\centering
\footnotesize
\setlength{\tabcolsep}{3pt}
\begin{tabular}{@{}lrlrrr@{}}
\toprule
Panel & Units & Framings & Pairs & \makecell[r]{Same\\verdict} & \makecell[r]{Same\\prob.} \\
\midrule
RiceChem & 14 & B, N, C & 3\,$\times$\,95 & 100 & 100 \\
HealthBench & 20 & B, N, C, I & 4\,$\times$\,42 & 100 & 100 \\
\midrule
ELLIPSE & 26 & C, S & 2\,$\times$\,156 & 100 & 100 \\
FED-Turn & 8 & C, S & 2\,$\times$\,64 & 100 & 100 \\
FED-Dialogue & 3 & C, S & 2\,$\times$\,30 & 100 & 100 \\
HelpSteer2 & 9 & C, S & 2\,$\times$\,36 & 100 & 100 \\
LFQA & 12 & C, S & 2\,$\times$\,36 & 100 & 100 \\
USR-TC & 8 & C, S & 2\,$\times$\,40 & 100 & 100 \\
USR-PC & 6 & C, S & 2\,$\times$\,30 & 100 & 100 \\
\midrule
All & 106 & & 1{,}237 & 100 & 100 \\
\bottomrule
\end{tabular}
\caption{The Decider's repeat: the share (\%) of repeated pairs with the same verdict (level, MET or UNMET, or abstention) and with the same returned probabilities in both runs, minutes apart. Framings: B, bare Noul; N, Noul; C, Choice; S, Score; I, Noul with the physician-written ideal. Pairs: framings times pairs per framing. Over the ordinal panels, each framing's mean QWK between the two runs is 1.0 (25 Choice and 27 Score criteria), and Cohen's $\kappa$ on the 39 embedded binary pairs is 1.0.}
\label{tab:r-repeat}
\end{table}

\paragraph{The repeat.}
The repeat resent unchanged every request that the full run had made for the repeat units (Section~\ref{sec:llm}), 266 requests covering 1,237 pairs, each matched to its first sending by the hash of its body. A repeated request went out 98 to 924 seconds after the first (median 811), so the repeat tests stability over minutes only. Every verdict and every returned probability came back the same (Table~\ref{tab:r-repeat}). So did the answers to all 659 duplicate requests of the study, 3,865 criterion answers in all. Besides the repeat, these include the RiceChem panel's requests, sent a second time 85 to 99 seconds later as part of the full set, and the 30 requests of the smoke run. The client keeps no response cache, the 7,000 request identifiers that the API returned are all distinct, and repeated requests took longer than first sendings (median 0.41 against 0.31 seconds), though a cache on the server with ordinary latency cannot be ruled out. Perplexity documents that identical requests usually get identical answers and occasionally differ in the second decimal. In design, the repeat differs from Jev's five interleaved passes (Appendix~\ref{app:protocol}), which we do not set beside it. For scale, the LLM judges' whole-rubric repeat runs reproduced 86.2\% to 92.6\% of their verdicts (Table~\ref{tab:wr-agreement}).

\begin{table*}[!tp]
\centering
\footnotesize
\setlength{\tabcolsep}{3.5pt}
\begin{tabular}{@{}>{\raggedright\arraybackslash}p{3.0cm}>{\raggedright\arraybackslash}p{4.1cm}>{\raggedright\arraybackslash}p{5.7cm}>{\raggedright\arraybackslash}p{2.3cm}@{}}
\toprule
Claim & Holds if (fixed before the run) & Decider's copy & Outcome \\
\midrule
LLM judges cost far more & every LLM judge at least 10 times Decider Choice's nine-panel cost, per-criterion runs & 32, 357, and 73 times (16, 178, and 36 at the launch price) & holds \\
\addlinespace[2pt]
LLM judges take far longer & every LLM judge at least 5 times Decider Choice's wall time & 5.8, 6.4, and 42.4 times; lowest in any condition 5.4 (whole-rubric Luna) & holds, narrowly for Luna \\
\addlinespace[2pt]
The decision model is close to the LLM judges & at most 9 of 27 comparisons separated, deficits only on ordinal panels & 13 of 27 separated (14, 10, and 13 in the other conditions); the five deficits ordinal by the intervals, one on HealthBench by the unit sign test & missed \\
\addlinespace[2pt]
A shared departure on ordinal criteria & judge--judge above judge--label QWK on 7 of 7 panels; offsets negative on at least 5 of 7 & 7 of 7; every matched offset negative on 4, Decider Choice's positive on HelpSteer2, LFQA, and USR-TC & missed for offsets \\
\addlinespace[2pt]
LLM judges repeat the confident errors & at least 80\% of LLM verdicts repeat, against about half expected & 86.5\% against 47.9\%; 88.1\%, 86.9\%, and 85.3\% in the whole-rubric, high, and medium conditions; 79.8\% against 47.8\% in the higher tier & missed in the higher tier, by one verdict \\
\addlinespace[2pt]
Cascades add little & largest oracle gain at most 3 points, Decider first, flash-tier fallbacks & 1.7 points (LFQA) & holds \\
\addlinespace[2pt]
Confidence is calibrated & Decider Noul's ECE at most 0.10 on the binary panels & 0.084 on RiceChem, 0.148 on HealthBench & missed on HealthBench \\
\bottomrule
\end{tabular}
\caption{The thresholds recorded before the Decider's first answer on a study item, each evaluated on its own copy of the study, with the outcome. Per-criterion runs unless a condition is named.}
\label{tab:r-claims}
\end{table*}

\paragraph{Claims fixed before the run.}
We recorded the claims to check on the Decider's copy at 01:49:40 UTC on 7 October 2026, each with a threshold (Table~\ref{tab:r-claims}) or with the test that would decide its wording in the paper. Until then the Decider had answered only the two probe requests. It first answered on a study item at 04:19:49 UTC, in the smoke run, and the full run started at 04:21:22. Four of the seven thresholds were missed. Three were missed outright, and the one on repeated confident errors only in the higher tier. After the run, we tested ten statements of the paper against the Decider's copy, each by the test recorded with the thresholds. Three hold as worded for both decision models, those on cost, wall time, and denominators. Every LLM judge cost more than Decider Choice on every panel in every condition, at either price, and took longer on every panel whose time was validly measured, and the denominators needed only the Decider's own counts. The other seven hold for the Decider in another form, and the text names the model wherever it states them. They are how often a decision model differs from the LLM judges (Section~\ref{sec:accuracy}), the shared shortfall on ordinal criteria (Section~\ref{sec:agreement}), the repeated confident errors (Section~\ref{sec:shared-errors}), the ranking of errors by confidence and the calibration on binary panels (Section~\ref{sec:confidence}), the bound on cascade gains (Sections~\ref{sec:intro} and~\ref{sec:cascades}), the claim that the answers hold in all four conditions (Table~\ref{tab:conditions}), and the stability of repeated answers (see \ref{sec:limitations}). After testing these ten, we added an eleventh, the account through unstated conventions of Appendix~\ref{app:conventions}. It does not fit the Decider, so Section~\ref{sec:conventions} confines it to Jev and the LLM judges and reports the Decider as an exception.

\paragraph{Analyses run for Jev alone.}
Only Jev took part in the five-pass stability study (Appendix~\ref{app:protocol}), and the expectations that Appendix~\ref{app:binaryobs} re-examines were formed before the Decider was run. The appendices that analyze Jev's copy in depth add the Decider's values where its copy bears on a statement made there, but several of their analyses remain Jev's alone. Among them are the threshold set on RiceChem's training portion (Appendix~\ref{app:binary}), the exchange of decoding rules and the middle level on FED-Turn (Appendix~\ref{app:choicescore}), the reading of individual abstained responses (Appendix~\ref{app:abstention}), the AUROC of every criterion and Jev Score's error rate by its own confidence (Appendix~\ref{app:confidence}), and the per-panel comparisons of the whole-rubric, effort, and higher-tier conditions (Appendices~\ref{app:wholerubric} to~\ref{app:highertier}).